\documentclass[oneside,11pt]{article}

\usepackage{times,url,mathrsfs}
\usepackage{amsmath,wrapfig,color}
\usepackage{amsfonts}
\usepackage{graphicx}
\usepackage{subfigure}

\newtheorem{lemma}{Lemma}

\begin{document}

\title{One Permutation Hashing for Efficient Search and Learning}

\author{ Ping Li \\
         Dept. of Statistical Science\\
       Cornell University\\
         Ithaca, NY 14853\\
       pingli@cornell.edu
       \and
       Art Owen\\
         Dept. of Statistics\\
         Stanford University\\
         Stanford, CA 94305\\
         owen@stanford.edu
         \and
         Cun-Hui Zhang\\
         Dept. of Statistics\\
         Rutgers University\\
         New Brunswick, NJ 08901\\
         czhang@stat.rutgers.edu
        }
\date{}
\maketitle

\begin{abstract}

\noindent Minwise hashing is a standard procedure in the context of search, for efficiently estimating set similarities in massive binary data such as text. Recently, the method of $b$-bit minwise hashing has been applied to large-scale linear learning (e.g., linear SVM or logistic regression) and sublinear time near-neighbor search. The major drawback of minwise hashing is the  expensive preprocessing cost, as the method requires applying (e.g.,) $k=200$ to 500 permutations on the data. The testing time can also be expensive if a new data point (e.g., a new document or image) has not been processed, which might be a significant issue in user-facing applications. While it is true that the preprocessing step can be parallelized, it comes at the cost of additional hardware \& implementation and is not an energy-efficient solution.\\

\noindent We develop a very simple solution based on \textbf{one permutation hashing}. Conceptually, given a massive binary data matrix, we permute the columns only once and divide the permuted columns evenly into $k$ bins; and we simply store, for each data vector, the smallest nonzero location in each bin. The interesting probability analysis (which is validated by experiments)   reveals that our one permutation scheme should perform very similarly to the original ($k$-permutation) minwise hashing. In fact, the one permutation scheme can be even slightly more accurate, due to the ``sample-without-replacement'' effect. \\

\noindent Our experiments with training linear SVM and logistic regression on the {\em webspam} dataset demonstrate that this one permutation hashing scheme can achieve the same  (or even slightly better) accuracies compared to  the original $k$-permutation scheme. To test the robustness of our method, we also experiment with the small {\em news20} dataset which is very sparse and has merely on average 500 nonzeros in each data vector. Interestingly, our one permutation scheme noticeably outperforms the $k$-permutation scheme when $k$ is not too small on the {\em news20} dataset. In summary, our method can achieve at least the same accuracy as the original $k$-permutation scheme, at merely $1/k$ of the original preprocessing cost.


\end{abstract}

\section{Introduction}

Minwise hashing~\cite{Proc:Broder_WWW97,Proc:Broder_STOC98} is a standard technique for efficiently computing  set similarities, especially in the context of search. Recently, $b$-bit minwise hashing~\cite{Article:Li_Konig_CACM11}, which stores only the lowest $b$ bits of each hashed value,   has been applied to sublinear time near neighbor search~\cite{Proc:Shrivastava_ECML12} and linear learning (linear SVM and logistic regression)~\cite{Proc:HashLearning_NIPS11}, on large-scale high-dimensional binary data (e.g., text), which are common in practice.  The major drawback of minwise hashing and $b$-bit minwise hashing is that they require an expensive preprocessing step, by conducting $k$ (e.g., 200 to 500) permutations on the entire dataset.

\subsection{Massive High-Dimensional Binary Data}

In the context of search, text data are often processed to be binary in extremely high dimensions. A standard procedure is to represent documents (e.g., Web pages) using $w$-shingles (i.e., $w$ contiguous words), where  $w\geq5$ in several studies~\cite{Proc:Broder_WWW97,Proc:Fetterly_WWW03}. This means the size of the dictionary needs to be substantially increased, from (e.g.,) $10^5$ common English words to $10^{5w}$ ``super-words''. In current practice, it seems sufficient to set the total dimensionality to be $D = 2^{64}$, for convenience.  Text data generated by $w$-shingles are often treated as binary. In fact, for $w\geq 3$, it is expected that most of the $w$-shingles will occur at most one time in a document. Also, note that the idea of shingling can be naturally extended to images in Computer Vision, either at the pixel level (for simple aligned images) or at the Vision  feature level~\cite{Proc:Sivic_ICCV03}.

In machine learning practice, the use of extremely high-dimensional data has become common.  For example, \cite{GoogleBlog} discusses training datasets with (on average) $n=10^{11}$ items and $D=10^9$ distinct features.  \cite{Proc:Weinberger_ICML2009} experimented with a dataset of potentially $D=16$ trillion ($1.6\times10^{13}$) unique features.

\subsection{Minwise Hashing}\label{sec_minwise}

Minwise hashing is mainly designed for binary data. A binary (0/1) data vector can be equivalently viewed as a set (locations of the nonzeros).  Consider sets $S_i \subseteq \Omega=\{0, 1, 2, ..., D-1\}$, where $D$, the size of the space, is often set to be  $D = 2^{64}$ in industrial applications. The similarity between two sets $S_1$ and $S_2$ is commonly measured by the {\em resemblance}, which is a normalized version of the inner product:
\begin{align}
R = \frac{|S_1\cap S_2|}{|S_1\cup S_2|}  = \frac{a}{f_1+f_2-a},\hspace{0.2in} \text{where } f_1 = |S_1|, \ f_2 = |S_2|, \ a = |S_1\cap S_2|
\end{align}

For large-scale applications, the cost of computing   resemblances exactly can be prohibitive in time,  space, and energy-consumption. The minwise hashing method was proposed for efficient computing resemblances. The method requires applying $k$ independent random permutations on the data.

Denote $\pi$ a random permutation: $\pi: \Omega \rightarrow \Omega$. The hashed values are the two minimums of the sets after applying the permutation $\pi$ on $S_1$ and $S_2$. The  probability at which the two hashed  values are equal is
\begin{align}
&\mathbf{Pr}\left(\min(\pi(S_1)) = \min(\pi(S_2))\right) = \frac{|S_1\cap S_2|}{|S_1\cup S_2|} = R
\end{align}
One can then estimate $R$ from $k$  independent permutations, $\pi_1$, ..., $\pi_k$:
\begin{align}\label{eqn_Var_M}
&\hat{R}_{M} = \frac{1}{k}\sum_{j=1}^{k}1\{{\min}({\pi_j}(S_1)) =
  {\min}({\pi_j}(S_2))\}, \hspace{0.3in}
\text{Var}\left(\hat{R}_{M}\right) = \frac{1}{k}R(1-R)
\end{align}

Because the indicator function $1\{{\min}({\pi_j}(S_1)) =  {\min}({\pi_j}(S_2))\}$ can  be written as an inner product between two binary vectors (each having only one 1)  in $D$ dimensions~\cite{Proc:HashLearning_NIPS11}:
\begin{align}
1\{{\min}({\pi_j}(S_1)) =  {\min}({\pi_j}(S_2))\} = \sum_{i=0}^{D-1} 1\{{\min}({\pi_j}(S_1)) = i\} \times 1\{{\min}({\pi_j}(S_2)) = i\}
\end{align}
we know that minwise hashing can be potentially used for training linear SVM and logistic regression on high-dimensional binary data by converting the permuted data into a new data matrix in $D\times k$ dimensions. This of course would not be realistic if $D=2^{64}$.

The method of $b$-bit minwise hashing~\cite{Article:Li_Konig_CACM11} provides a simple solution by storing only the lowest $b$ bits of each hashed data. This way, the dimensionality of the expanded data matrix from the hashed data would be only $2^b \times k$ as opposed to $2^{64}\times k$. \cite{Proc:HashLearning_NIPS11} applied this idea to large-scale learning on the {\em webspam} dataset (with about 16 million features) and demonstrated that using $b=8$ and $k=200$ to 500 could achieve very similar accuracies as using the original data. More recently, \cite{Proc:Shrivastava_ECML12} directly used the bits generated by $b$-bit minwise hashing for building hash tables to achieve sublinear time near neighbor search. We will briefly review these two important applications in Sec.~\ref{sec_bbit}. Note that both applications require the hashed data to be ``aligned'' in that only the hashed data generated by the same permutation are interacted. For example, when computing the inner products, we simply concatenate the results from $k$ permutations.

\subsection{The Cost of Preprocessing and Testing}

Clearly, the preprocessing step of minwise hashing can be very costly. For example,  in our experiments, loading the  {\em webspam} dataset (350,000 samples, about 16 million features, and about 24GB in Libsvm/svmlight format) used in~\cite{Proc:HashLearning_NIPS11} took about $1000$ seconds when the data are stored in Libsvm/svmlight (text) format, and took about $150$ seconds after we converted the data into binary. In contrast, the preprocessing cost for $k=500$ was about 6000 seconds (which is $\gg150$). Note that, compared to industrial applications~\cite{GoogleBlog}, the {\em webspam}  dataset is very  small. For larger datasets, the preprocessing step will be much more expensive.

In the testing phrase (in search or learning), if a new data point (e.g., a new document or a new image) has not processed, then the cost will be expensive if it includes the preprocessing cost. This may raise significant issues in user-facing applications where the testing efficiency is crucial.\\

Intuitively, the standard practice of minwise hashing ought to be very ``wasteful'' in that all the nonzero elements in one set are scanned (permuted) but only the smallest one will be used.

\subsection{Our Proposal: One Permutation Hashing}

\begin{figure}[h!]
\begin{center}
\includegraphics[width=3in]{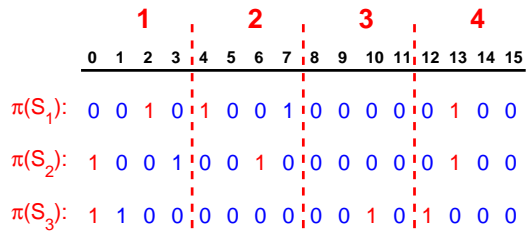}
\end{center}
\vspace{-0.3in}
\caption{\textbf{Fixed-length hashing scheme}. Consider $S_1, S_2, S_3\subseteq\Omega = \{0, 1, ..., 15\}$ (i.e., $D=16$). We apply one permutation $\pi$ on the three sets and present $\pi(S_1)$, $\pi(S_2)$, and $\pi(S_3)$ as binary (0/1) vectors, where $\pi(S_1) = \{2,4,7,13\}$, $\pi(S_2) = \{0, 6, 13\}$, and $\pi(S_3) = \{0, 1, 10, 12\}$. We divide the space $\Omega$ evenly into $k=4$ bins, select the smallest nonzero in each bin, and \textbf{re-index} the selected elements as three  samples: $[2,\ 0,\ *,\ 1]$, $[0,\ 2,\ *,\ 1]$, and $[0,\ *,\ 2,\ 0]$.  For now, we use `*' for empty bins, which occur rarely unless the  number of nonzeros is  small compared to $k$.  }\label{fig_one_permutation}\vspace{-0in}
\end{figure}

As illustrated in Figure~\ref{fig_one_permutation}, the idea of  {\em one permutation hashing} is very  simple. We view sets as 0/1 vectors in $D$ dimensions so that we can treat a collection of sets as a binary data matrix in $D$ dimensions. After we permute the columns (features) of the data matrix, we divide the columns evenly into $k$ parts (bins) and we simply take, for each data vector, the smallest nonzero element in each bin.

\vspace{-0.1in}

In the example in Figure~\ref{fig_one_permutation} (which concerns 3 sets), the sample selected from $\pi(S_1)$ is $[2,4,*,13]$, where we use '*' to denote an empty bin, for the time being. Since only want to compare elements with the same bin number (so that we can obtain an inner product), we can actually re-index the elements of each bin to use the smallest possible representations. For example, for $\pi(S_1)$, after re-indexing, the sample $[2,4,*,13]$  becomes $[2-4\times0,4-4\times1,*,13-4\times3] = [2,0,*,1]$. Similarly, for $\pi(S_2)$, the original sample $[0,6,*,13]$ becomes $[0,6-4\times1,*,13-4\times3] =[0,2,*,1]$, etc.

Note that, when there are no empty bins, similarity estimation is equivalent to computing an inner product, which is crucial for taking advantage of the modern linear learning algorithms~\cite{Proc:Joachims_KDD06,Proc:Shalev-Shwartz_ICML07,Article:Fan_JMLR08,Proc:Hsieh_ICML08}. We will show that empty bins occur rarely unless the total number of nonzeros for some set is small compared to $k$, and we will present strategies on how to deal with empty bins should they occur.

\subsection{Summary of the Advantages of One Permutation Hashing}

\begin{itemize}
\item Reducing $k$ (e.g., 500) permutations to just one permutation (or a few) is much more computationally efficient. From the perspective of energy consumption, this scheme is highly desirable, especially considering that minwise hashing is deployed in the search industry.
\item While it is true that the preprocessing can be parallelized, it comes at the cost of additional hardware and software implementation.
\item In the testing phase, if a new data point (e.g., a new document or a new image) has to be first processed with $k$ permutations, then the testing performance may not meet the demand in for example user-facing applications such as search or interactive visual analytics.
\item It should be much easier to implement the one permutation hashing than the original $k$-permutation scheme, from the perspective of random number generation. For example, if a dataset has one billion features ($D=10^9$), we can simply generate a ``permutation vector'' of length $D=10^9$, the memory cost of which (i.e., 4GB) is not significant.  On the other hand, it would not be realistic to store a ``permutation matrix'' of size $D\times k$ if $D=10^9$ and $k=500$; instead, one usually has to resort to approximations such as using universal hashing~\cite{Proc:Carter_STOC77} to approximate permutations. Universal hashing often works well in practice although theoretically there are always worst cases. Of course, when $D=2^{64}$, we  have to use universal hashing, but it is always much easier to generate just one permutation.
\item One permutation hashing is a better matrix sparsification scheme than the original $k$-permutation. In terms of the original binary data matrix, the one permutation scheme simply makes many nonzero entries be zero, without further ``damaging'' the original data matrix. With the original $k$-permutation scheme, we store, for each permutation and each row, only the first nonzero and make all the other nonzero entries be zero; and then we have to concatenate $k$ such data matrices. This will significantly change the structure of the original data matrix. As a consequence, we expect that our one permutation scheme will produce at least the same or even more accurate results, as later verified by experiments.

\end{itemize}

\subsection{Related Work }

One of the authors worked on another ``one permutation'' scheme named  {\em Conditional Random Sampling (CRS)}~\cite{Proc:Li_Church_EMNLP,Proc:Li_Church_Hastie_NIPS08} since 2005. Basically, CRS works by continuously taking the first $k$ nonzeros after applying one permutation on the data, then it uses a simple ``trick'' to construct a random sample for each pair with the effective sample size determined at the estimation stage. By taking the nonzeros continuously, however, the samples are no longer ``aligned'' and hence we can not write the estimator as an inner product in a unified fashion. In comparison, our new one permutation scheme works by first breaking the columns evenly into $k$ bins and then taking the first nonzero in each bin, so that the hashed data can be nicely aligned.

Interestingly, in the original ``minwise hashing'' paper~\cite{Proc:Broder_WWW97} (we use quotes because the scheme was not called ``minwise hashing'' at that time), only one permutation was used and a sample was the first $k$ nonzeros  after the permutation.  After the authors of~\cite{Proc:Broder_WWW97} realized that the estimators could not be written as an inner product and hence the scheme was not suitable for many applications such as sublinear time near neighbor search using hash tables, they quickly moved to the $k$-permutation minwise hashing scheme~\cite{Proc:Broder_STOC98}.
 In the context of large-scale linear learning, the importance of having estimators which are inner products should become  more obvious after~\cite{Proc:HashLearning_NIPS11} introduced the idea of using ($b$-bit) minwise hashing for  linear learning.

We are also inspired by the work on ``very sparse random projections''~\cite{Proc:Li_Hastie_Church_KDD06}. The regular random projection method also has the expensive preprocessing cost as it needs $k$ projections. The work of~\cite{Proc:Li_Hastie_Church_KDD06} showed that one can substantially reduce the preprocessing cost by using an extremely sparse projection matrix. The preprocessing cost of ``very sparse random projections'' can be as small as merely doing one projection.\footnote{See \url{http://www.stanford.edu/group/mmds/slides2012/s-pli.pdf} for the experimental results on clustering/classification/regression using very sparse random projections~\cite{Proc:Li_Hastie_Church_KDD06}.}

Figure~\ref{fig_one_permutation} presents the ``fixed-length'' scheme, while in Sec.~\ref{sec_variable} we will also develop a ``variable-length'' scheme. Two schemes are more or less equivalent, although we believe the fixed-length scheme is more convenient to implement (and it is slightly more accurate). The variable-length hashing scheme is to some extent related to the Count-Min (CM) sketch~\cite{Article:Cormode_05} and the Vowpal Wabbit (VW)~\cite{Article:Shi_JMLR09,Proc:Weinberger_ICML2009} hashing algorithms.

\section{Applications of Minwise Hashing on Efficient Search and Learning} \label{sec_bbit}

In this section, we will briefly review two important applications of the original ($k$-permutation) minwise hashing: (i) sublinear time near neighbor search~\cite{Proc:Shrivastava_ECML12}, and (ii) large-scale linear learning~\cite{Proc:HashLearning_NIPS11}.

\subsection{Sublinear Time Near Neighbor Search}\label{sec_sublinear}

The  task of {\em near neighbor search} is to identify a set of data points which are ``most similar''  to a query data point. Efficient algorithms for near neighbor search  have  numerous applications in the context of search, databases, machine learning,  recommending systems, computer vision, etc. It has been an active  research topic since the early days of modern computing (e.g,~\cite{Article:Friedman_75}).

In current practice,   methods for approximate near neighbor search often fall into the general framework of {\em Locality Sensitive Hashing (LSH)}~\cite{Proc:Indyk_STOC98,Article:Andoni_CACM08}. The performance of LSH solely depends on its underlying implementation. The idea in~\cite{Proc:Shrivastava_ECML12} is to directly use the bits generated by ($b$-bit) minwise hashing to construct hash tables, which allow us to search near neighbors in sublinear time (i.e., no need to scan all data points).

Specifically, we hash the data points using $k$ random permutations and store each hash value using $b$ bits (e.g., $b\leq 4$). For each data point, we concatenate the resultant $B = b\times k$ bits as a {\em signature}. The size of the space is $2^B = 2^{b\times k}$, which is not too large for small $b$ and $k$ (e.g., $bk=16$). This way, we create a table of   $2^{B}$ buckets, numbered from 0 to $2^{B}-1$; and  each bucket stores the pointers of the data points whose signatures match the bucket number.  In the testing phrase, we apply the same $k$  permutations to a  query data point to generate a $bk$-bit signature  and only search data points in the corresponding bucket. Since using only one hash table will likely miss many true near neighbors, as a remedy, we generate (using independent random permutations) $L$ hash tables. The query result is the union of  the data points retrieved in $L$  tables.

\begin{figure}[h!]
\begin{center}
\mbox{
\includegraphics[width=1.5in]{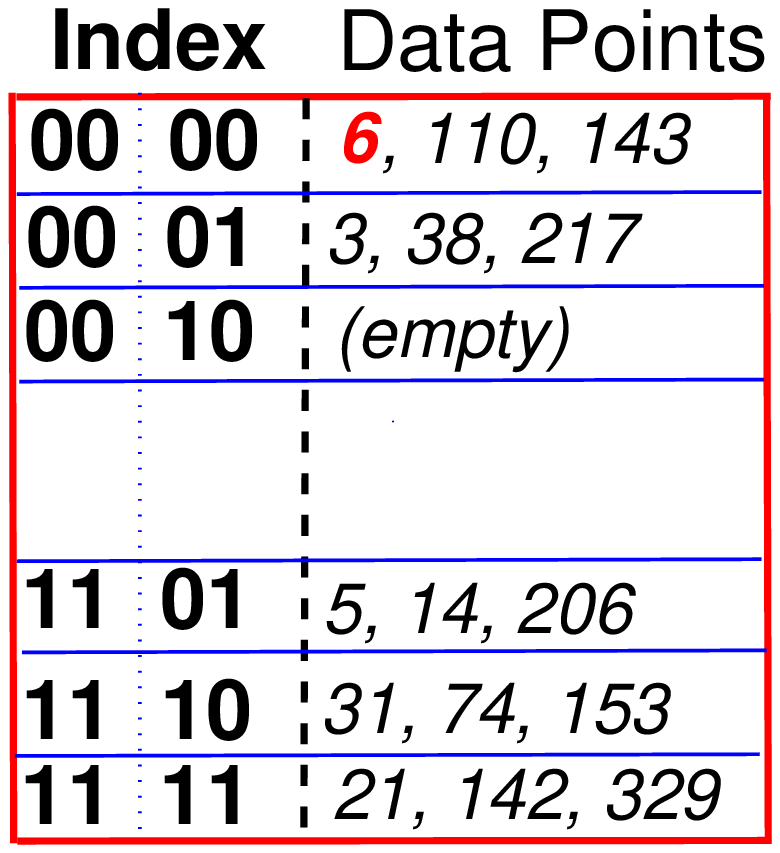}\hspace{0.2in}
\includegraphics[width=1.5in]{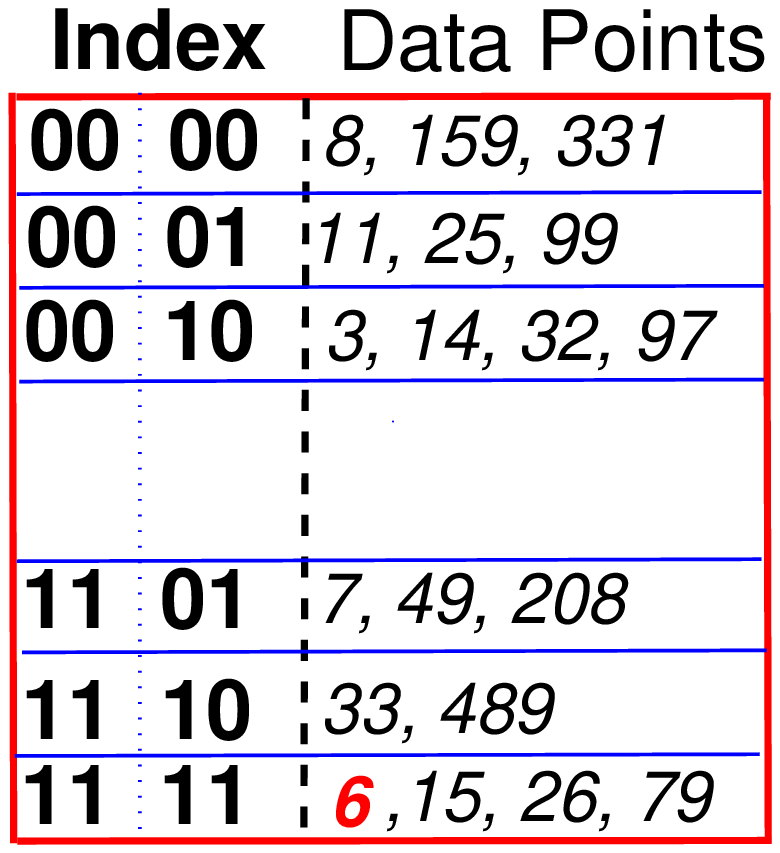}}
\end{center}
\vspace{-0.4in}
\caption{An example of hash tables, with $b=2$, $k=2$, and $L=2$. }\label{fig_hashtable}
\end{figure}

Figure~\ref{fig_hashtable} provides an example with $b=2$ bits,  $k=2$ permutations, and $L=2$ tables. The size of each hash table is $2^{4}$. Given $n$ data points, we apply $k=2$ permutations and store $b=2$ bits of each hashed value to generate $n$ (4-bit) signatures $L$ times. Consider data point 6. For Table 1 (left panel of Figure~\ref{fig_hashtable}), the lowest $b$-bits of its two hashed values are 00 and 00 and thus its signature is 0000 in binary; hence we place a pointer to data point 6 in bucket number 0. For Table 2 (right panel of Figure~\ref{fig_hashtable}), we apply another $k=2$ permutations. This time, the signature of data point 6 becomes 1111 in binary and hence we place it in the last bucket. Suppose in the testing phrase, the two (4-bit) signatures of a new data point are 0000 and 1111, respectively. We then only search for the near neighbors in the set $\{6, 15, 26, 79, 110, 143\}$, which is much smaller than the set of $n$ data points.

The experiments in~\cite{Proc:Shrivastava_ECML12} confirmed that this very simple strategy performed well.

\subsection{Large-Scale Linear Learning}\label{sec_linear_learning}

The recent development of highly efficient linear learning algorithms (such as linear SVM and logistic regression) is a major breakthrough in machine learning. Popular software packages include SVM$^\text{perf}$~\cite{Proc:Joachims_KDD06}, Pegasos~\cite{Proc:Shalev-Shwartz_ICML07}, Bottou's SGD SVM~\cite{URL:Bottou_SGD}, and LIBLINEAR~\cite{Article:Fan_JMLR08}.

Given a dataset $\{(\mathbf{x}_i, y_i)\}_{i=1}^n$, $\mathbf{x}_i\in\mathbb{R}^{D}$, $y_i\in\{-1,1\}$, the $L_2$-regularized logistic regression solves the following optimization  problem:
\begin{align}
\min_{\mathbf{w}}\ \ \frac{1}{2}\mathbf{w^Tw} + C \sum_{i=1}^n \log \left(1 + e^{-y_i\mathbf{w^Tx_i}}\right),
\end{align}
where $C>0$ is the regularization parameter. The $L_2$-regularized linear SVM solves a  similar problem:
\begin{align}
\min_{\mathbf{w}}\ \ \frac{1}{2}\mathbf{w^Tw} + C \sum_{i=1}^n \max \left\{1 - y_i\mathbf{w^Tx_i},\ 0\right\},
\end{align}

In their approach~\cite{Proc:HashLearning_NIPS11}, they  apply $k$  random permutations on each (binary) feature vector $\mathbf{x}_i$ and store the lowest $b$ bits of each hashed value, to obtain a new dataset which can be stored using merely $nbk$ bits. At run-time,  each new data point has to be expanded into a $2^b\times k$-length vector with exactly $k$ 1's.

To illustrate this simple procedure, \cite{Proc:HashLearning_NIPS11} provided a toy example with $k=3$ permutations. Suppose for one data vector, the  hashed values are $\{12013,\ 25964,\ 20191\}$, whose binary digits are respectively\\ $\{010111011101101, \ 110010101101100, 100111011011111\}$. Using  $b=2$ bits, the binary digits are stored as $\{01, 00, 11\}$ (which corresponds to $\{1, 0, 3\}$ in decimals). At run-time, the ($b$-bit) hashed data are expanded into a vector of length $2^bk = 12$, to be $\{ 0, 0, 1, 0, \ \ 0, 0, 0, 1, \ \ 1, 0, 0, 0\}$, which will be the new feature vector fed to a solver such as LIBLINEAR. The procedure for this feature vector is summarized as follows:
\begin{align}\notag
\begin{array}{lrrr}
\text{Original hashed values } (k=3): &12013 &25964 &20191\\
\text{Original binary representations}: &010111011101101& 110010101101100& 100111011011111\\
\text{Lowest $b=2$ binary digits}: &01& 00& 11\\
\text{Expanded $2^b=4$ binary digits }: &0 0 1 0 & 0 0 0 1 & 1 0 0 0\\
\text{New feature vector fed to a solver}: &&[0, 0, 1, 0, 0, 0, 0, 1, 1, 0, 0, 0]\times\frac{1}{\sqrt{k}}
\end{array}
\end{align}
The same procedure (with the same $k=3$ permutations) is then applied to all $n$ feature vectors. Very interestingly, we notice that the all-zero vector (0000 in this example) is never used when expanding the data. In our one permutation hashing scheme, we will actually take advantage of the all-zero vector to conveniently encode empty bins, a strategy which we will later refer to as the ``\textbf{zero coding}'' strategy.

The experiments in~\cite{Proc:HashLearning_NIPS11} confirmed that this simple procedure performed well. \\

Clearly, in both applications (near neighbor search and linear learning), the hashed data have to be ``aligned'' in that only the hashed data generated from the same permutation are compared with each other. With our one permutation scheme as presented in Figure~\ref{fig_one_permutation}, the hashed data are indeed aligned according to the bin numbers. The only caveat is that we need a practical strategy to deal with empty bins, although they occur rarely unless the number of nonzeros in one data vector is small compared to $k$, the number of bins.

\section{Theoretical Analysis of the Fixed-Length One  Permutation Scheme}~\label{sec_fixed}

\vspace{-0.2in}

While the one permutation hashing scheme, as demonstrated in Figure~\ref{fig_one_permutation}, is intuitive, we present in this section some interesting probability analysis to provide a rigorous theoretical foundation for this method.   Without loss of generality, we consider two sets $S_1$ and $S_2$. We first introduce two definitions, for the number of ``jointly empty bins'' and the number of ``matched bins,'' respectively:
\begin{align}
&N_{emp} = \sum_{j=1}^k I_{emp,j}, \hspace{0.6in} N_{mat} = \sum_{j=1}^k I_{mat,j}
\end{align}
where $I_{emp,j}$ and $I_{mat,j}$ are defined for the $j$-th bin, as
\begin{align}
&I_{emp,j} = \left\{
\begin{array}{ll}
1 &\text{if both } \pi(S_1) \text{ and } \pi(S_2) \text{ are  empty in the } j\text{-th bin}\\
0 &\text{otherwise}
\end{array}
\right.\\\notag\\
&I_{mat,j} = \left\{
\begin{array}{ll}
1 &\text{if both }  \pi(S_1) \text{ and } \pi(S_1) \text{ are not empty and the smallest element of }\pi(S_1) \\
& \text{ matches the smallest element of } \pi(S_2),\hspace{0.05in} \text{in the } j\text{-th bin}\\
0 &\text{otherwise}
\end{array}
\right.
\end{align}

Later we will also use $I_{emp,j}^{(1)}$ (or $I_{emp,j}^{(2)}$) to indicate whether $\pi(S_1)$ (or $\pi(S_2)$) is empty in the $j$-th bin.

\subsection{Expectation, Variance, and Distribution of the Number of Jointly Empty Bins}

Recall the notation: $f_1 = |S_1|, \ f_2 = |S_2|, \ a = |S_1\cap S_2|$. We also use $f =|S_1\cup S_2| = f_1 + f_2 -a$.
\begin{lemma}\label{lem_Nemp}
 Assume $D\left(1-\frac{1}{k}\right)\geq f = f_1+f_2-a$,
\begin{align}\label{eqn_Nemp_mean}
\frac{E\left(N_{emp}\right)}{k} =& \prod_{j=0}^{f-1} \frac{D\left(1-\frac{1}{k}\right)-j}{D-j} \leq \left(1-\frac{1}{k}\right)^f
\end{align}
Assume $D\left(1-\frac{2}{k}\right)\geq f = f_1+f_2-a$,
\begin{align}\label{eqn_Nemp_var}
\frac{Var\left(N_{emp}\right)}{k^2}
=&\frac{1}{k}\left(\frac{E(N_{emp})}{k}\right)\left(1-\frac{E(N_{emp})}{k}\right)\\\notag
&-\left(1-\frac{1}{k}\right)\left( \left(\prod_{j=0}^{f-1} \frac{D\left(1-\frac{1}{k}\right)-j}{D-j}\right)^2 - \prod_{j=0}^{f-1} \frac{D\left(1-\frac{2}{k}\right)-j}{D-j}\right)\\\label{eqn_Nemp_var_ineq}
<&\frac{1}{k}\left(\frac{E(N_{emp})}{k}\right)\left(1-\frac{E(N_{emp})}{k}\right)
\end{align}
\textbf{Proof:}\ See Appendix~\ref{app_lem_Nemp}. $\Box$
\end{lemma}

The inequality (\ref{eqn_Nemp_var_ineq}) says that the variance of $\frac{N_{emp}}{k}$ is  smaller than its ``binomial analog.''  \\

In practical scenarios, the data are often sparse, i.e., $f = f_1 + f_2 -a \ll D$. In this case,  Lemma~\ref{lem_Nemp_approx} illustrates that in (\ref{eqn_Nemp_mean}) the  upper bound $\left(1-\frac{1}{k}\right)^f$ is a good approximation to the true value of $\frac{E\left(N_{emp}\right)}{k}$.  Since $\left(1-\frac{1}{k}\right)^f \approx e^{-f/k}$, we know that the chance of  empty bins is small when $f\gg k$. For example, if $f/k=5$ then $\left(1-\frac{1}{k}\right)^f \approx 0.0067$; if $f/k=1$, then $\left(1-\frac{1}{k}\right)^f \approx 0.3679$. For practical applications, we would expect that $f\gg k$ (for most data pairs), otherwise hashing probably would not be too useful anyway. This is why we do not expect empty bins will significantly impact (if at all) the performance in practical settings.

\begin{lemma}\label{lem_Nemp_approx}

Assume $D\left(1-\frac{1}{k}\right)\geq f=f_1+f_2-a$.
\begin{align}
\frac{E\left(N_{emp}\right)}{k}  =&\left(1-\frac{1}{k}\right)^{f}\exp\left(\frac{-D\log \frac{D+1}{D - f+1} + f\left(1-\frac{1}{2(D-f+1)}\right)}{k-1} + ... \right)
\end{align}
Under the reasonable assumption that the data are sparse, i.e., $f_1 + f_2 - a =f\ll D$, we  obtain
\begin{align}\label{eqn_Nemp_mean_approx}
\frac{E\left(N_{emp}\right)}{k}
 =&\left(1-\frac{1}{k}\right)^{f}\left(1-O\left(\frac{f^2}{kD}\right)\right)
\\\label{eqn_Nemp_var_approx}
\frac{Var\left(N_{emp}\right)}{k^2}
=&\frac{1}{k}\left(1-\frac{1}{k}\right)^{f}\left(1-\left(1-\frac{1}{k}\right)^{f}\right)\\\notag
&-\left(1-\frac{1}{k}\right)^{f+1} \left(\left(1-\frac{1}{k}\right)^{f}  - \left(1-\frac{1}{k-1}\right)^{f}\right)+ O\left(\frac{f^2}{kD}\right)
\end{align}
\textbf{Proof:}\ See Appendix~\ref{app_lem_Nemp_approx}. $\Box$
\end{lemma}

In addition to its mean and variance,  we can also write down the distribution of $N_{emp}$.

\begin{lemma}\label{lem_Nemp_Pr}
\begin{align}
\mathbf{Pr}\left(N_{emp}=j\right) = \sum_{s=0}^{k-j}(-1)^s\frac{k!}{j!s!(k-j-s)!}\prod_{t=0}^{f-1}\frac{D\left(1-\frac{j+s}{k}\right)-t}{D-t}
\end{align}
\textbf{Proof:}\ See Appendix~\ref{app_lem_Nemp_Pr}. $\Box$
\end{lemma}

Because $E\left(N_{emp}\right) = \sum_{j=0}^{k-1} j \mathbf{Pr}\left(N_{emp}=j\right)$, this yields  an interesting combinatorial identity:
\begin{align}
k\prod_{j=0}^{f-1} \frac{D\left(1-\frac{1}{k}\right)-j}{D-j}  = \sum_{j=0}^{k-1} j\sum_{s=0}^{k-j}(-1)^s\frac{k!}{j!s!(k-j-s)!}\prod_{t=0}^{f-1}\frac{D\left(1-\frac{j+s}{k}\right)-t}{D-t}
\end{align}

\subsection{Expectation and Variance of the Number of  Matched Bins}

\begin{lemma}\label{lem_Nmat}
Assume $D\left(1-\frac{1}{k}\right)\geq f= f_1+f_2-a$.
\begin{align}\label{eqn_Nmat_mean}
\frac{E\left(N_{mat}\right)}{k} =  R\left(1-\frac{E\left(N_{emp}\right)}{k}\right) = R\left(1-\prod_{j=0}^{f-1}\frac{D\left(1-\frac{1}{k}\right)-j}{D-j}\right)
\end{align}
Assume $D\left(1-\frac{2}{k}\right)\geq f=f_1+f_2-a$.
\begin{align}\label{eqn_Nmat_var}
\frac{Var(N_{mat})}{k^2} =& \frac{1}{k}\left(\frac{E(N_{mat})}{k}\right)\left(1-\frac{E(N_{mat})}{k}\right)\\\notag
&+\left(1-\frac{1}{k}\right)R\frac{a-1}{f-1}\left(1-2\prod_{j=0}^{f-1}\frac{D\left(1-\frac{1}{k}\right)-j}{D-j}
+ \prod_{j=0}^{f-1}\frac{D\left(1-\frac{2}{k}\right)-j}{D-j}\right)\\\notag
&-\left(1- \frac{1}{k}\right)R^2
\left(1-\prod_{j=0}^{f-1}\frac{D\left(1-\frac{1}{k}\right)-j}{D-j}\right)^2\\
<&\frac{1}{k}\left(\frac{E(N_{mat})}{k}\right)\left(1-\frac{E(N_{mat})}{k}\right)
\end{align}
\textbf{Proof:}\ See Appendix~\ref{app_lem_Nmat}. $\Box$
\end{lemma}


\subsection{Covariance of $N_{mat}$ and $N_{emp}$}

Intuitively, $N_{mat}$ and $N_{emp}$ should be negatively correlated, as confirmed by the following Lemma:
\begin{lemma}\label{lem_CovN}
 Assume $D\left(1-\frac{2}{k}\right)\geq f = f_1+f_2-a$.
\begin{align}\notag
\frac{Cov\left(N_{mat},\ N_{emp}\right)}{k^2}
=&R\left(\prod_{j=0}^{f-1}\frac{D\left(1-\frac{1}{k}\right)-j}{D-j}\right)\left(\prod_{j=0}^{f-1}\frac{D\left(1-\frac{1}{k}\right)-j}{D-j}
-\prod_{j=0}^{f-1}\frac{D\left(1-\frac{2}{k}\right)-j}{D\left(1-\frac{1}{k}\right)-j}\right)\\\label{eqn_CovN}
&-\frac{1}{k}R\left(1-\prod_{j=0}^{f-1}\frac{D\left(1-\frac{2}{k}\right)-j}{D\left(1-\frac{1}{k}\right)-j}\right)\left(\prod_{j=0}^{f-1}
\frac{D\left(1-\frac{1}{k}\right)-j}{D-j}\right)
\end{align}
and
\begin{align}
Cov\left(N_{mat},\ N_{emp}\right)\leq 0
\end{align}
\textbf{Proof:}\ See Appendix~\ref{app_lem_CovN}. $\Box$
\end{lemma}

\subsection{An Unbiased Estimator of $R$ and the Variance}

Lemma~\ref{lem_Rmat} shows  the following estimator $\hat{R}_{mat}$ of the resemblance is  unbiased:
\begin{lemma}\label{lem_Rmat}
\begin{align}\label{eqn_Rmat}
&\hat{R}_{mat} = \frac{N_{mat}}{k - N_{emp}},\hspace{0.2in} E\left(\hat{R}_{mat}\right) = R\\\label{eqn_Rmat_Var}
&Var\left(\hat{R}_{mat}\right) = R(1-R)\left(E\left(\frac{1}{k-N_{emp}}\right)\left(1+\frac{1}{f-1}\right) - \frac{1}{f-1}\right)\\\label{eqn_Nemp_ineq}
&E\left(\frac{1}{k-N_{emp}}\right) = \sum_{j=0}^{k-1}\frac{\mathbf{Pr}\left(N_{emp}=j\right)}{k-j}  \geq \frac{1}{k-E(N_{emp})}
\end{align}
\textbf{Proof:}\ See Appendix~\ref{app_lem_Rmat}. The right-hand side of the inequality (\ref{eqn_Nemp_ineq}) is actually a very good approximation (see Figure~\ref{fig_Rmat}). The exact expression for $\mathbf{Pr}\left(N_{emp}=j\right)$ is already derived in Lemma~\ref{lem_Nemp_Pr}. $\Box$
\end{lemma}

The fact that $E\left(\hat{R}_{mat}\right) = R$ may seem surprising as in general  ratio estimators are not unbiased.  Note that $k-N_{emp}>0$ always because we assume the original data vectors are not completely empty (all-zero).

As expected, when {\small$k\ll f = f_1+f_2-a$, $N_{emp}$} is essentially zero and hence {\small$Var\left(\hat{R}_{mat}\right)\approx \frac{R(1-R)}{k}$}. In fact, {\small$Var\left(\hat{R}_{mat}\right)$} is  somewhat smaller  than {\small$\frac{R(1-R)}{k}$}, which can be seen from the  approximation:
\begin{align}
\frac{Var\left(\hat{R}_{mat}\right)}{R(1-R)/k} \approx g(f;k) = \frac{1}{1-\left(1-\frac{1}{k}\right)^{f}}\left(1+\frac{1}{f-1}\right) -\frac{k}{f-1}
\end{align}
\begin{lemma}\label{lem_var_ratio}
\begin{align}\vspace{-0.2in}
g(f;k) \leq 1
\end{align}
\textbf{Proof}:\ See Appendix~\ref{app_lem_var_ratio}. $\Box$
\end{lemma}

It is probably not surprising that our one permutation scheme may (slightly) outperform the original $k$-permutation scheme (at merely $1/k$ of its preprocessing cost), because one permutation hashing can be viewed as a ``sample-without-replacement'' scheme.

\subsection{Experiments for Validating the Theoretical Results }

This set of experiments is for validating the theoretical results.  The Web crawl dataset (in Table \ref{tab_15pairs}) consists of 15 (essentially randomly selected) pairs of word vectors (in $D=2^{16}$ dimensions) of a range of similarities and sparsities. For each word vector, the $j$-th element is whether the word appeared in the $j$-th Web page.

\begin{table}[h]
\caption{15 pairs of English words. For example, ``RIGHTS'' and ``RESERVED'' correspond to the two sets of document IDs which contained  word ``RIGHTS'' and  word ``RESERVED'' respectively.
 }
\begin{center}{
\begin{tabular}{l l l l l l }
\hline \hline
Word 1 & Word 2 &$f_1$  &$f_2$  &$f=f_1+f_2-a$ &$R$  \\\hline
RIGHTS & RESERVED  &12234 &11272 & 12526 &0.877 \\
OF & AND &37339 &36289 &41572 &0.771\\
THIS &HAVE &27695 &17522 &31647 &0.429\\
ALL &MORE &26668 &17909 &31638 &0.409\\
CONTACT &INFORMATION &16836 &16339 &24974 &0.328\\
MAY & ONLY &12067 &11006 &17953 &0.285\\
CREDIT & CARD &2999 &2697 & 4433 &0.285 \\
SEARCH & WEB &1402 &12718 &21770 &0.229\\
RESEARCH & UNIVERSITY &4353 &4241 &7017 &0.225\\
FREE & USE &12406 &11744 &19782 &0.221\\
TOP &BUSINESS &9151 &8284 &14992 &0.163\\
BOOK &TRAVEL &5153 &4608 &8542 &0.143\\
TIME & JOB &12386        &3263 & 13874 &0.128 \\
REVIEW &PAPER &3197 &1944 &4769 &0.078\\
A  & TEST &  39063       &2278        &2060 &0.052\\
\hline\hline
\end{tabular}
}
\end{center}
\label{tab_15pairs}\vspace{-0.in}
\end{table}

We vary $k$ from  $2^3$ to $2^{15}$. Although $k=2^{15}$ is  probably way too large in practice, we  use it for the purpose of thorough validations. Figures~\ref{fig_Nemp_mean} to~\ref{fig_Rmat} present the empirical results based on $10^5$ repetitions.

\subsubsection{$E(N_{emp})$ and $Var(N_{emp})$}

\begin{figure}[h!]
\begin{center}
\mbox{
\includegraphics[width=1.5in]{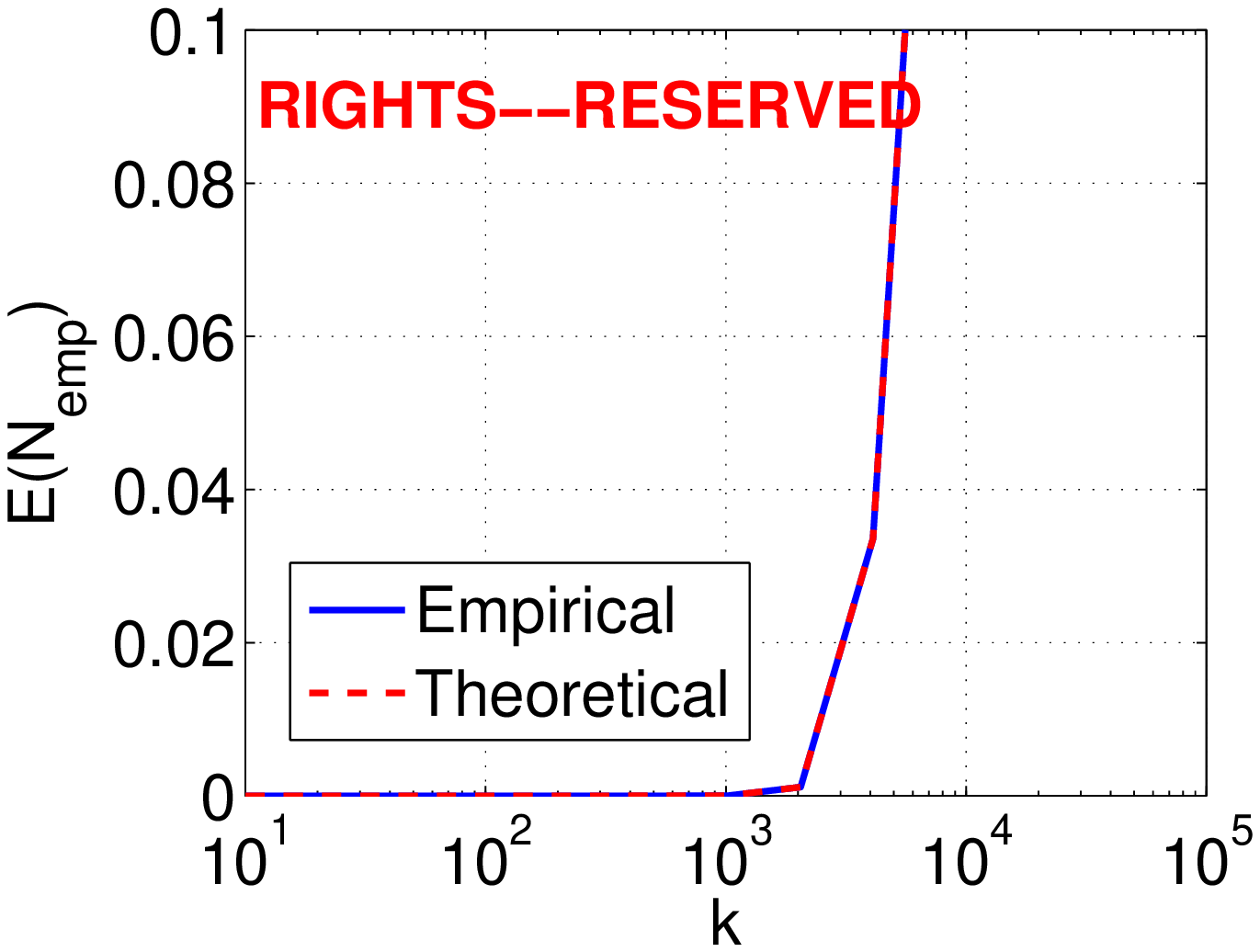}\hspace{-0.1in}
\includegraphics[width=1.5in]{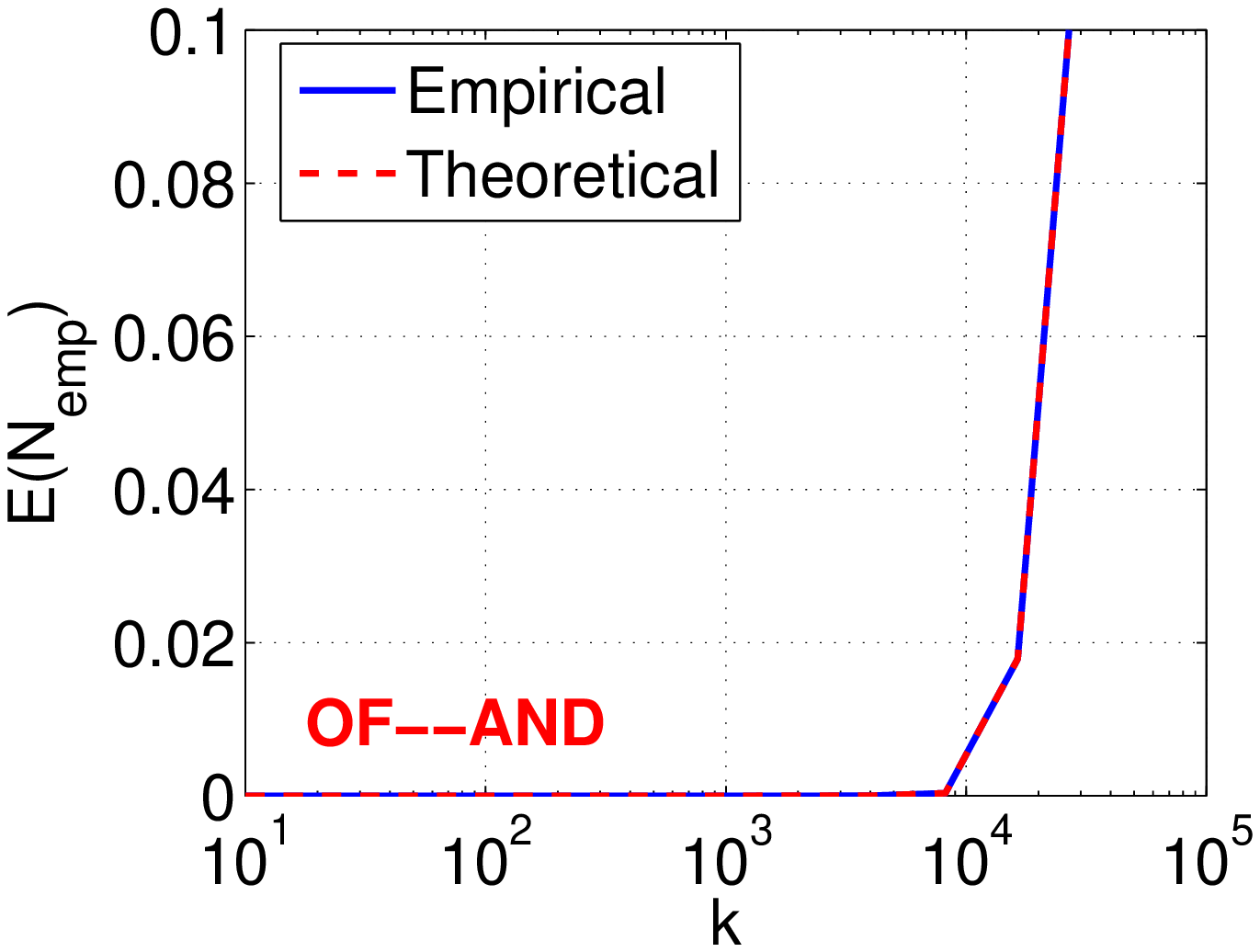}\hspace{-0.1in}
\includegraphics[width=1.5in]{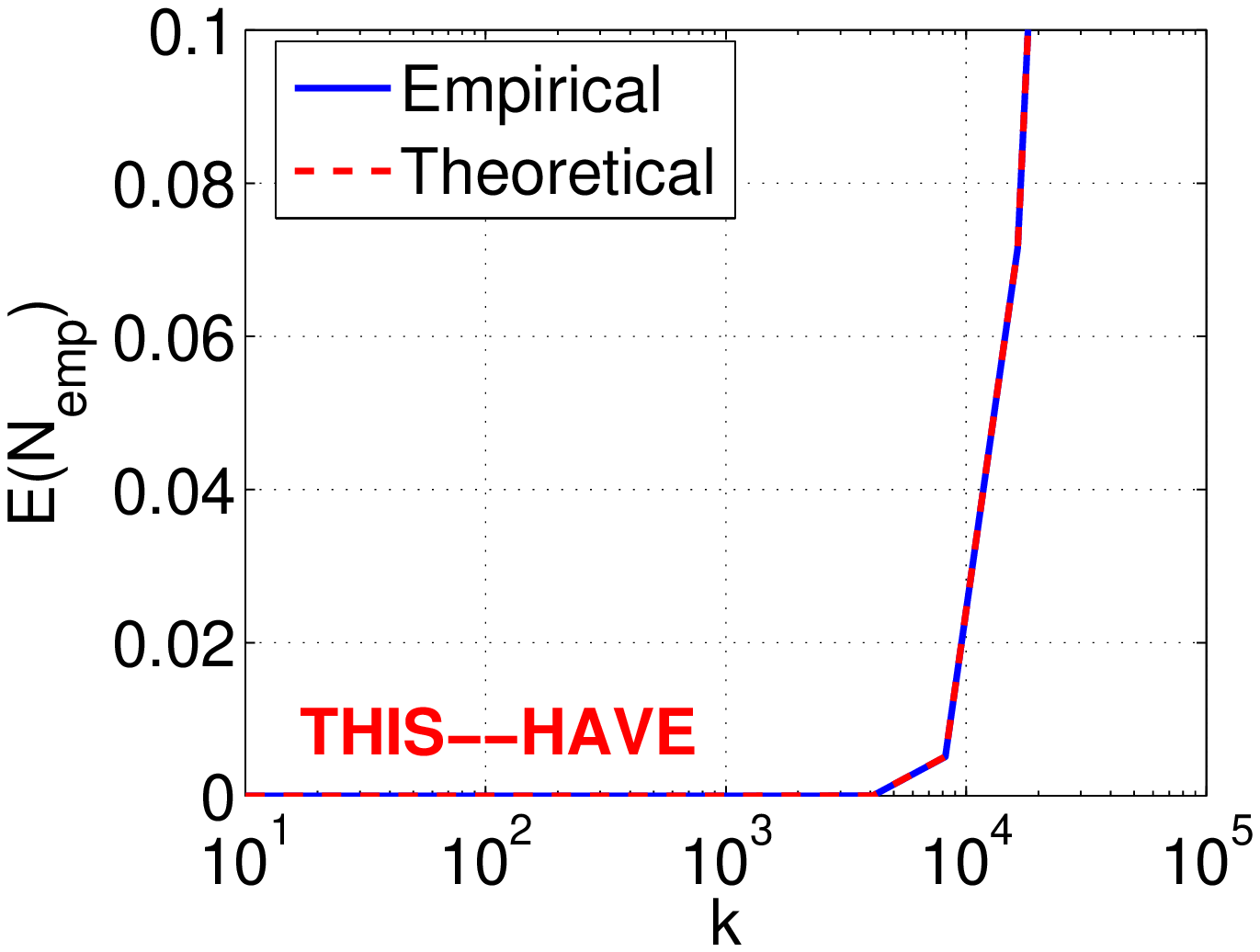}\hspace{-0.1in}
\includegraphics[width=1.5in]{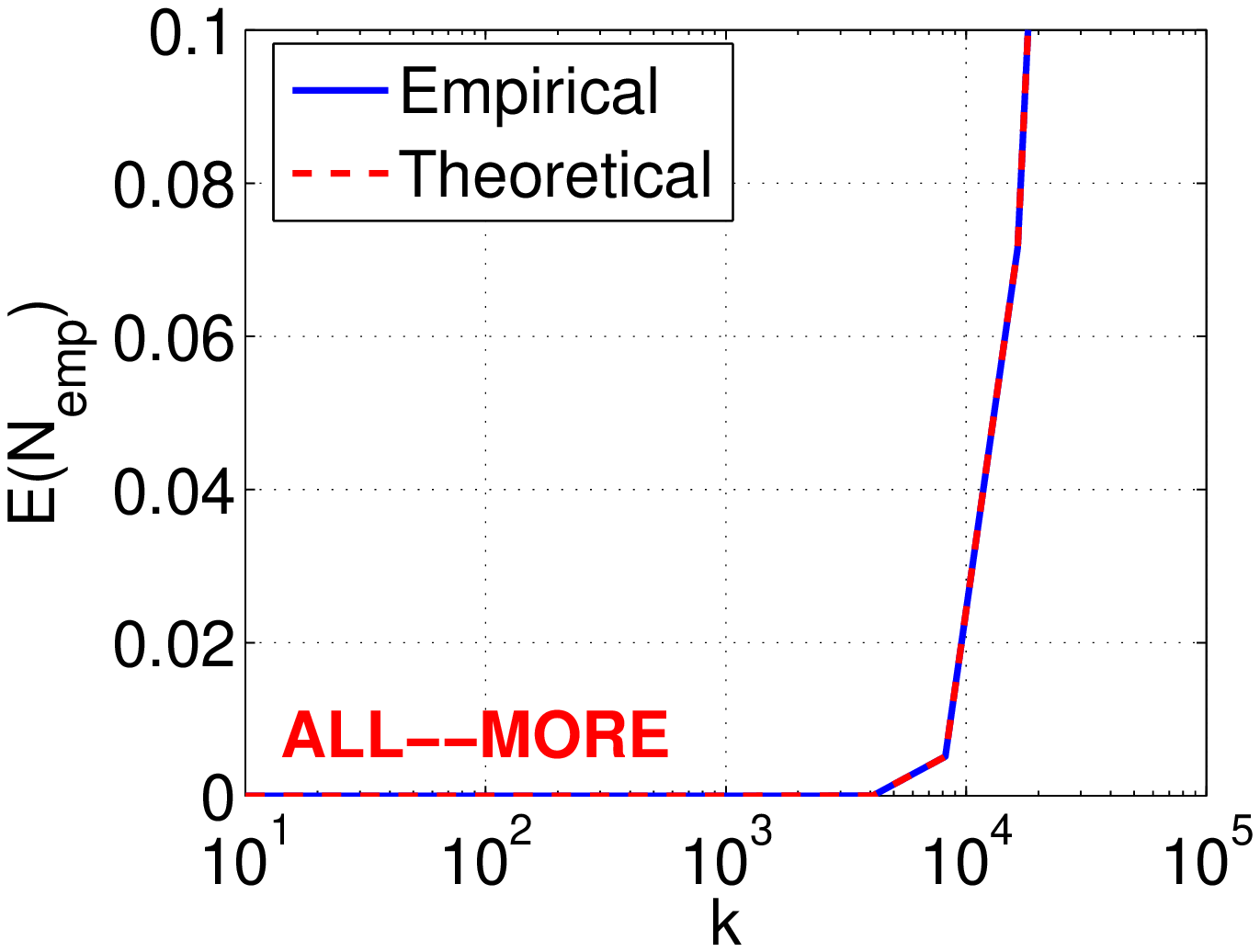}\hspace{-0.1in}
\includegraphics[width=1.5in]{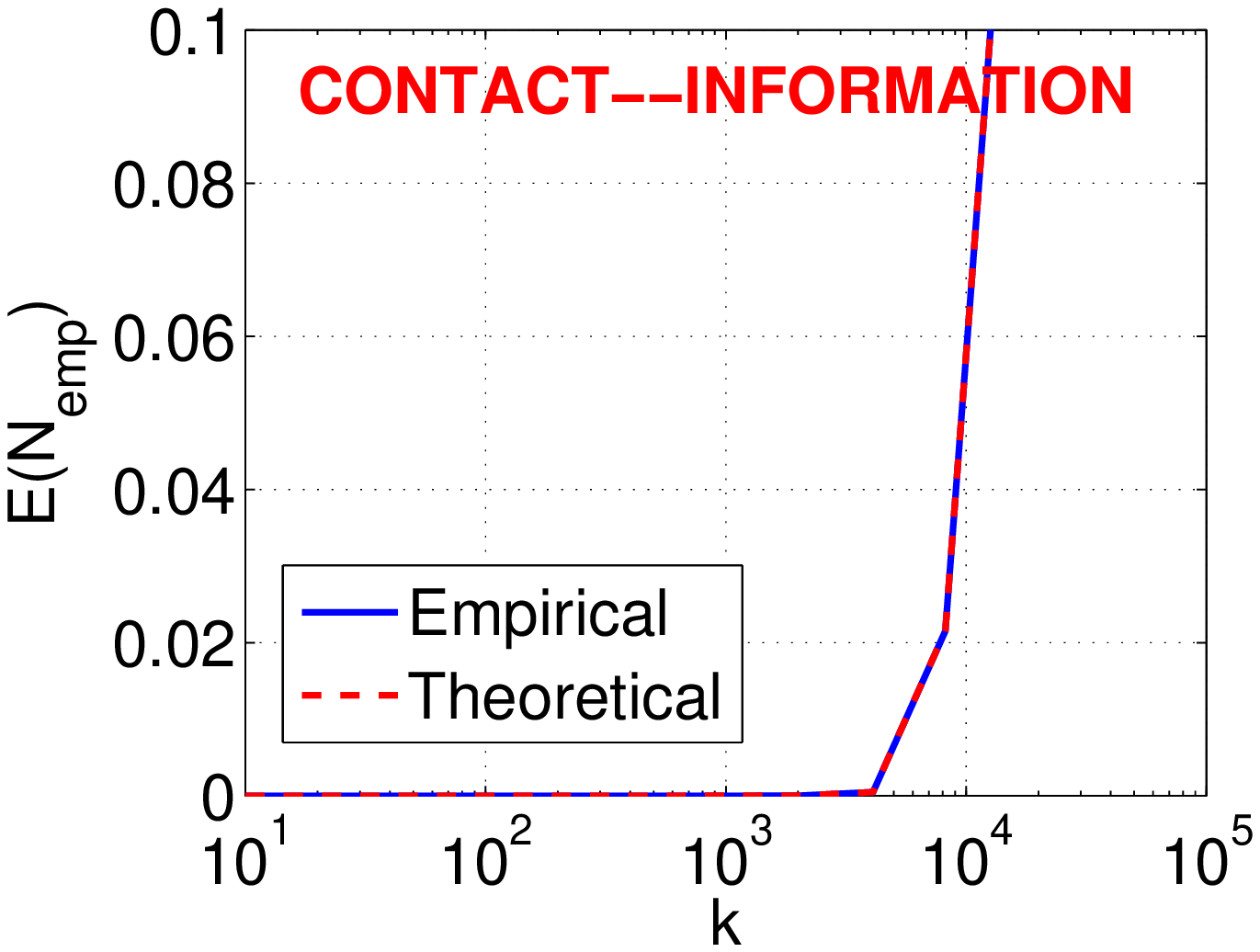}
}
\mbox{
\includegraphics[width=1.5in]{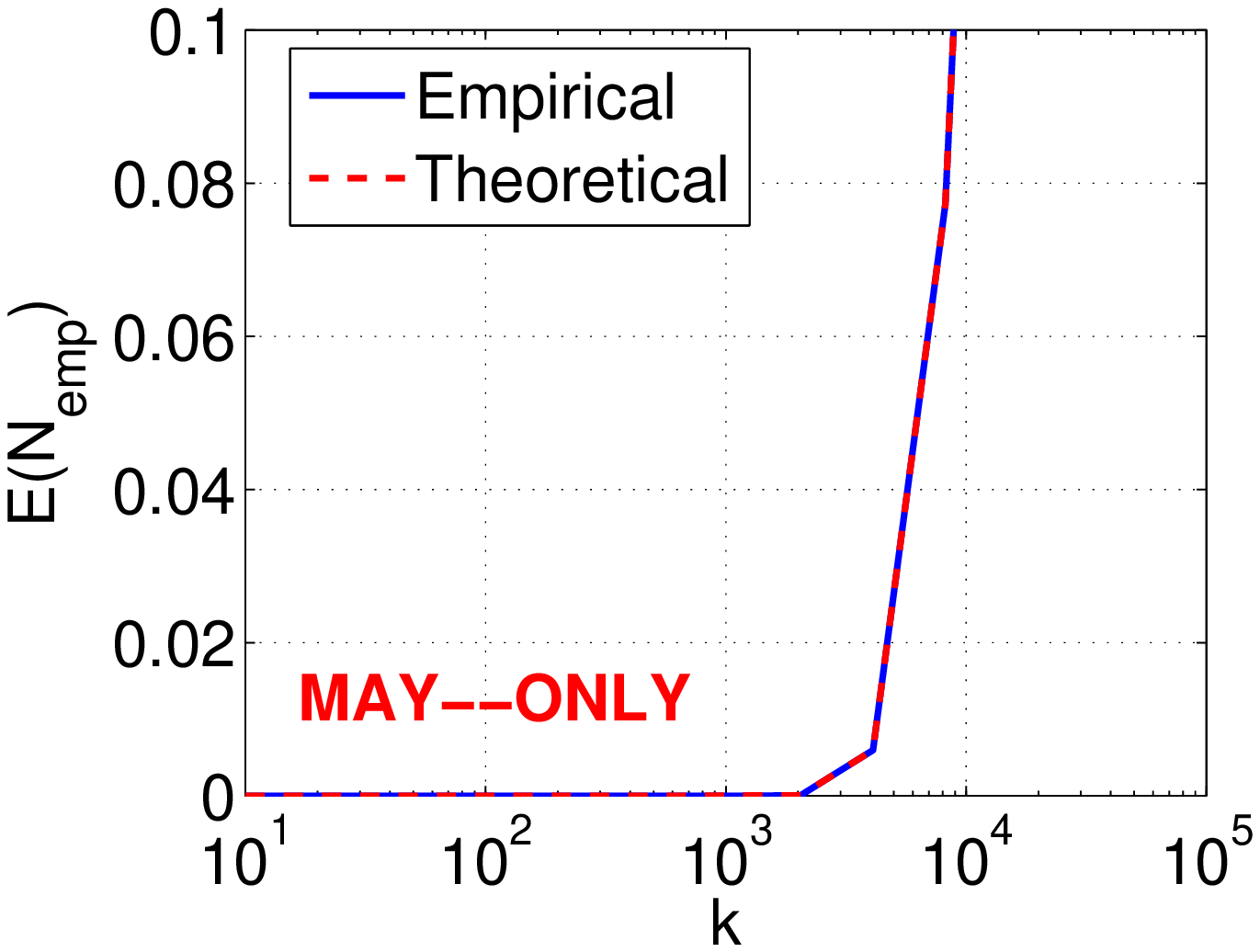}\hspace{-0.1in}
\includegraphics[width=1.5in]{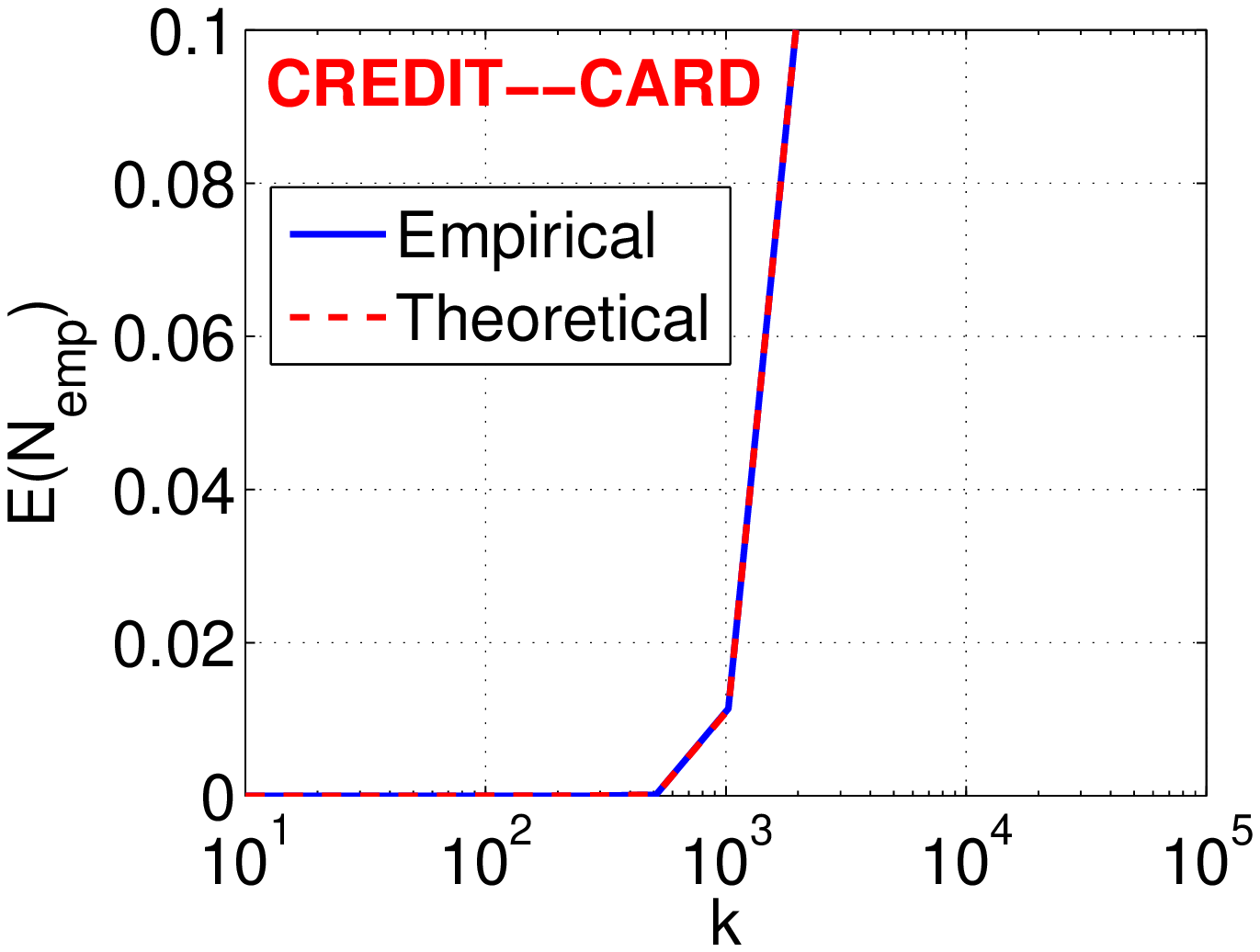}\hspace{-0.1in}
\includegraphics[width=1.5in]{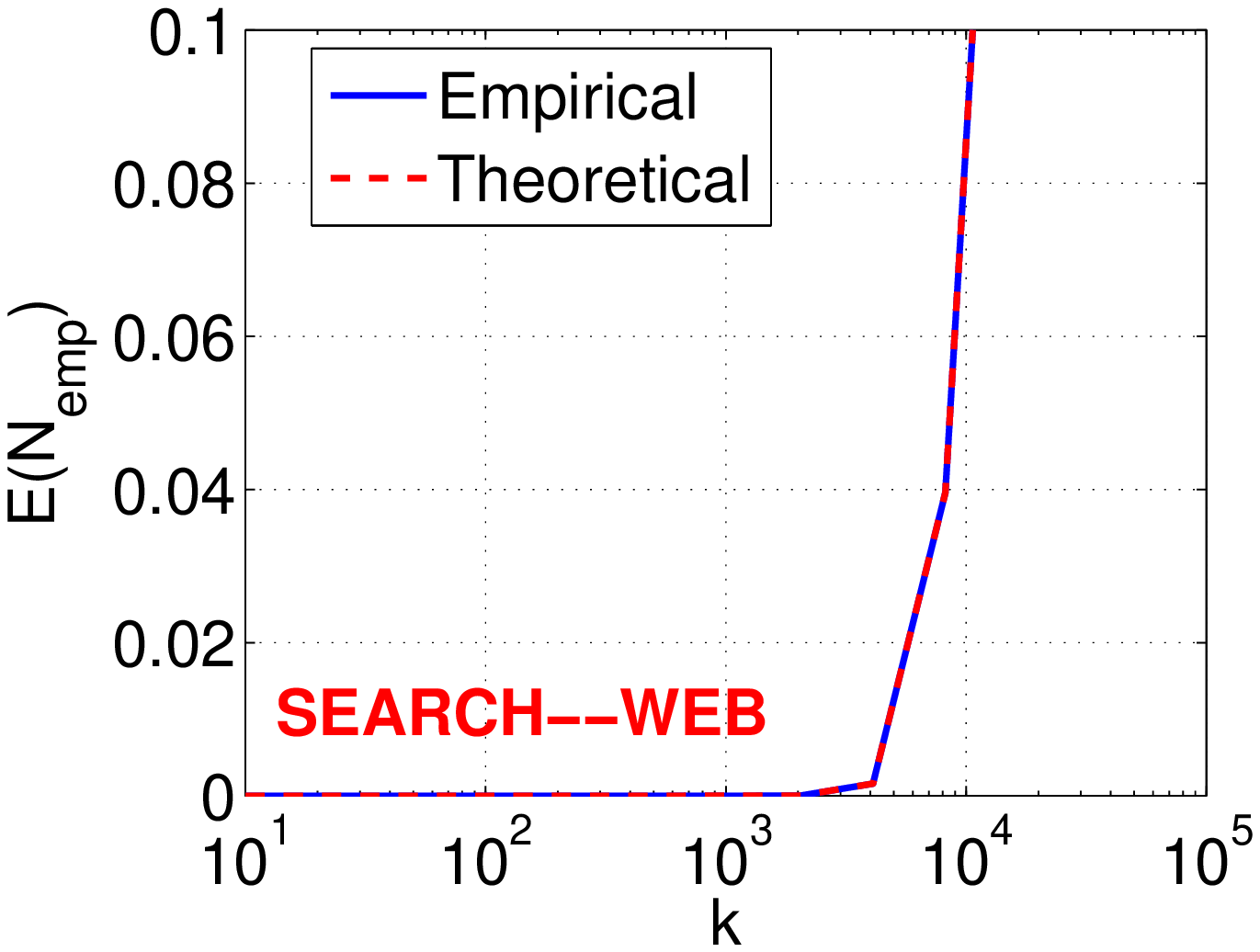}\hspace{-0.1in}
\includegraphics[width=1.5in]{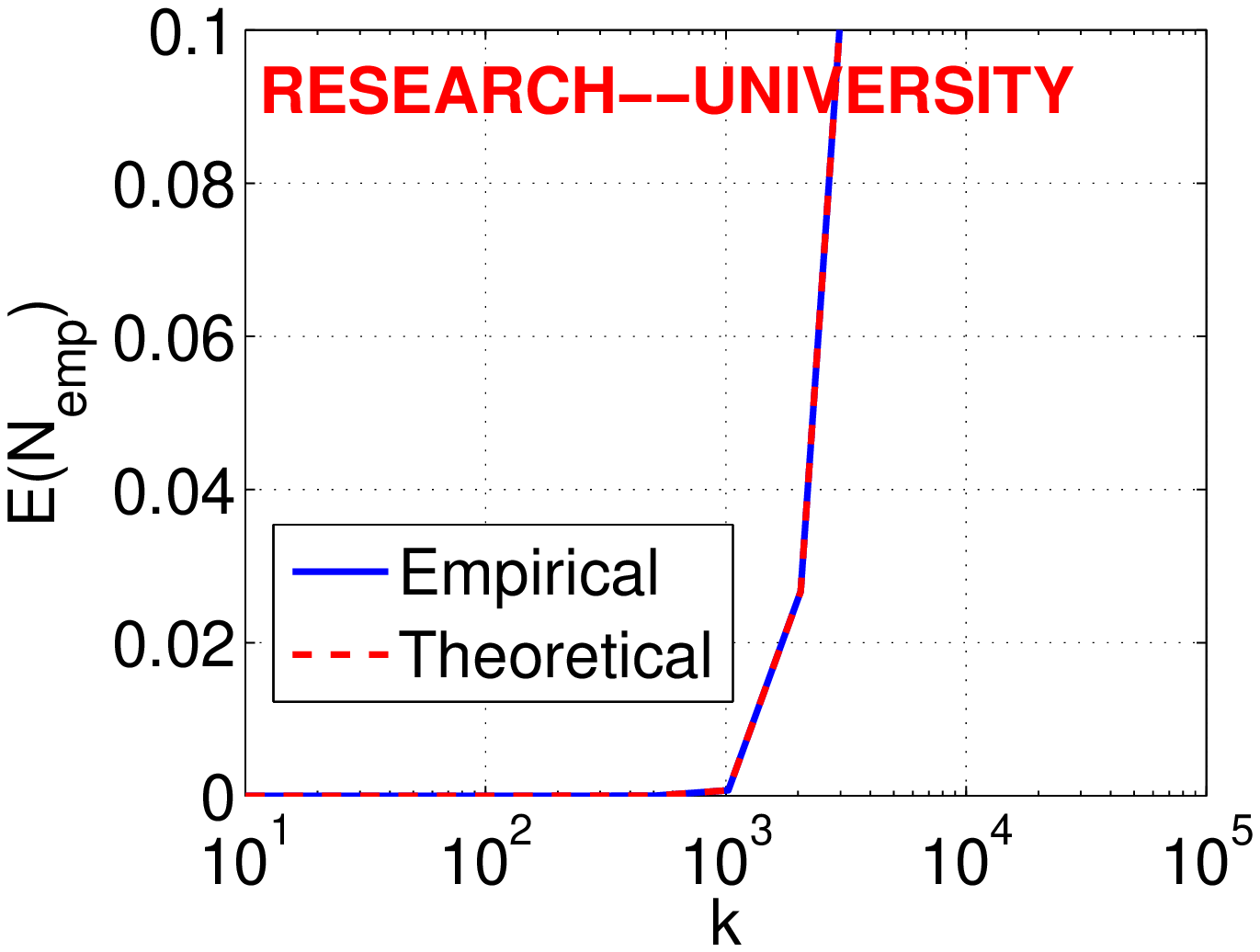}\hspace{-0.1in}
\includegraphics[width=1.5in]{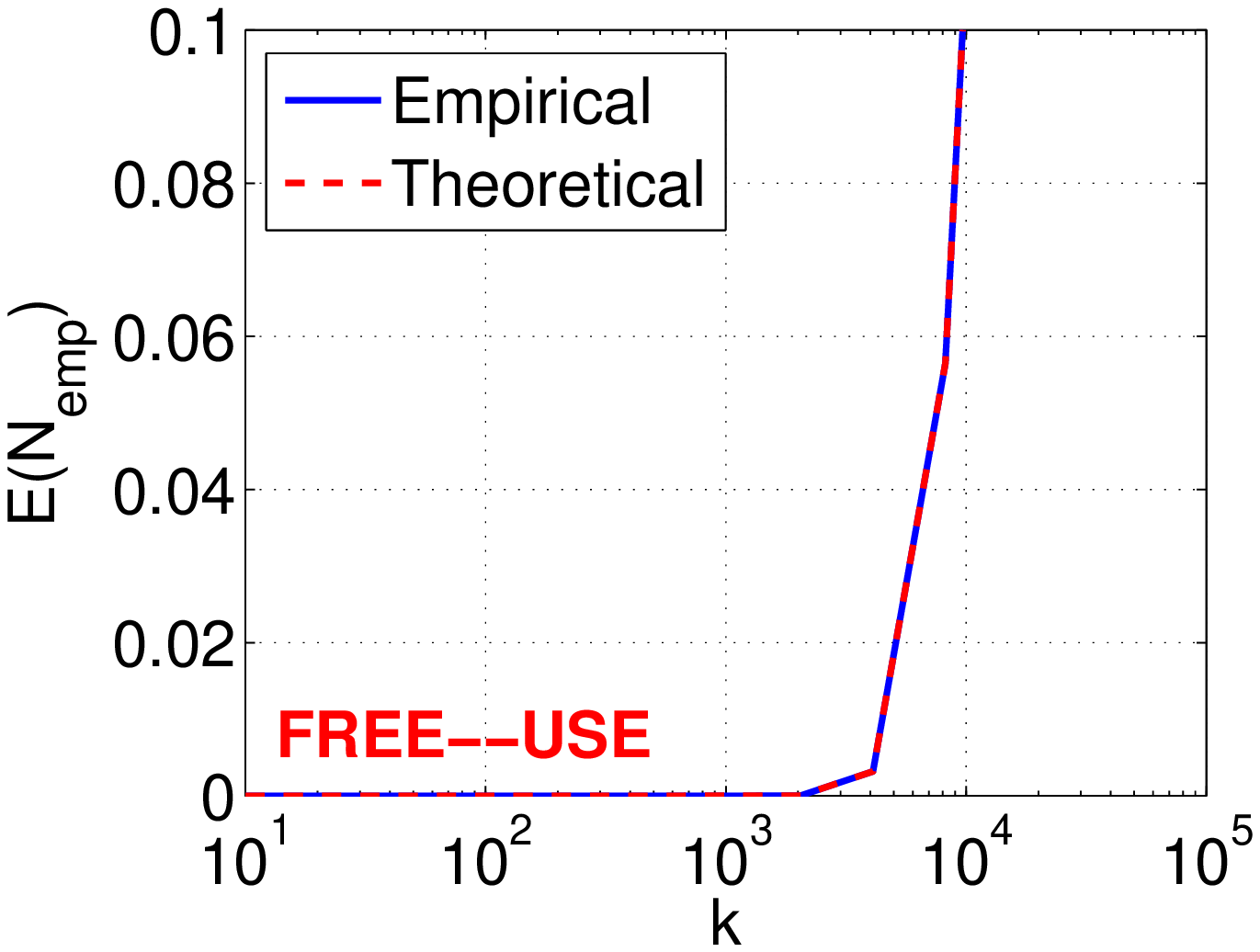}
}
\mbox{
\includegraphics[width=1.5in]{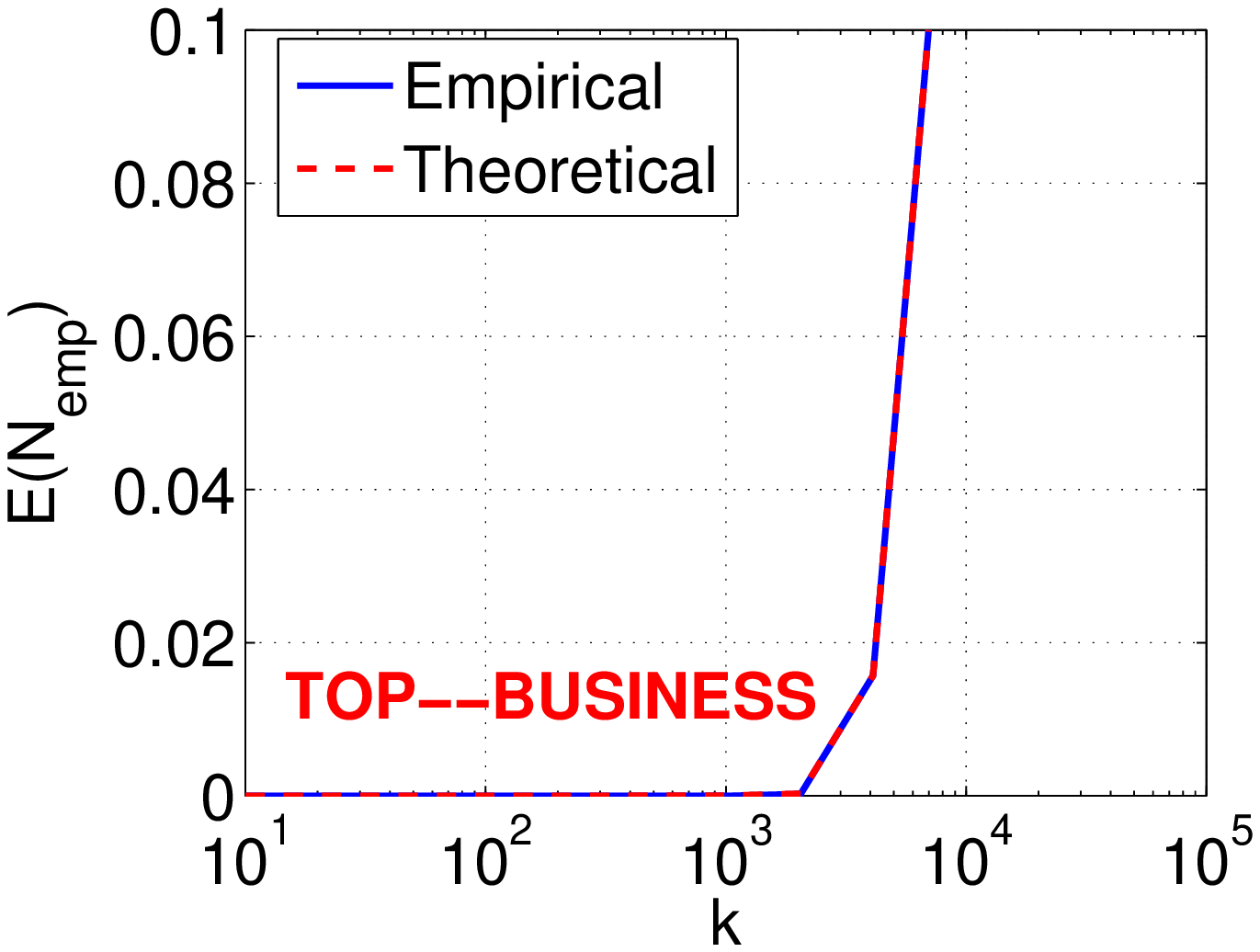}\hspace{-0.1in}
\includegraphics[width=1.5in]{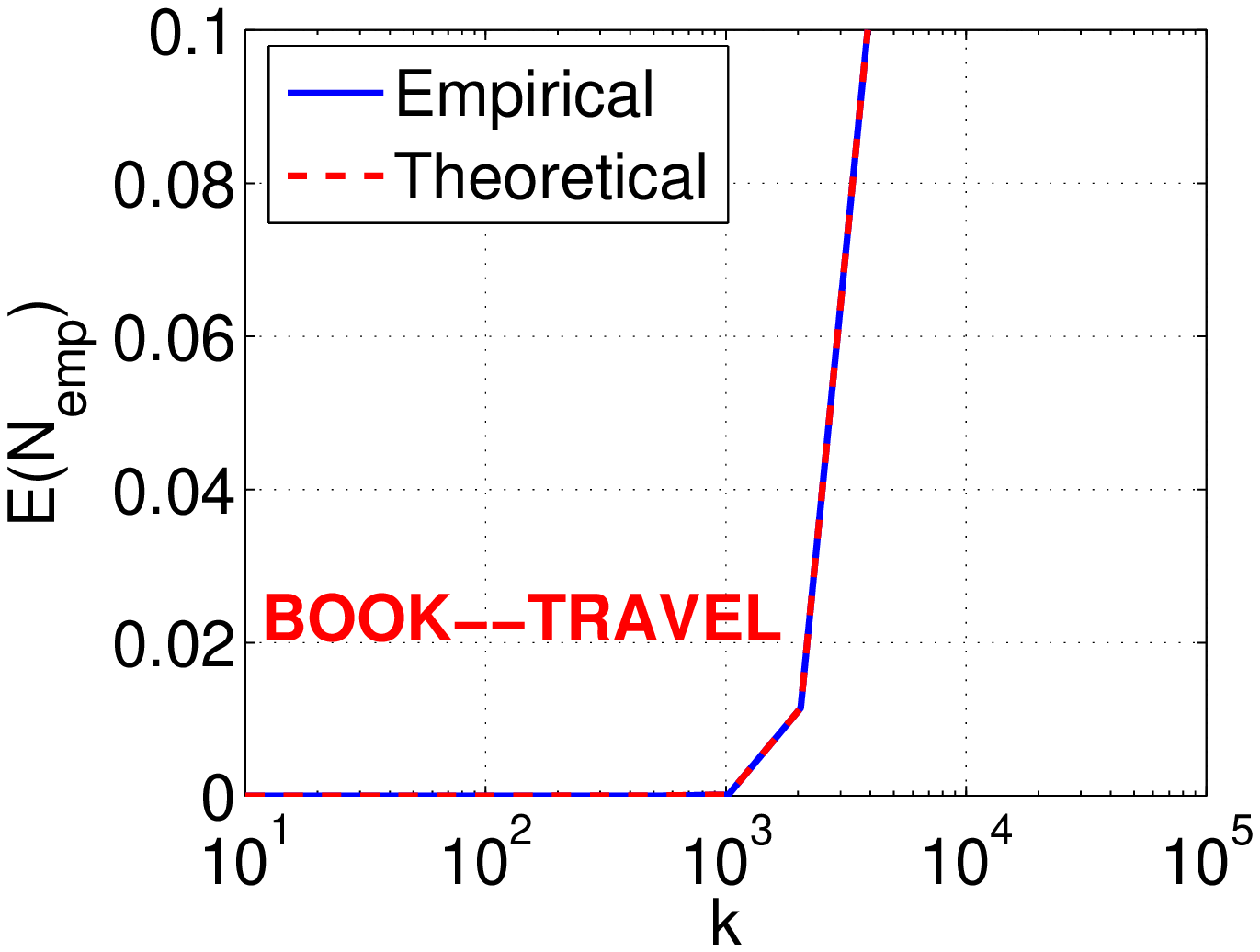}\hspace{-0.1in}
\includegraphics[width=1.5in]{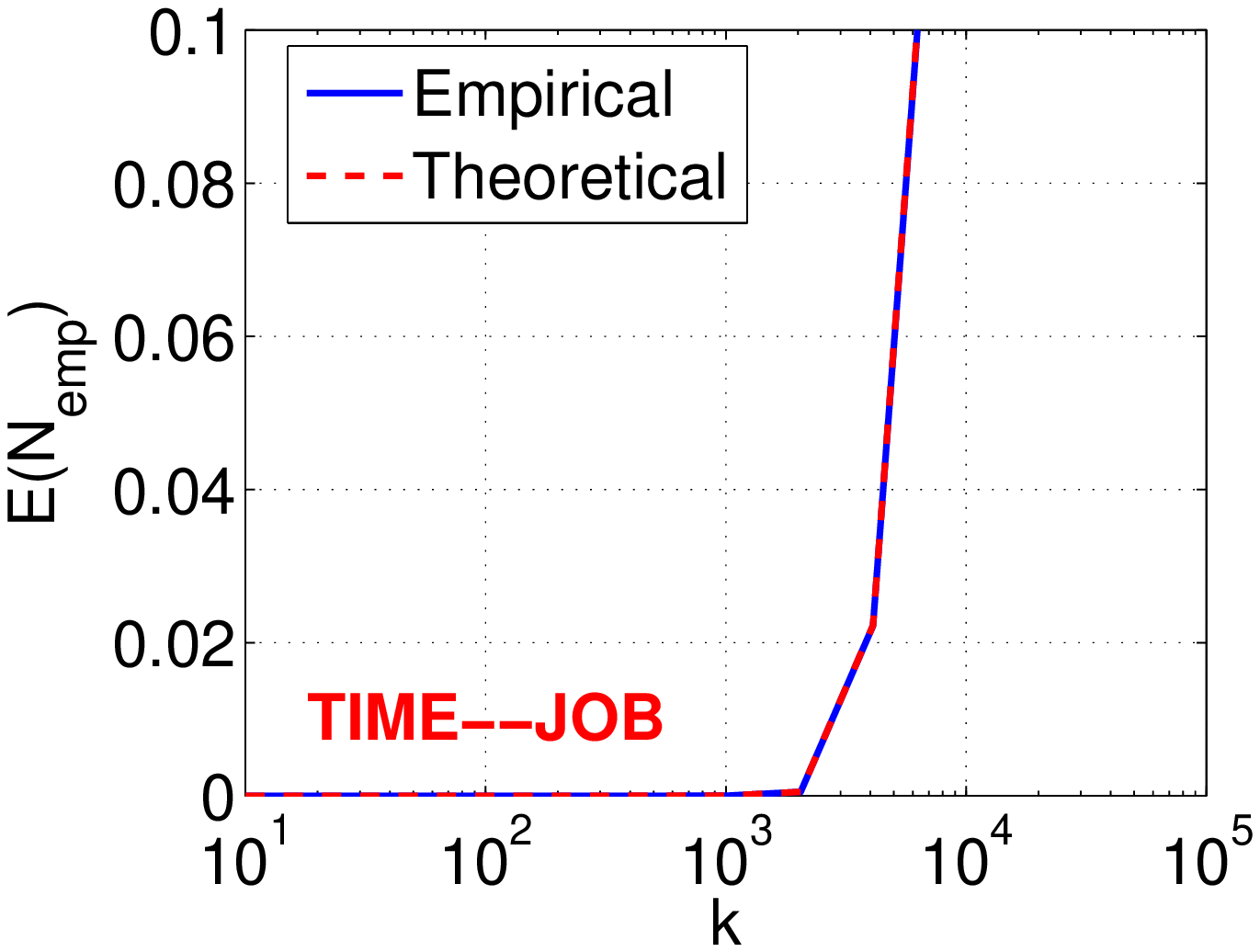}\hspace{-0.1in}
\includegraphics[width=1.5in]{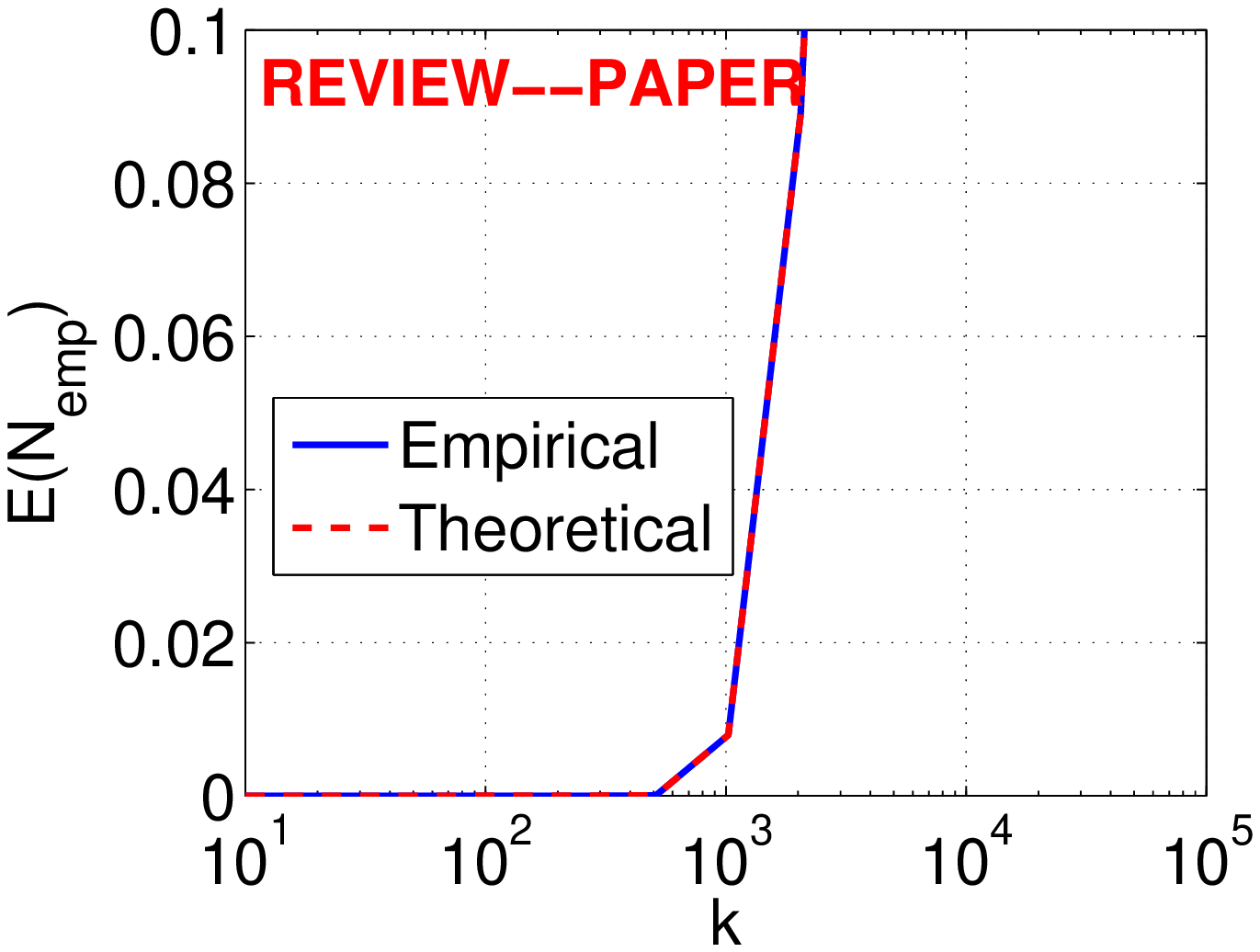}\hspace{-0.1in}
\includegraphics[width=1.5in]{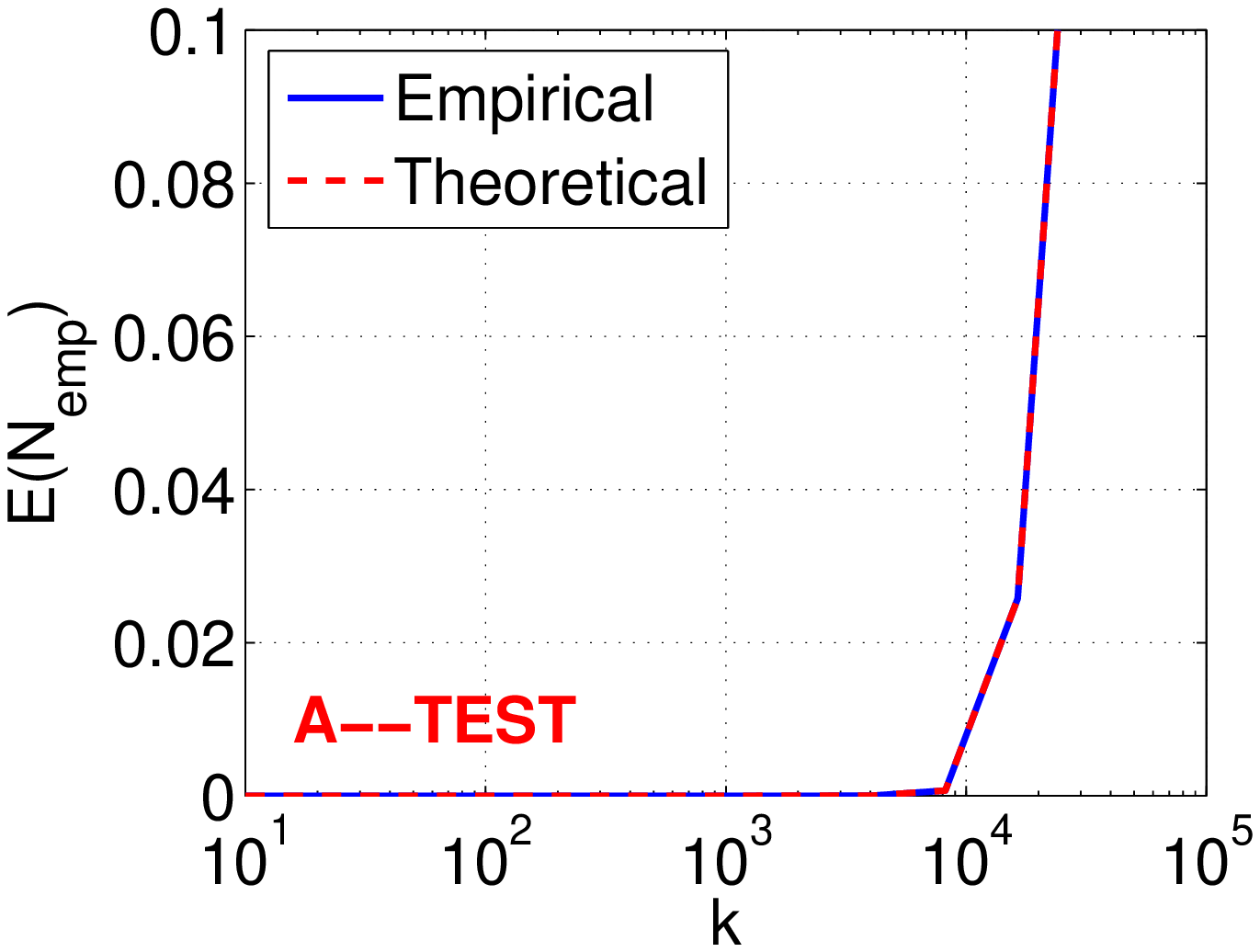}
}
\end{center}
\vspace{-0.3in}
\caption{$E(N_{emp})/k$. The empirical curves essentially overlap the theoretical curves as derived in Lemma~\ref{lem_Nemp}, i.e., (\ref{eqn_Nemp_mean}). The occurrences of empty bins become noticeable only at relatively large sample size $k$. }\label{fig_Nemp_mean}
\end{figure}

\begin{figure}[h!]
\begin{center}

\mbox{
\includegraphics[width=1.5in]{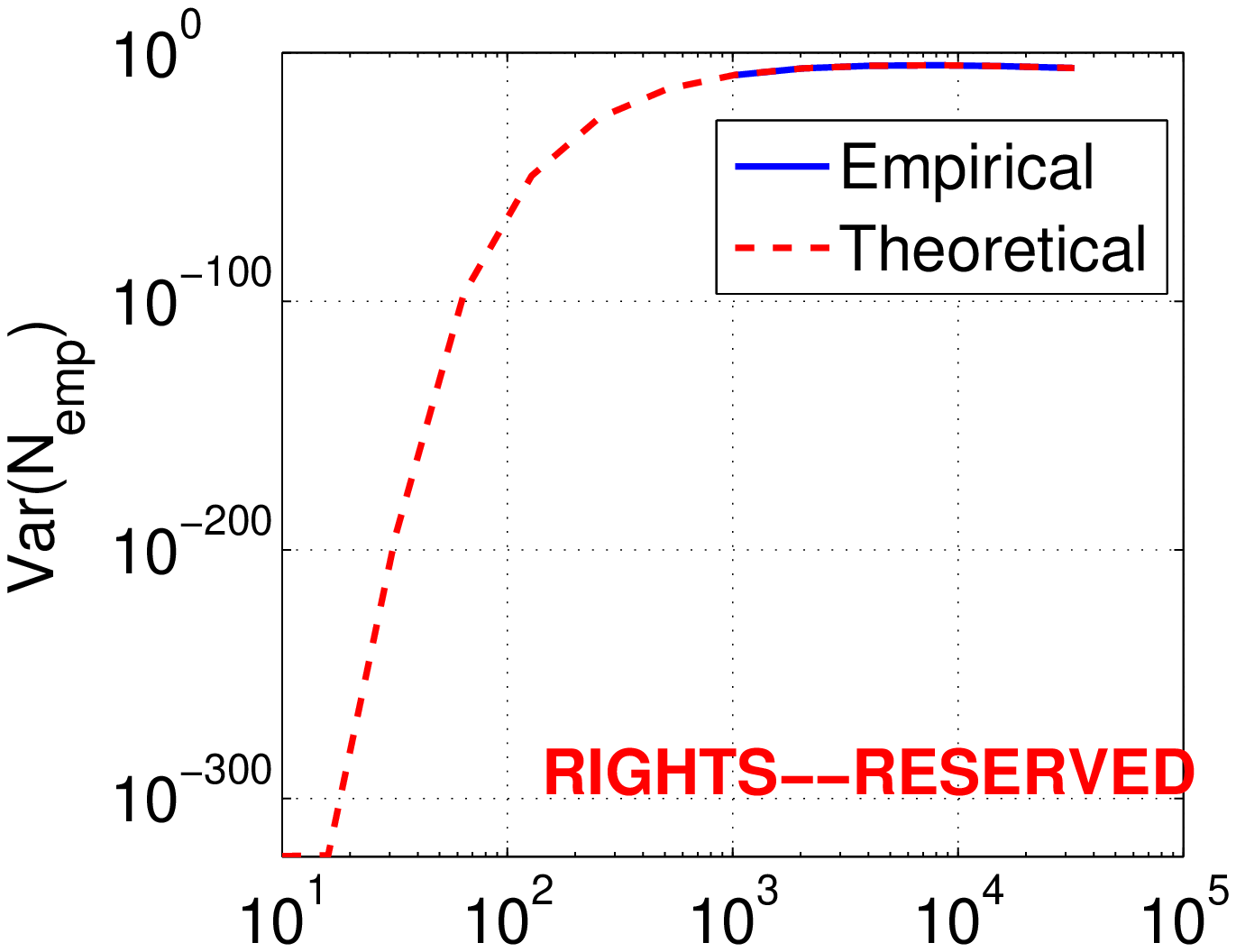}\hspace{-0.1in}
\includegraphics[width=1.5in]{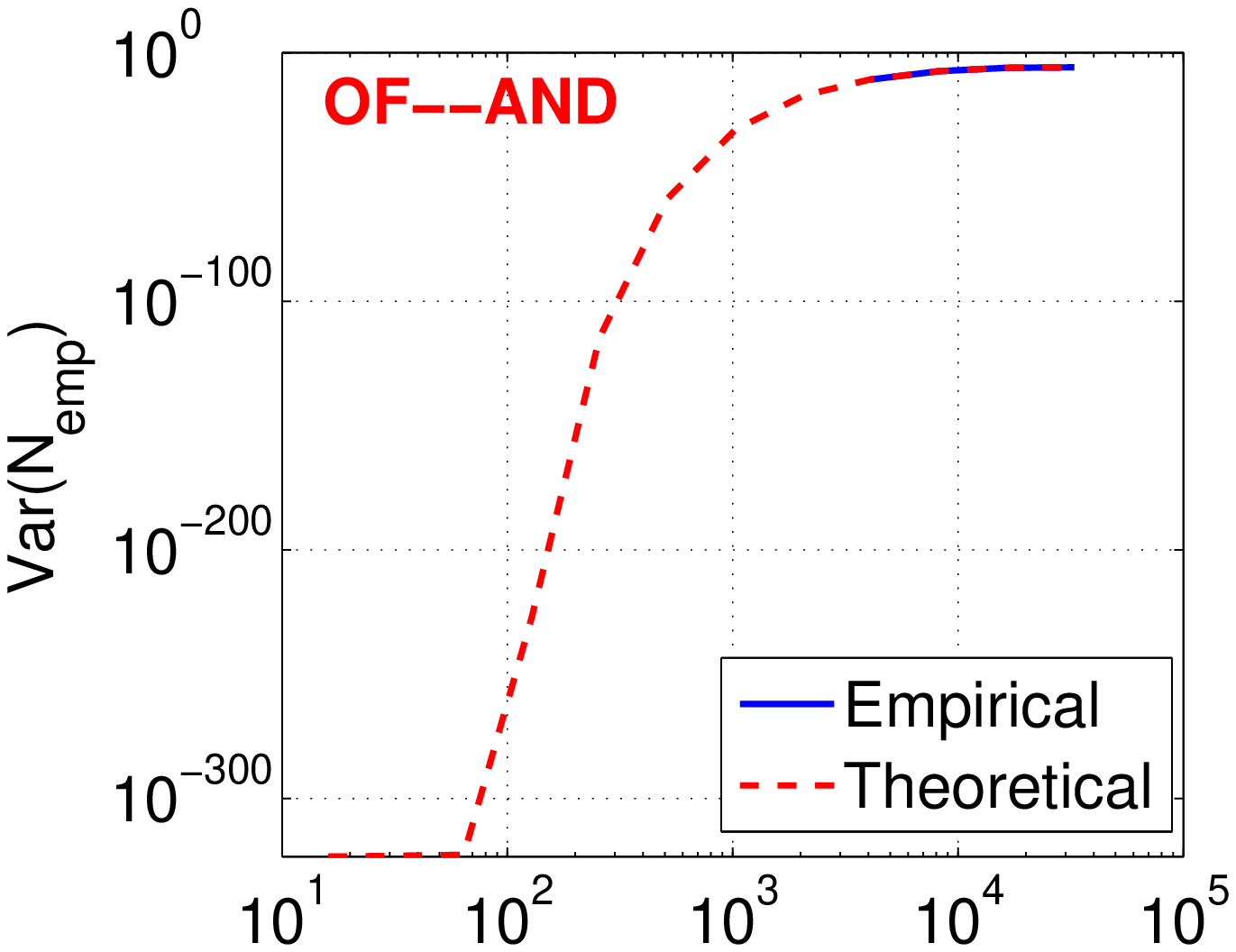}\hspace{-0.1in}
\includegraphics[width=1.5in]{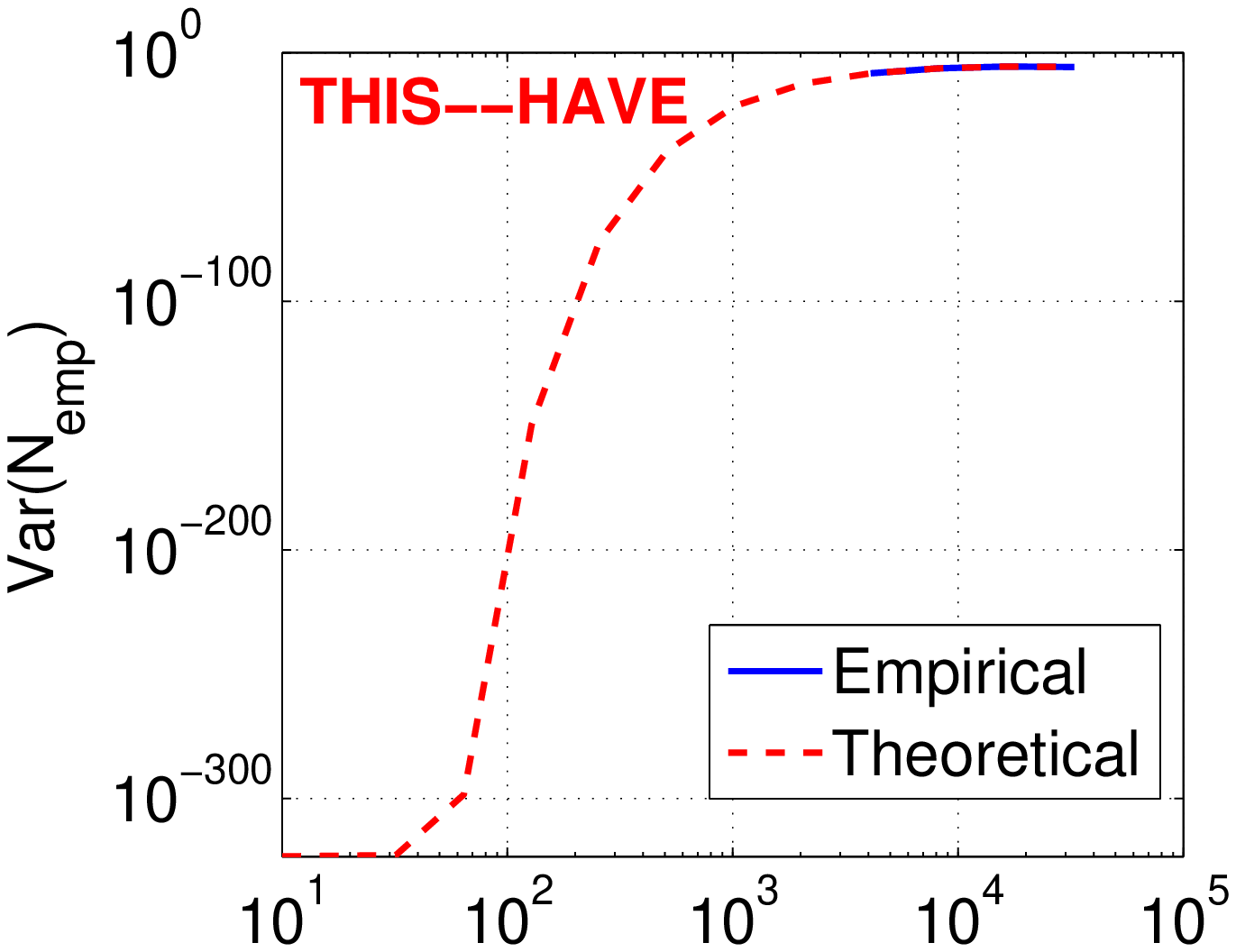}\hspace{-0.1in}
\includegraphics[width=1.5in]{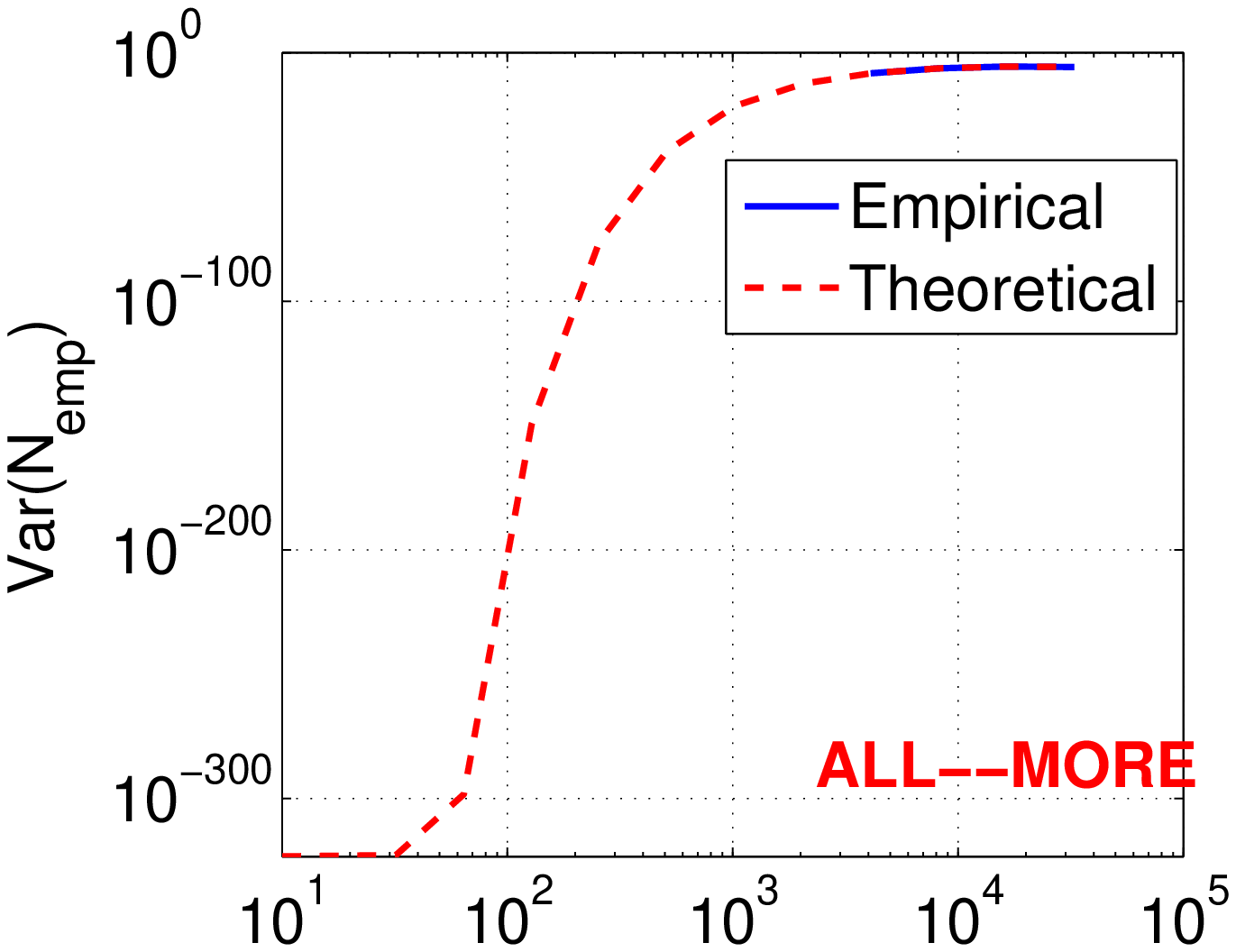}\hspace{-0.1in}
\includegraphics[width=1.5in]{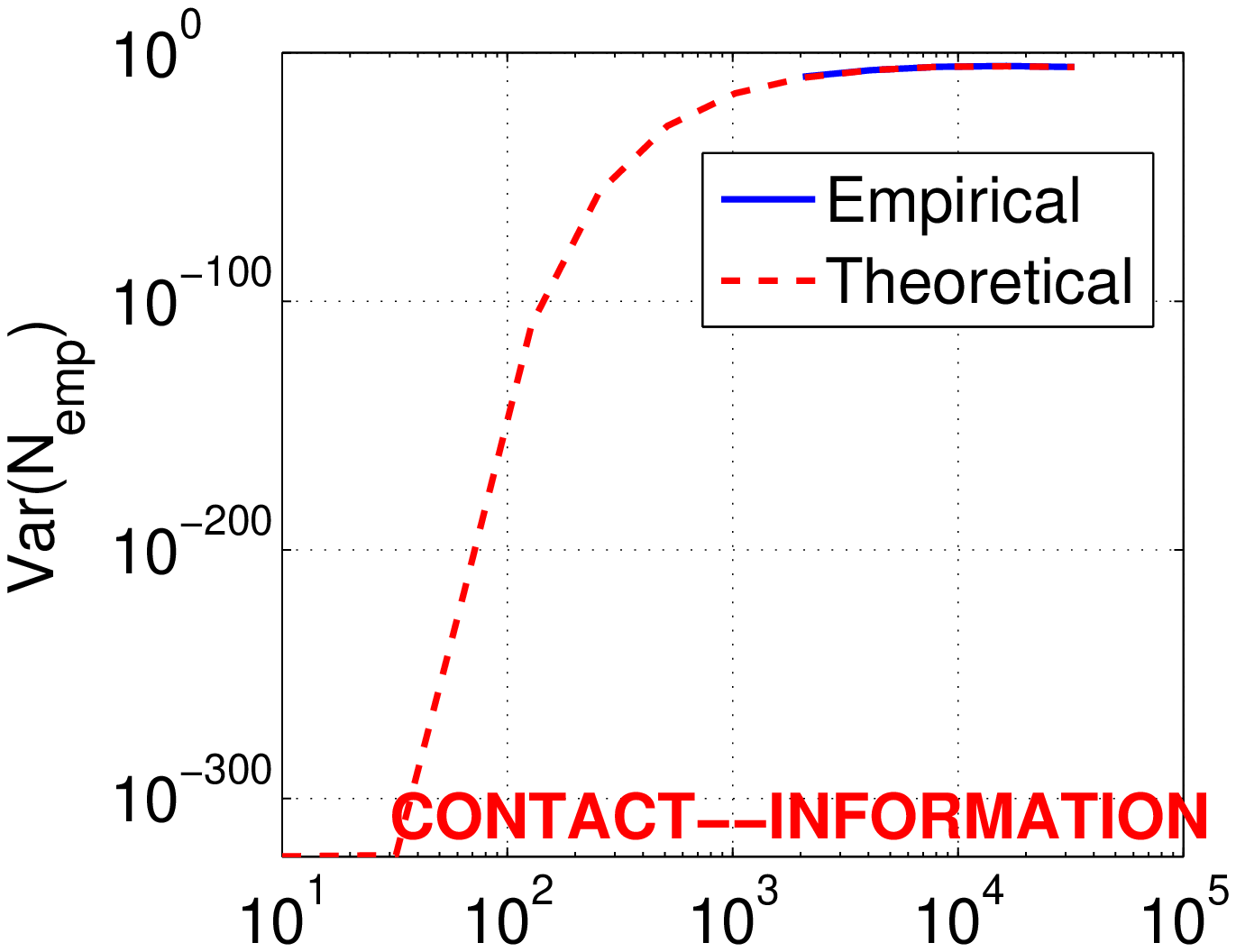}
}
\mbox{
\includegraphics[width=1.5in]{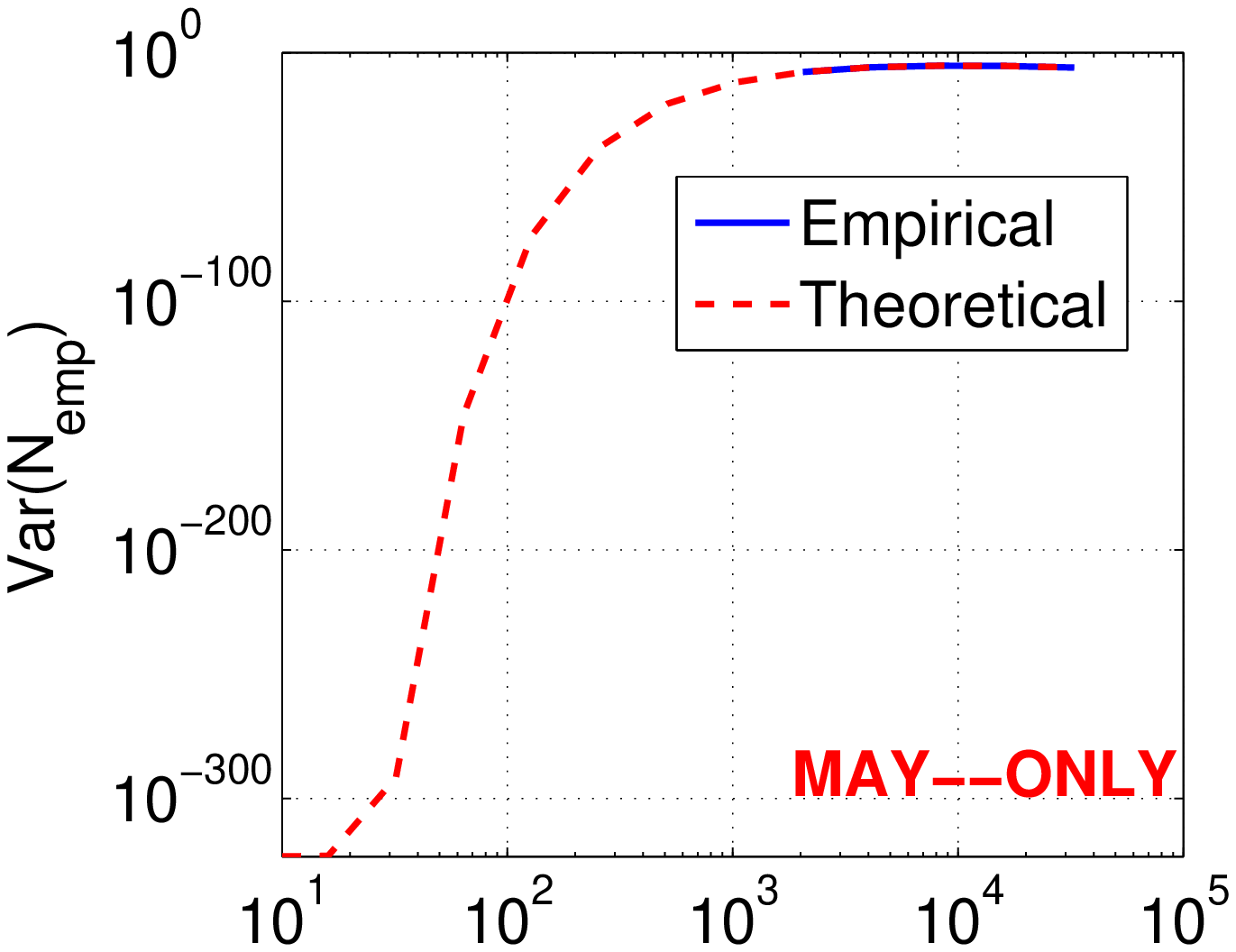}\hspace{-0.1in}
\includegraphics[width=1.5in]{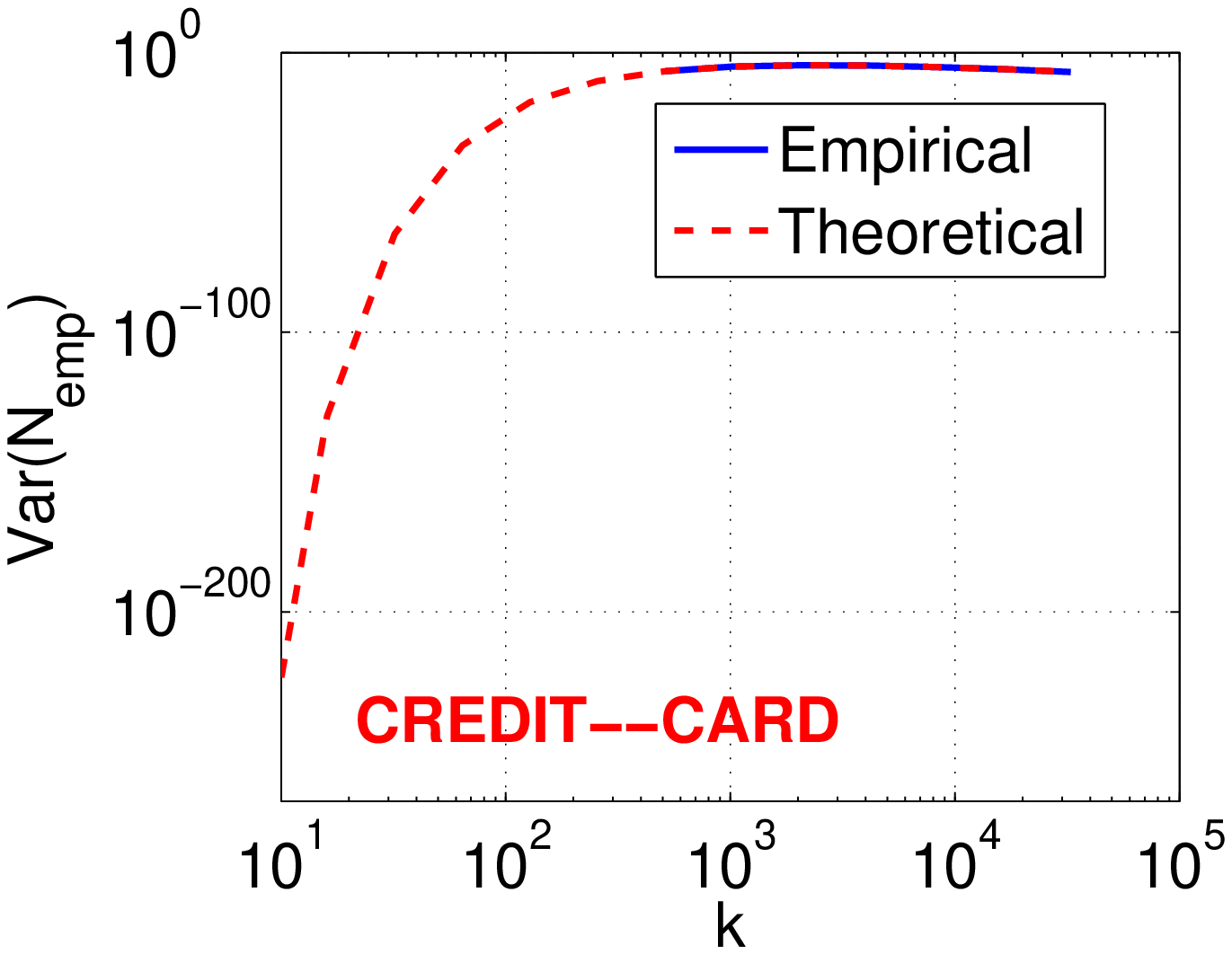}\hspace{-0.1in}
\includegraphics[width=1.5in]{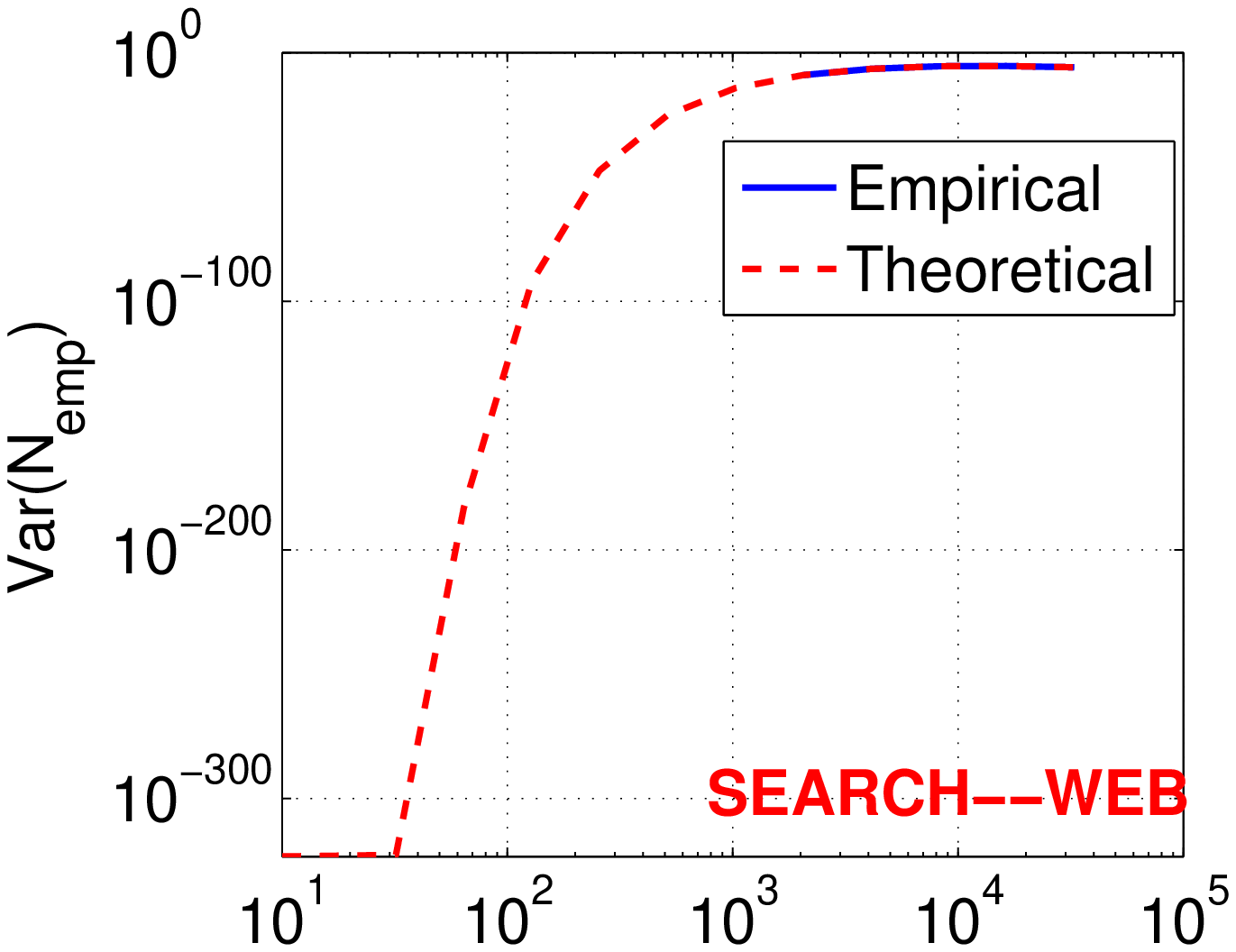}\hspace{-0.1in}
\includegraphics[width=1.5in]{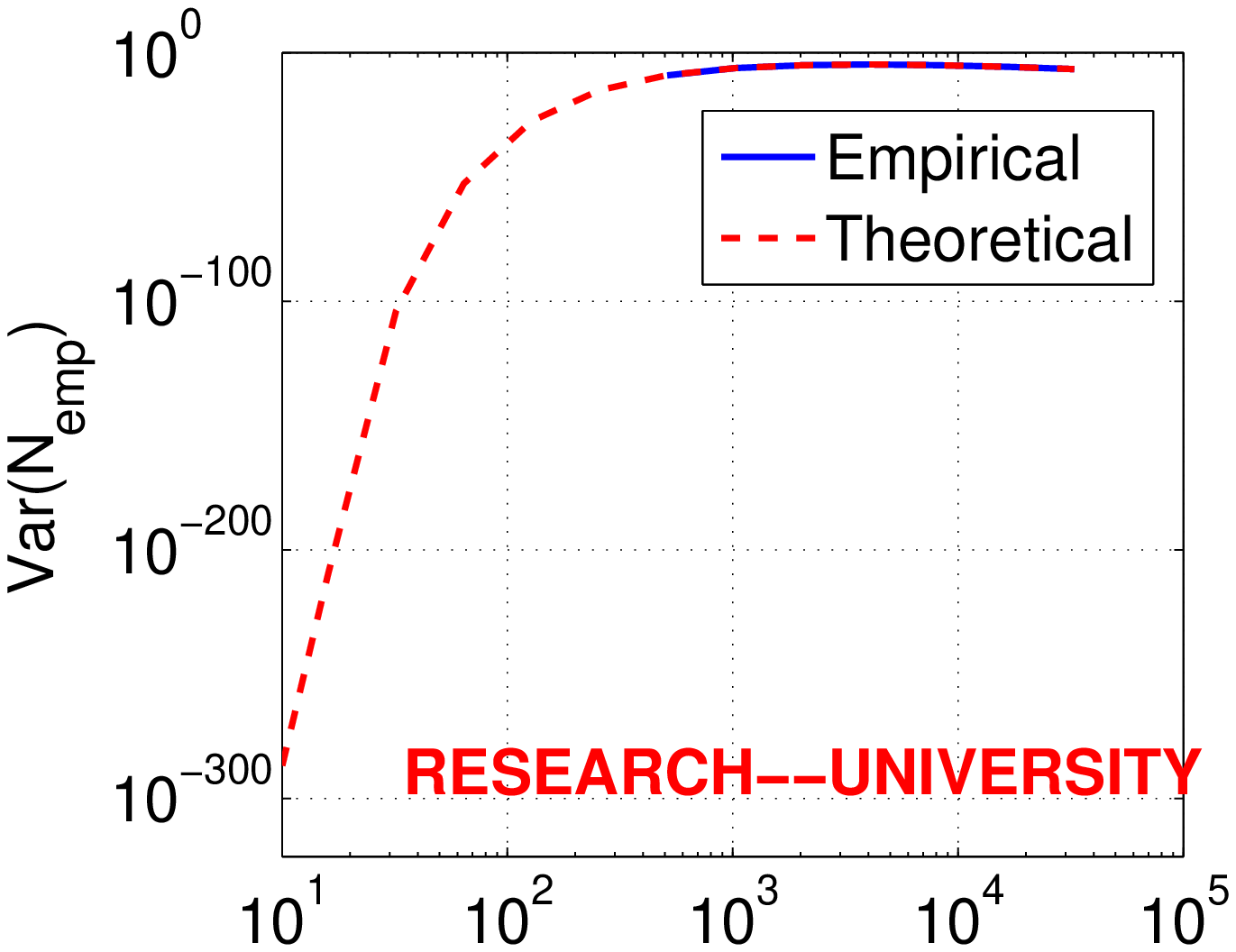}\hspace{-0.1in}
\includegraphics[width=1.5in]{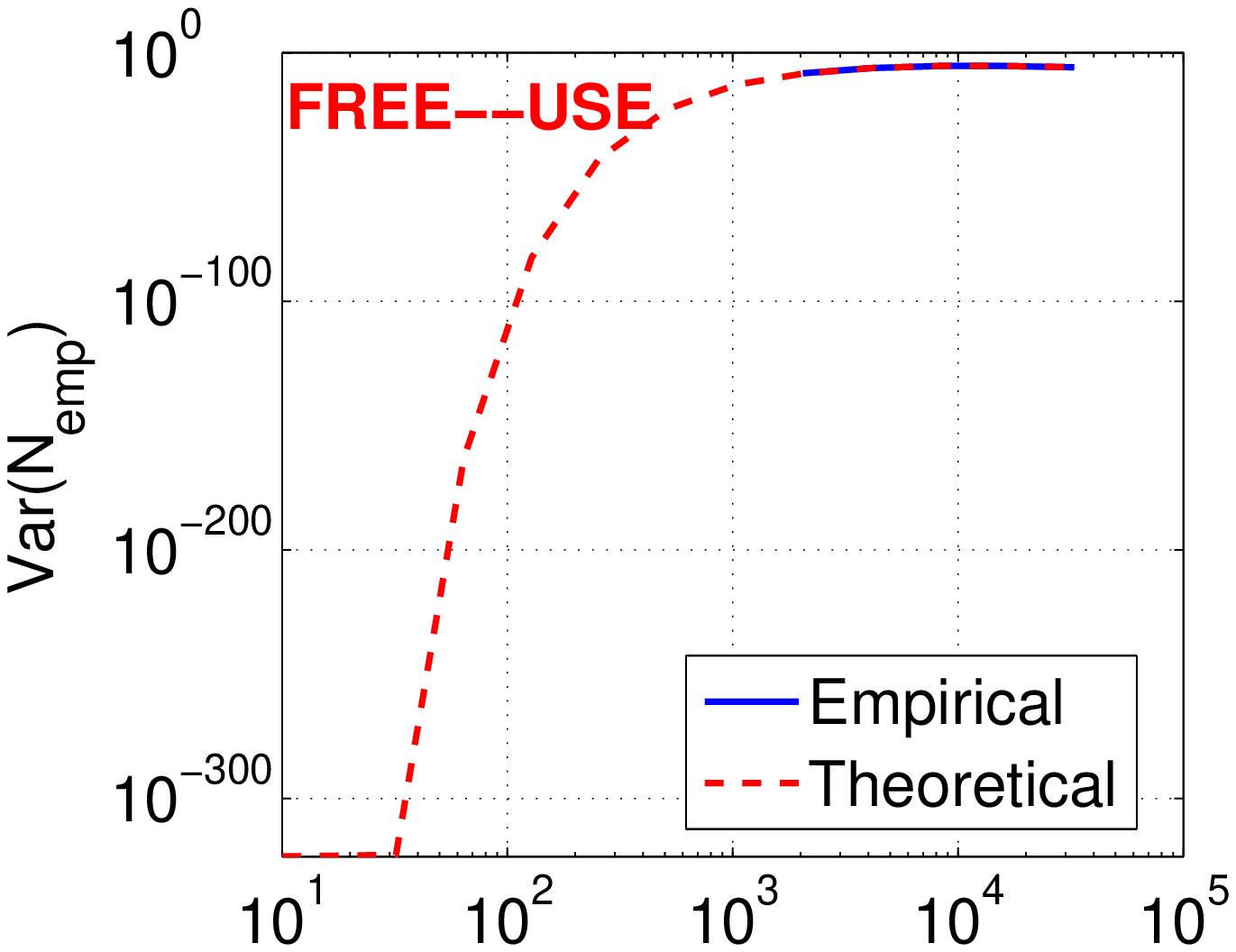}
}
\mbox{
\includegraphics[width=1.5in]{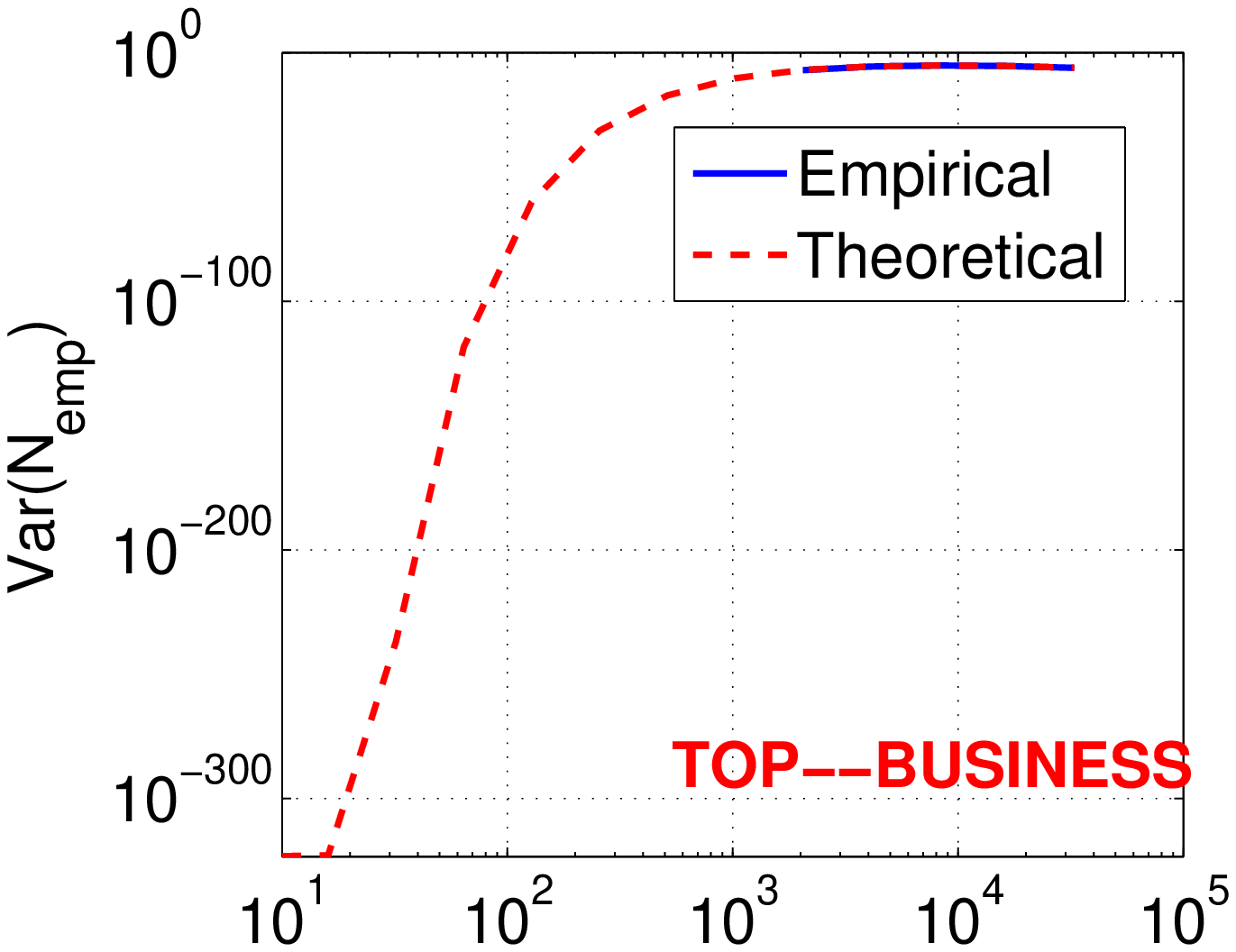}\hspace{-0.1in}
\includegraphics[width=1.5in]{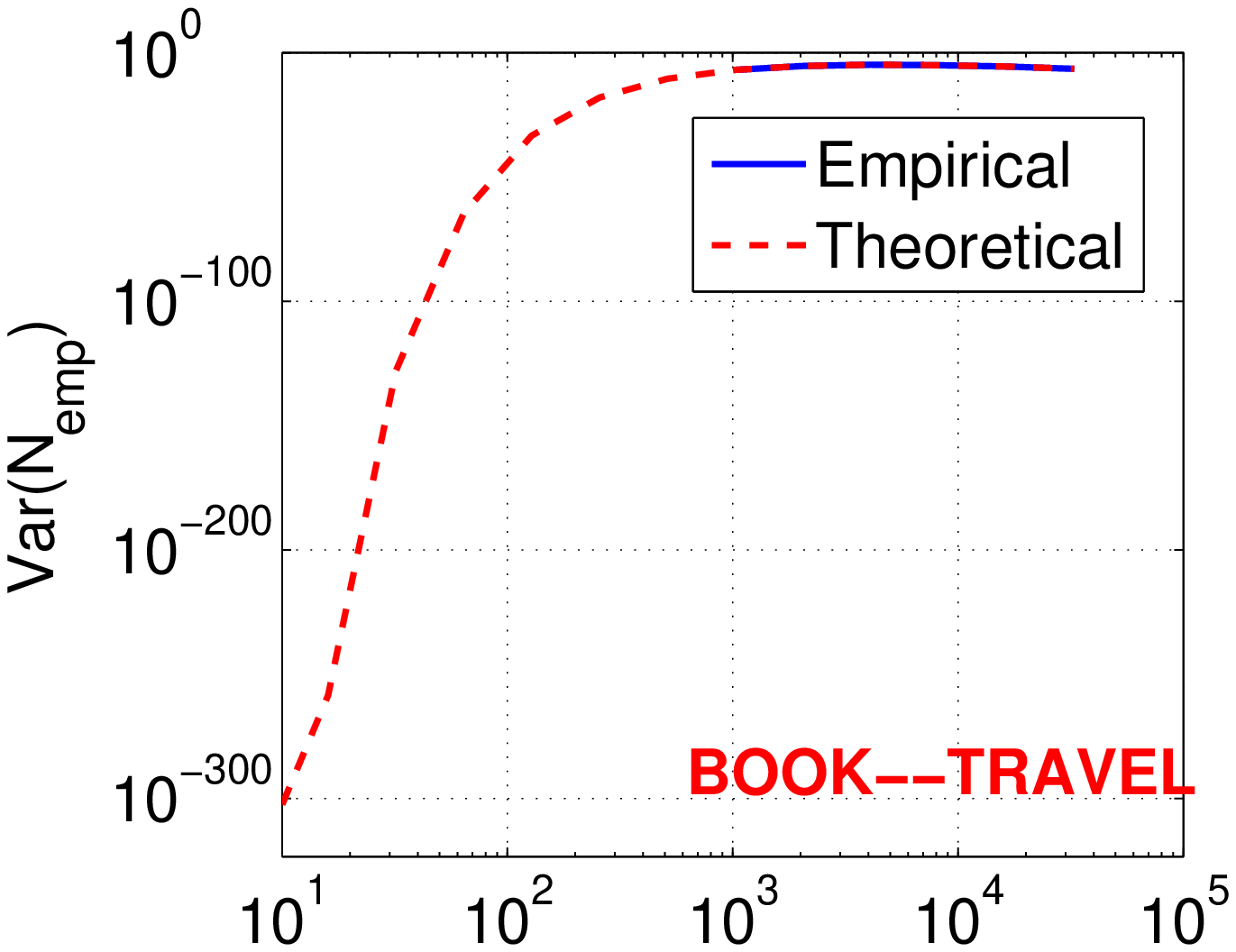}\hspace{-0.1in}
\includegraphics[width=1.5in]{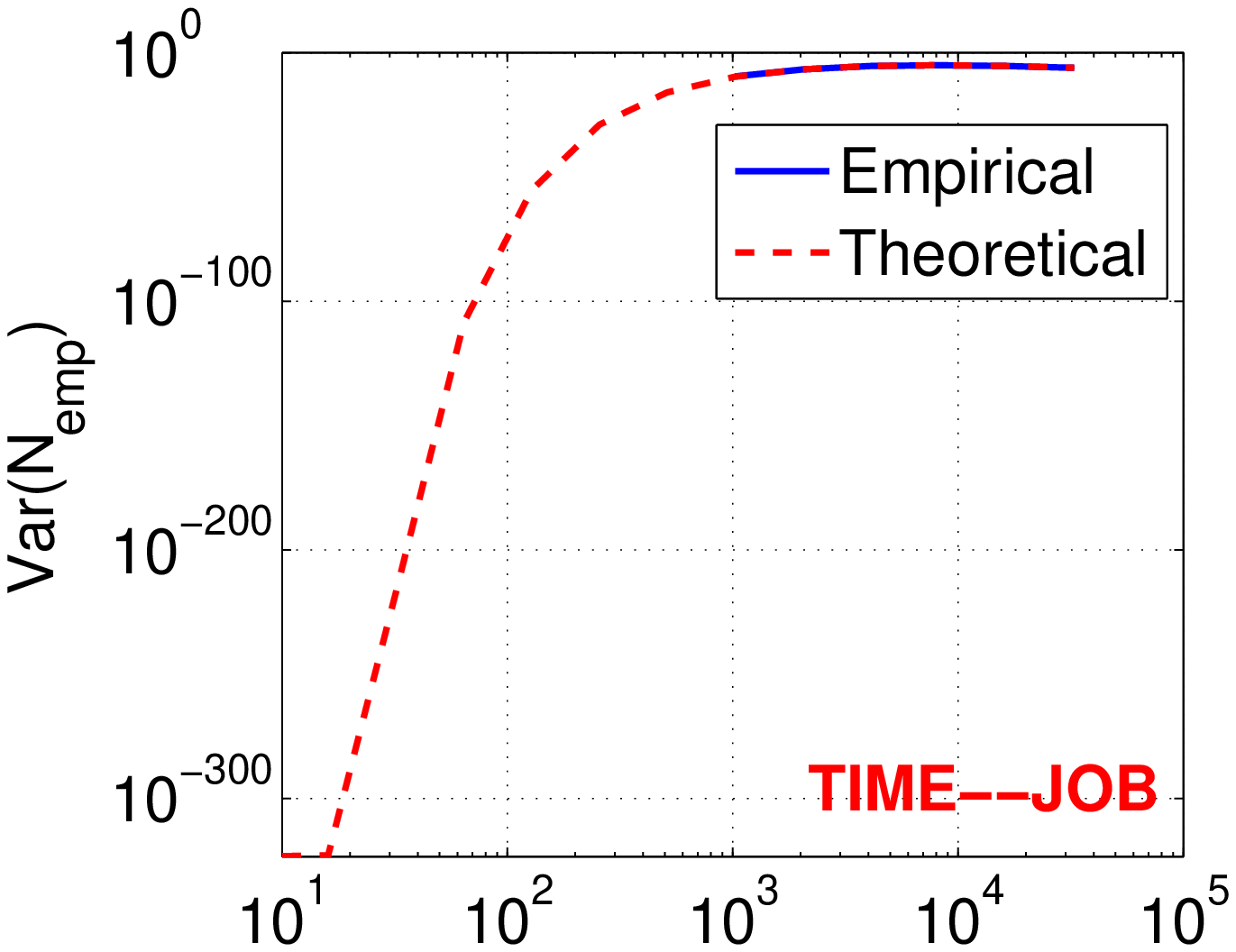}\hspace{-0.1in}
\includegraphics[width=1.5in]{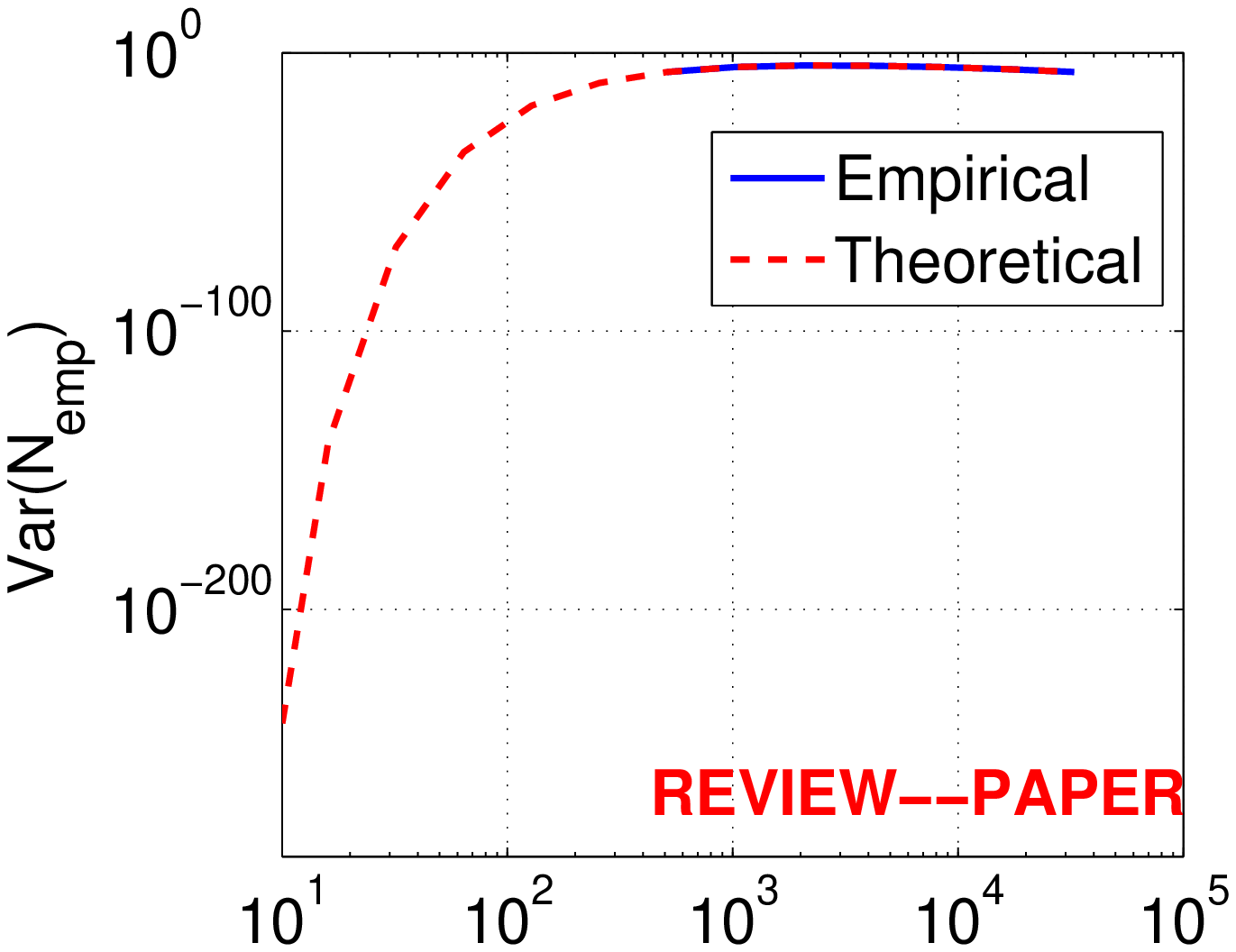}\hspace{-0.1in}
\includegraphics[width=1.5in]{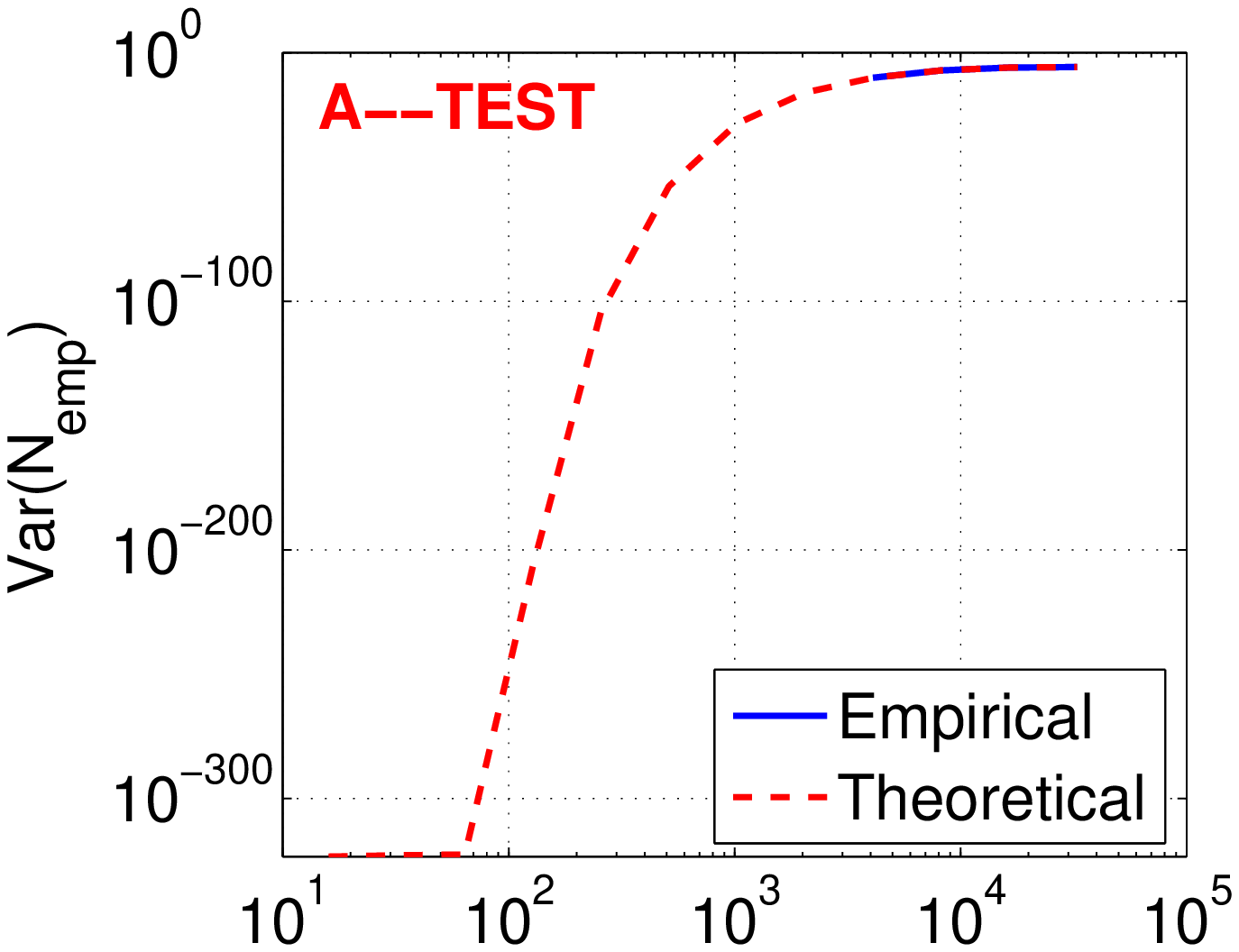}
}

\end{center}\vspace{-0.3in}
\caption{$Var(N_{emp})/k^2$. The empirical curves essentially overlap the theoretical curves as derived in Lemma~\ref{lem_Nemp}, i.e., (\ref{eqn_Nemp_var}).} \label{fig_Nemp_var}
\end{figure}

Figure~\ref{fig_Nemp_mean} and Figure~\ref{fig_Nemp_var} respectively verify $E(N_{emp})$ and $Var(N_{emp})$ as derived in Lemma~\ref{lem_Nemp}. Clearly, the theoretical curves  overlap the empirical curves.

Note that $N_{emp}$ is essentially 0 when $k$ is not large. Roughly when $k/f > 1/5$, the number of empty bins becomes  noticeable, which is expected because $E(N_{emp})/k\approx \left(1-\frac{1}{k}\right)^f \approx e^{-f/k}$ and $e^{-5} = 0.0067$. Practically speaking, as we often use minwise hashing to substantially reduce the number of nonzeros in massive datasets, we would expect that usually $f\gg k$ anyway. See Sec.~\ref{sec_empty_bin} for more discussion about strategies for dealing with empty bins.

\subsubsection{$E(N_{mat})$ and $Var(N_{mat})$}

\begin{figure}[h!]
\begin{center}
\mbox{
\includegraphics[width=1.5in]{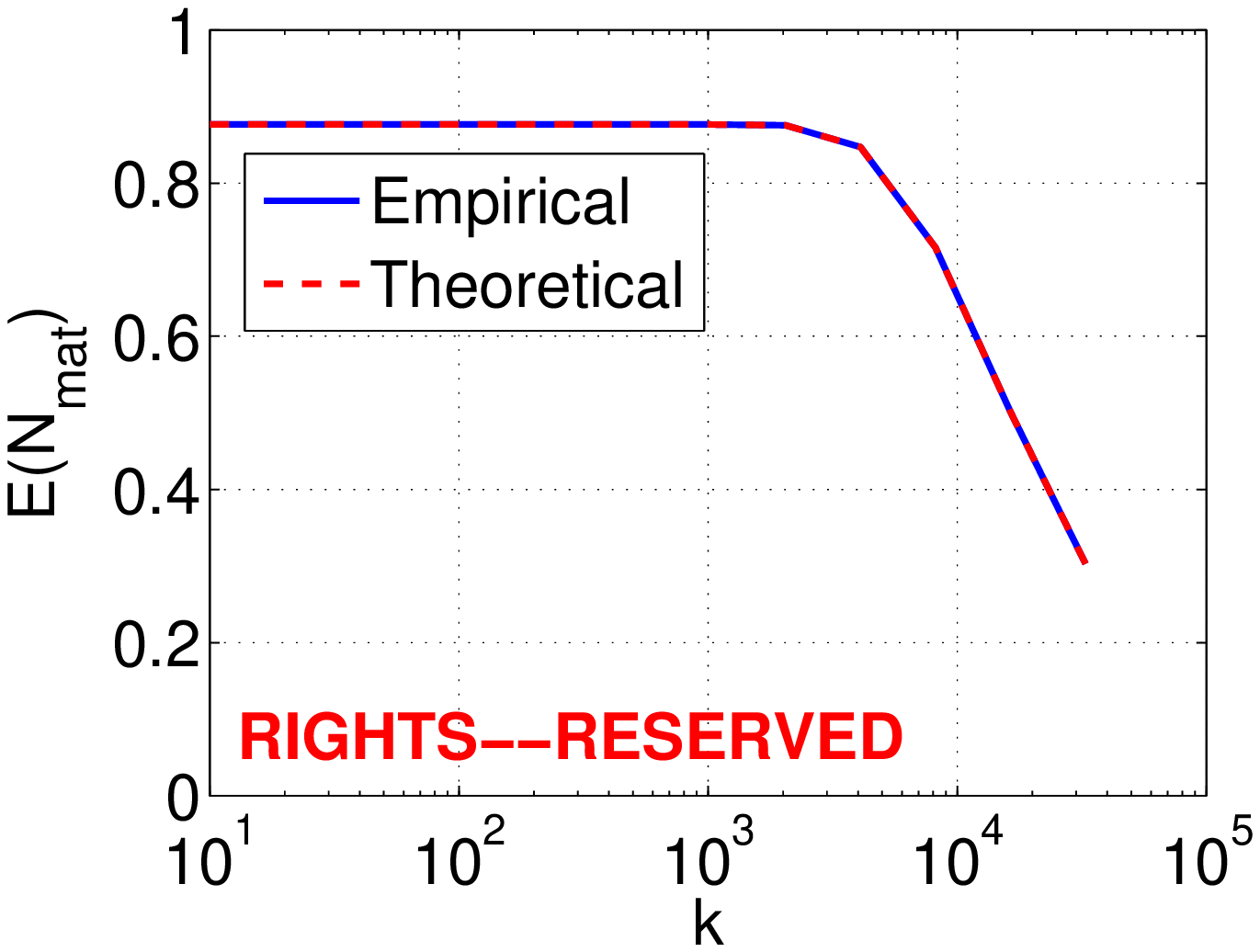}\hspace{-0.1in}
\includegraphics[width=1.5in]{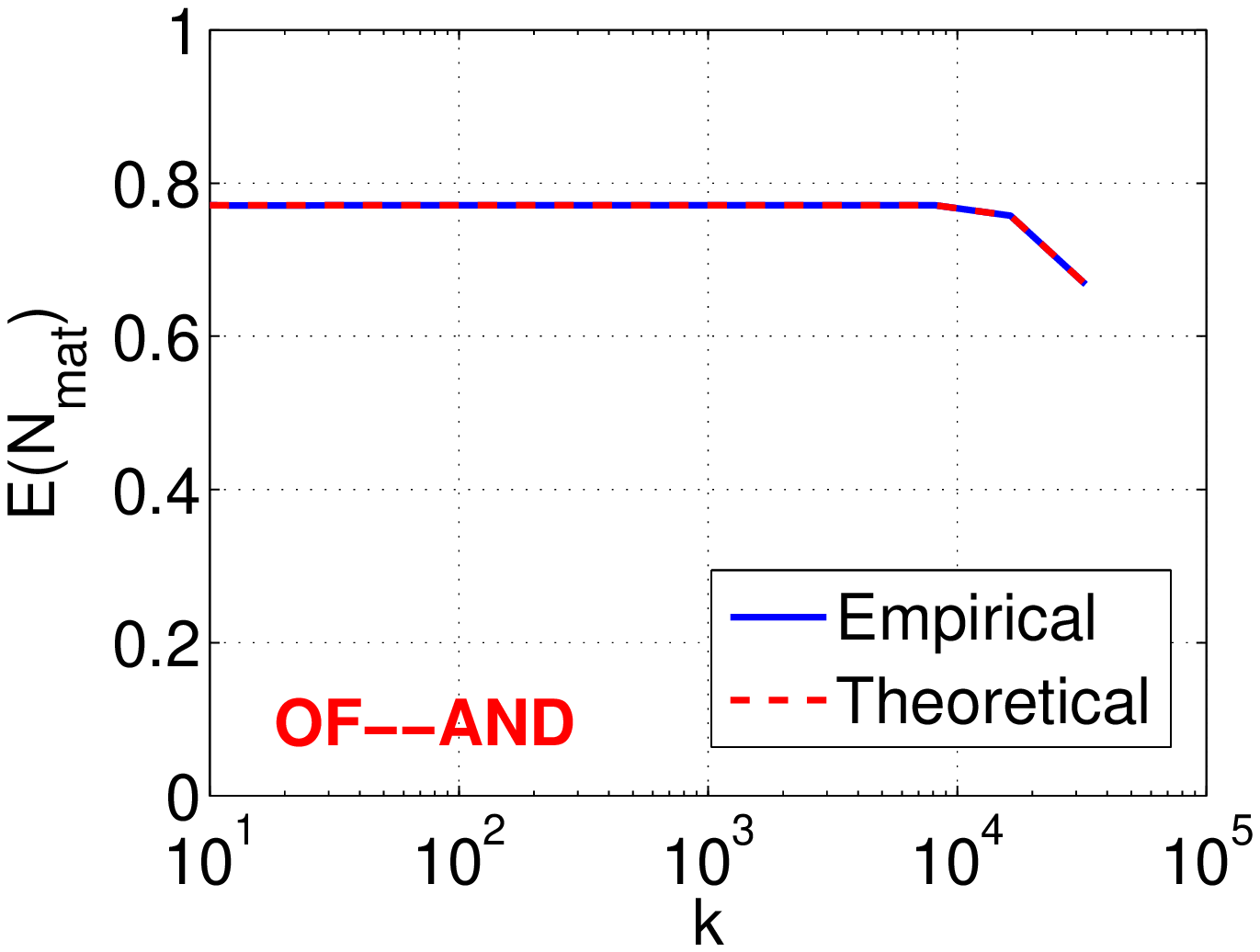}\hspace{-0.1in}
\includegraphics[width=1.5in]{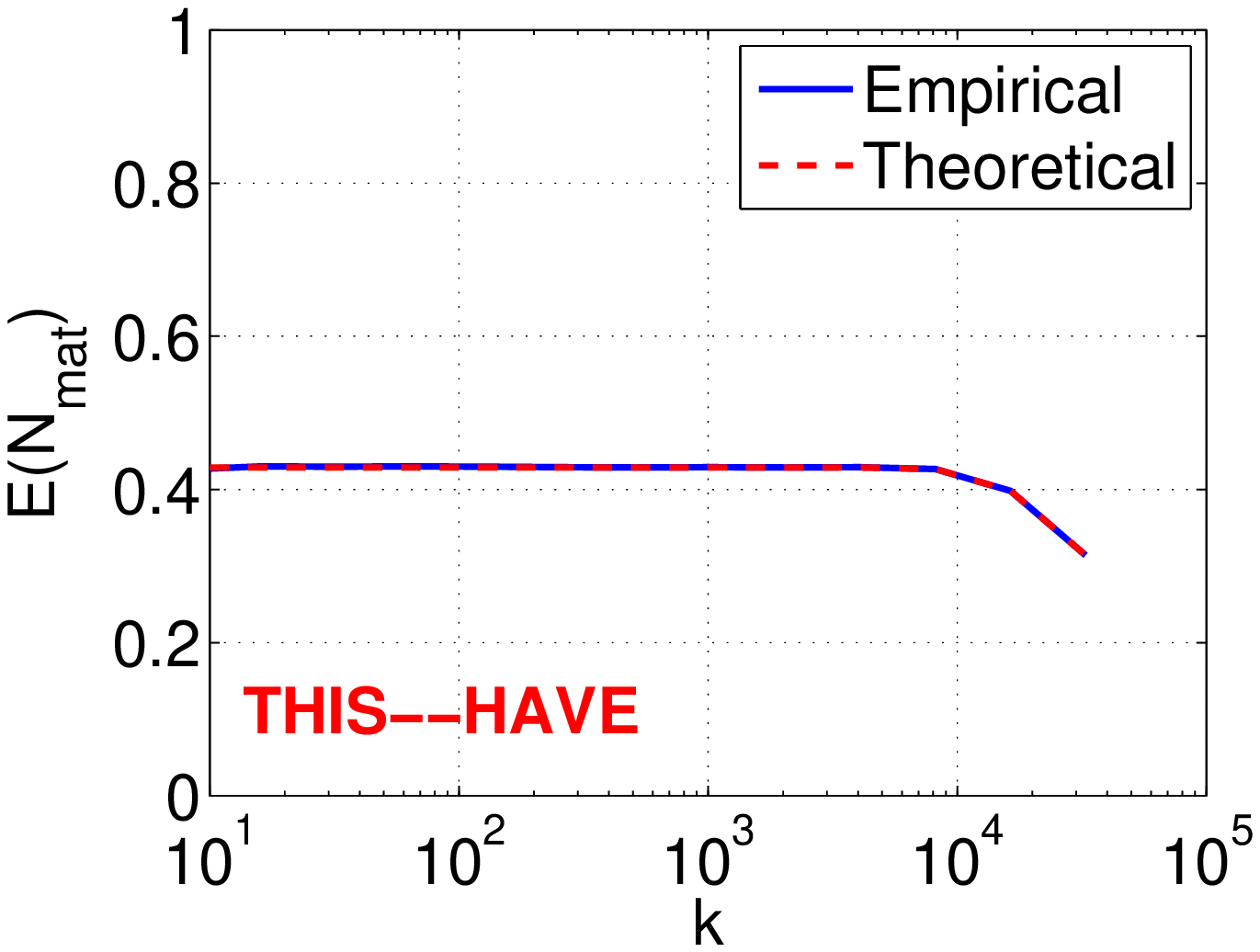}\hspace{-0.1in}
\includegraphics[width=1.5in]{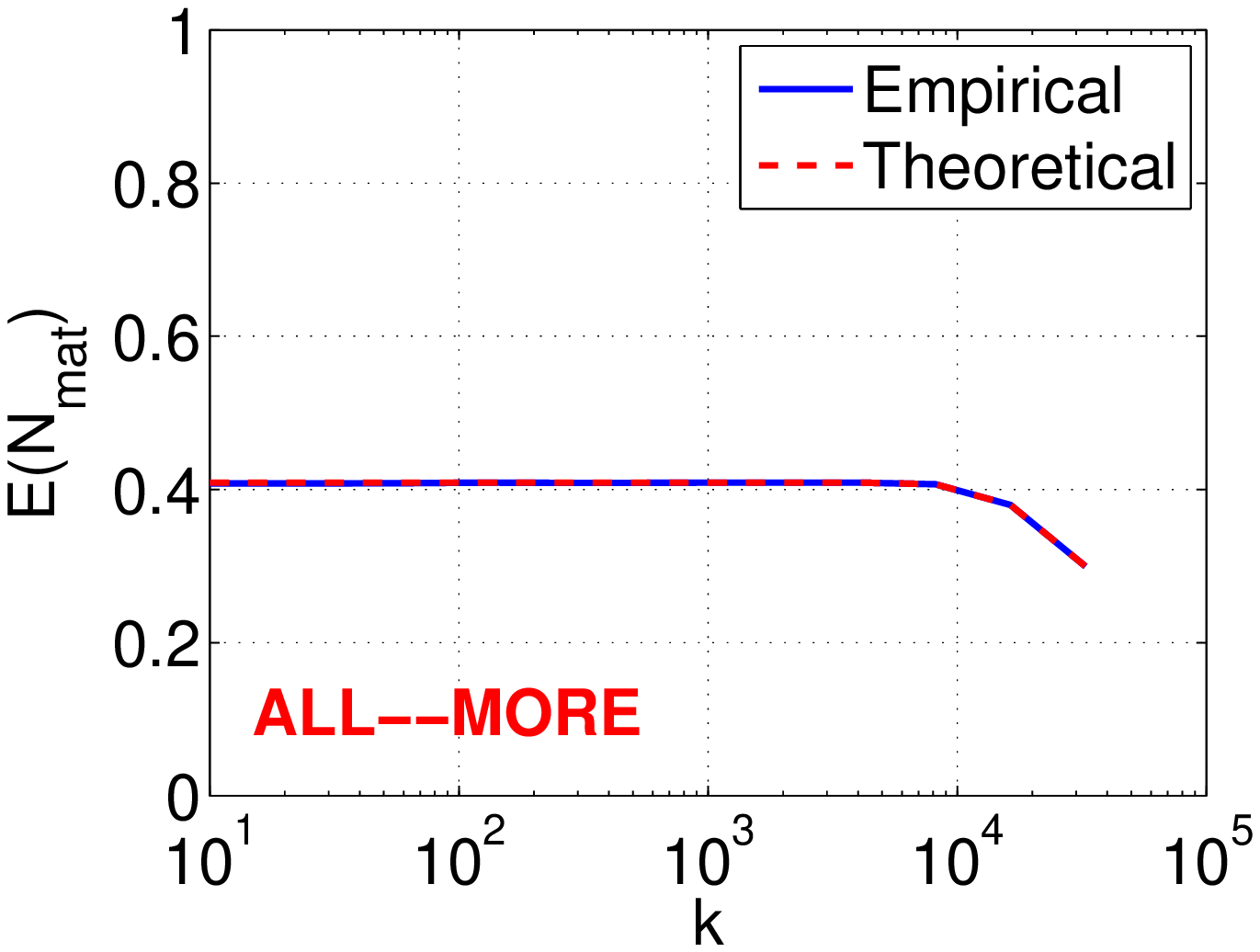}\hspace{-0.1in}
\includegraphics[width=1.5in]{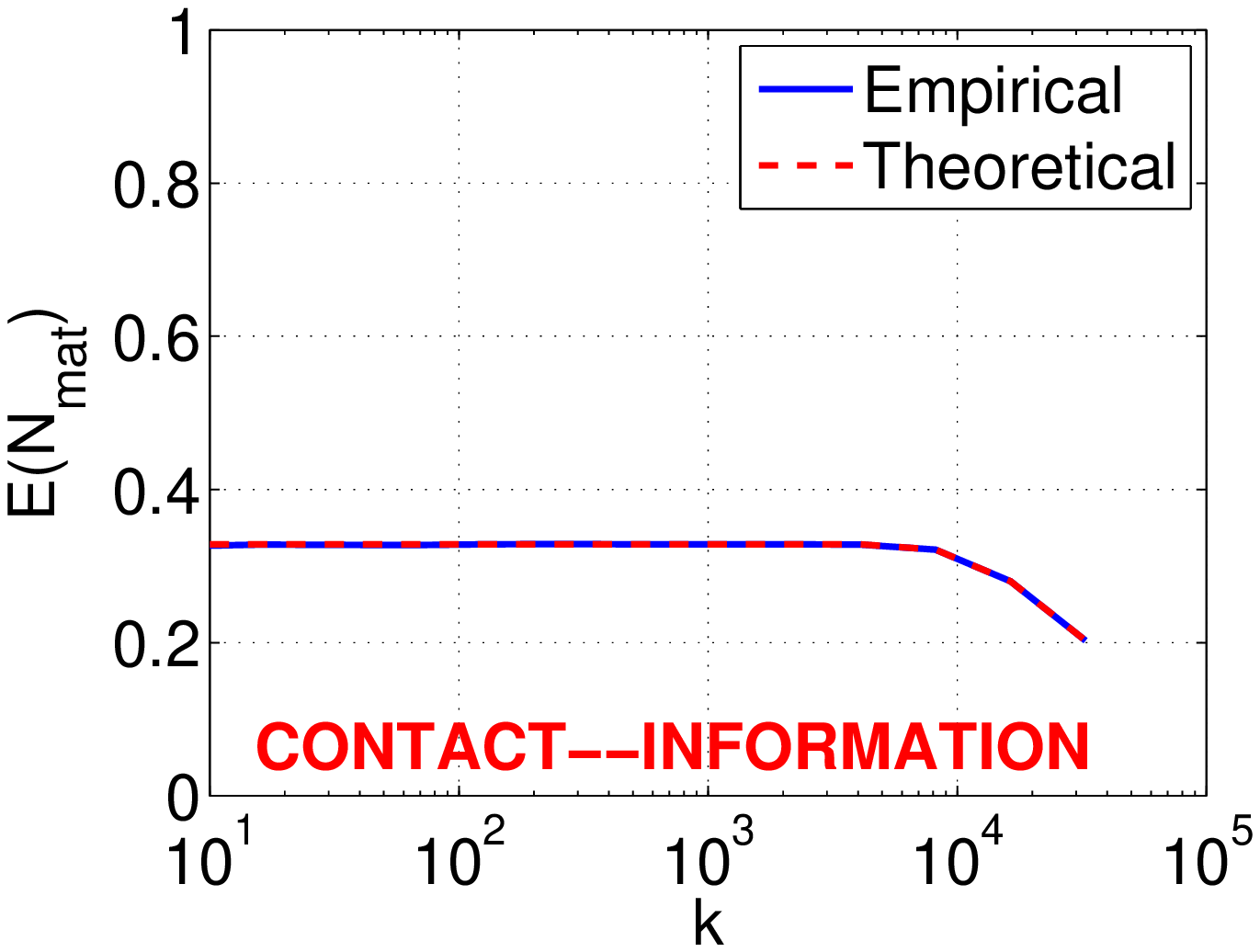}
}
\mbox{
\includegraphics[width=1.5in]{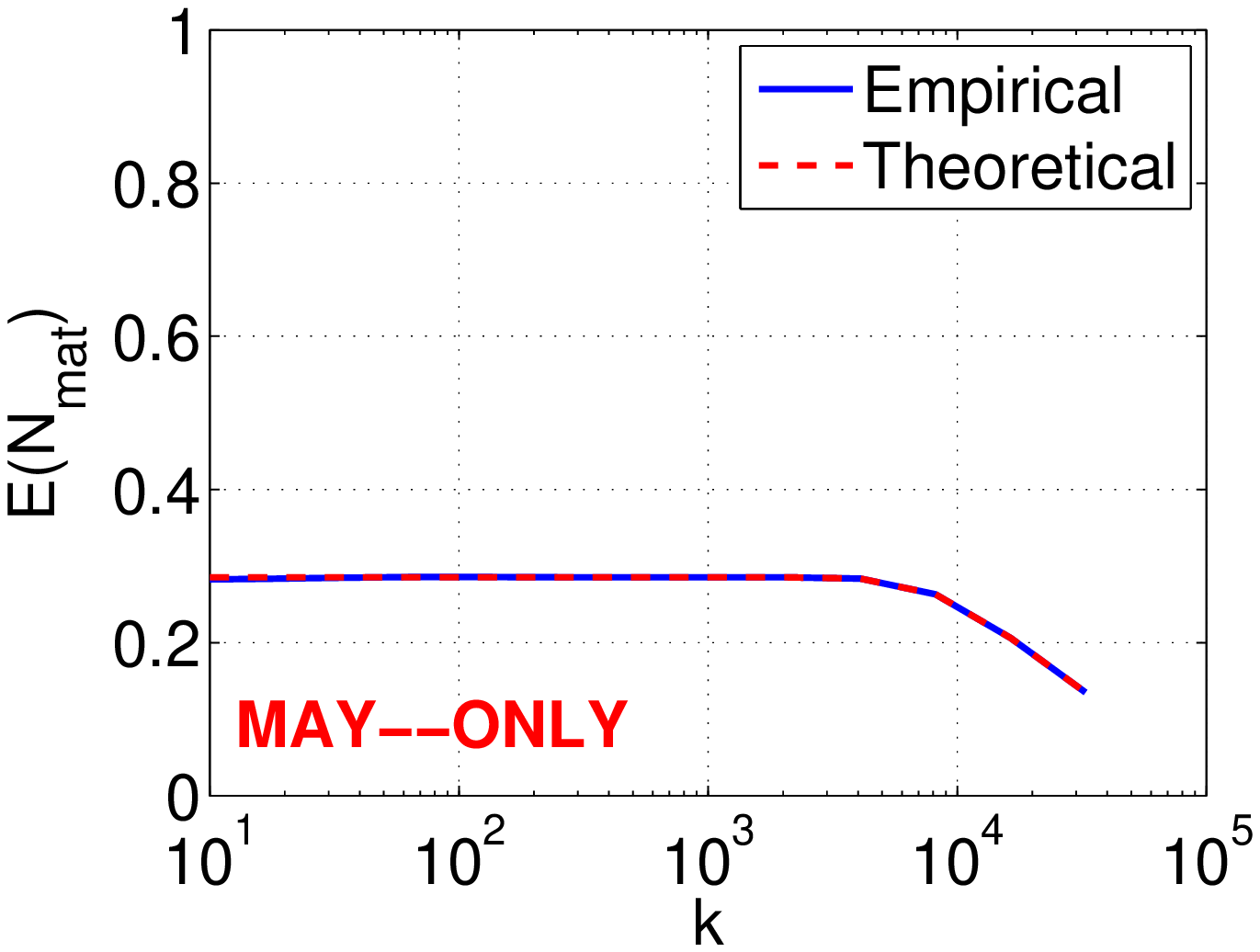}\hspace{-0.1in}
\includegraphics[width=1.5in]{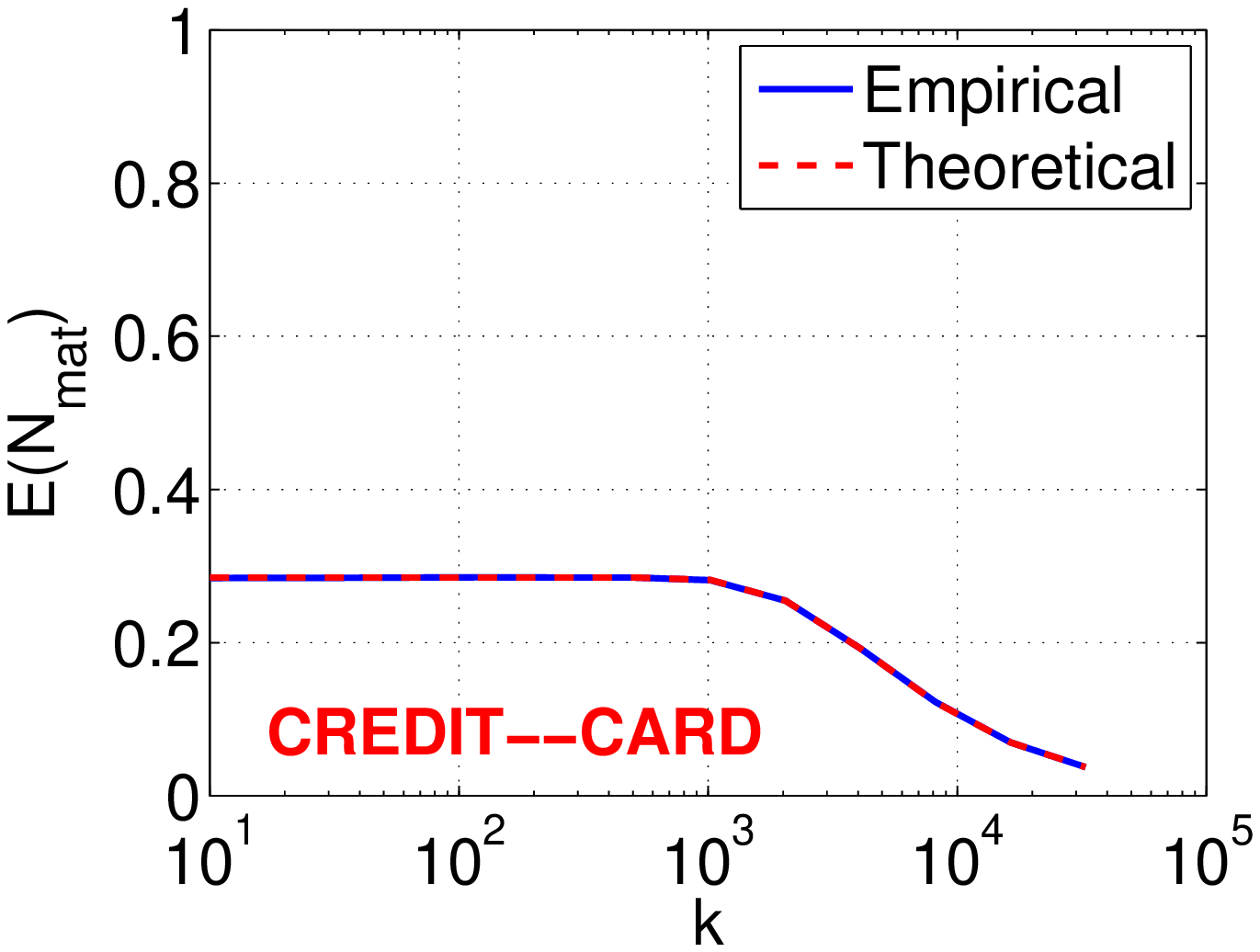}\hspace{-0.1in}
\includegraphics[width=1.5in]{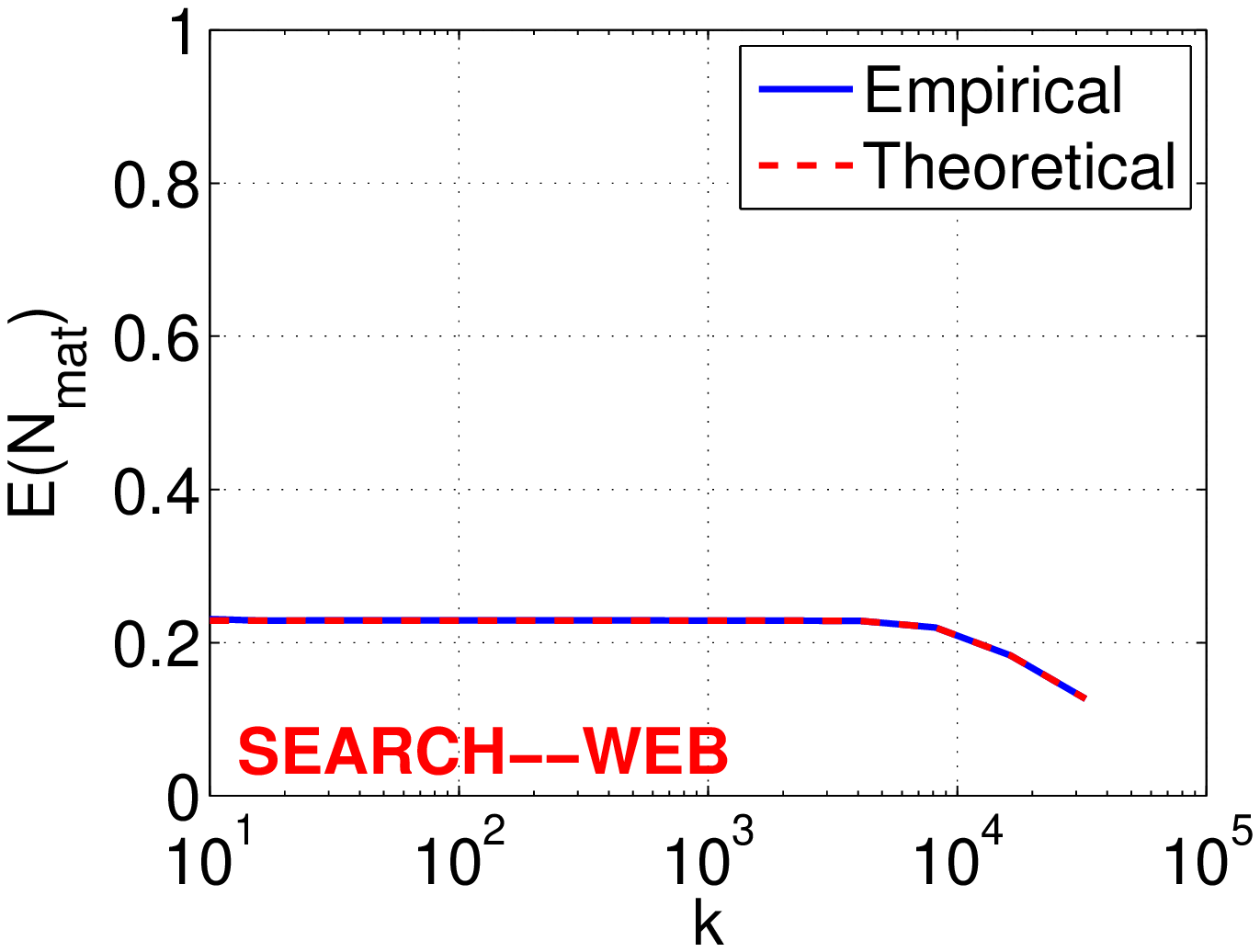}\hspace{-0.1in}
\includegraphics[width=1.5in]{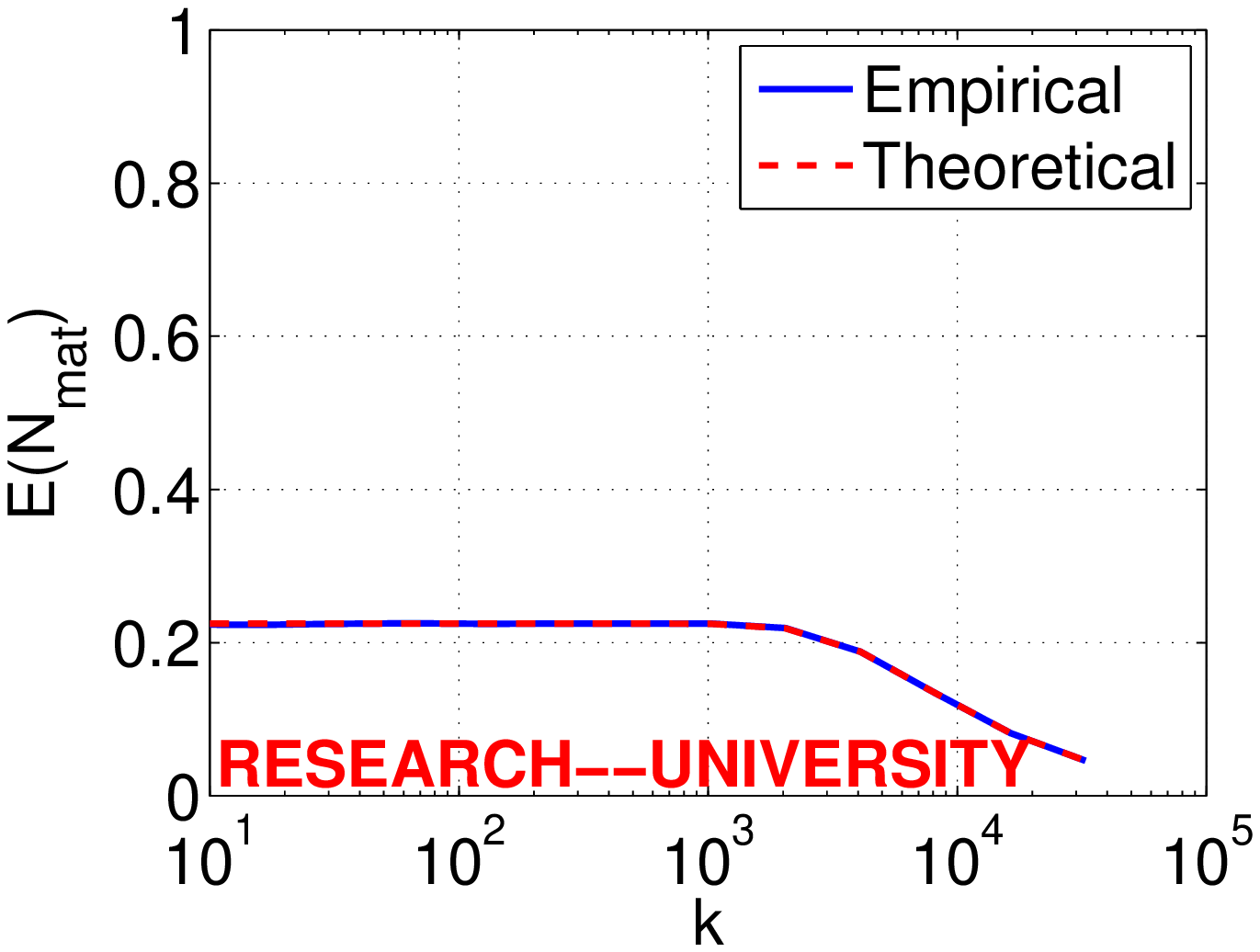}\hspace{-0.1in}
\includegraphics[width=1.5in]{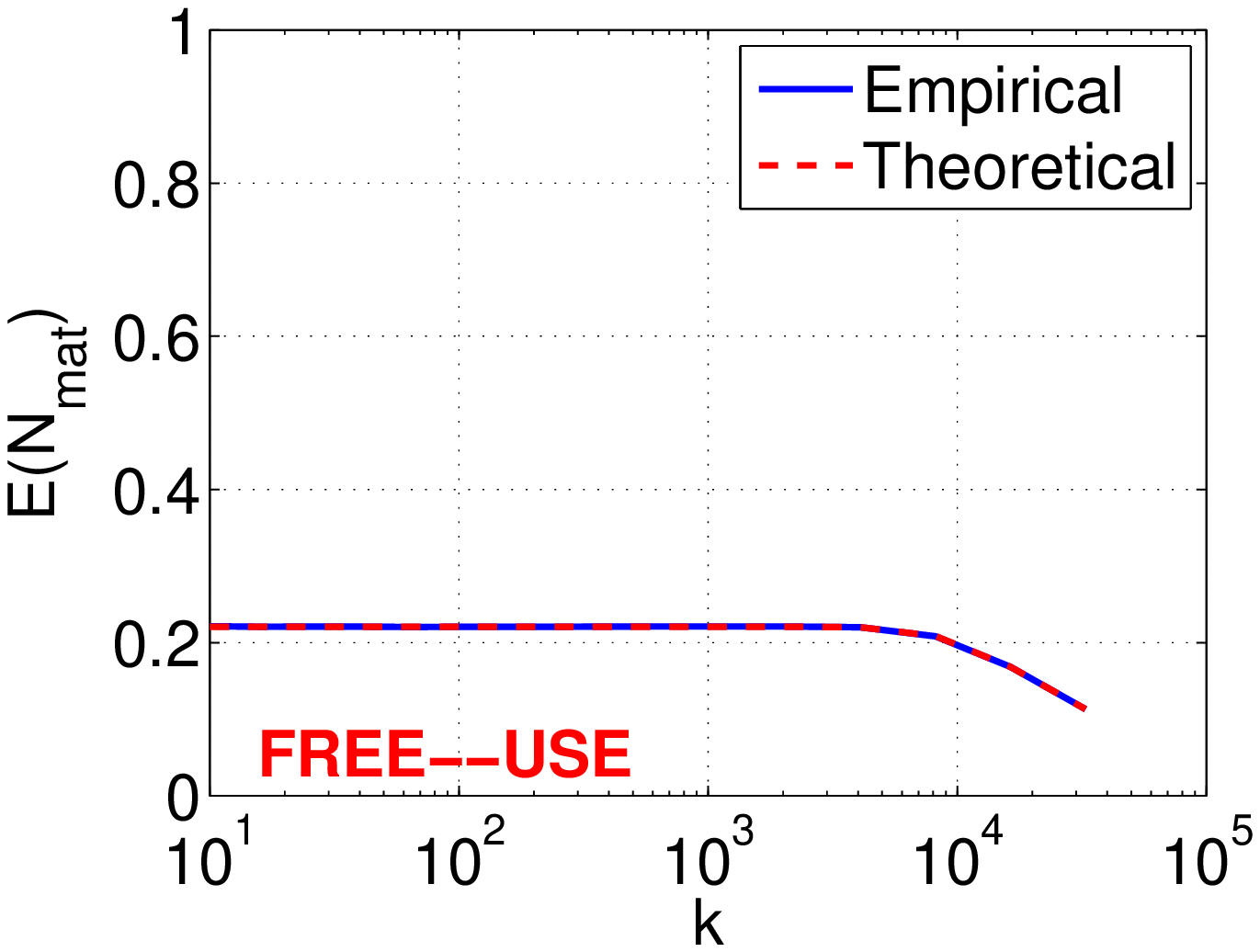}
}
\mbox{
\includegraphics[width=1.5in]{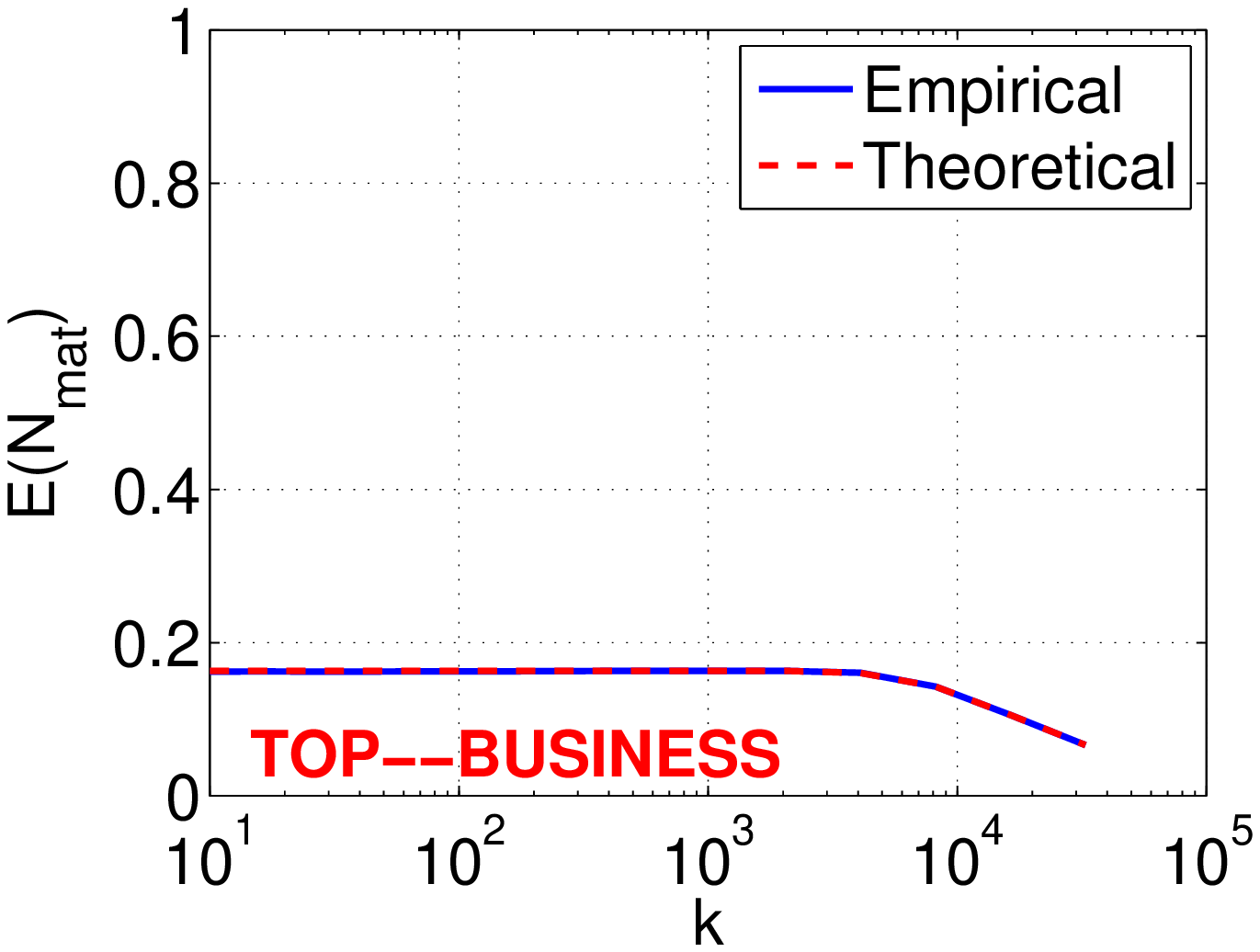}\hspace{-0.1in}
\includegraphics[width=1.5in]{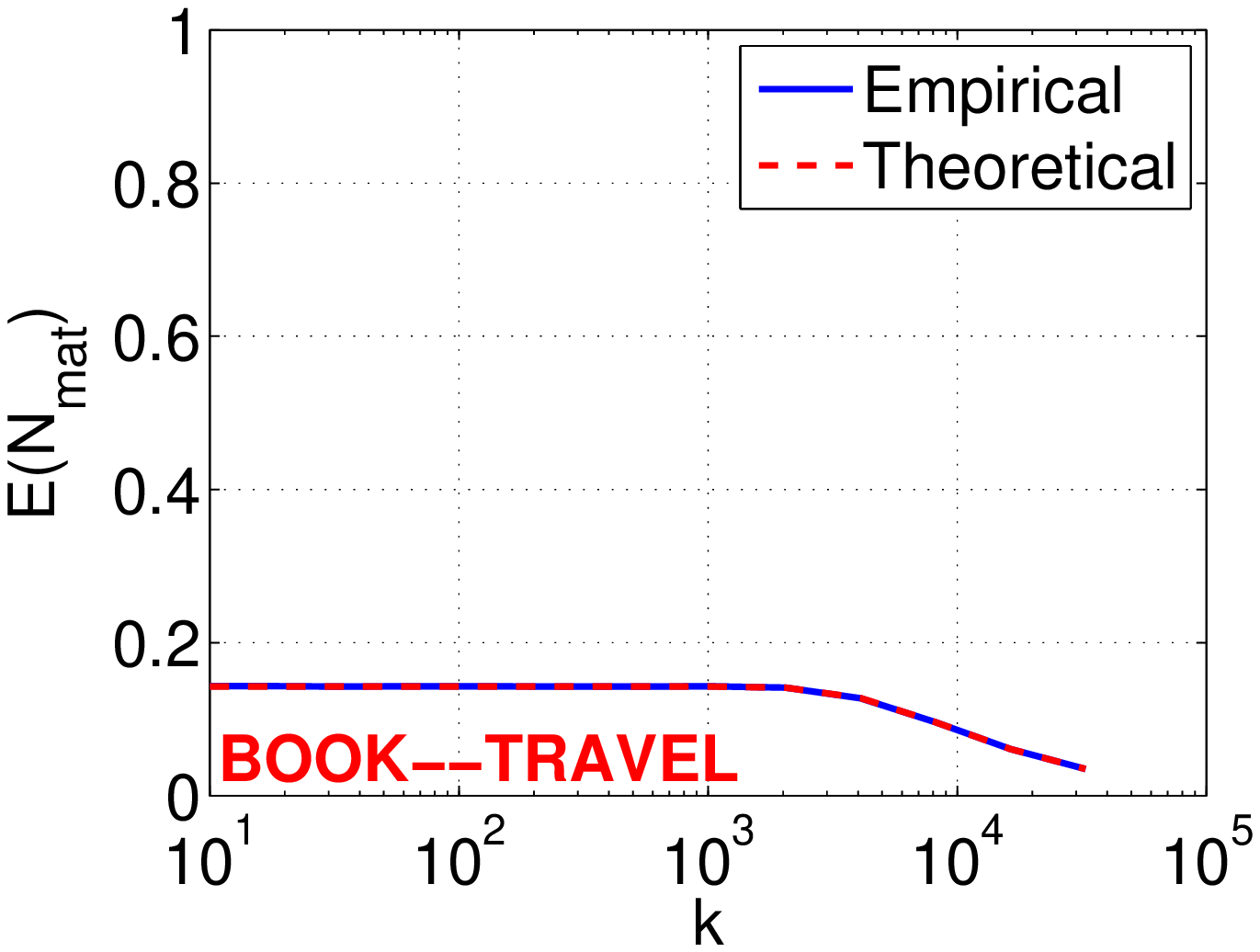}\hspace{-0.1in}
\includegraphics[width=1.5in]{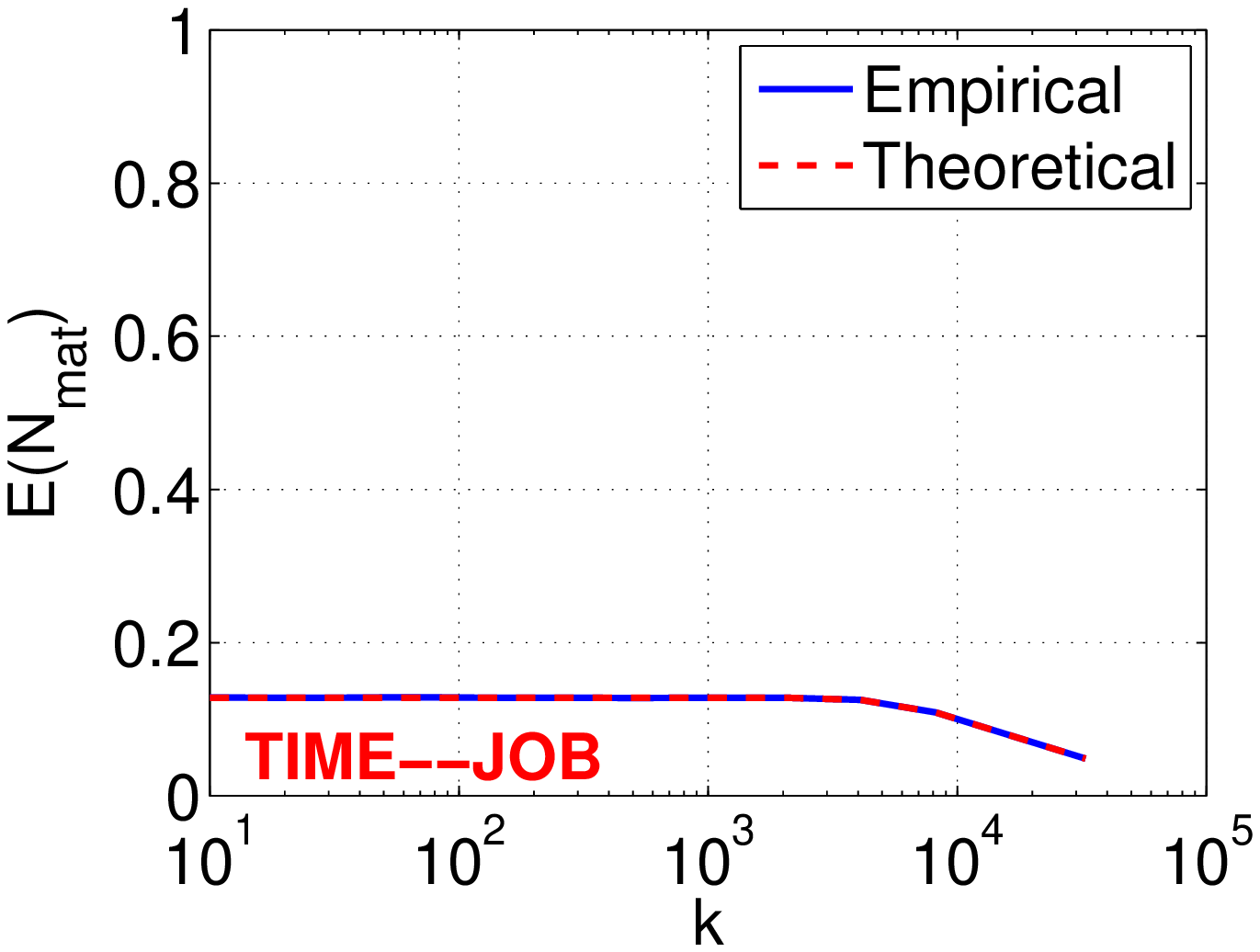}\hspace{-0.1in}
\includegraphics[width=1.5in]{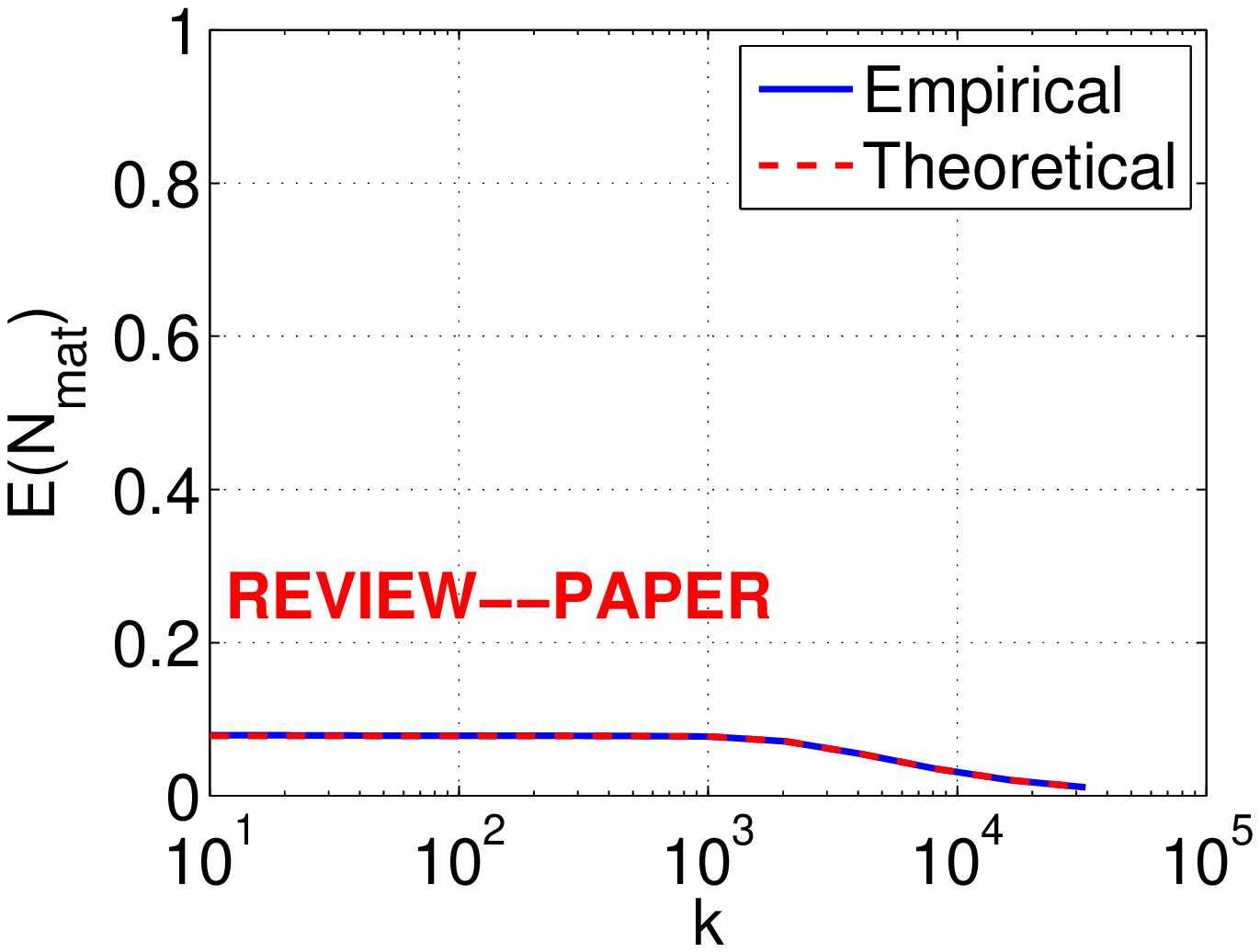}\hspace{-0.1in}
\includegraphics[width=1.5in]{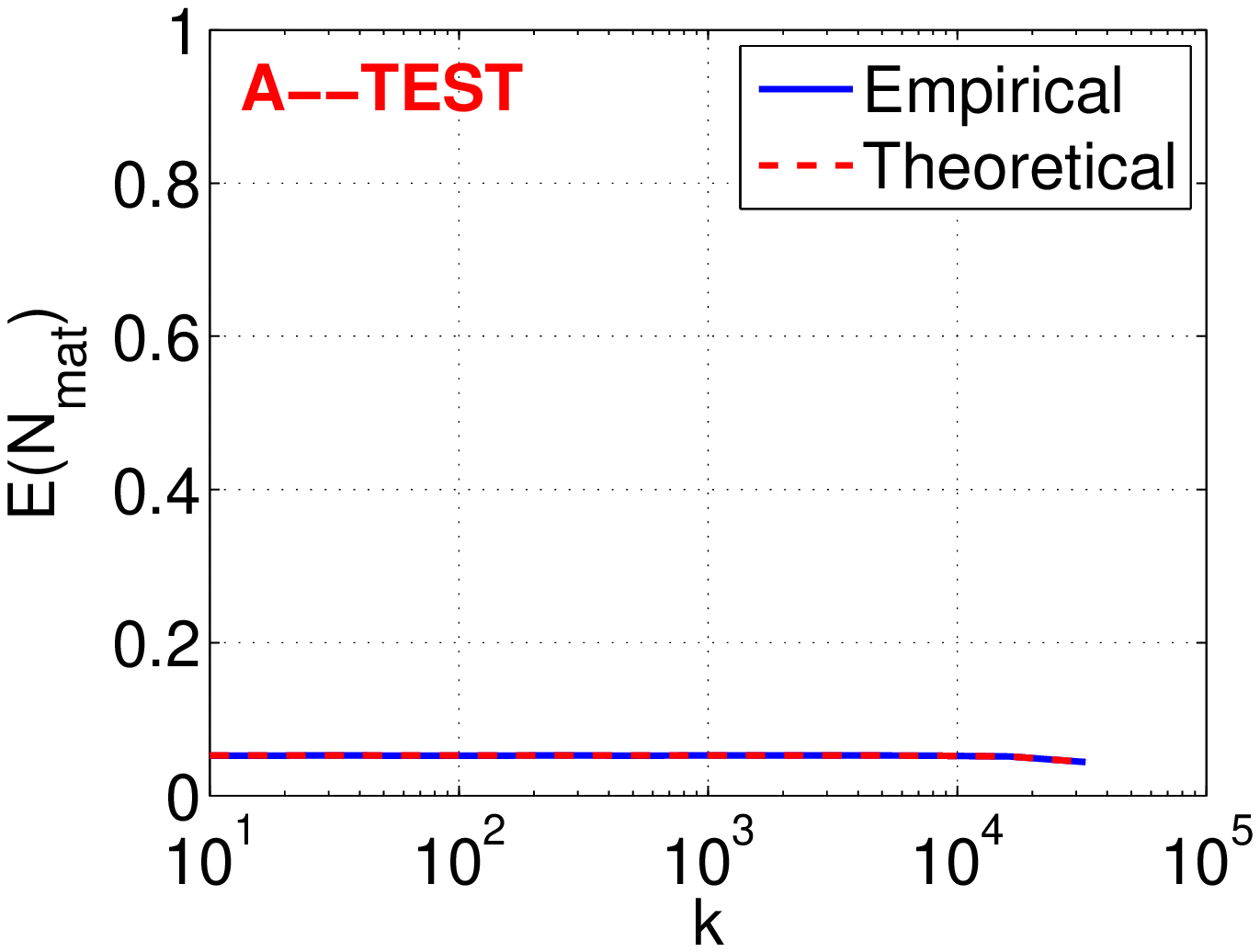}
}
\end{center}
\vspace{-0.3in}
\caption{$E(N_{mat})/k$. The empirical curves essentially overlap the theoretical curves as derived in Lemma~\ref{lem_Nmat}, i.e., (\ref{eqn_Nmat_mean}). }\label{fig_Nmat_mean}
\end{figure}

\begin{figure}[h!]
\begin{center}

\mbox{
\includegraphics[width=1.5in]{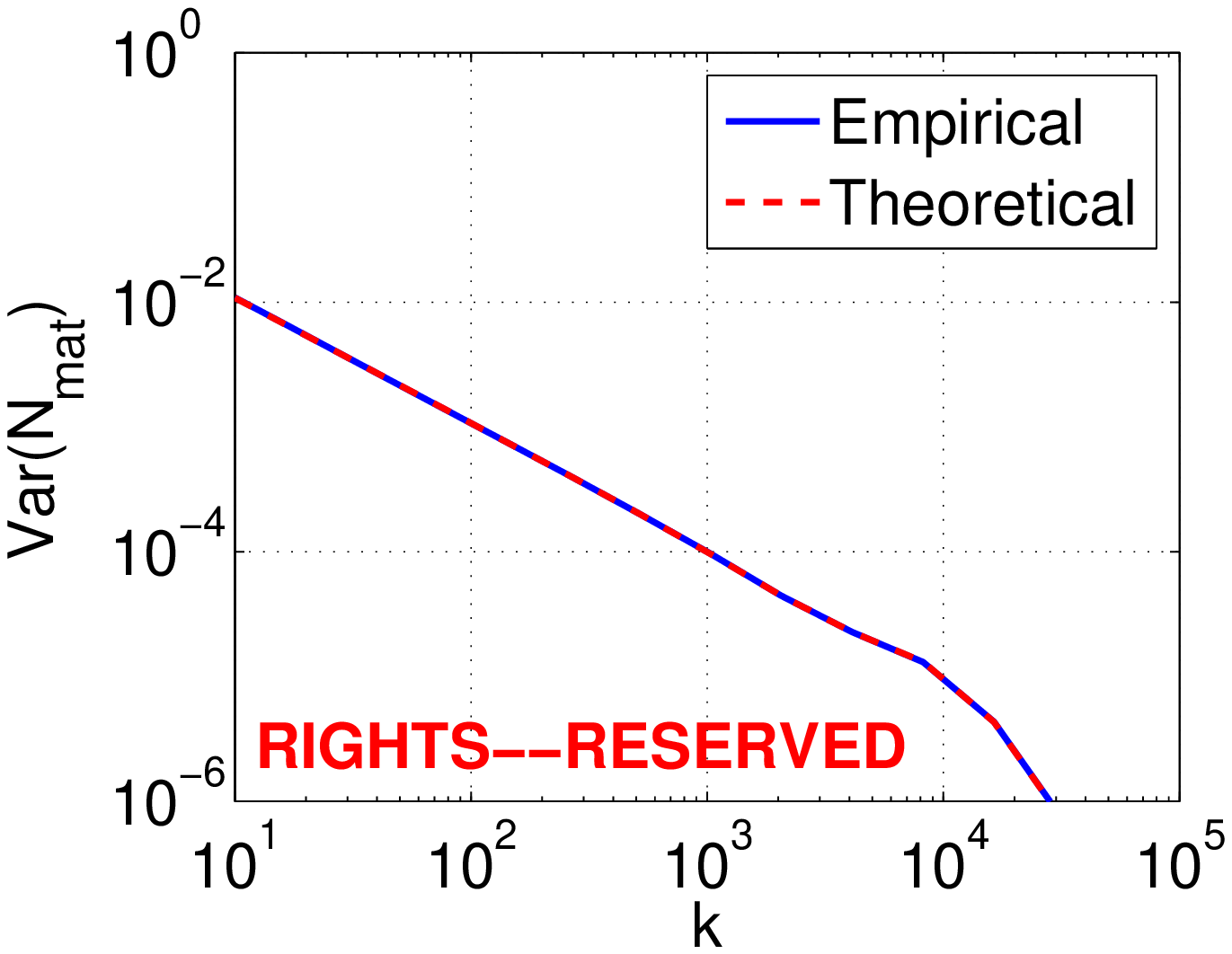}\hspace{-0.1in}
\includegraphics[width=1.5in]{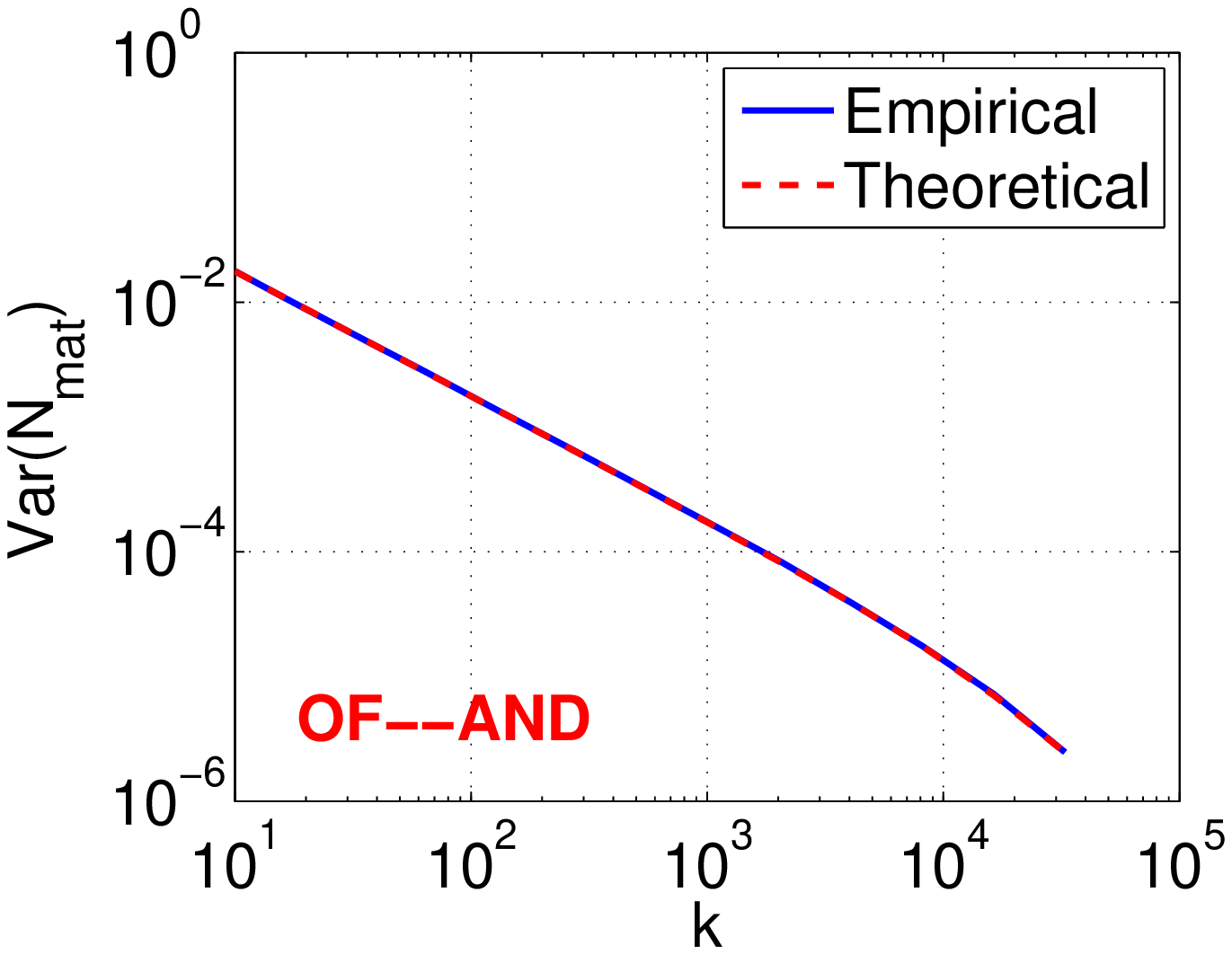}\hspace{-0.1in}
\includegraphics[width=1.5in]{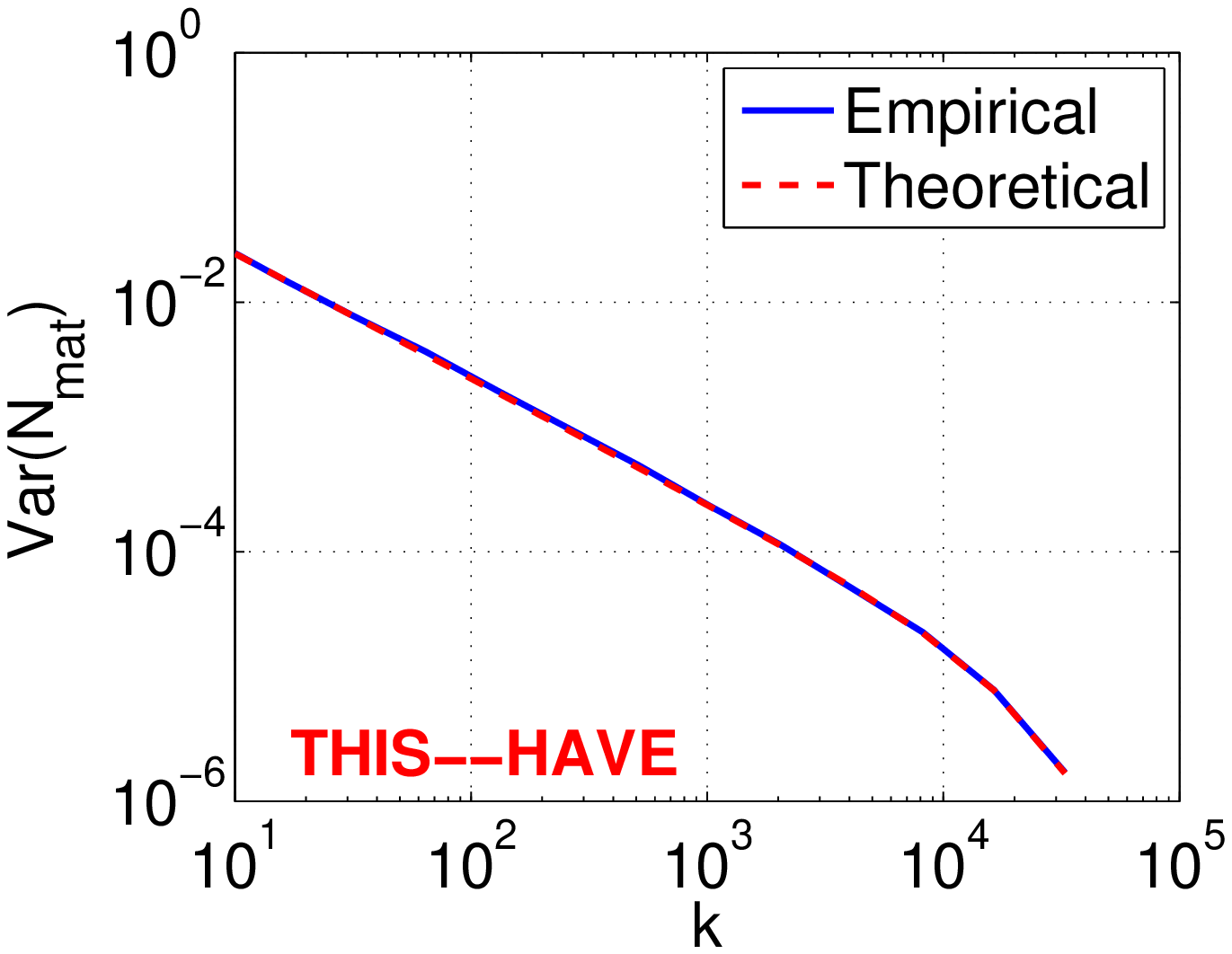}\hspace{-0.1in}
\includegraphics[width=1.5in]{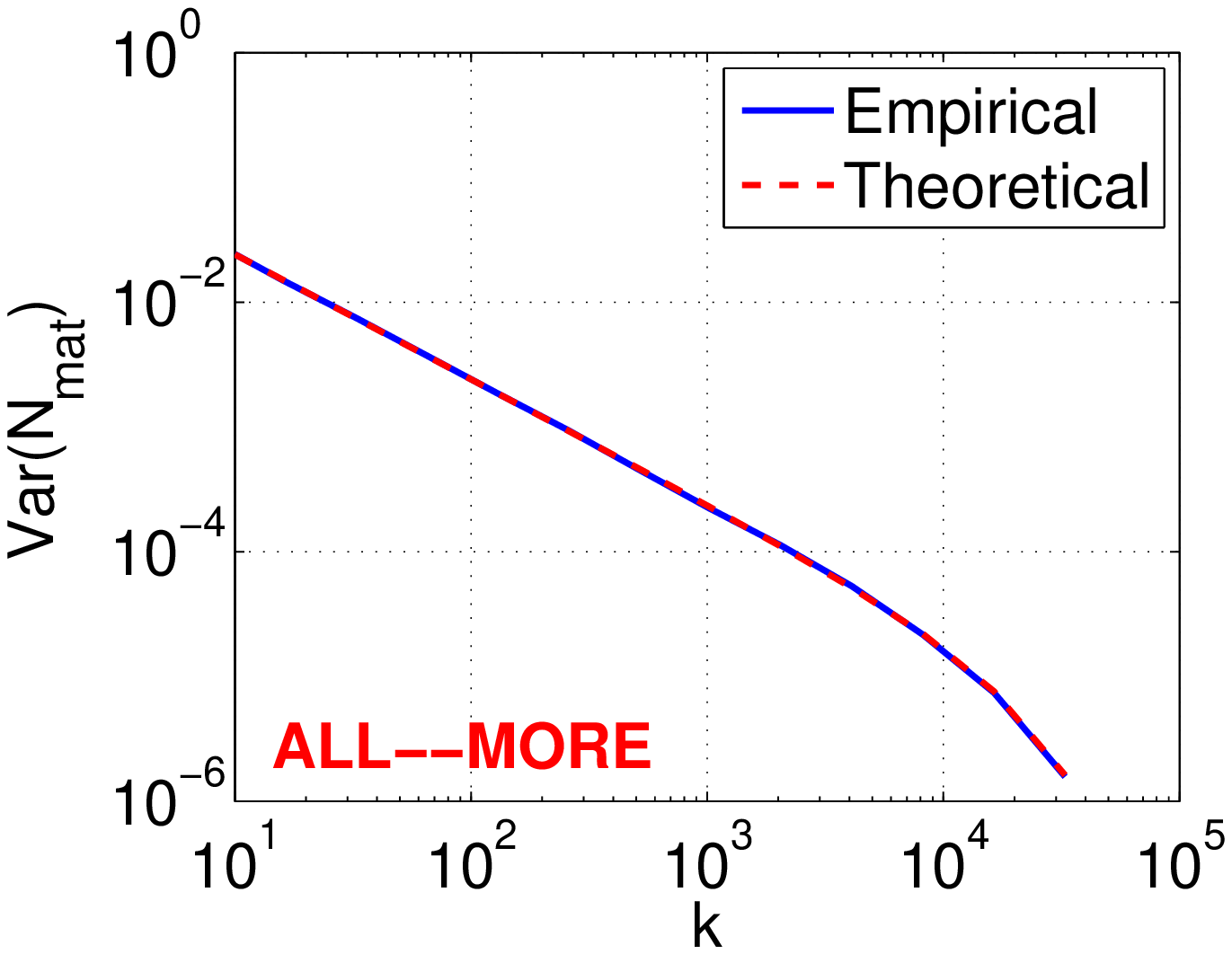}\hspace{-0.1in}
\includegraphics[width=1.5in]{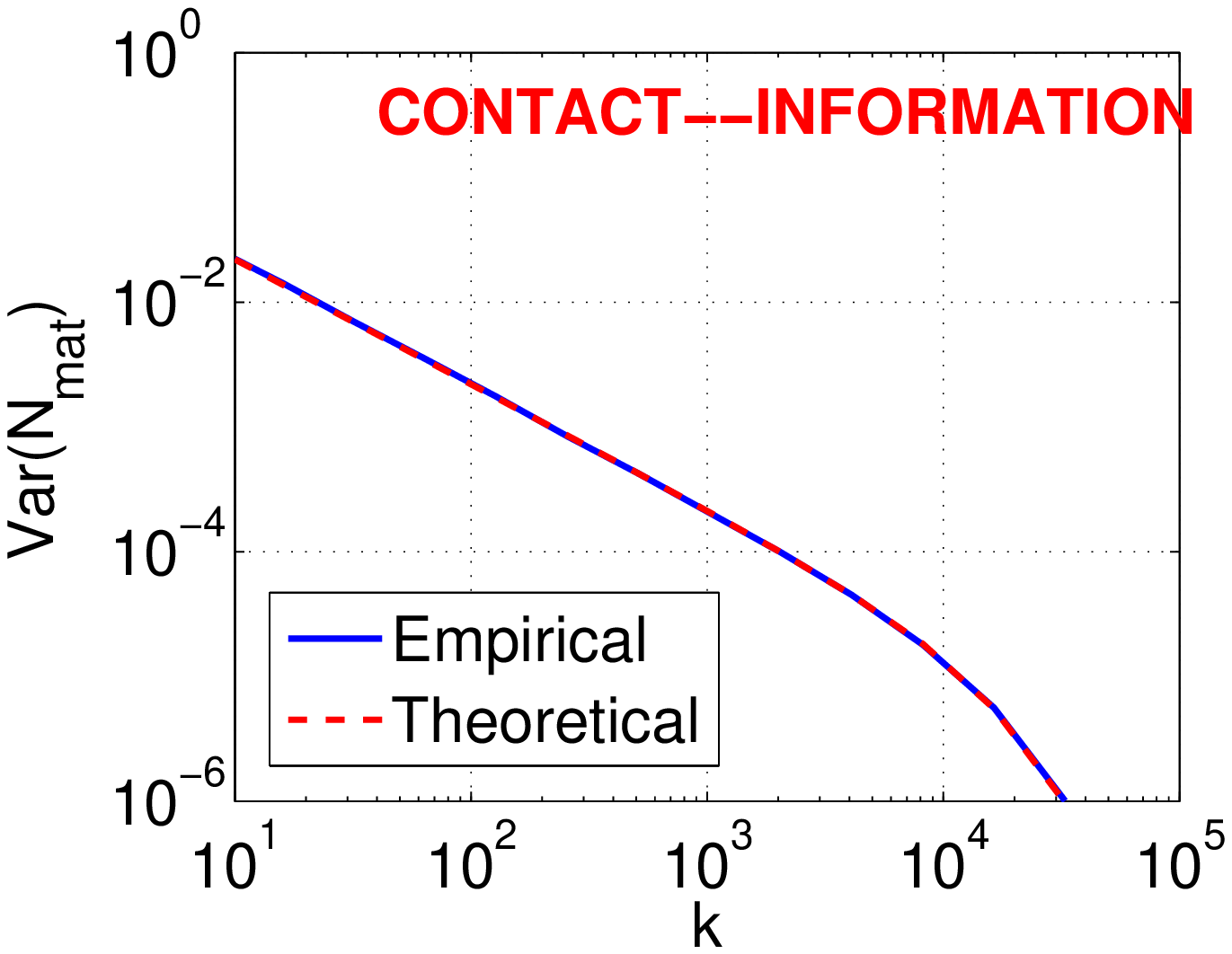}
}
\mbox{
\includegraphics[width=1.5in]{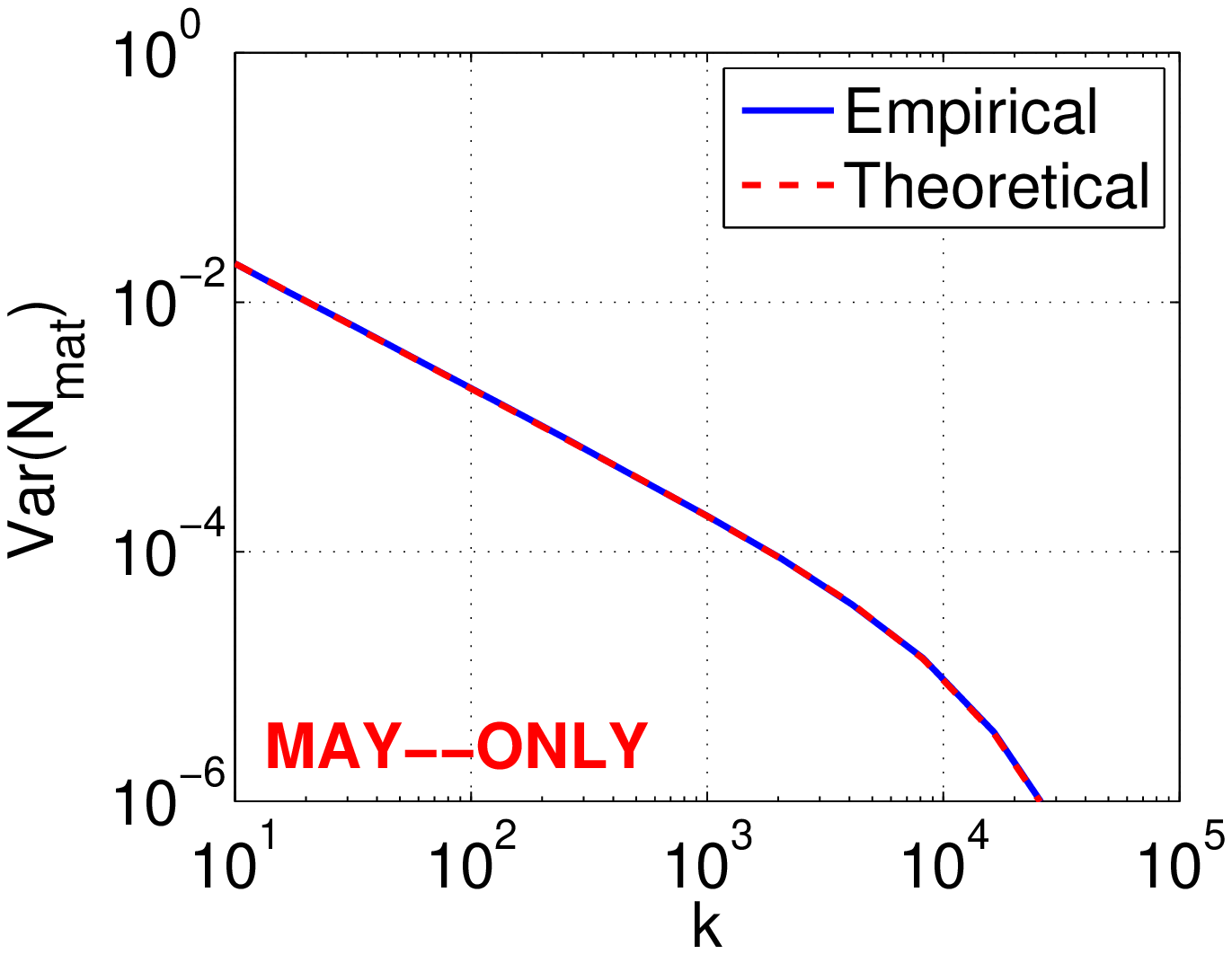}\hspace{-0.1in}
\includegraphics[width=1.5in]{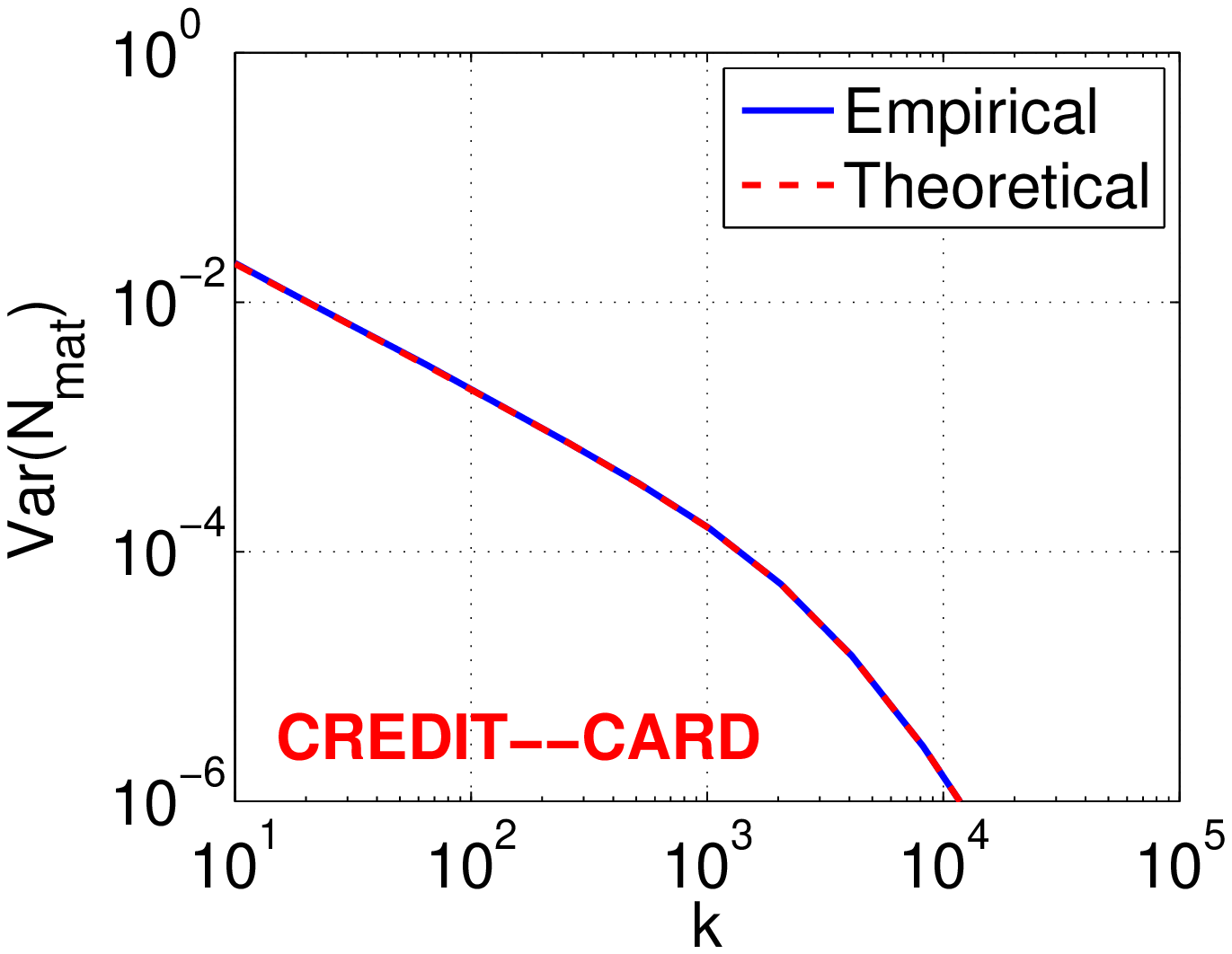}\hspace{-0.1in}
\includegraphics[width=1.5in]{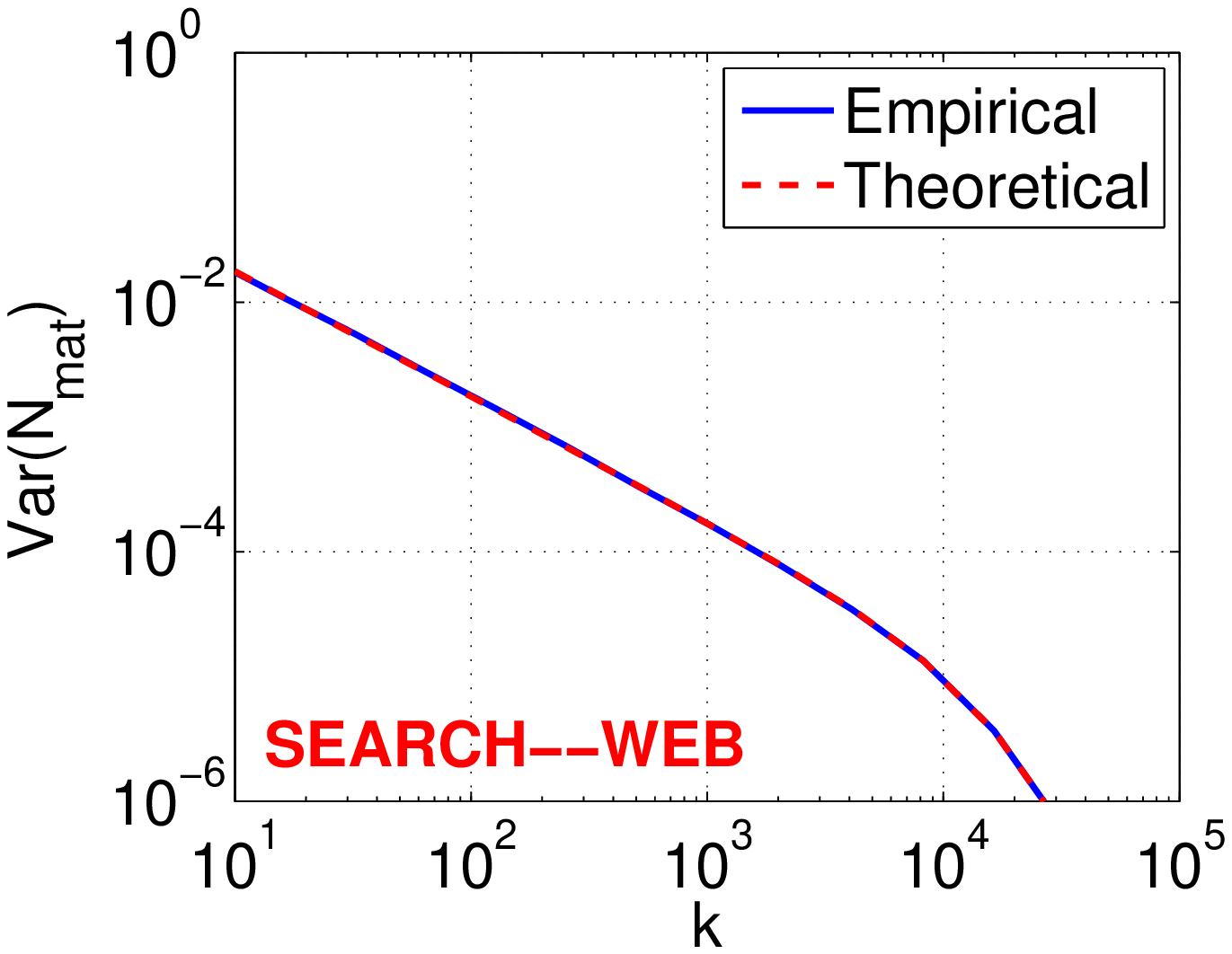}\hspace{-0.1in}
\includegraphics[width=1.5in]{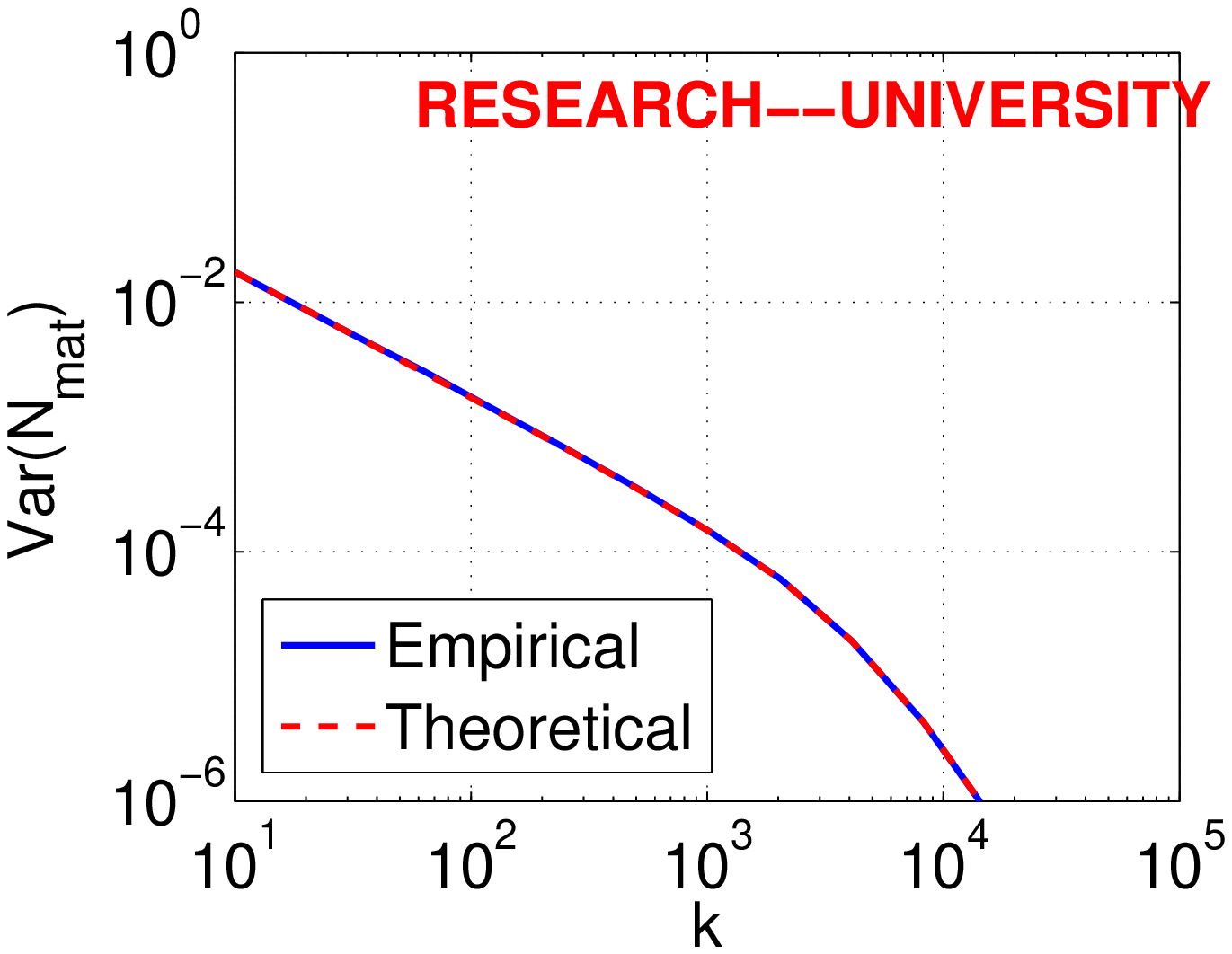}\hspace{-0.1in}
\includegraphics[width=1.5in]{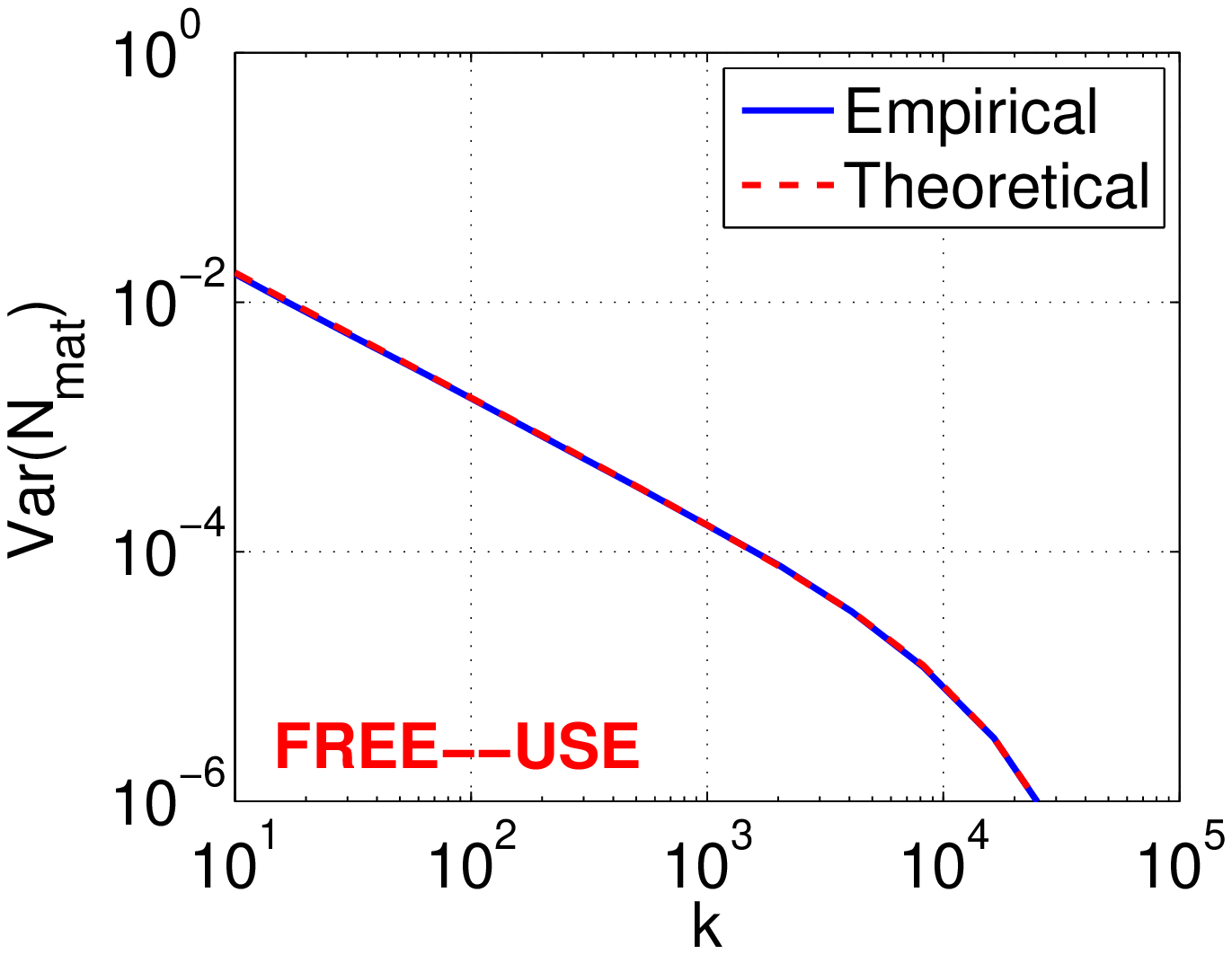}
}
\mbox{
\includegraphics[width=1.5in]{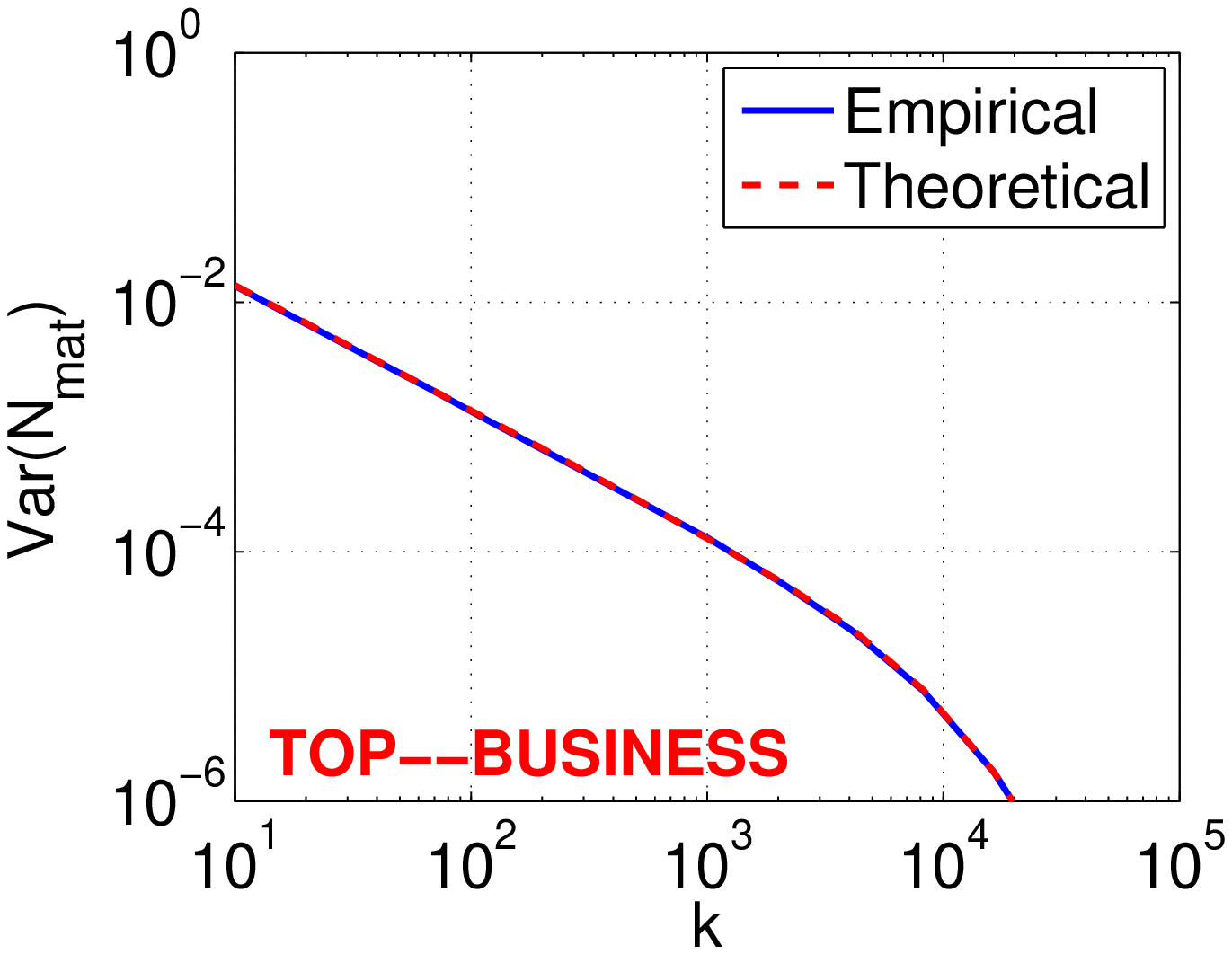}\hspace{-0.1in}
\includegraphics[width=1.5in]{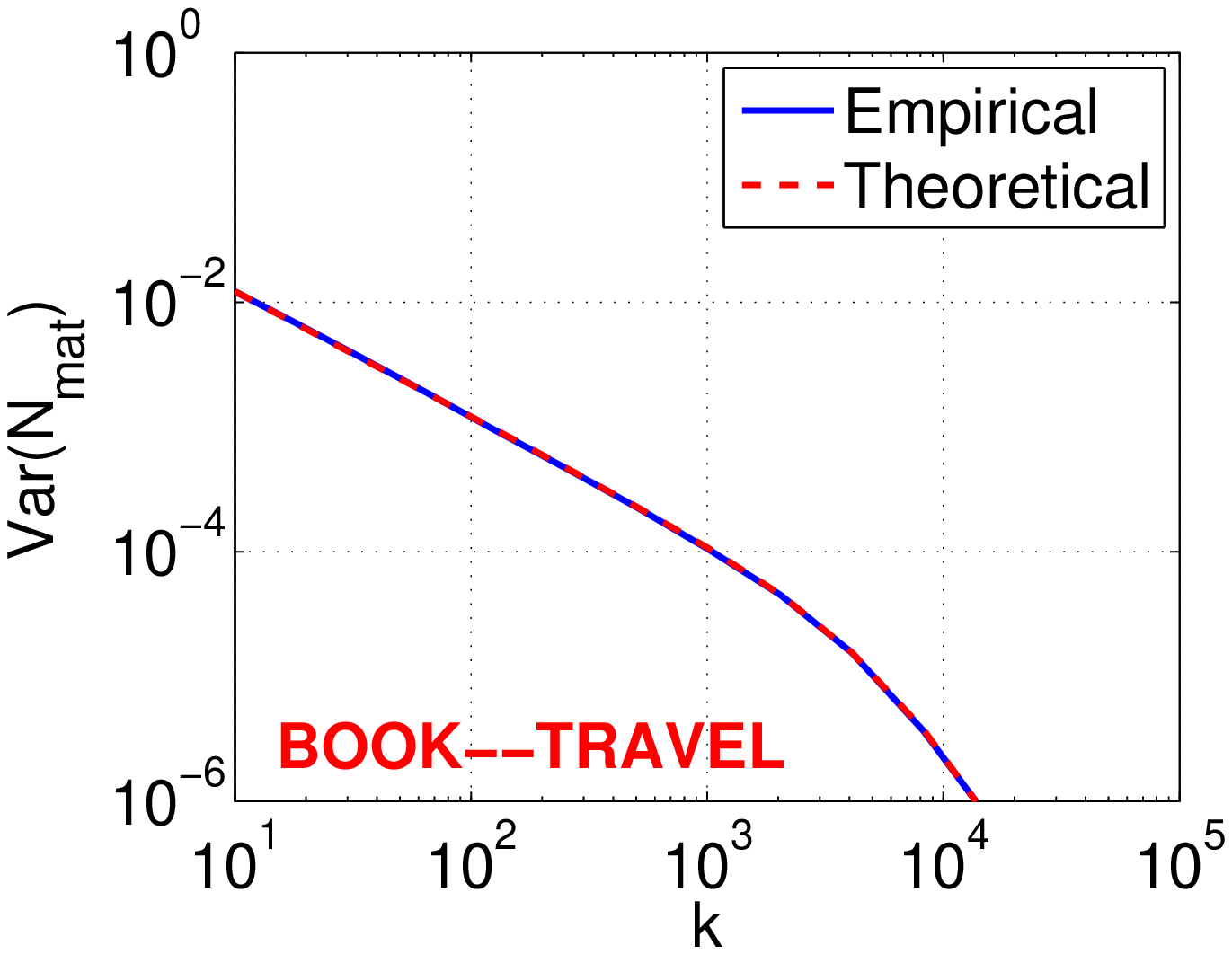}\hspace{-0.1in}
\includegraphics[width=1.5in]{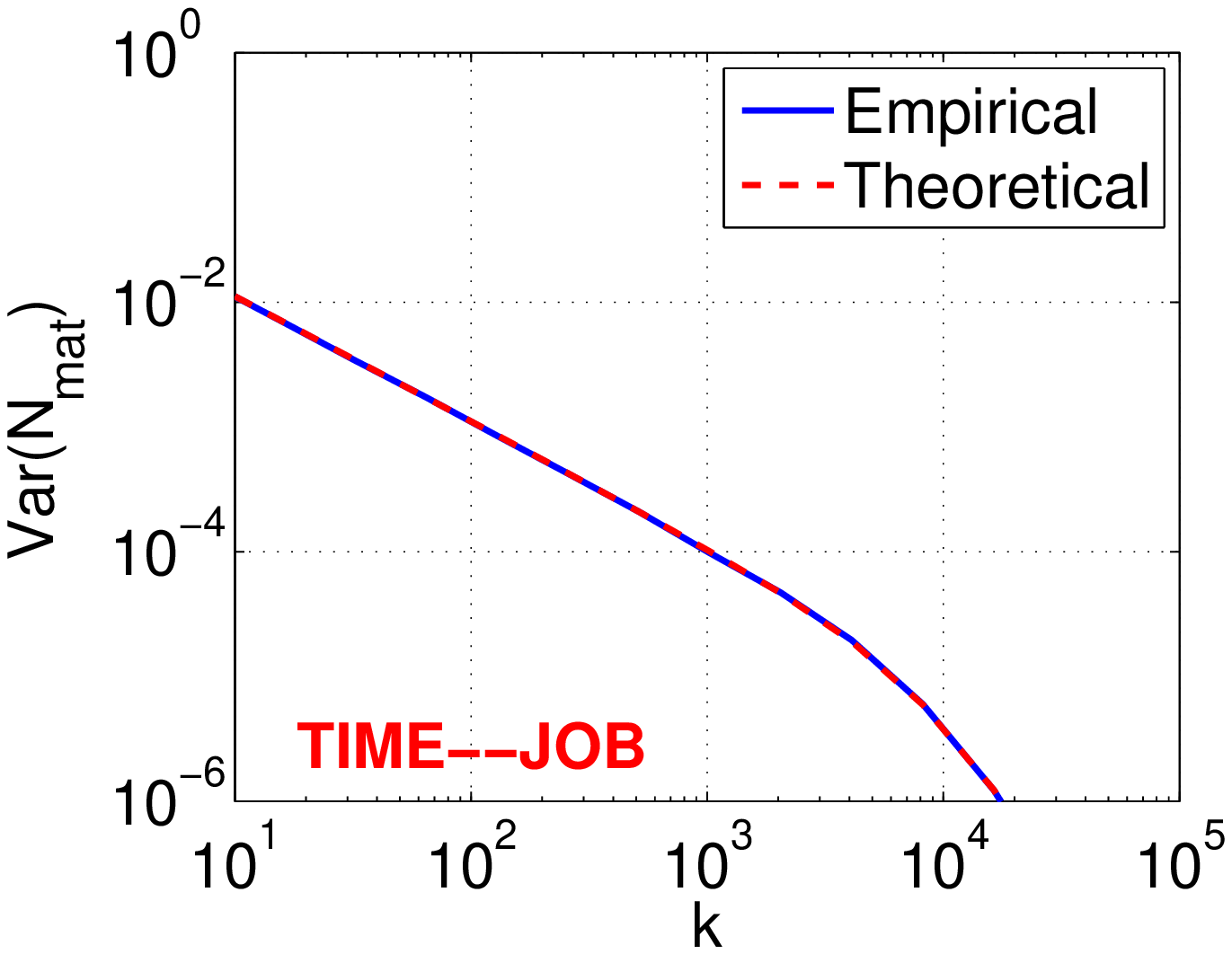}\hspace{-0.1in}
\includegraphics[width=1.5in]{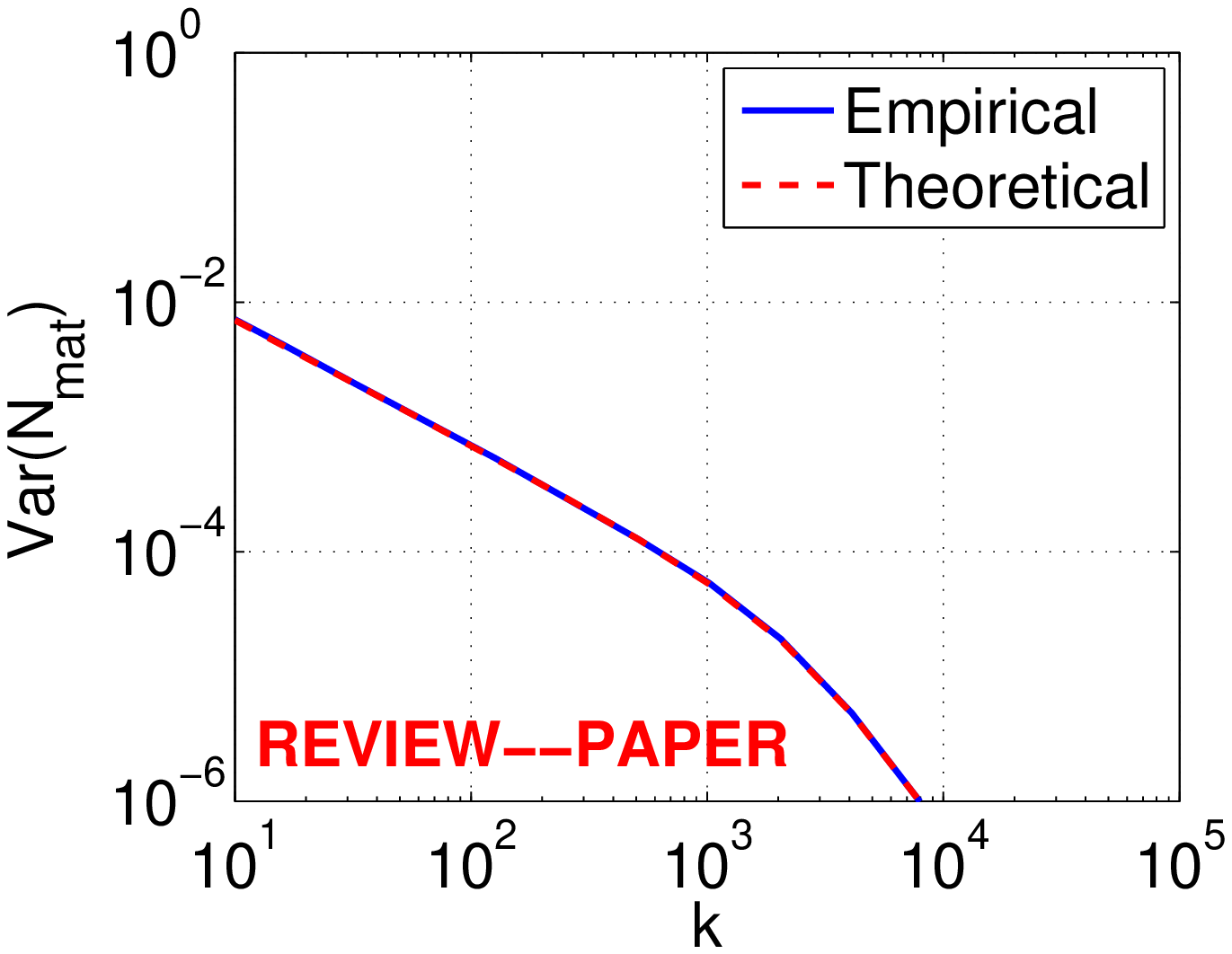}\hspace{-0.1in}
\includegraphics[width=1.5in]{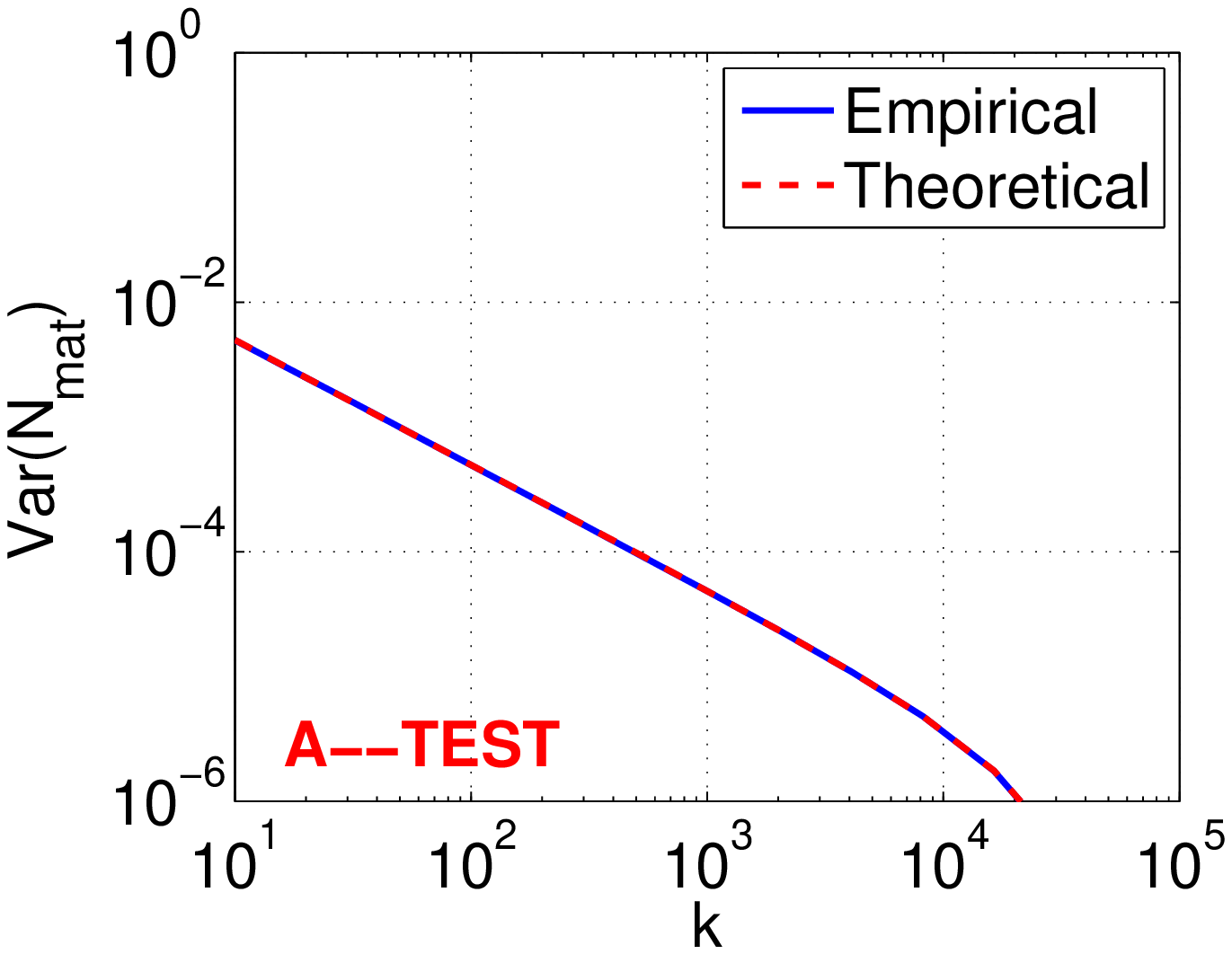}
}

\end{center}\vspace{-0.3in}
\caption{$Var(N_{mat})/k^2$.  The empirical curves essentially overlap the theoretical curves as derived in Lemma~\ref{lem_Nmat}, i.e., (\ref{eqn_Nmat_var}). } \label{fig_Nmat_var}
\end{figure}

Figure~\ref{fig_Nmat_mean} and Figure~\ref{fig_Nmat_var} respectively verify $E(N_{mat})$ and $Var(N_{mat})$ as derived in Lemma~\ref{lem_Nmat}. Again, the theoretical curves match the empirical ones and the curves start to change shapes at the point where the occurrences of empty bins are more noticeable.

\newpage\clearpage

\subsubsection{$Cov(N_{emp}, N_{mat})$}

\begin{figure}[h!]
\begin{center}

\mbox{
\includegraphics[width=1.5in]{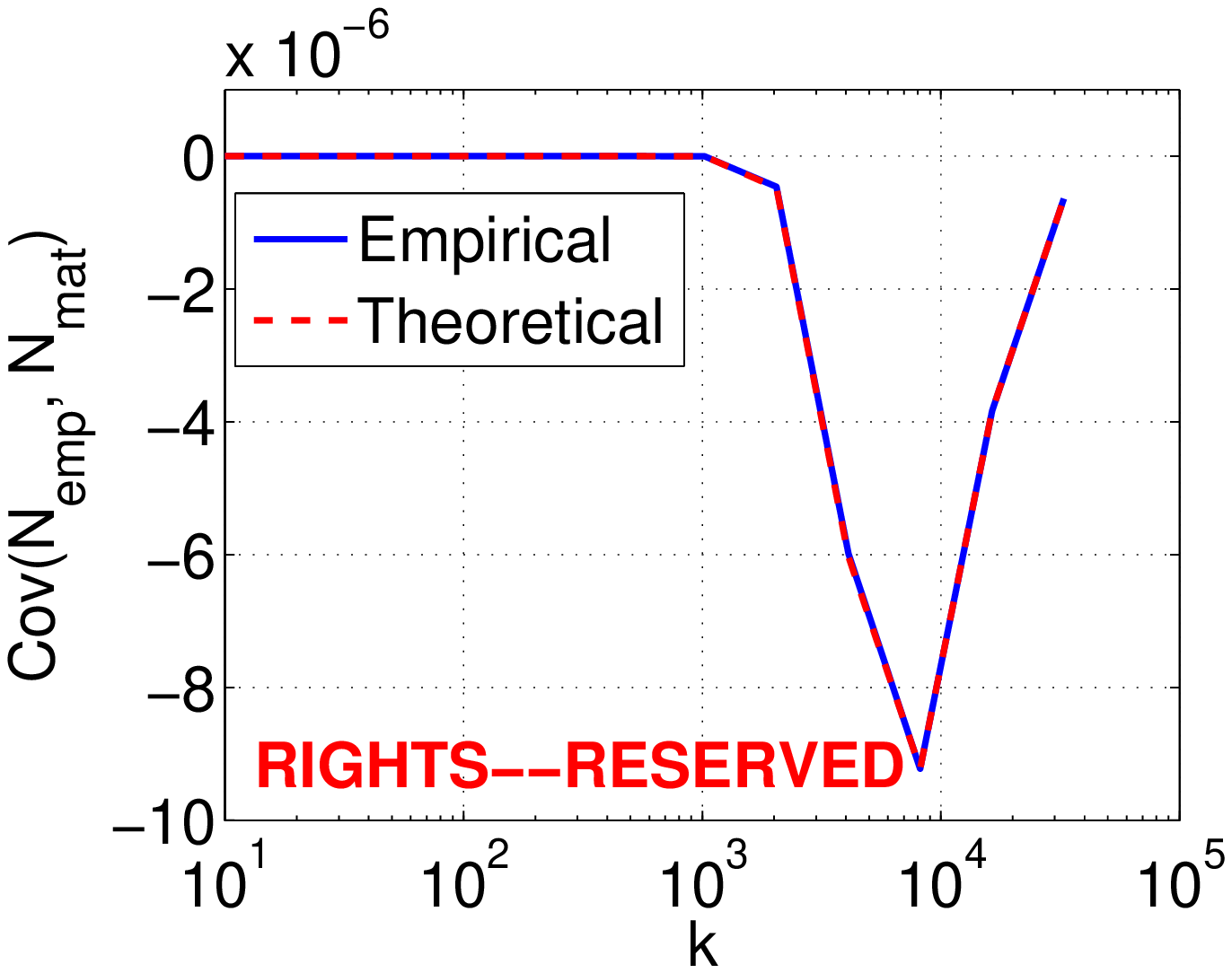}\hspace{-0.1in}
\includegraphics[width=1.5in]{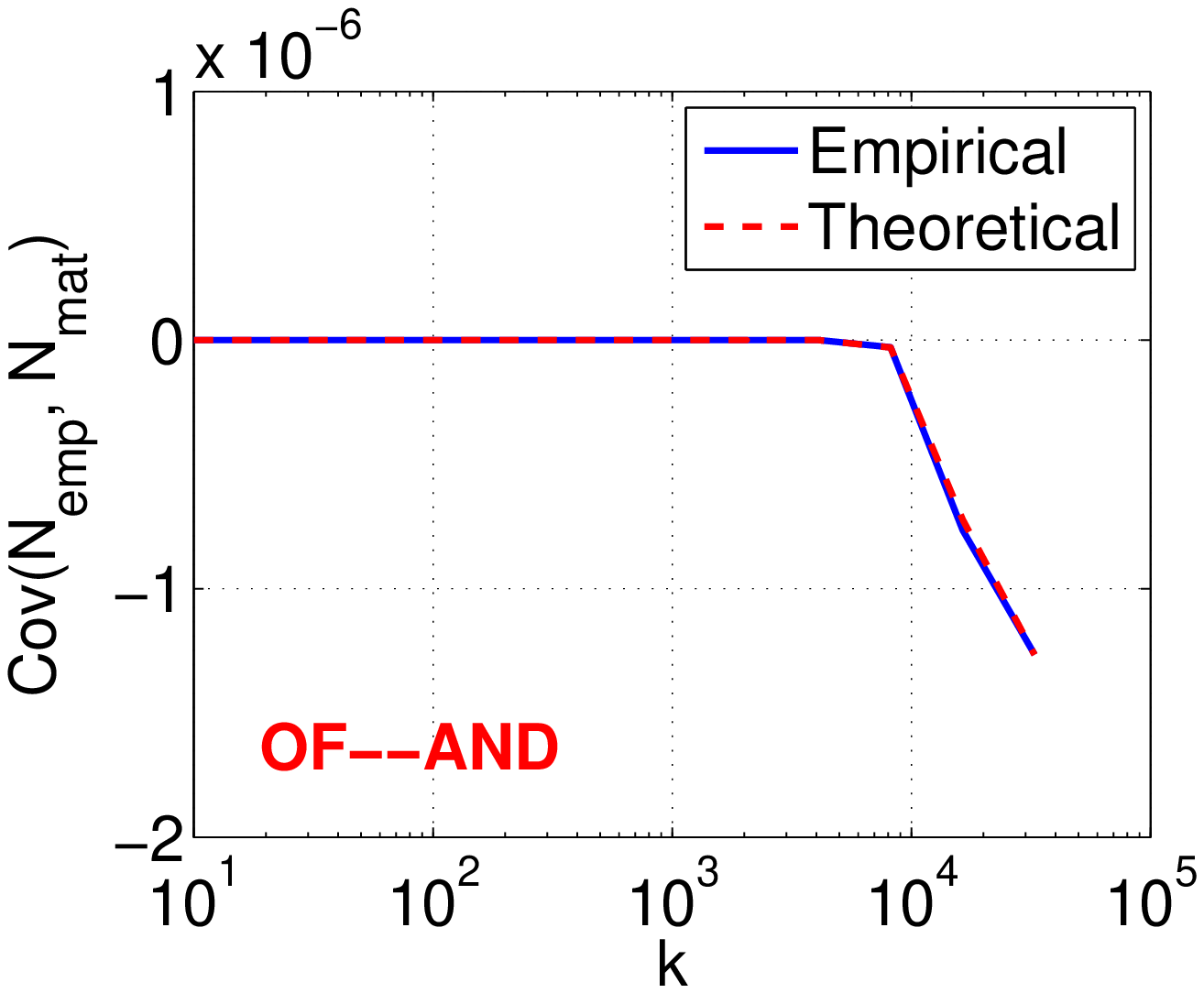}\hspace{-0.1in}
\includegraphics[width=1.5in]{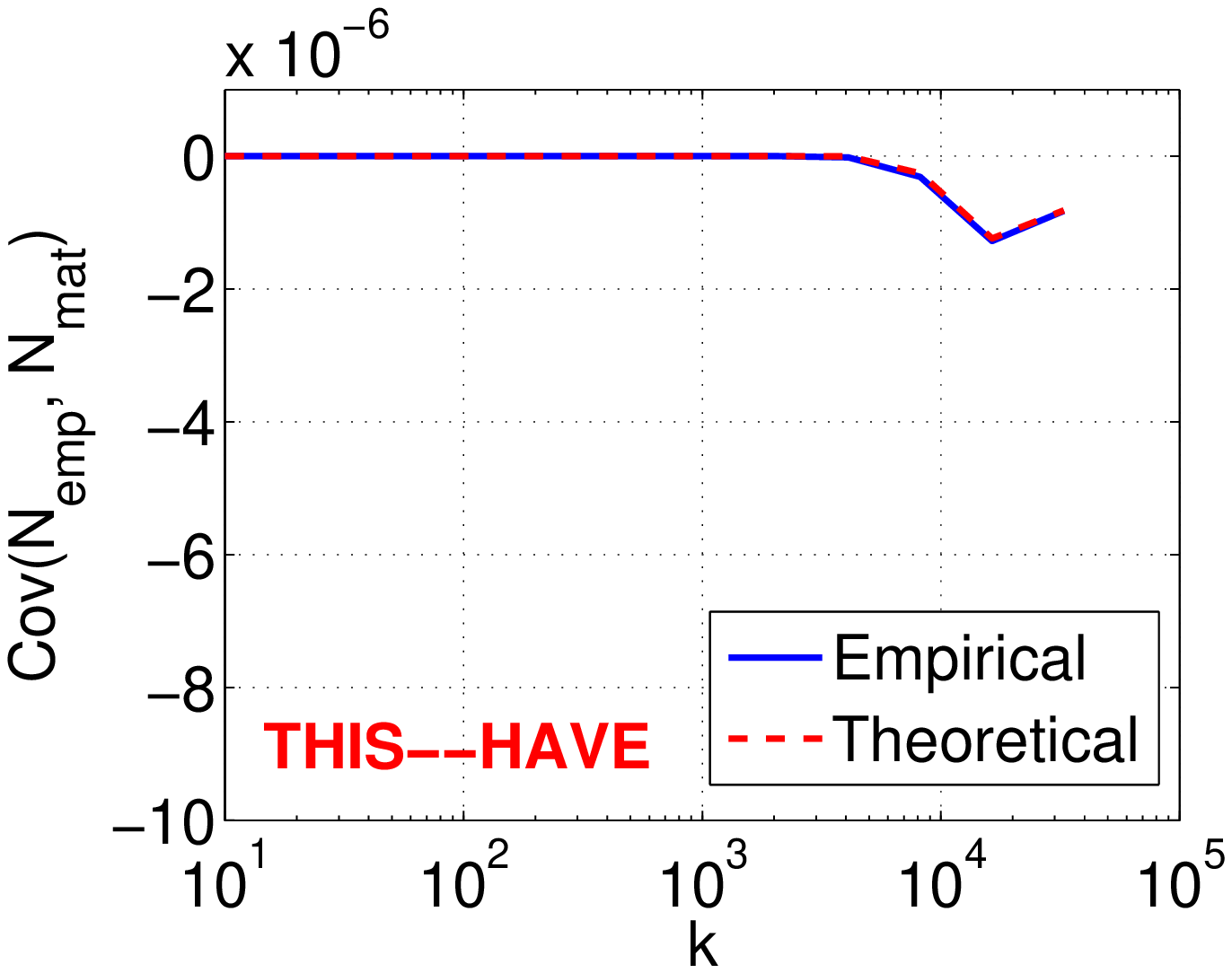}\hspace{-0.1in}
\includegraphics[width=1.5in]{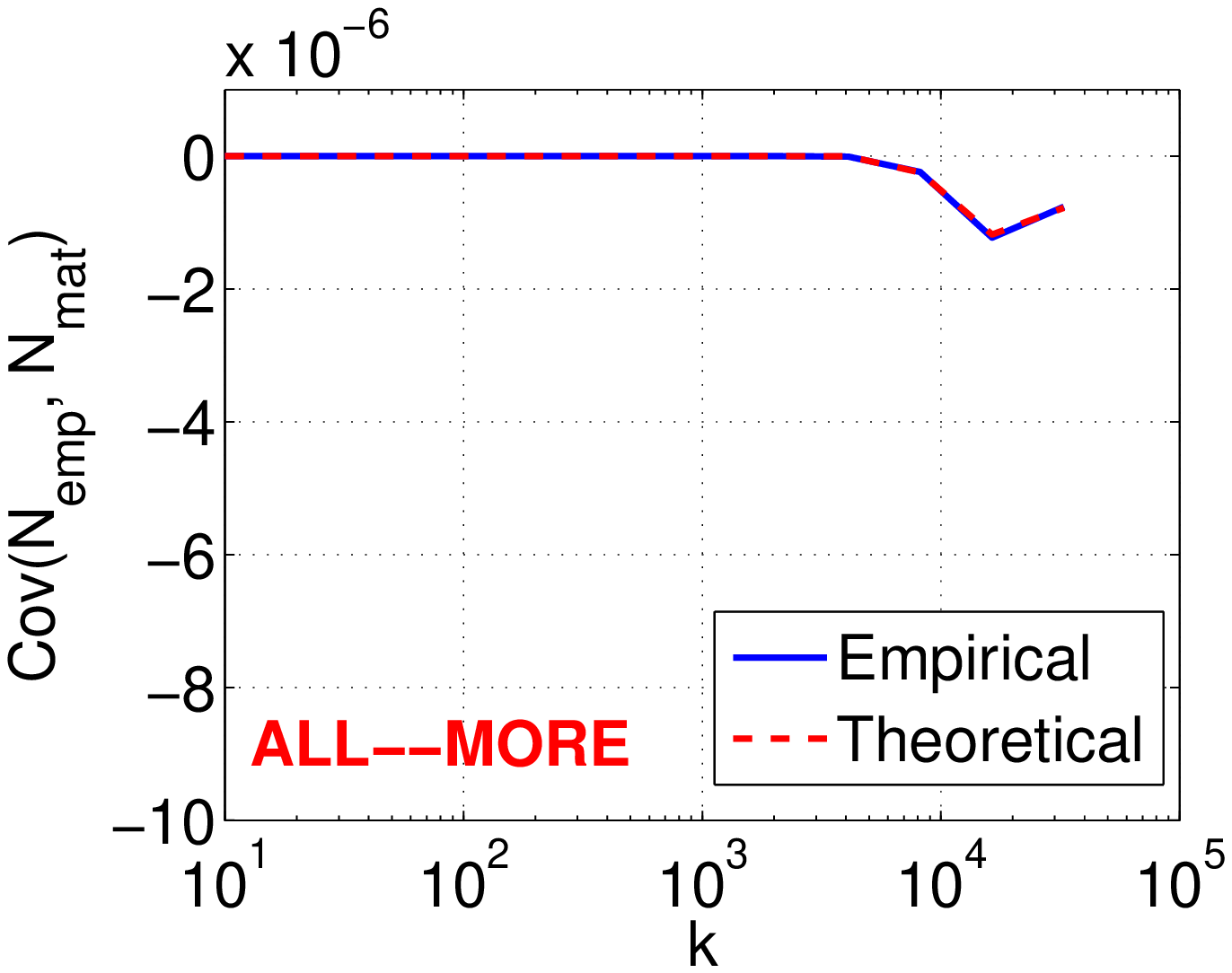}\hspace{-0.1in}
\includegraphics[width=1.5in]{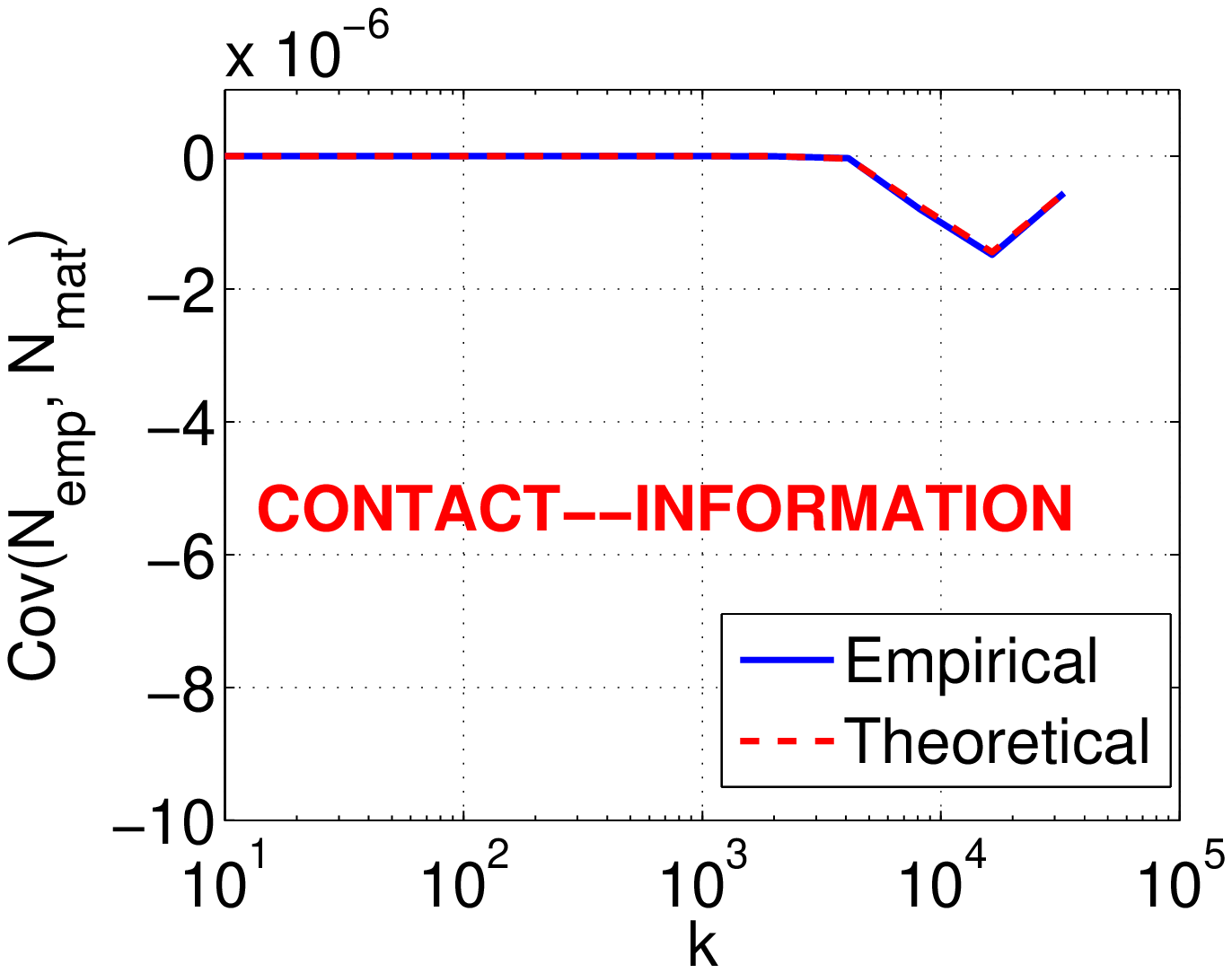}
}
\mbox{
\includegraphics[width=1.5in]{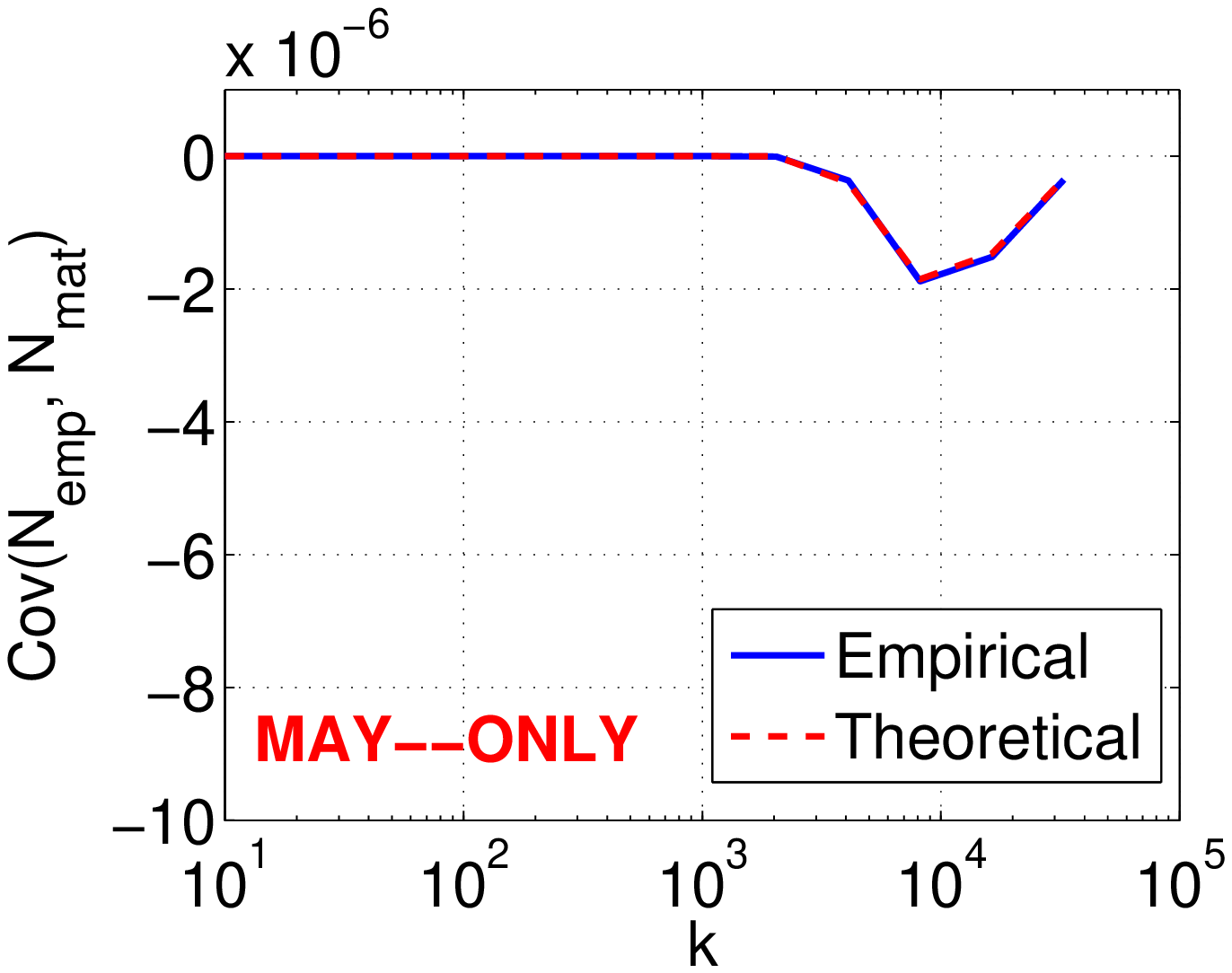}\hspace{-0.1in}
\includegraphics[width=1.5in]{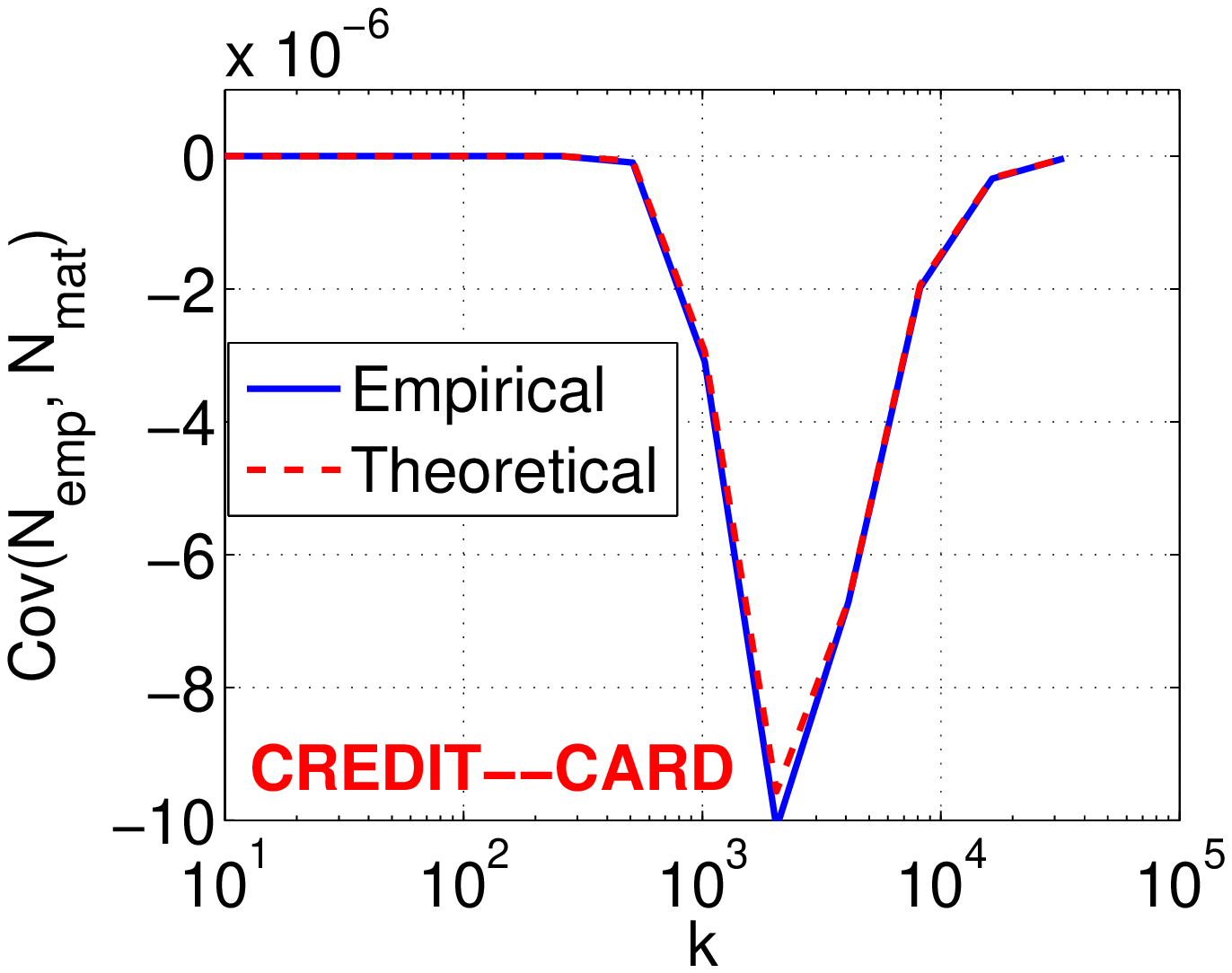}\hspace{-0.1in}
\includegraphics[width=1.5in]{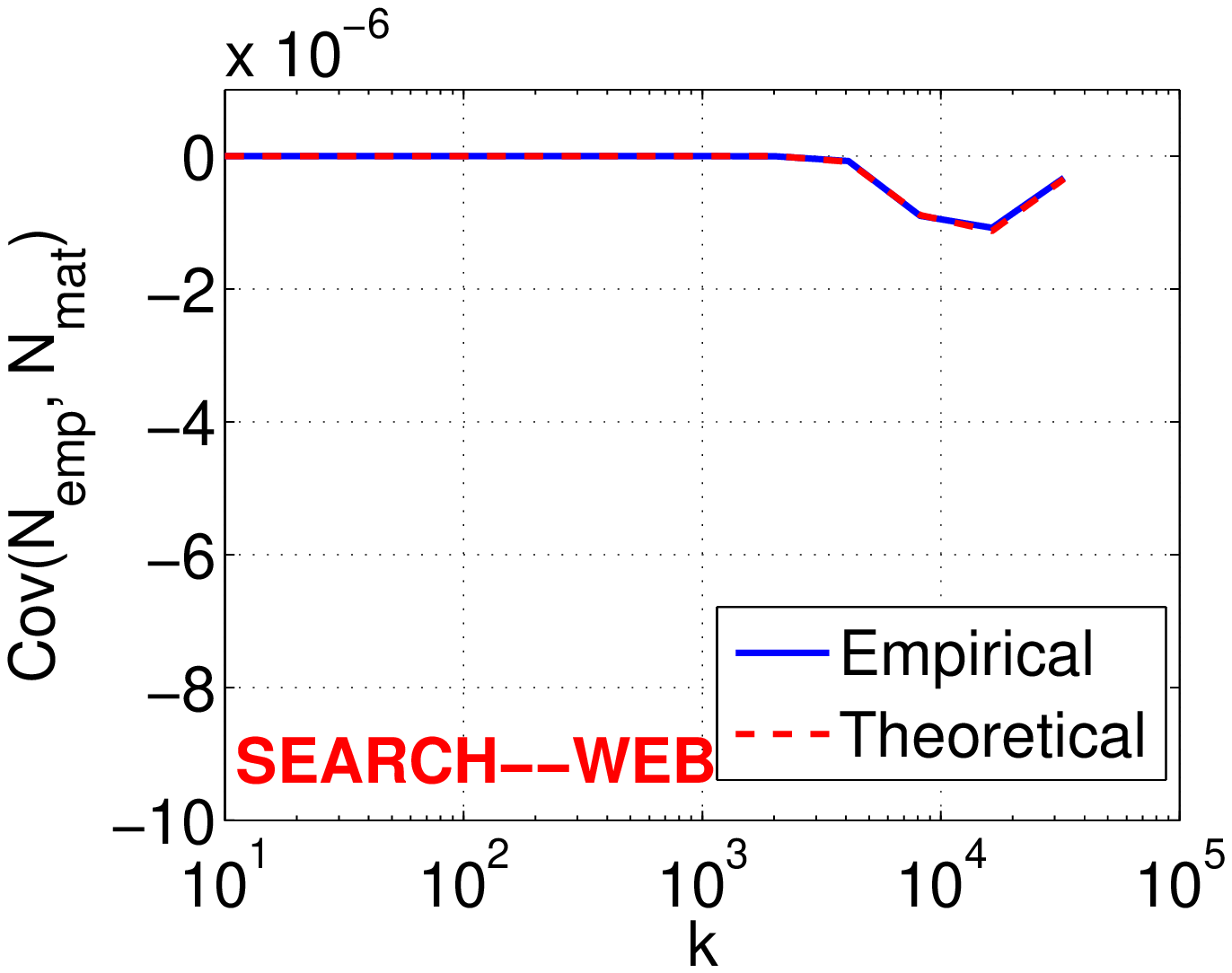}\hspace{-0.1in}
\includegraphics[width=1.5in]{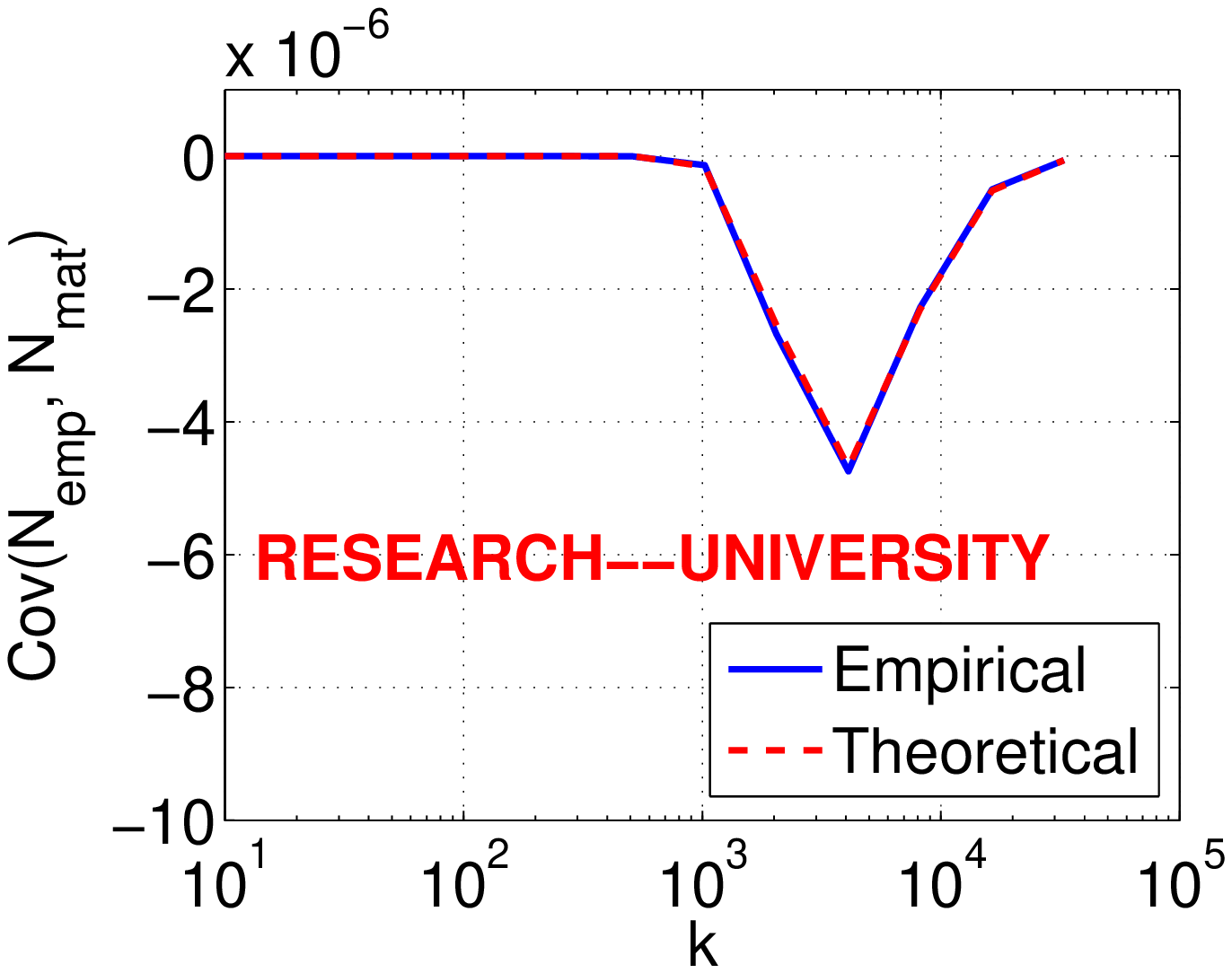}\hspace{-0.1in}
\includegraphics[width=1.5in]{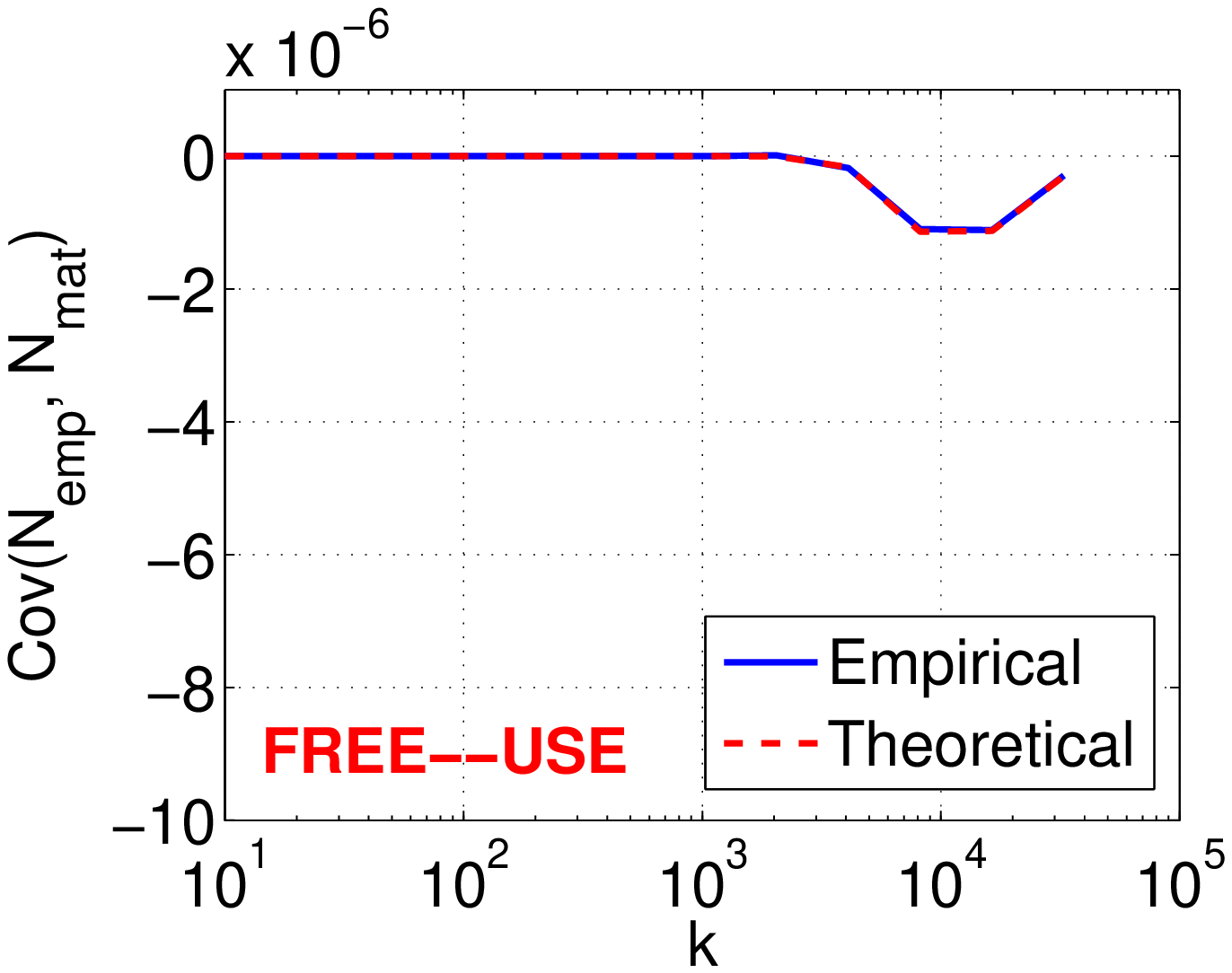}
}
\mbox{
\includegraphics[width=1.5in]{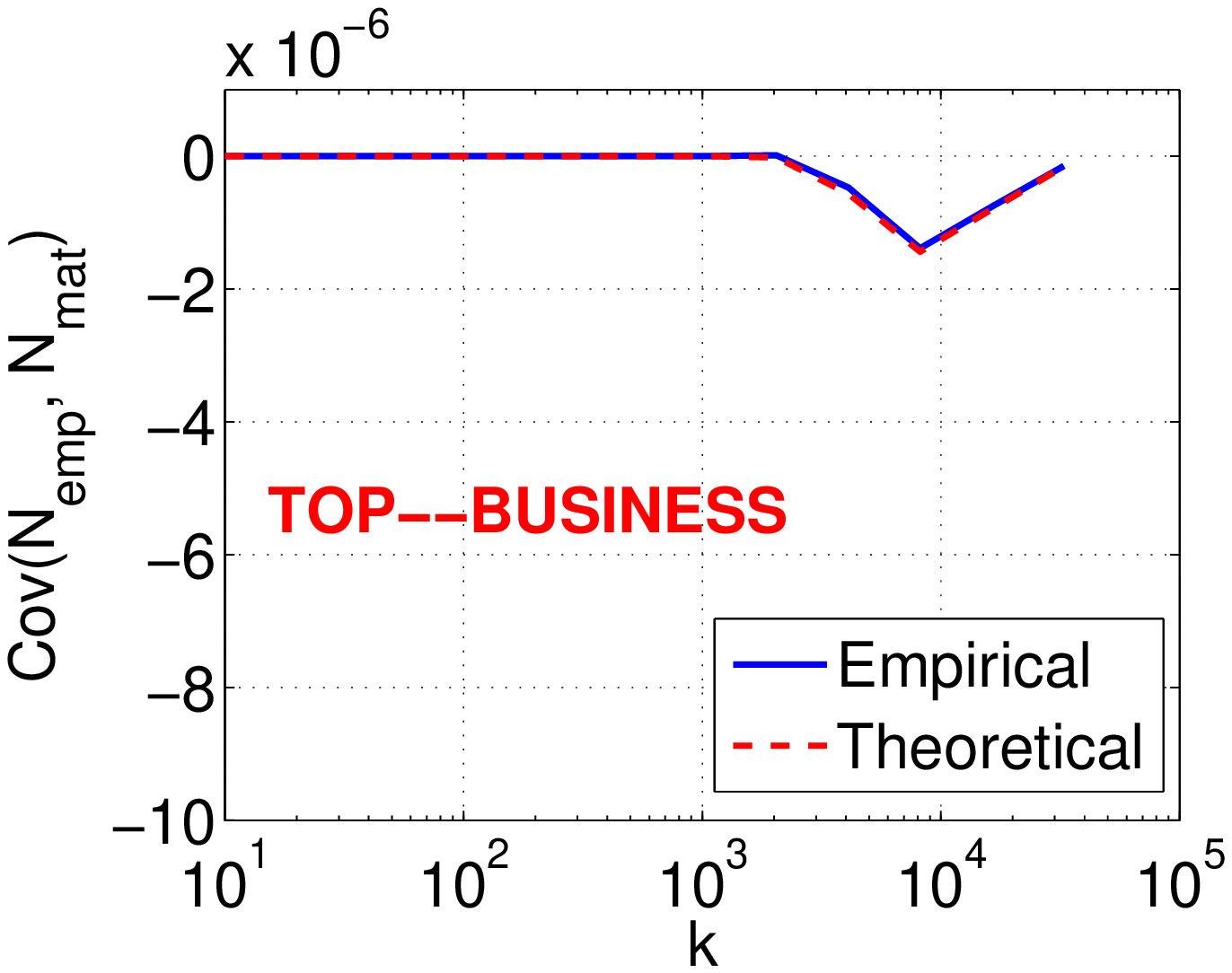}\hspace{-0.1in}
\includegraphics[width=1.5in]{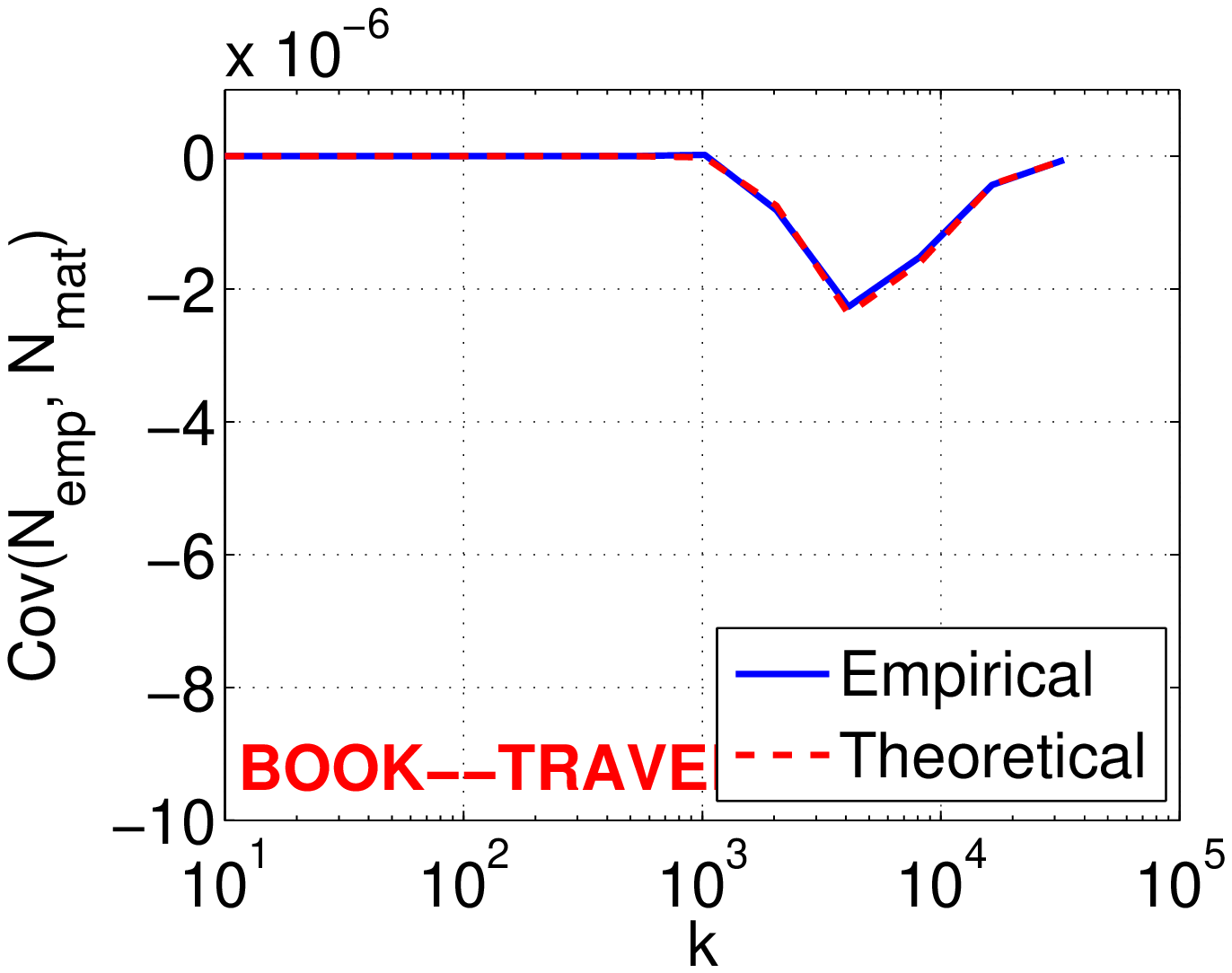}\hspace{-0.1in}
\includegraphics[width=1.5in]{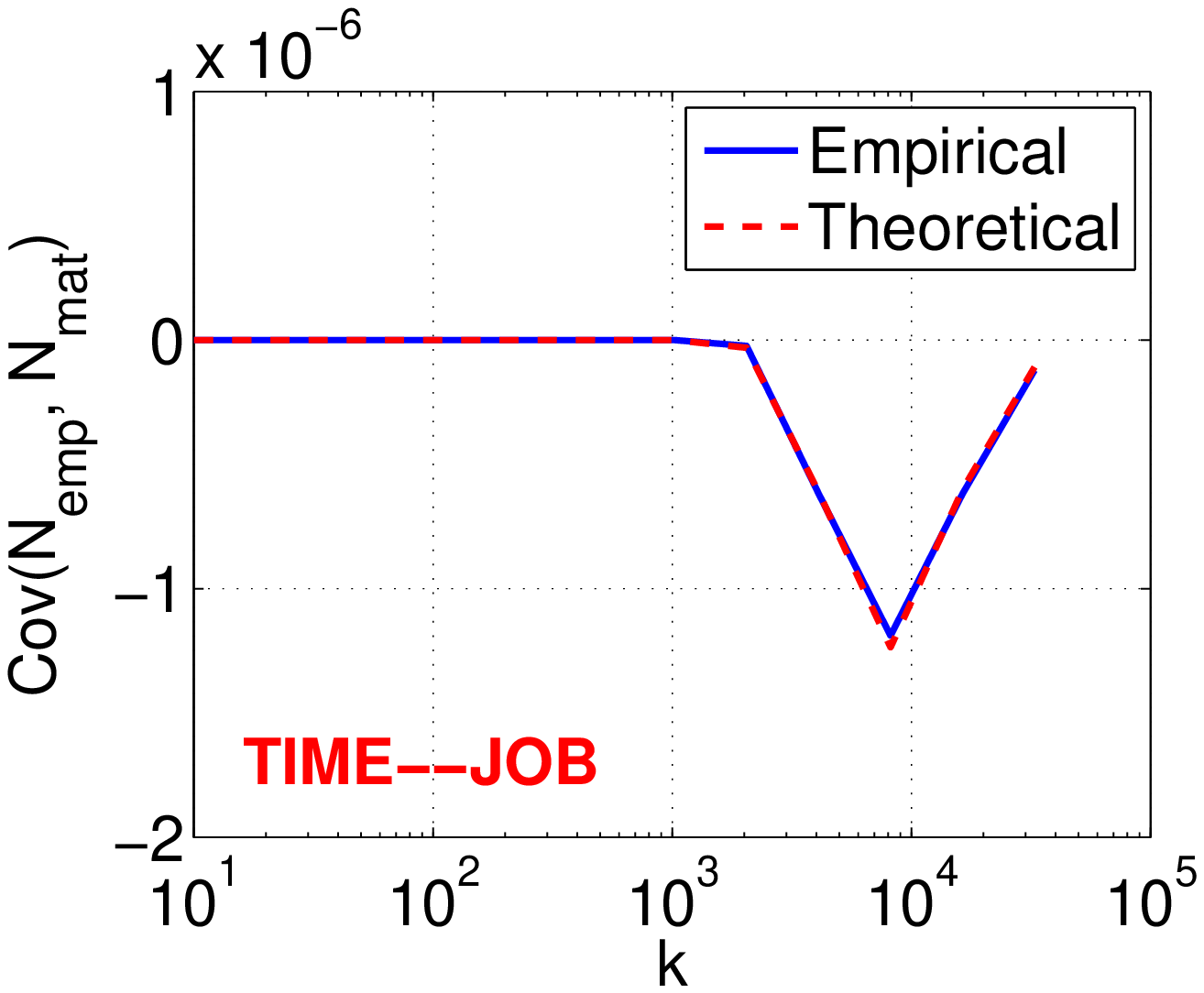}\hspace{-0.1in}
\includegraphics[width=1.5in]{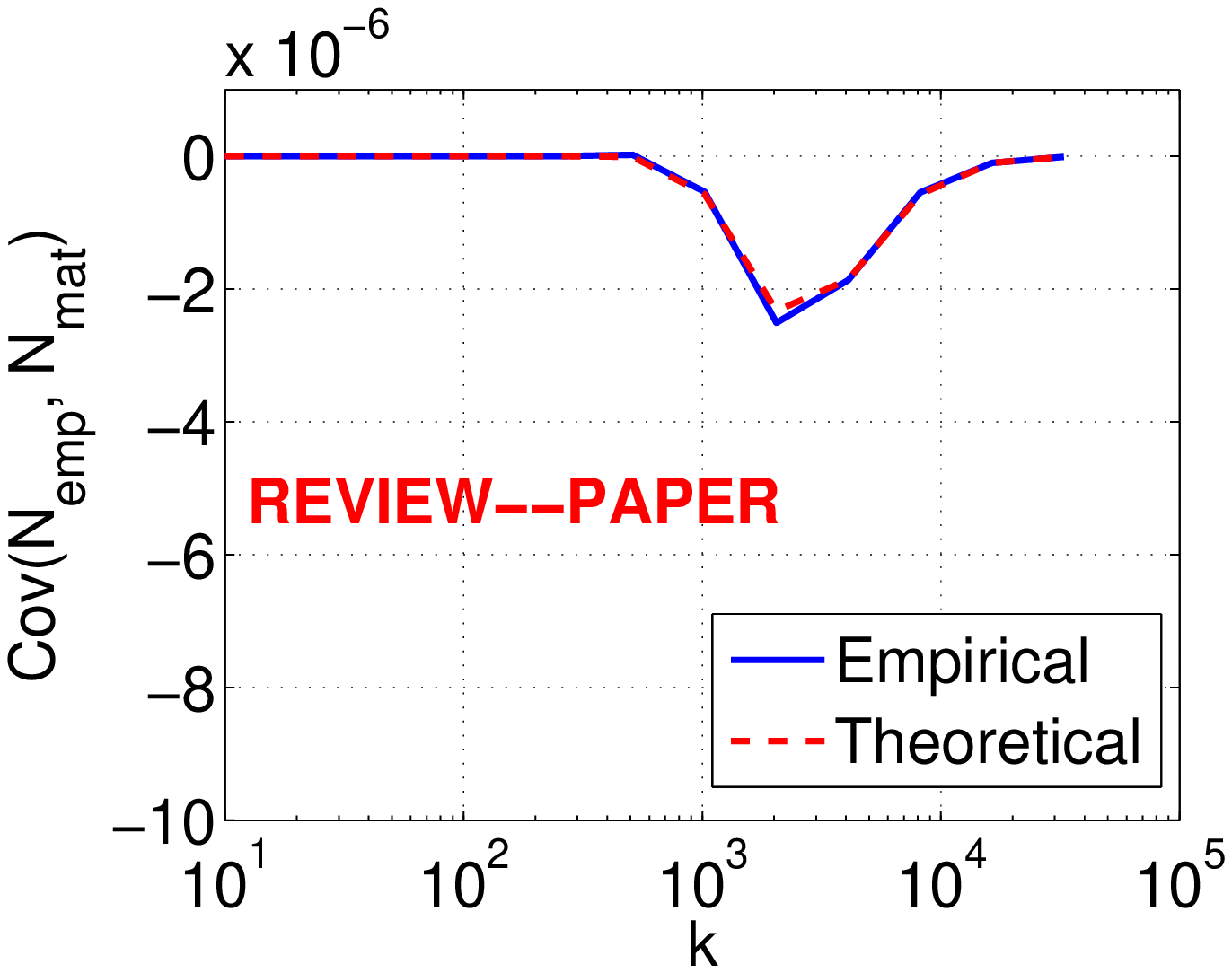}\hspace{-0.1in}
\includegraphics[width=1.5in]{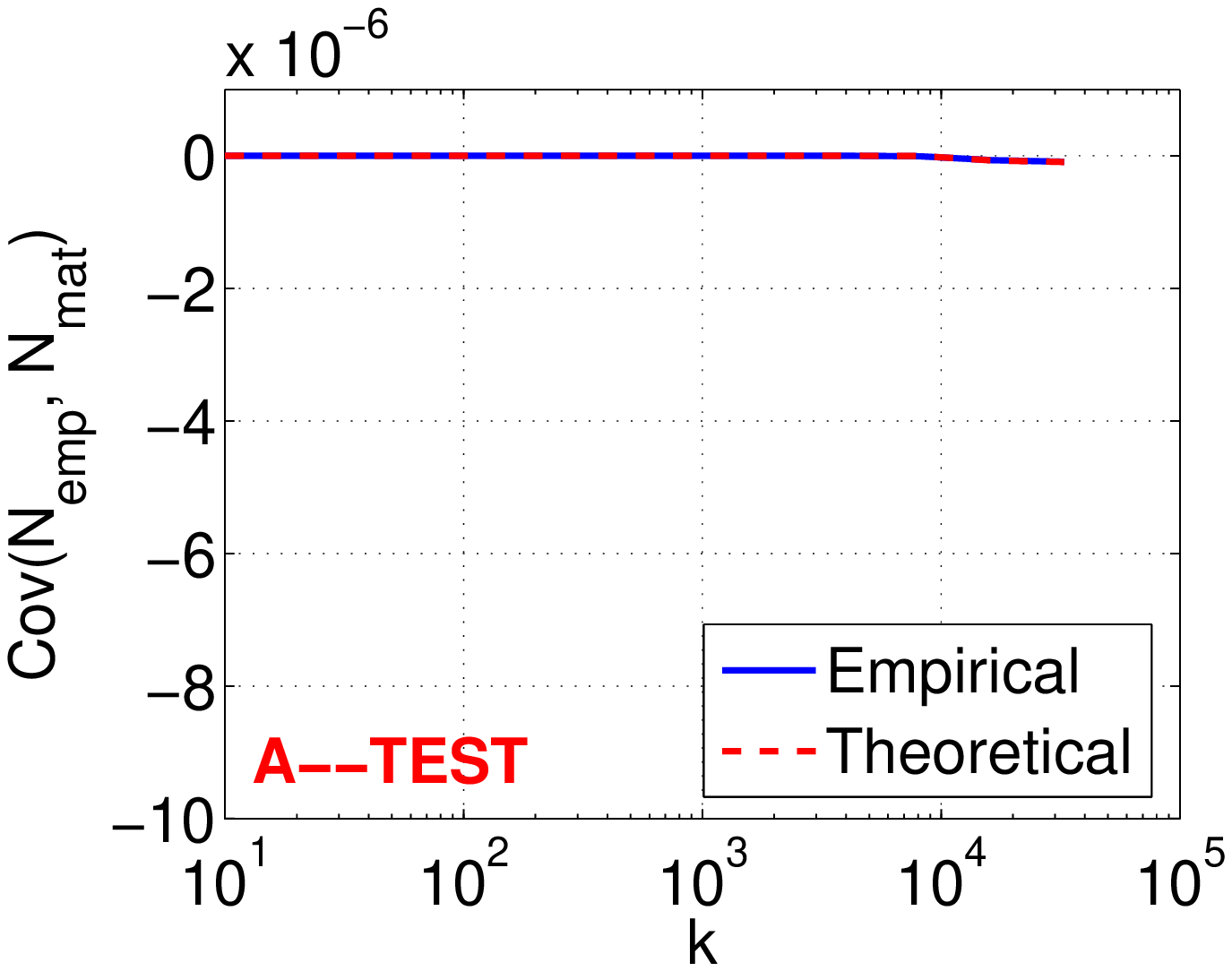}
}

\end{center}\vspace{-0.3in}
\caption{$Cov(N_{emp},N_{mat})/k^2$. The empirical curves essentially overlap the theoretical curves as derived in Lemma~\ref{lem_CovN}, i.e., (\ref{eqn_CovN}). The experimental results also confirm that  the covariance is  non-positive as theoretically shown in Lemma~\ref{lem_CovN}.}\label{fig_NempNmat_Cov}\vspace{-0.15in}
\end{figure}

To verify Lemma~\ref{lem_CovN}, Figure~\ref{fig_NempNmat_Cov} presents the theoretical and empirical covariances of $N_{emp}$ and $N_{mat}$. Note that $Cov\left(N_{emp},\ N_{mat}\right) \leq 0$ as shown in Lemma~\ref{lem_CovN}.

\subsubsection{$E(\hat{R}_{mat})$ and $Var(\hat{R}_{mat})$}

\begin{figure}[h!]
\begin{center}

\mbox{
\includegraphics[width=1.5in]{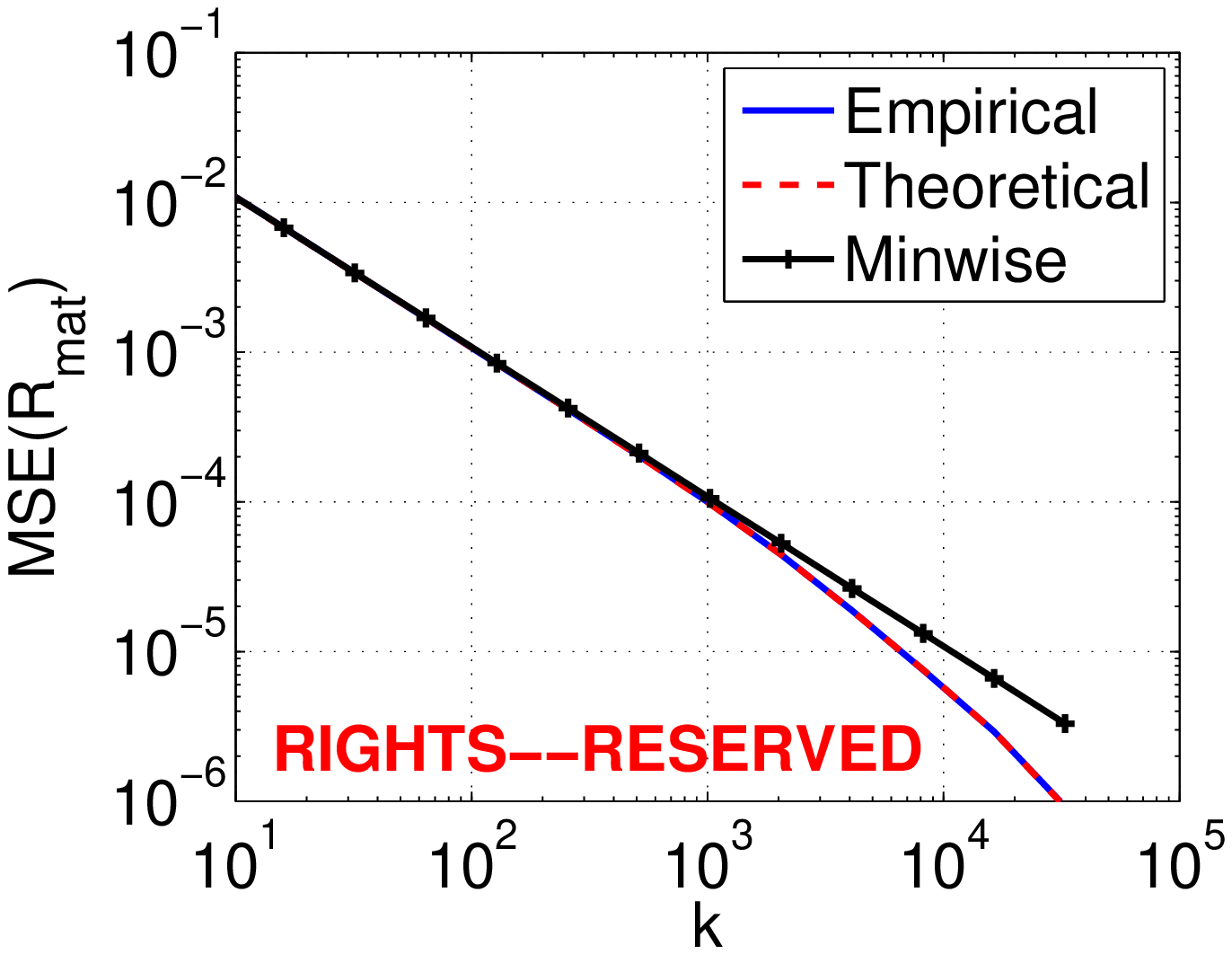}\hspace{-0.1in}
\includegraphics[width=1.5in]{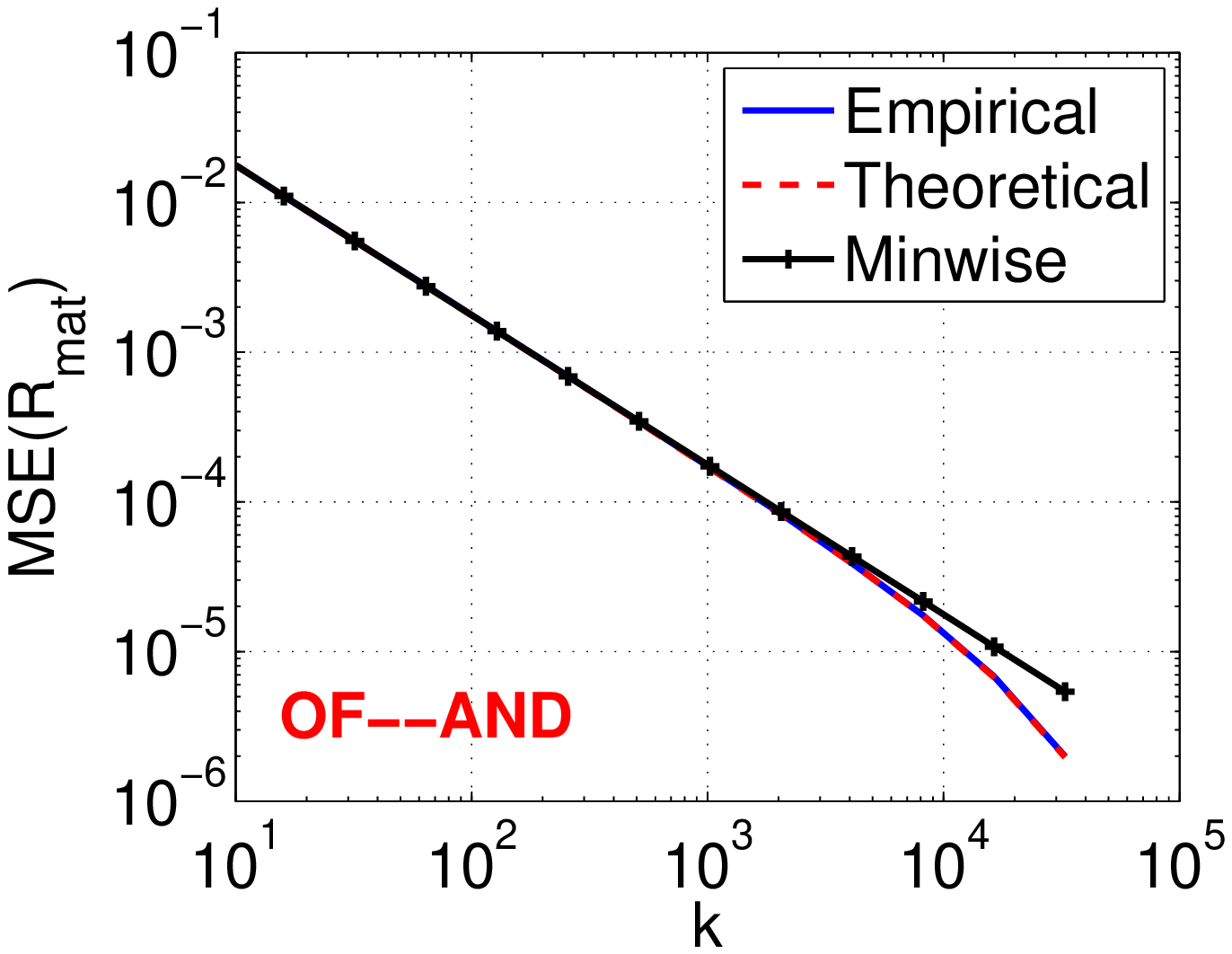}\hspace{-0.1in}
\includegraphics[width=1.5in]{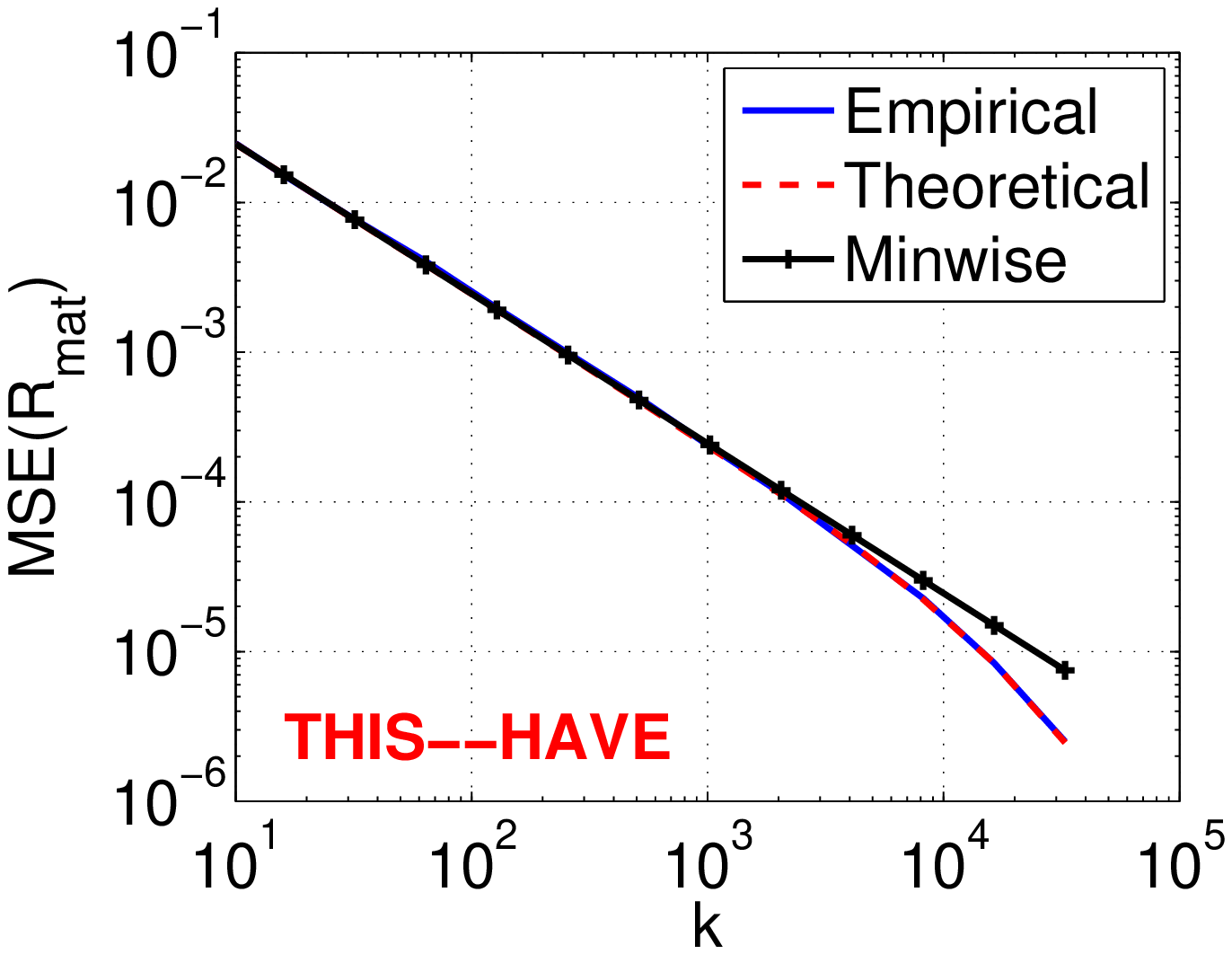}\hspace{-0.1in}
\includegraphics[width=1.5in]{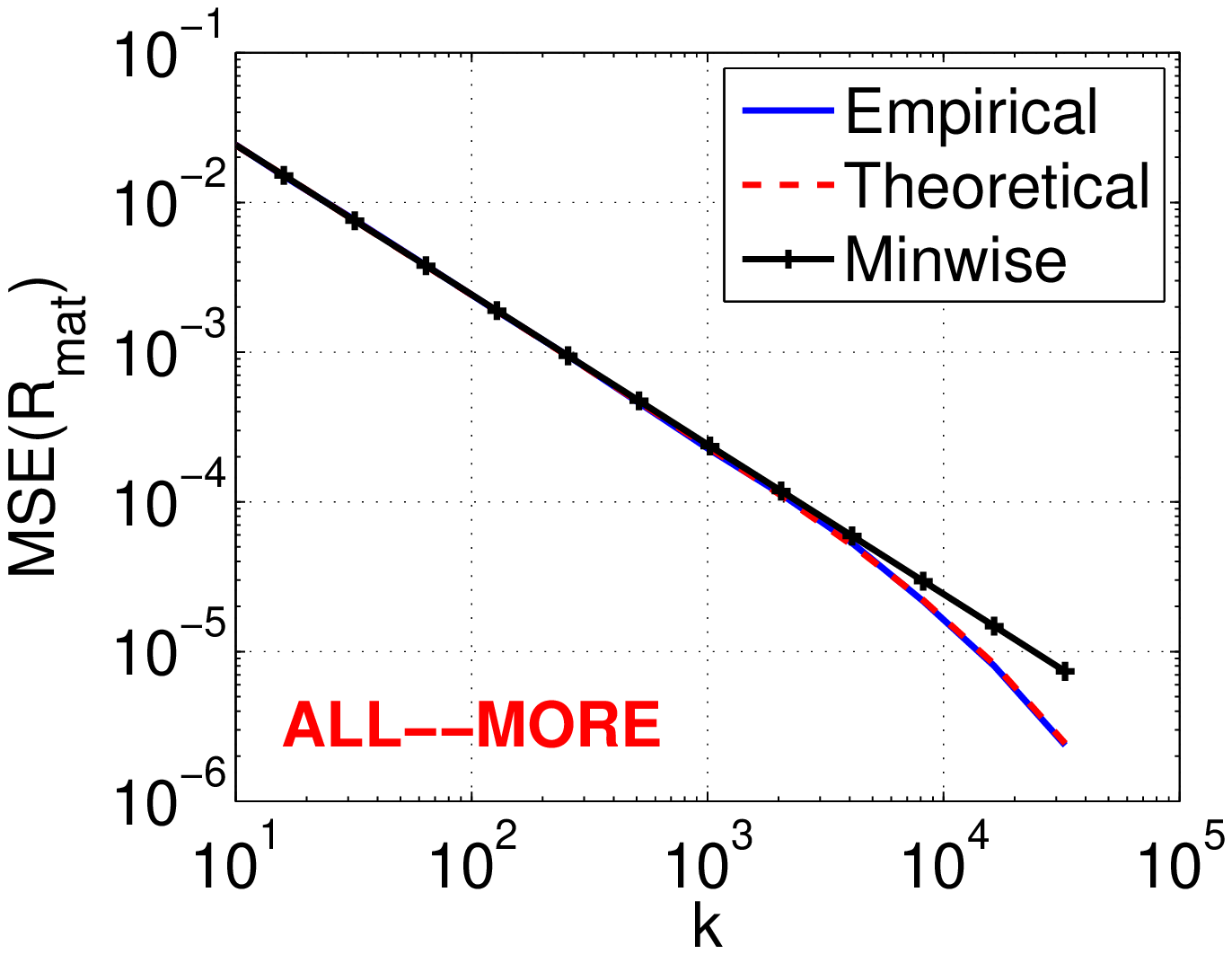}\hspace{-0.1in}
\includegraphics[width=1.5in]{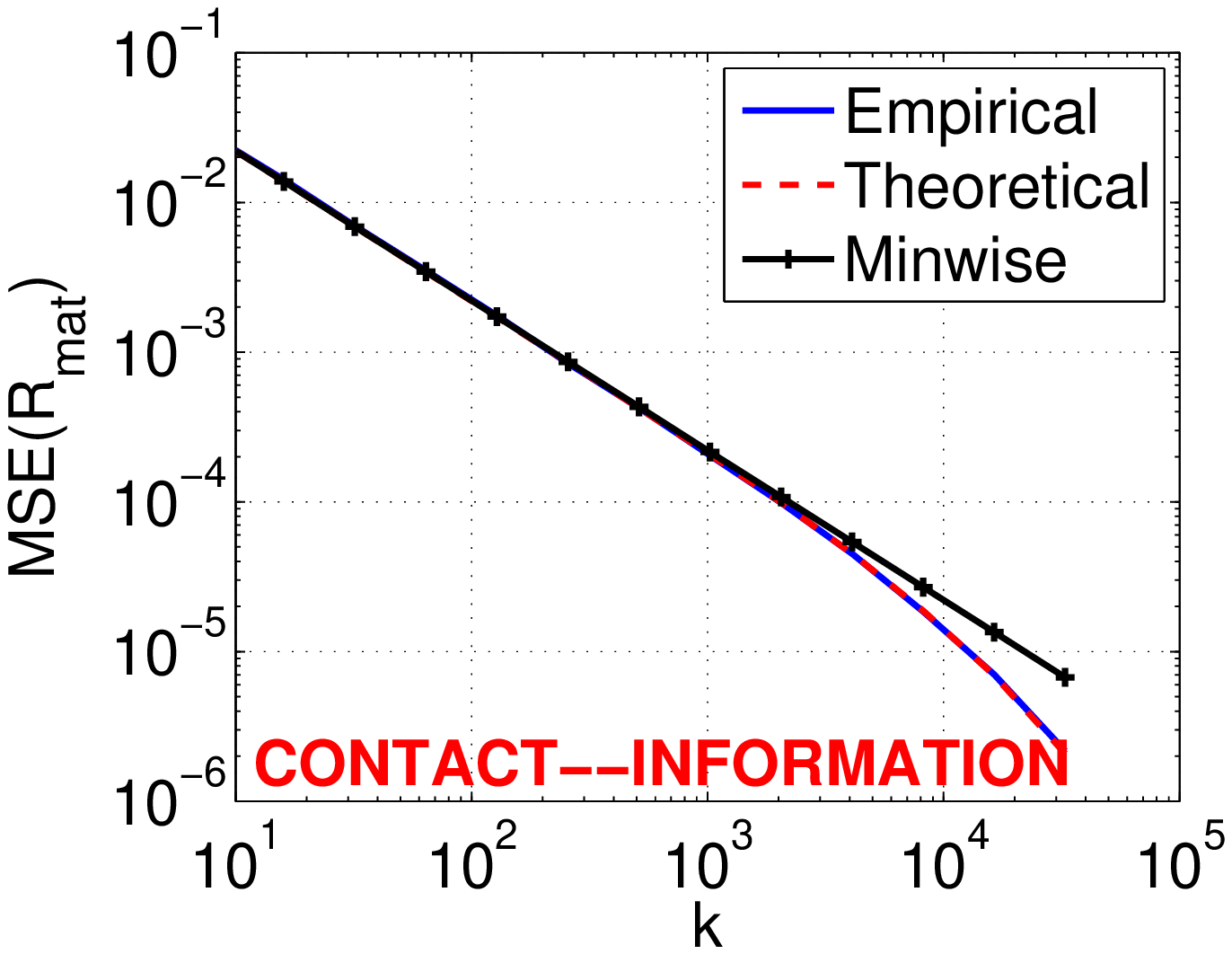}
}
\mbox{
\includegraphics[width=1.5in]{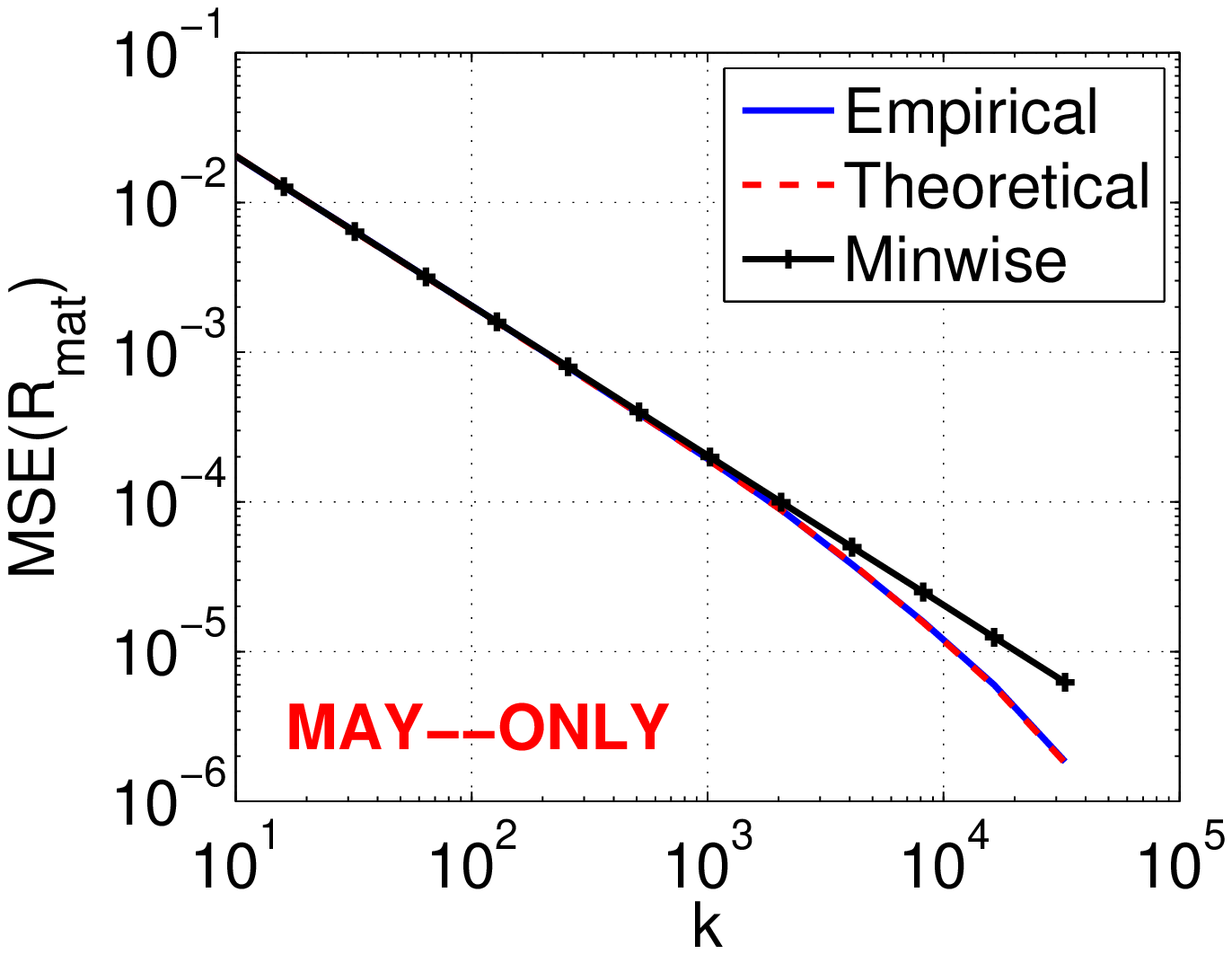}\hspace{-0.1in}
\includegraphics[width=1.5in]{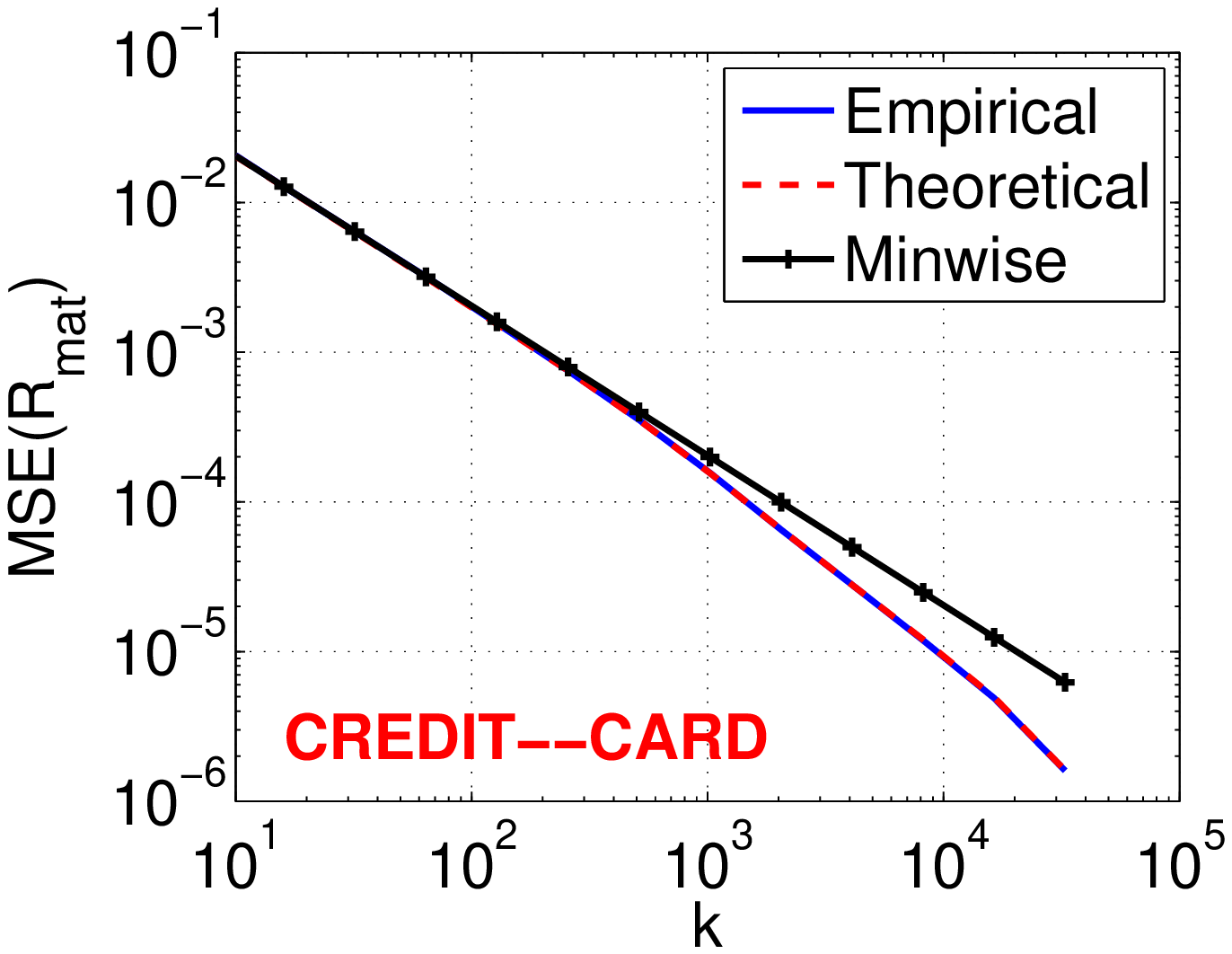}\hspace{-0.1in}
\includegraphics[width=1.5in]{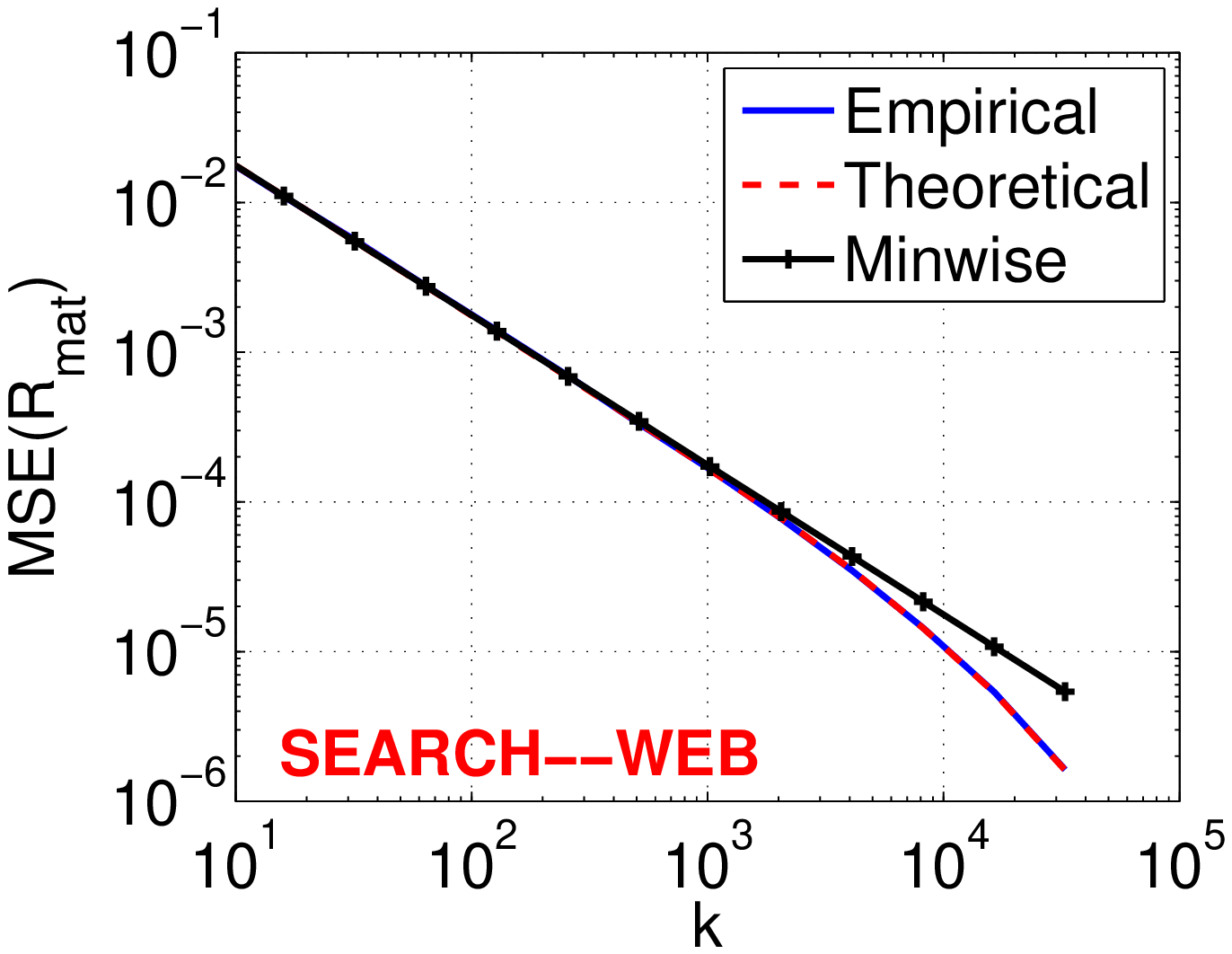}\hspace{-0.1in}
\includegraphics[width=1.5in]{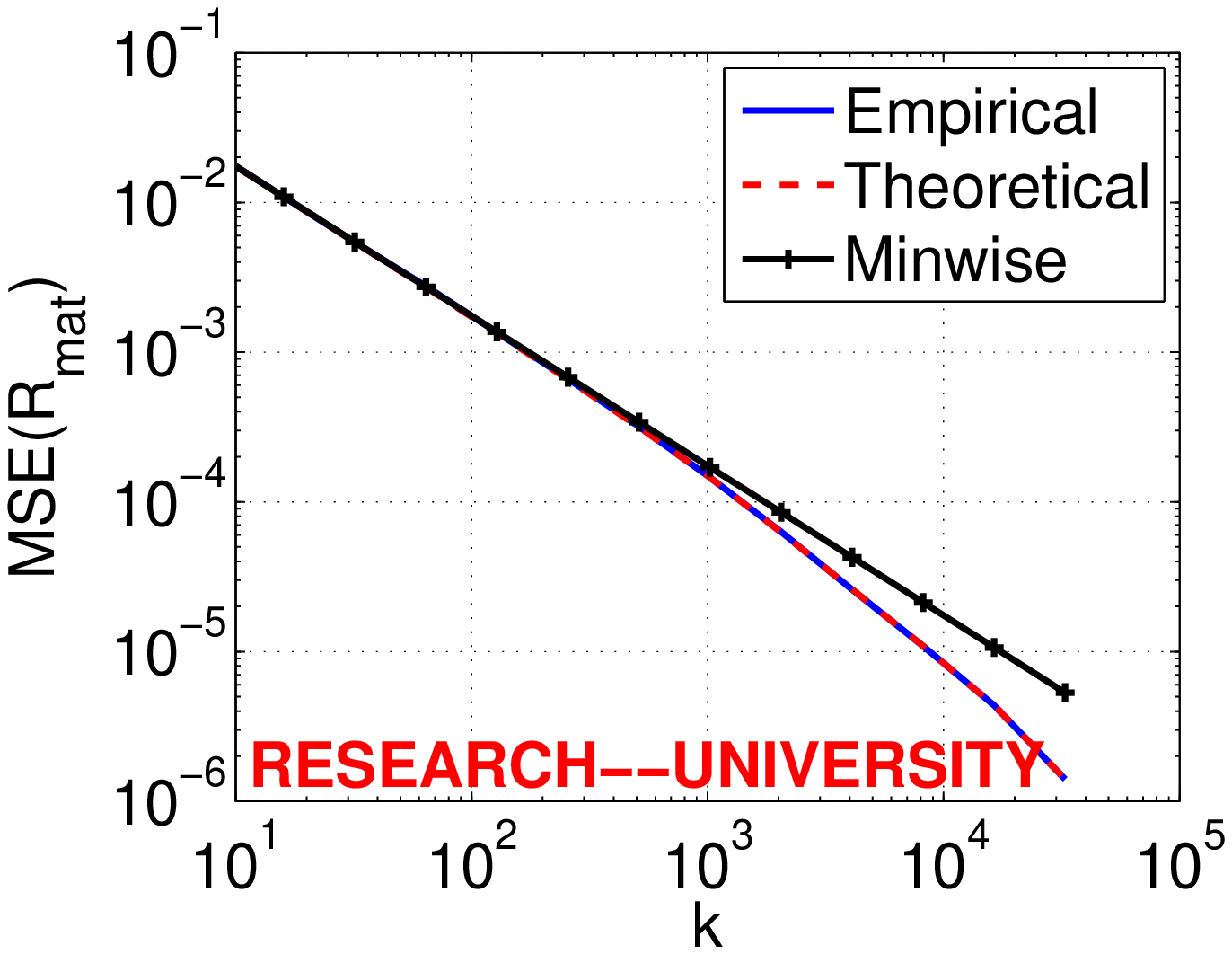}\hspace{-0.1in}
\includegraphics[width=1.5in]{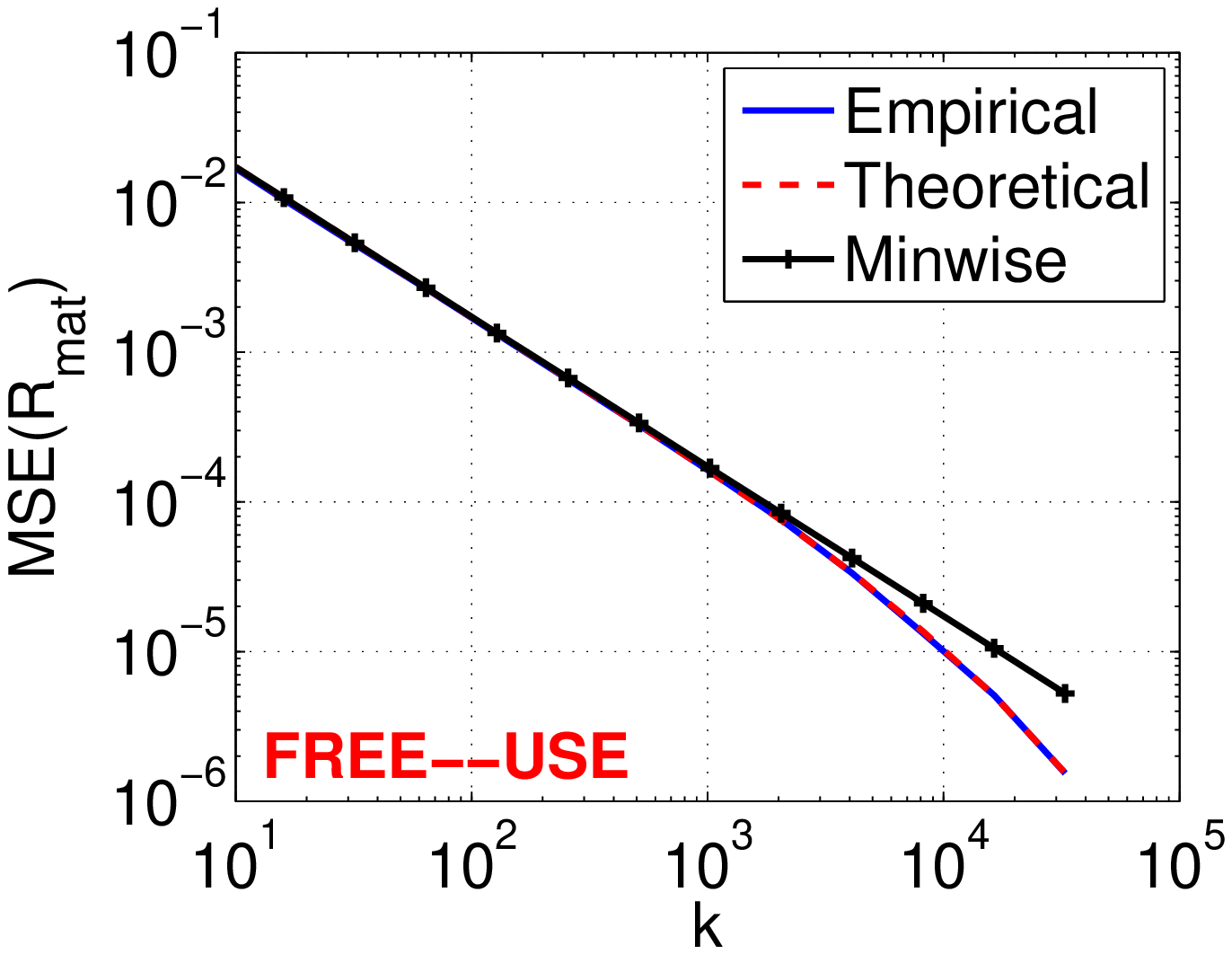}
}
\mbox{
\includegraphics[width=1.5in]{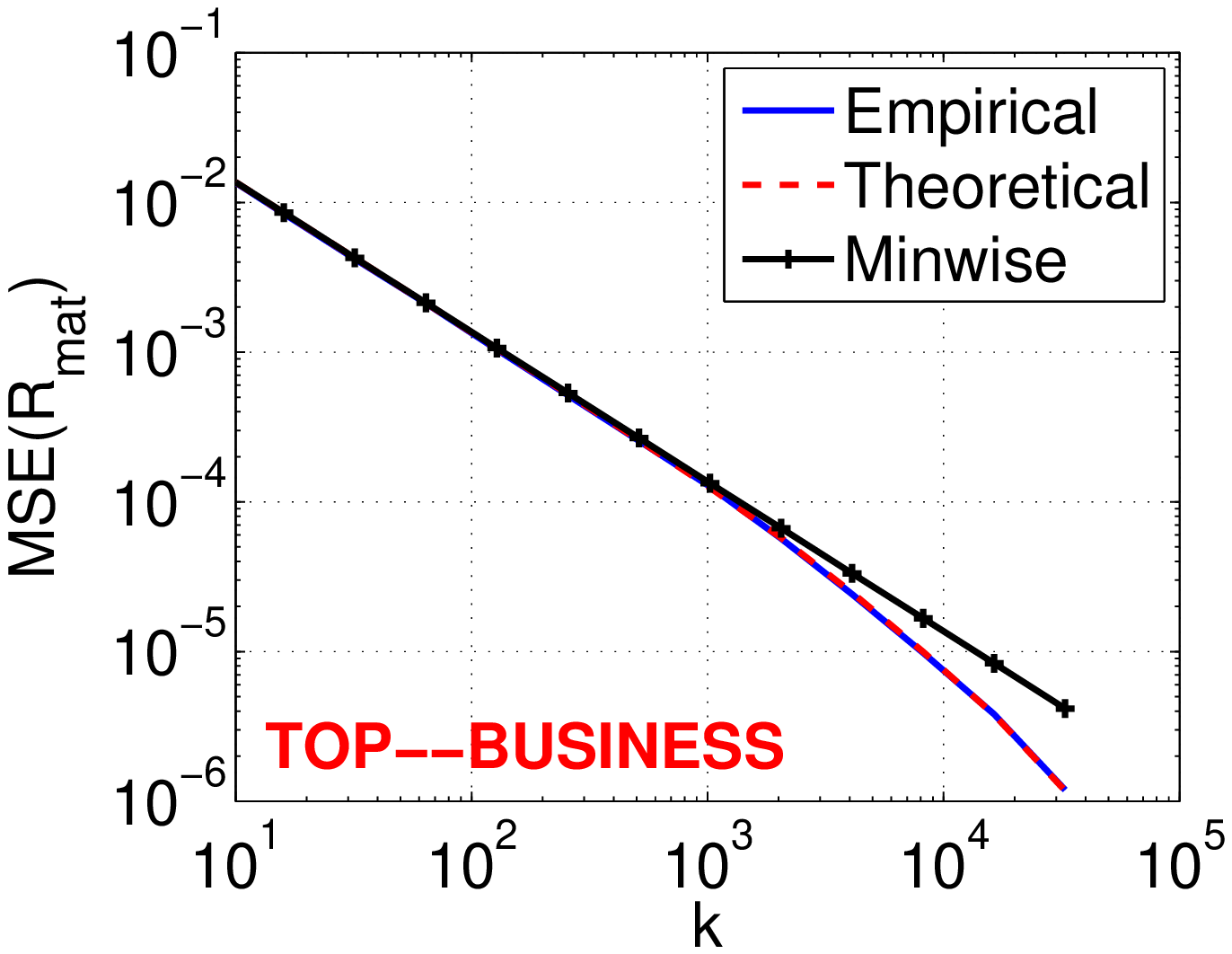}\hspace{-0.1in}
\includegraphics[width=1.5in]{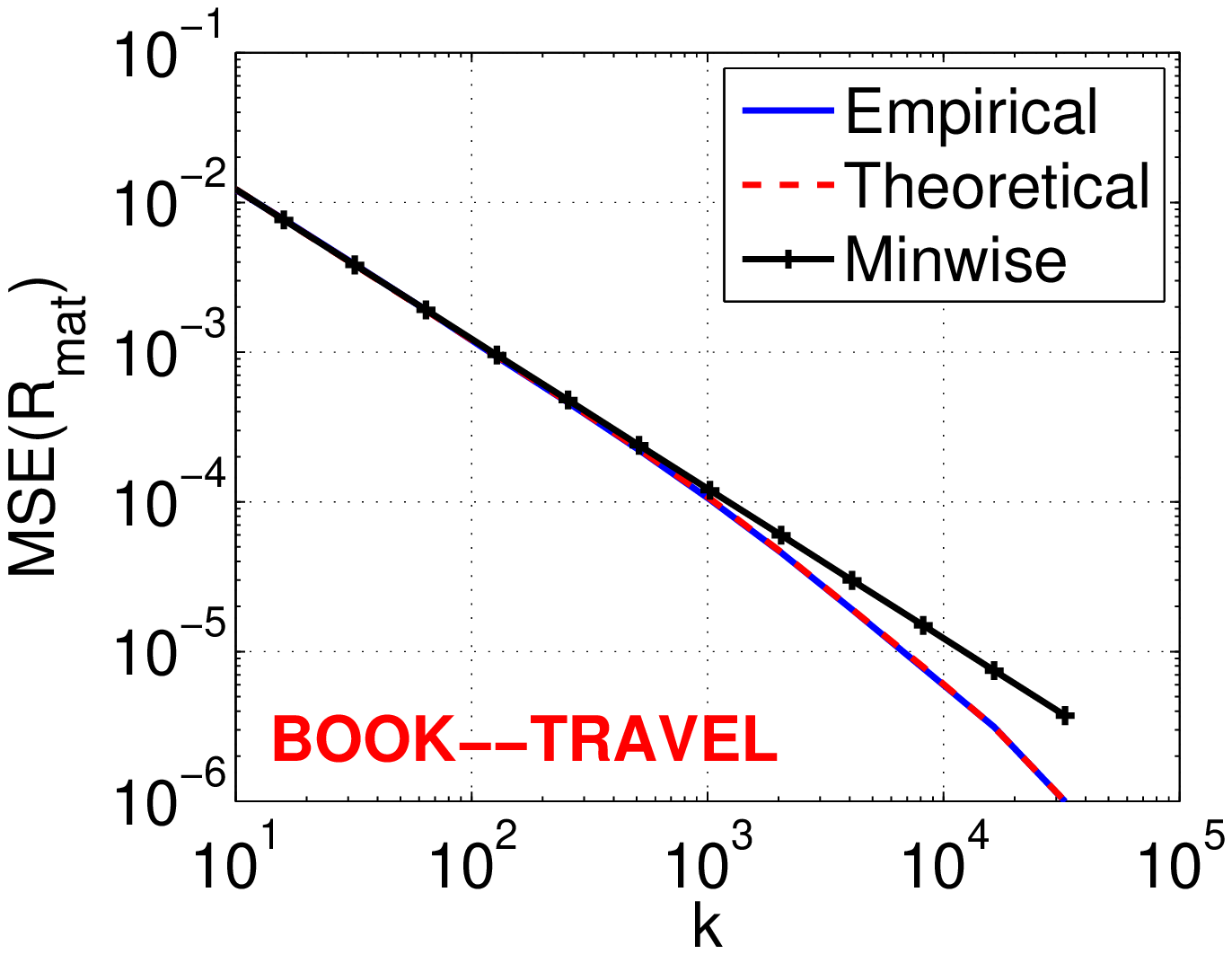}\hspace{-0.1in}
\includegraphics[width=1.5in]{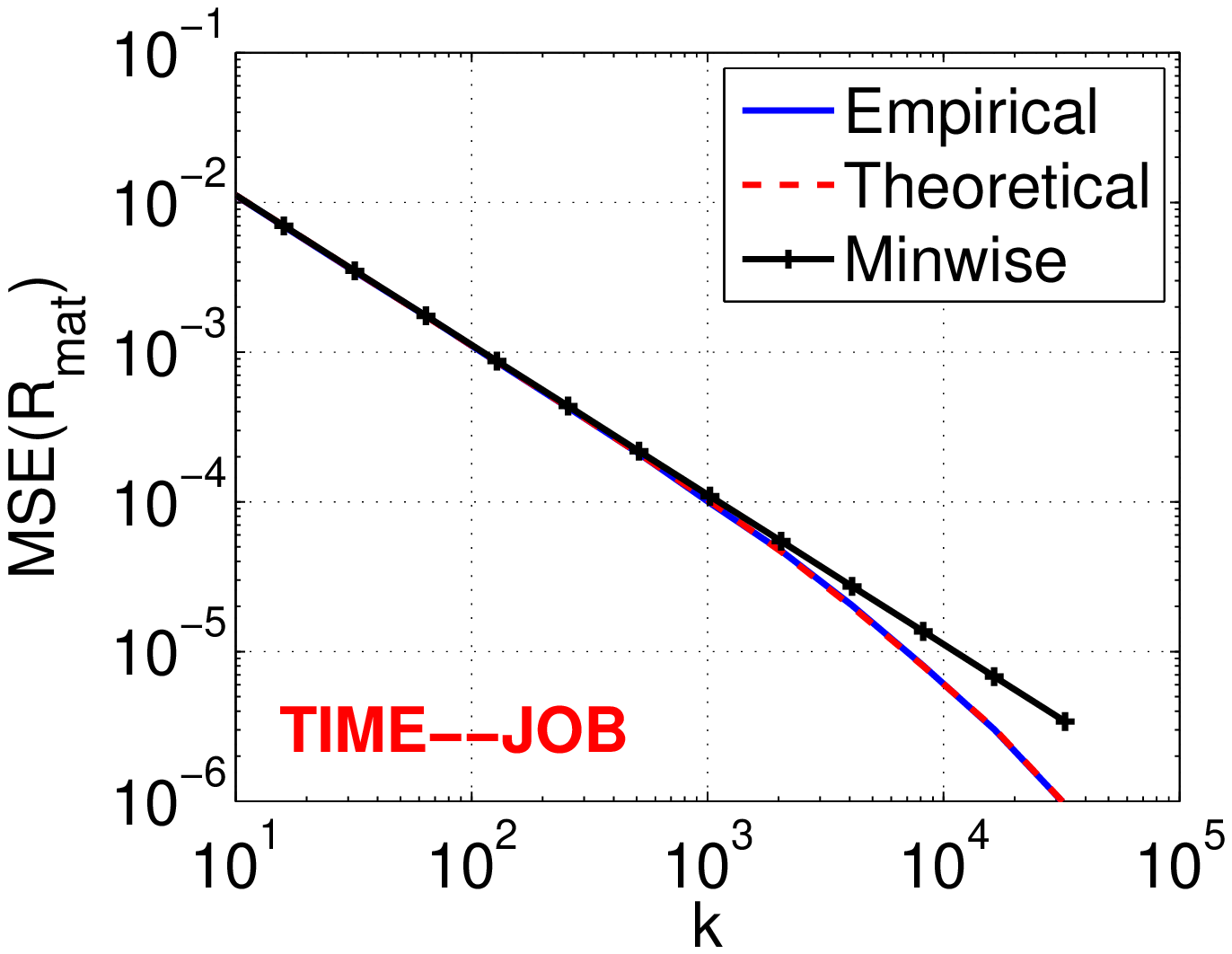}\hspace{-0.1in}
\includegraphics[width=1.5in]{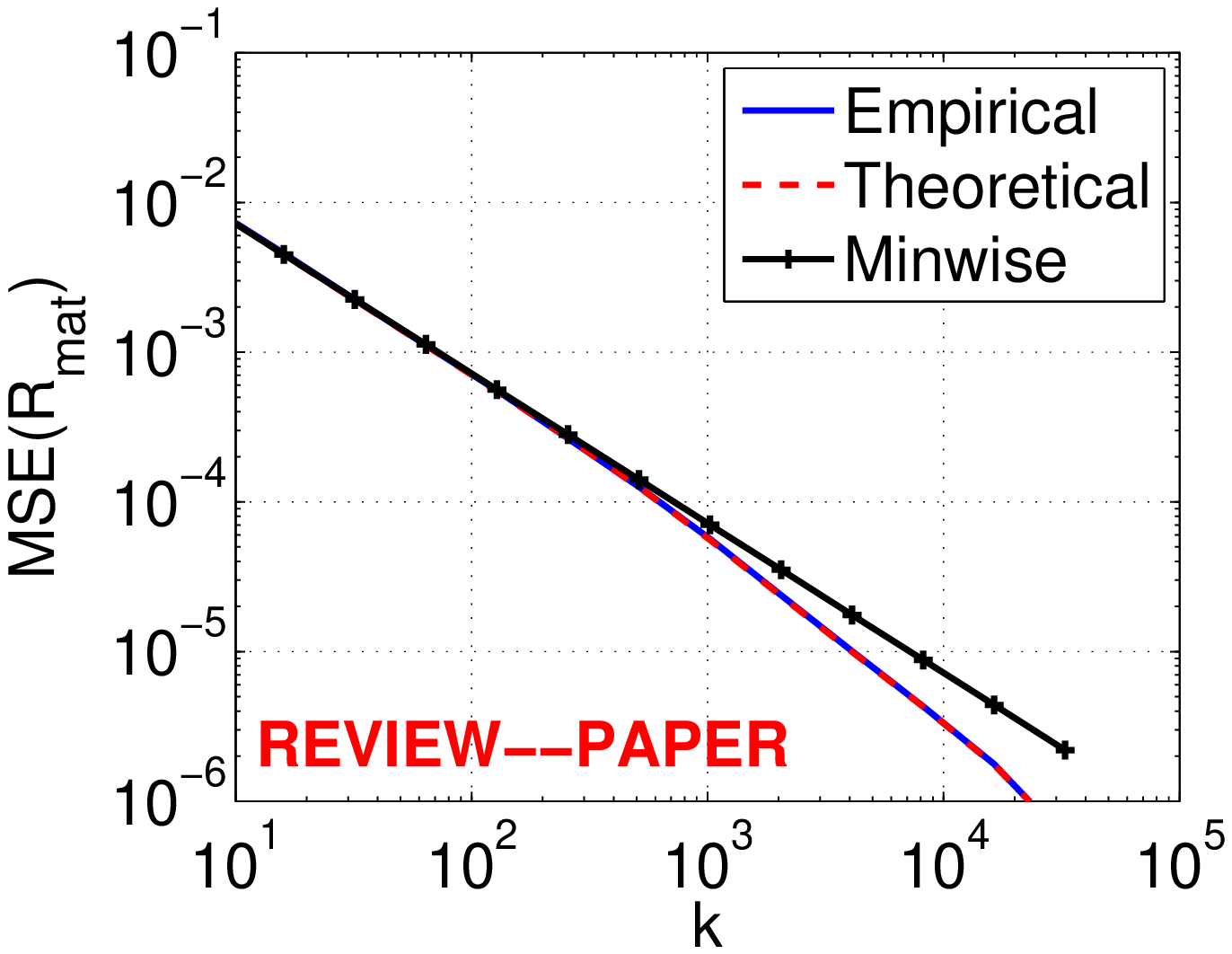}\hspace{-0.1in}
\includegraphics[width=1.5in]{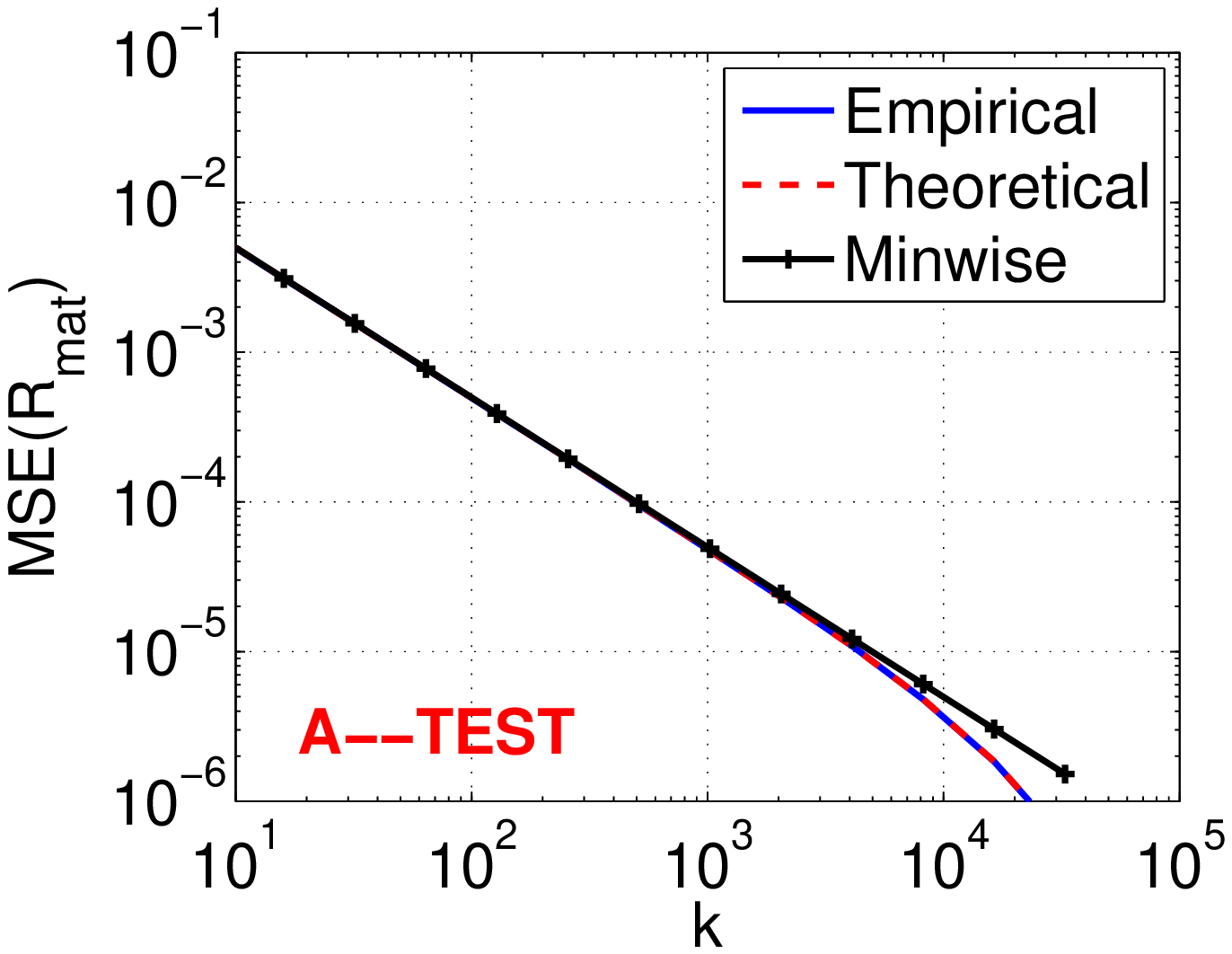}
}

\end{center}\vspace{-0.3in}
\caption{$MSE(\hat{R}_{mat})$, to verify the theoretical results of Lemma~\ref{lem_Rmat}. Note that the theoretical variance curves use the approximation (\ref{eqn_Nemp_ineq}), for convenience. The experimental results confirm that: (i) the estimator $\hat{R}_{mat}$ is unbiased, (ii) the variance formula (\ref{eqn_Rmat_Var}) and the approximation (\ref{eqn_Nemp_ineq}) are  accurate; (iii) the variance of $\hat{R}_{mat}$ is somewhat smaller than $R(1-R)/k$, the variance of the original $k$-permutation minwise hashing. }\label{fig_Rmat}\vspace{-.in}
\end{figure}

Finally, Figure~\ref{fig_Rmat} plots the empirical MSEs (MSE = bias$^2$ + variance) and the theoretical variances (\ref{eqn_Rmat_Var}), where the term $E\left(\frac{1}{k-N_{emp}}\right)$ is approximated by $\frac{1}{k-E(N_{emp})}$ as in  (\ref{eqn_Nemp_ineq}).

The experimental results confirm Lemma~\ref{lem_Rmat}: (i) the estimator $\hat{R}_{mat}$ is unbiased; (ii) the variance formula (\ref{eqn_Rmat_Var}) and the approximation (\ref{eqn_Nemp_ineq}) are  accurate; (iii) the variance of $\hat{R}_{mat}$ is somewhat smaller than $R(1-R)/k$, which is the variance of the original $k$-permutation minwise hashing, due to the ``sample-without-replacement'' effect.

\vspace{0.3in}

\noindent\textbf{Remark:}\ The empirical results presented in Figures~\ref{fig_Nemp_mean} to~\ref{fig_Rmat} have clearly validated the theoretical results for our one permutation hashing scheme. Note that we did not add the empirical results of  the original $k$-permutation minwise hashing scheme because they would simply overlap the theoretical curves. The fact that the original $k$-permutation scheme provides the unbiased estimate of $R$ with variance $\frac{R(1-R)}{k}$ has been well-validated in prior literature, for example~\cite{Article:Li_Konig_CACM11}.

\newpage

\section{Strategies for Dealing with Empty Bins}\label{sec_empty_bin}

In general, we expect that empty bins should not occur often because $E(N_{emp})/k \approx e^{-f/k}$, which is very close to zero if $f/k>5$. (Recall $f=|S_1\cup S_2|$.) If the goal of using minwise hashing is for data reduction, i.e., reducing the number of nonzeros, then we would expect that $f\gg k$ anyway.

Nevertheless, in applications where we need the estimators to be inner products, we  need  strategies to deal with empty bins in case they occur. Fortunately, we realize a  (in retrospect)  simple strategy which can be very nicely integrated with linear learning algorithms and performs very well. \\

\begin{wrapfigure}{r}{0.5\textwidth}
\vspace{-0.3in}
\begin{center}
\includegraphics[width = 1.8in]{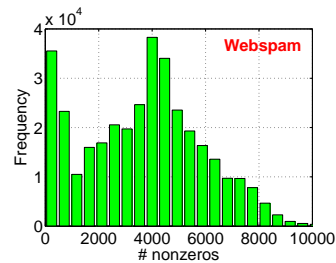}
\end{center}
\vspace{-0.3in}
\caption{Histogram of the numbers of nonzeros in the {\em webspam} dataset (350,000 samples).}\label{fig_f_hist}\vspace{-0.2in}
\end{wrapfigure}

Figure~\ref{fig_f_hist} plots the histogram of the numbers of nonzeros in the {\em webspam} dataset, which has 350,000 samples. The average number of nonzeros is about 4000 which should be much larger than the $k$ (e.g., 200 to 500) for the hashing procedure. On the other hand, about $10\%$ (or $2.8\%$) of the samples have $<500$ (or $<200$) nonzeros. Thus, we must deal with empty bins if we do not want to exclude those data points. For example, if $f=k=500$, then $N_{emp}\approx e^{-f/k} =0.3679$, which is not small. \\

The first (obvious) idea is \textbf{random coding}. That is, we simply replace an empty bin (i.e., ``*'' as in Figure~\ref{fig_one_permutation}) with a random number. In terms of the original unbiased estimator $\hat{R}_{mat} = \frac{N_{mat}}{k-N_{emp}}$, the random coding scheme will almost not change the numerator $N_{mat}$. The drawback of random coding is that the denominator will effectively become $k$. Of course, in most practical scenarios, we expect $N_{emp}\approx 0$ anyway. \\

The strategy we recommend for linear learning is \textbf{zero coding}, which is tightly coupled with the strategy of hashed data expansion~\cite{Proc:HashLearning_NIPS11} as reviewed in Sec.~\ref{sec_linear_learning}. More details will be elaborated in Sec.~\ref{sec_svm}. Basically, we can encode ``*'' as ``zero'' in the expanded space, which means $N_{mat}$ will remain the same (after taking the inner product in the expanded space). A very nice property of this strategy is that it is \textbf{sparsity-preserving}. This strategy essentially corresponds to the following modified estimator:
\begin{align}\label{eqn_Rmat0}
\hat{R}_{mat}^{(0)} = \frac{N_{mat}}{\sqrt{k-N_{emp}^{(1)}} \sqrt{k-N_{emp}^{(2)}} }
\end{align}
where $N_{emp}^{(1)} = \sum_{j=1}^k I_{emp,j}^{(1)}$ and $N_{emp}^{(2)}=\sum_{j=1}^k I_{emp,j}^{(2)}$ are the numbers of empty bins in $\pi(S_1)$ and $\pi(S_2)$, respectively. This modified estimator actually makes a lot of sense, after some careful thinking.

Basically, since each data vector is processed and coded separately, we actually do not know $N_{emp}$ (the number of {\em jointly} empty bins) until we see both $\pi(S_1)$ and $\pi(S_2)$. In other words, we can not really compute $N_{emp}$ if we want to use linear estimators. On the other hand, $N_{emp}^{(1)}$ and $N_{emp}^{(2)}$ are always available. In fact, the use of $\sqrt{k-N_{emp}^{(1)}} \sqrt{k-N_{emp}^{(2)}}$ in the denominator corresponds to the normalizing step which is usually needed before feeding the data to a solver. This point will probably become more clear in Sec.~\ref{sec_svm}.\\

When $N_{emp}^{(1)} = N_{emp}^{(2)} = N_{emp}$, (\ref{eqn_Rmat0}) is equivalent to the original $\hat{R}_{mat}$. When two original vectors are very similar (e.g., large $R$), $N_{emp}^{(1)}$ and $N_{emp}^{(2)}$ will be close to $N_{emp}$.   When two sets are highly unbalanced, using (\ref{eqn_Rmat0}) will likely overestimate $R$; however, in this case, $N_{mat}$ will be so small that the absolute error will not be  large. In any case,  we do not expect the existence of empty bins will significantly affect the performance in practical settings.

\subsection{The $m$-Permutation Scheme with $1<m\ll k$}

In case some readers would like to further (significantly) reduce the chance of the occurrences of empty bins, here we shall mention that one  does not really have to strictly follow ``one permutation,'' since one can always conduct $m$ permutations with $k^\prime = k/m$ and concatenate the hashed data. Once the preprocessing is no longer the bottleneck, it  matters less  whether we use 1 permutation or (e.g.,) $m=3$ permutations. The chance of having empty bins decreases exponentially  with increasing $m$.

\subsection{An Example of The ``Zero Coding'' Strategy for Linear Learning}\label{sec_svm}

Sec.~\ref{sec_linear_learning} has already reviewed the data-expansion strategy used by~\cite{Proc:HashLearning_NIPS11} for integrating ($b$-bit) minwise hashing with linear learning. We will adopt a similar strategy with modifications for considering empty bins.

We  use a similar example as in Sec.~\ref{sec_linear_learning}. Suppose we apply our one permutation hashing scheme and use $k=4$ bins. For the first data vector, the hashed values are  $[12013,\ 25964,\ 20191,\ *]$ (i.e., the 4-th bin is empty). Suppose again we use $b=2$ bits.  With the ``zero coding'' strategy, our procedure is summarized as follows:
\begin{align}\notag
&\begin{array}{lrrrr}
\text{Original hashed values } (k=4): &12013 &25964 &20191 &*\\
\text{Original binary representations}: &010111011101101& 110010101101100& 100111011011111 &*\\
\text{Lowest $b=2$ binary digits}: &01& 00& 11 &*\\
\text{Expanded $2^b=4$ binary digits }: &0 0 1 0 & 0 0 0 1 & 1 0 0 0 &{\color{red}0 0 0 0}\\
\end{array}\\\notag
&\text{New feature vector fed to a solver}: \frac{1}{\sqrt{\color{red}4-1}}\times[0, 0, 1, 0, 0, 0, 0, 1, 1, 0, 0, 0, 0, 0, 0, 0]
\end{align}

We apply the same procedure to all  feature vectors in the data matrix to generate a new  data matrix. The normalization factor $\frac{1}{\sqrt{k-N_{emp}^{(i)}}}$ varies, depending on the number of empty bins in the $i$-th feature vector.

We believe  zero coding  is an ideal strategy for dealing with empty bins in the context of linear learning as it is very convenient and produces accurate results (as we will show by experiments). If we use the ``random coding'' strategy (i.e., replacing a ``*'' by a random number in $[0,\ 2^b-1]$), we  need to add artificial nonzeros (in the expanded space) and the normalizing factor is always $\frac{1}{\sqrt{k}}$ (i.e., no longer ``sparsity-preserving'').

We apply both the zero coding and random coding strategies on the {\em webspam} dataset, as presented in Sec.~\ref{sec_webspam} Basically, both strategies produce similar results even when $k=512$, although the zero coding strategy is slightly better. We also compare the results with the original $k$-permutation scheme. On the {\em webspam} dataset, our one permutation scheme achieves similar (or even slightly better) accuracies compared to the $k$-permutation scheme.

To test the robustness of one permutation hashing, we also experiment with the {\em news20} dataset, which has only 20,000 samples and 1,355,191 features, with merely about 500 nonzeros per feature vector on average. We purposely let $k$ be as large as $4096$. Interestingly, the experimental results show that the zero coding strategy can perform extremely well. The test accuracies consistently improve as $k$ increases. In comparisons, the random coding strategy performs  badly unless $k$ is small (e.g., $k\leq 256$). 

On the {\em news20} dataset, our one permutation scheme actually outperforms the original $k$-permutation scheme, quite noticeably when $k$ is large. This should be due to the benefits from the ``sample-without-replacement'' effect. One permutation hashing  provides a good matrix sparsification scheme without ``damaging'' the original data matrix too much.

\section{Experimental Results on the Webspam Dataset}\label{sec_webspam}

The {\em webspam} dataset has 350,000 samples and 16,609,143  features. Each feature vector has on average about 4000 nonzeros; see Figure~\ref{fig_f_hist}. Following~\cite{Proc:HashLearning_NIPS11}, we use $80\%$ of  samples for training and the remaining $20\%$ for testing. We conduct extensive experiments on linear SVM and logistic regression, using our proposed one permutation hashing scheme with $k \in \{2^5, 2^6, 2^7, 2^8, 2^9\}$ and $b\in\{1,2,4,6,8\}$. For convenience, we use $D = 2^{24}$, which is divisible by $k$ and is slightly larger than 16,609,143.

There is one regularization parameter $C$ in linear SVM and logistic regression. Since our purpose is to demonstrate the effectiveness of our proposed hashing scheme, we simply provide the results for a wide range of $C$ values and  assume that the best performance is achievable if we conduct cross-validations. This way, interested readers may be able to easily reproduce our experiments.

\subsection{One Permutation  v.s. k-Permutation }

\begin{figure}[h!]
\mbox{
\includegraphics[width=1.5in]{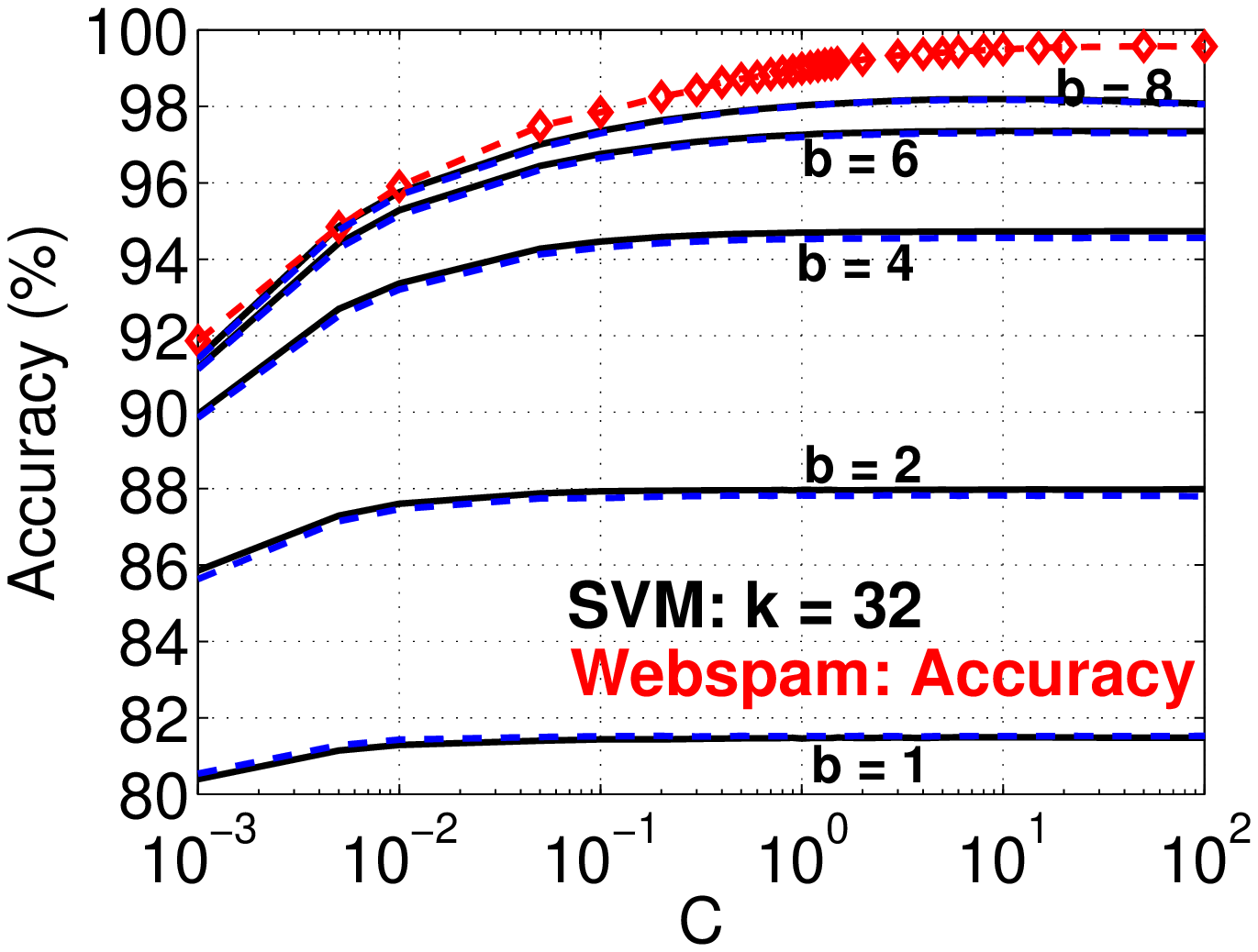}\hspace{-0.1in}
\includegraphics[width=1.5in]{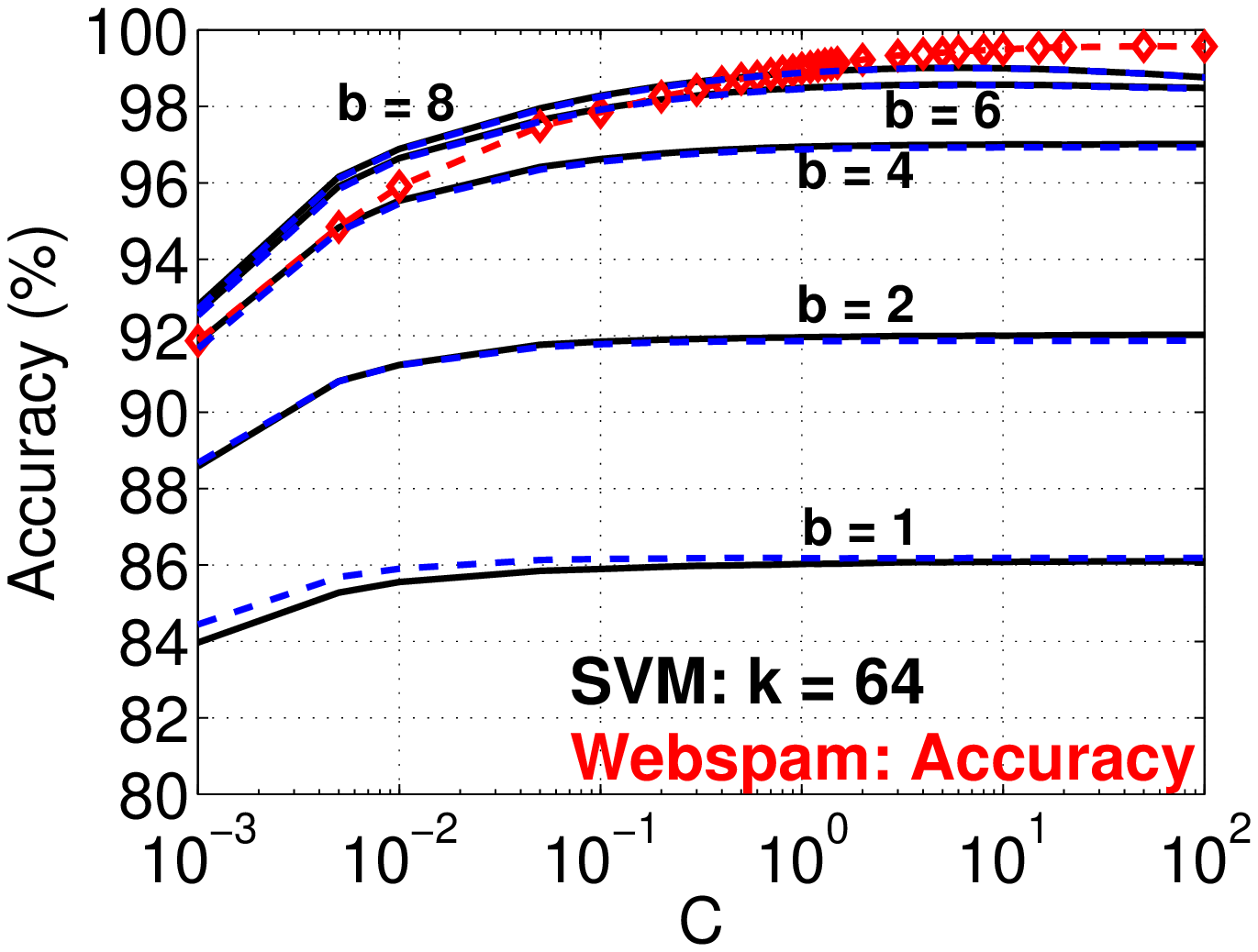}\hspace{-0.1in}
\includegraphics[width=1.5in]{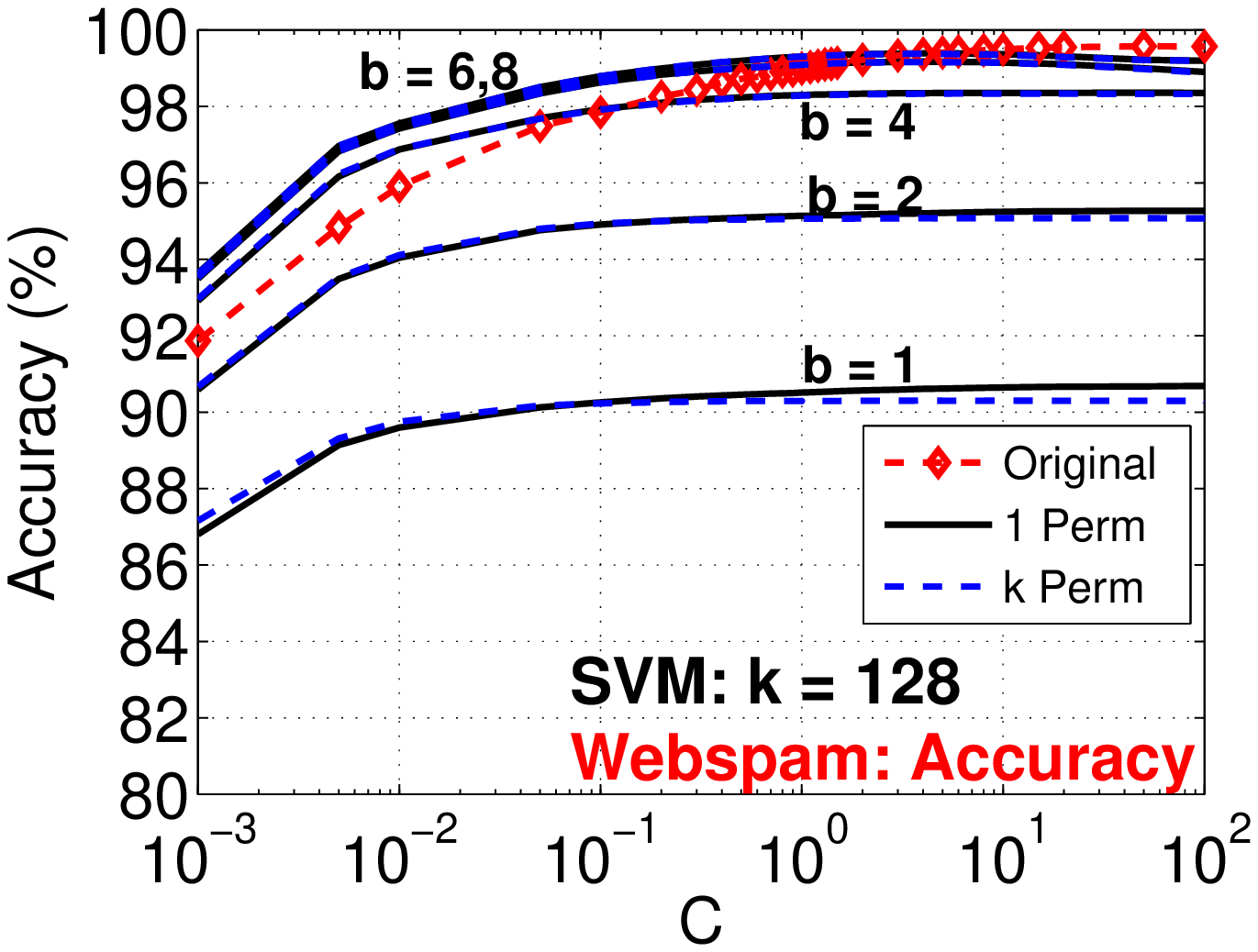}\hspace{-0.1in}
\includegraphics[width=1.5in]{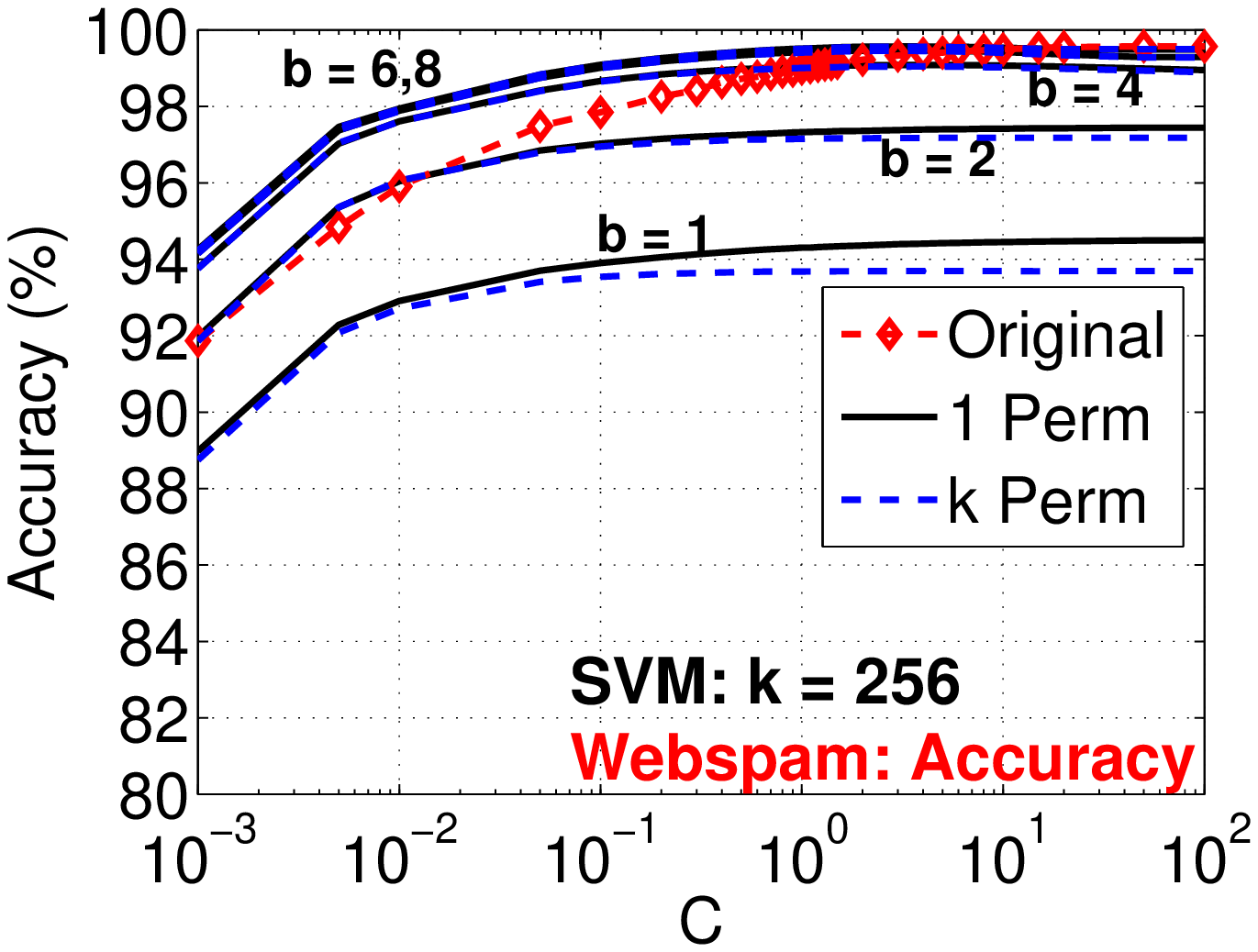}\hspace{-0.1in}
\includegraphics[width=1.5in]{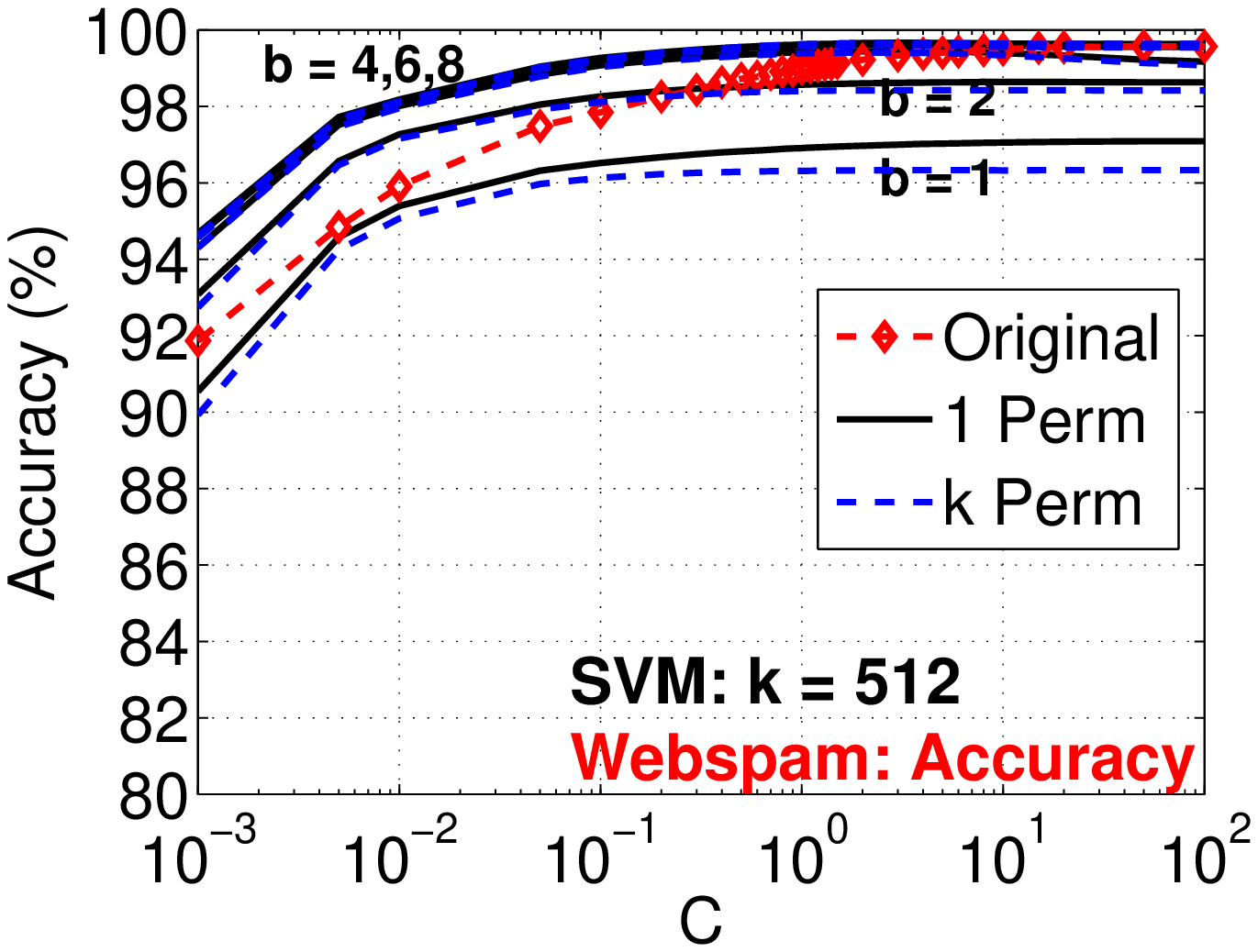}
}

\mbox{
\includegraphics[width=1.5in]{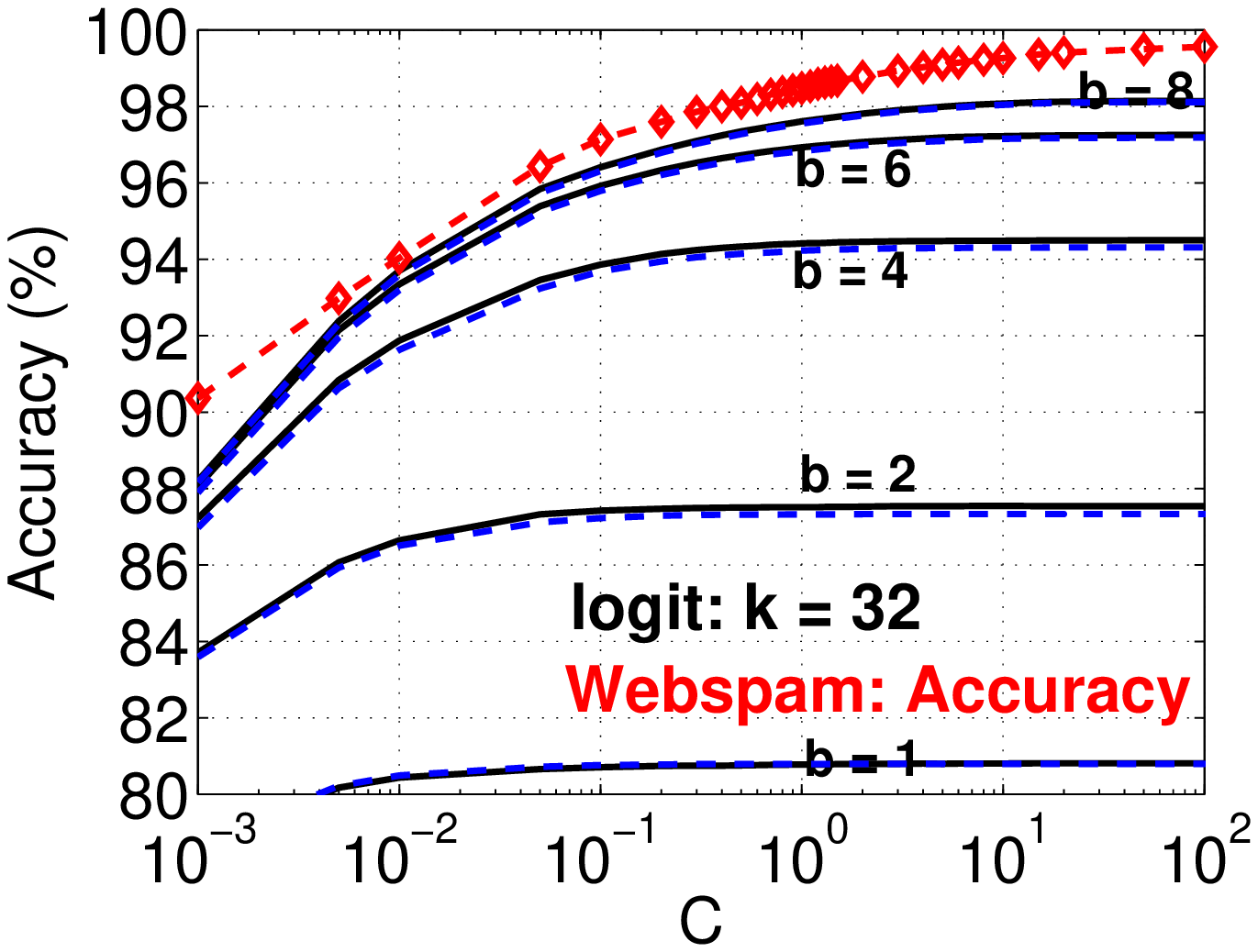}\hspace{-0.1in}
\includegraphics[width=1.5in]{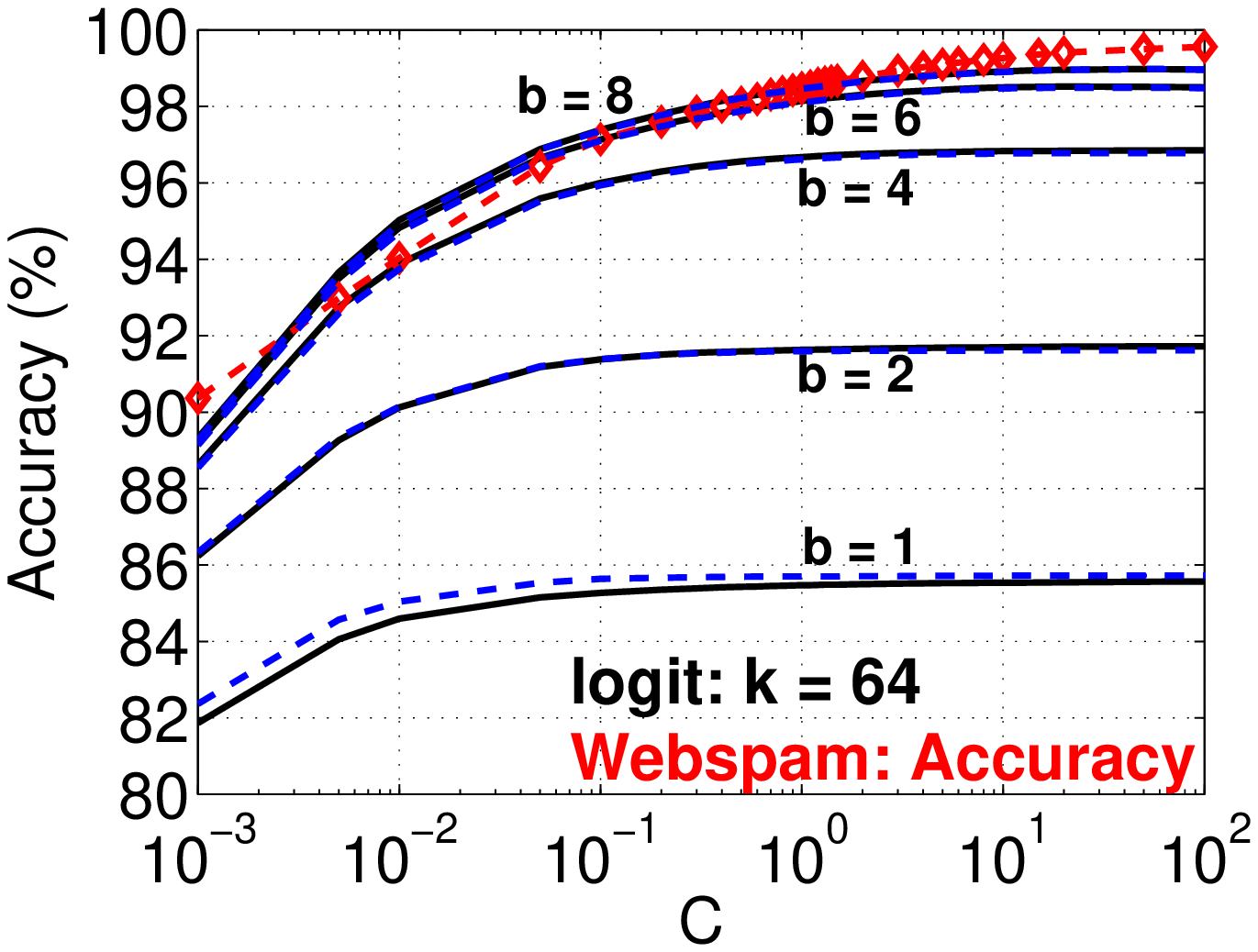}\hspace{-0.1in}
\includegraphics[width=1.5in]{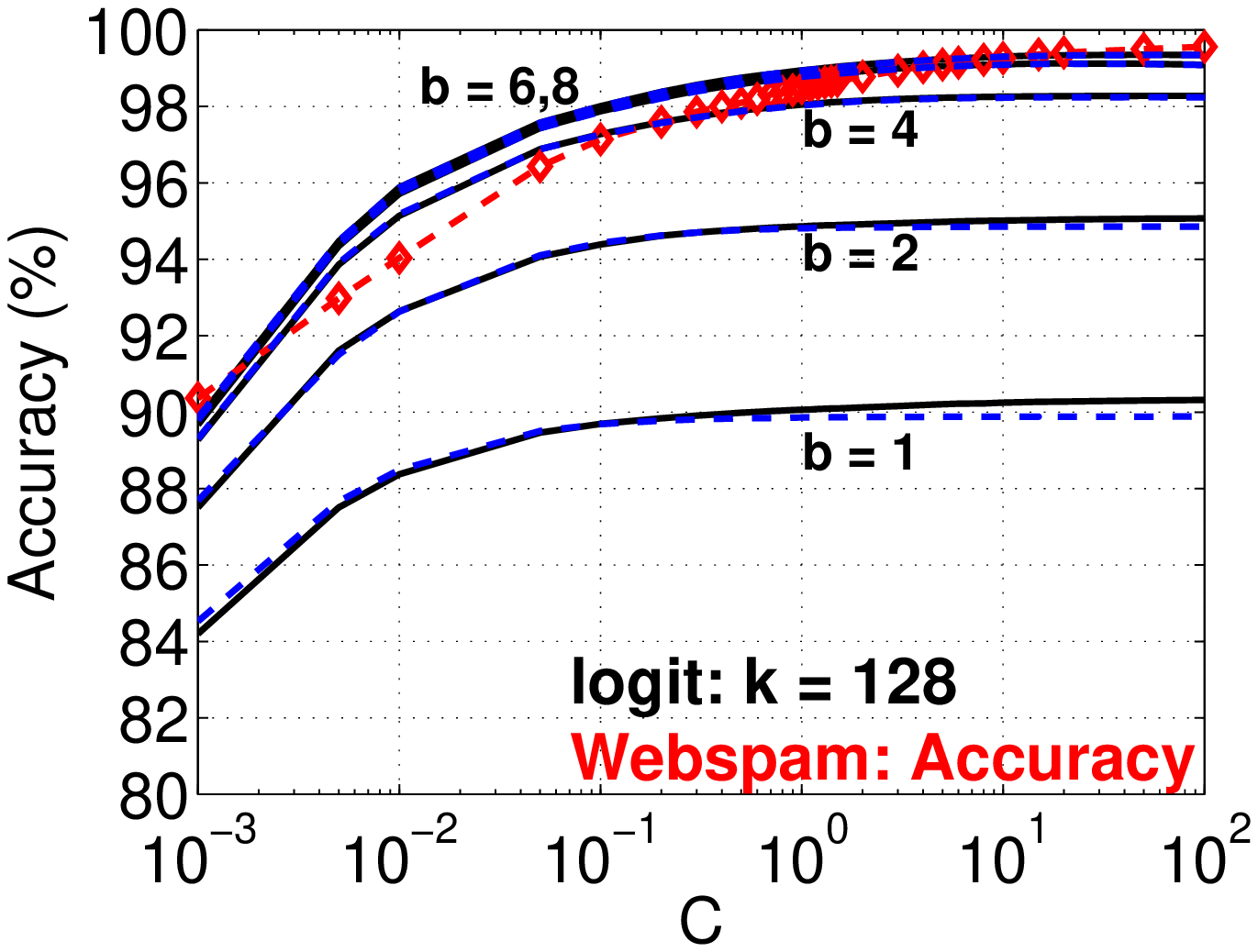}\hspace{-0.1in}
\includegraphics[width=1.5in]{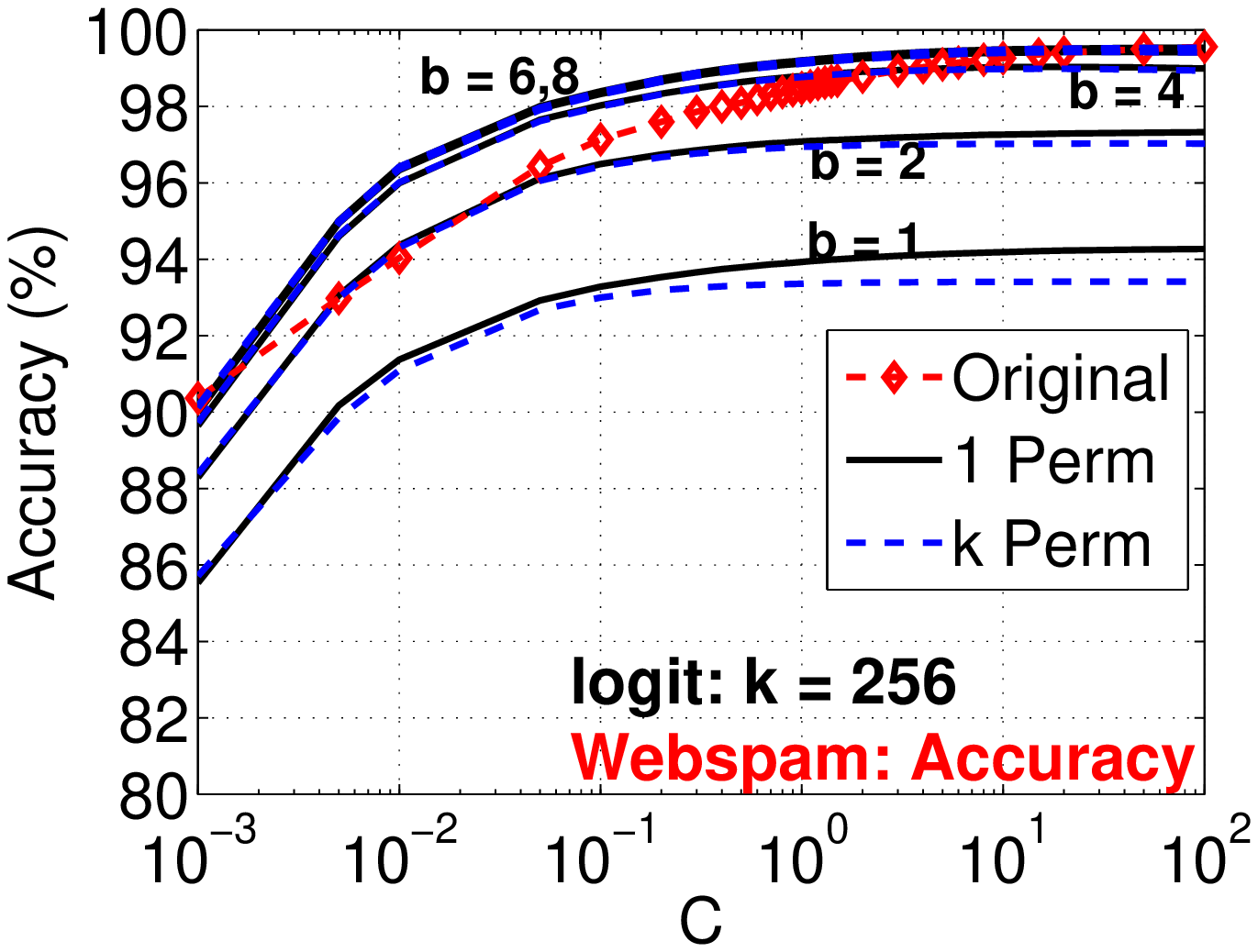}\hspace{-0.1in}
\includegraphics[width=1.5in]{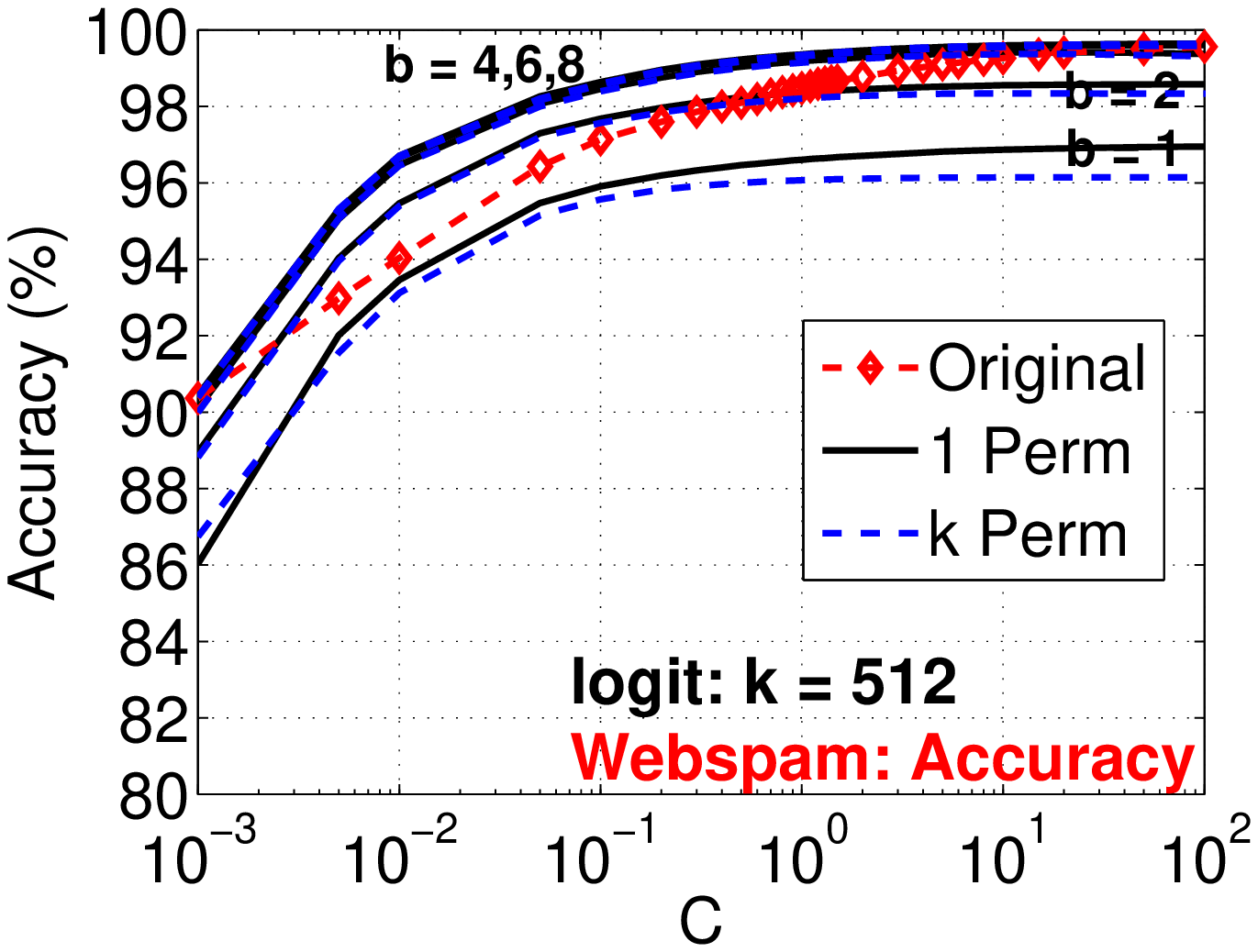}
}

\vspace{-0.2in}

\caption{Test accuracies of SVM (upper panels) and logistic regression (bottom panels),  averaged over 50 repetitions. The accuracies of using the original data  are plotted as dashed (red, if color is available) curves with ``diamond'' markers. $C$ is the regularization parameter. Compared with the original $k$-permutation minwise hashing scheme (dashed and blue if color is available), the proposed one permutation hashing scheme  achieves very similar accuracies, or even slightly better accuracies when $k$ is large. }\label{fig_spam_accuracy}
\end{figure}

Figure~\ref{fig_spam_accuracy} presents the test accuracies for both linear SVM (upper panels) and logistic regression (bottom panels).  Clearly, when $k=512$ (or even 256) and $b=8$, $b$-bit one permutation hashing achieves similar test accuracies as using the original data. Also, compared to the original $k$-permutation scheme as in~\cite{Proc:HashLearning_NIPS11}, our one permutation scheme achieves similar (or even very slightly better) accuracies.

\subsection{Preprocessing Time and Training Time}

The preprocessing cost for processing the data using $k=512$ independent permutations is about 6,000 seconds.  In contrast, the processing cost for the proposed one permutation scheme is only $1/k$ of the original cost, i.e., about 10 seconds. Note that {\em webspam} is merely a small dataset compared to industrial applications. We expect the (absolute) improvement will be even more substantial in much larger datasets.

The prior work~\cite{Proc:HashLearning_NIPS11} already presented the training time using the $k$-permutation hashing scheme.  With one permutation hashing, the training time remains essentially the same (for the same $k$ and $b$) on the {\em webspam} dataset. Note that, with the zero coding strategy, the new data matrix generated by one permutation hashing has potentially less nonzeros than the original minwise hashing scheme, due to the occurrences of empty bins.  This phenomenon in theory may bring additional advantages such as slightly reducing the training time. Nevertheless, the most significant advantage of one permutation hashing lies in the dramatic reduction of the preprocessing cost, which is what we focus on in this study.

\subsection{Zero Coding v.s. Random Coding for Empty Bins}

\begin{figure}[h!]
\mbox{
\includegraphics[width=1.5in]{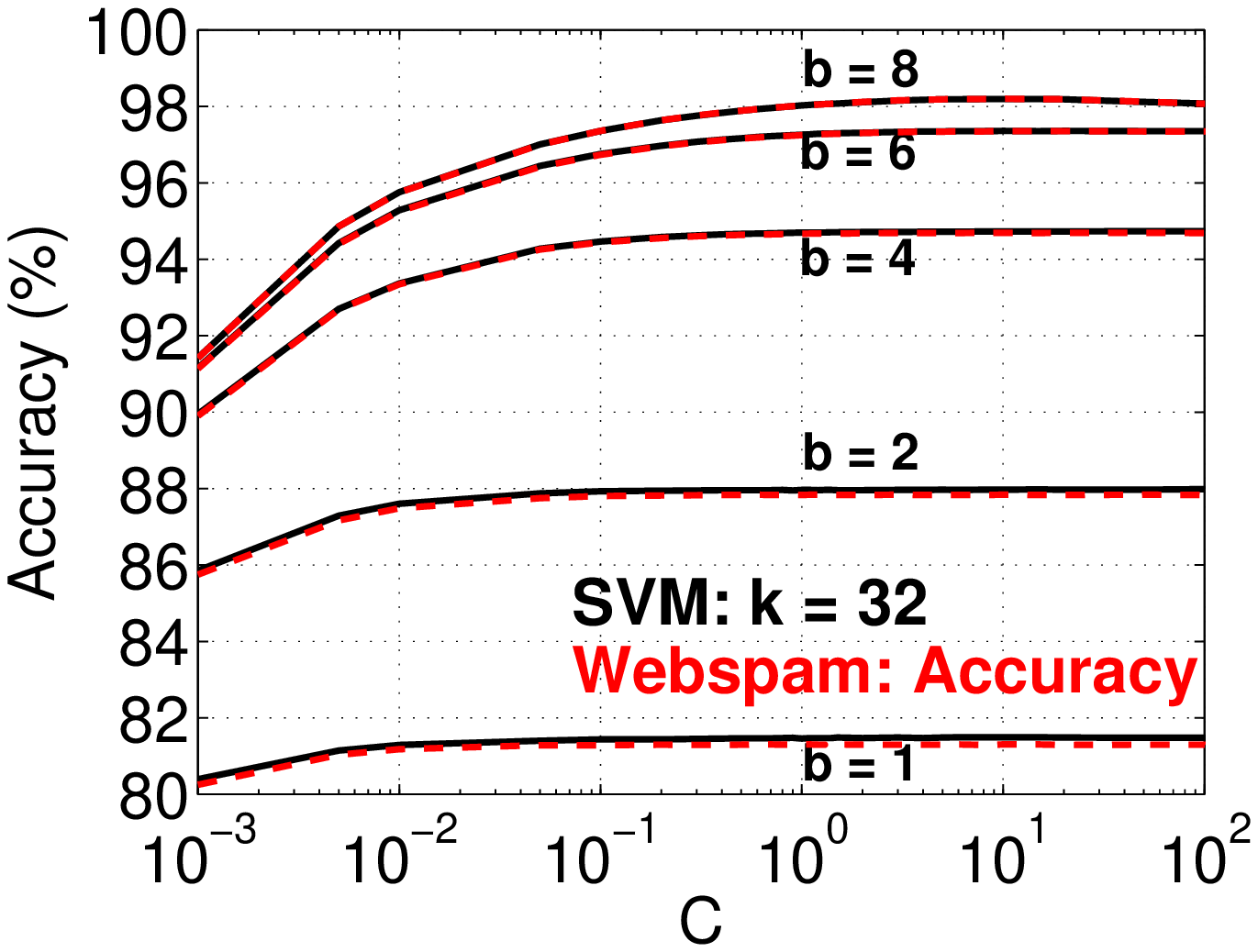}\hspace{-0.1in}
\includegraphics[width=1.5in]{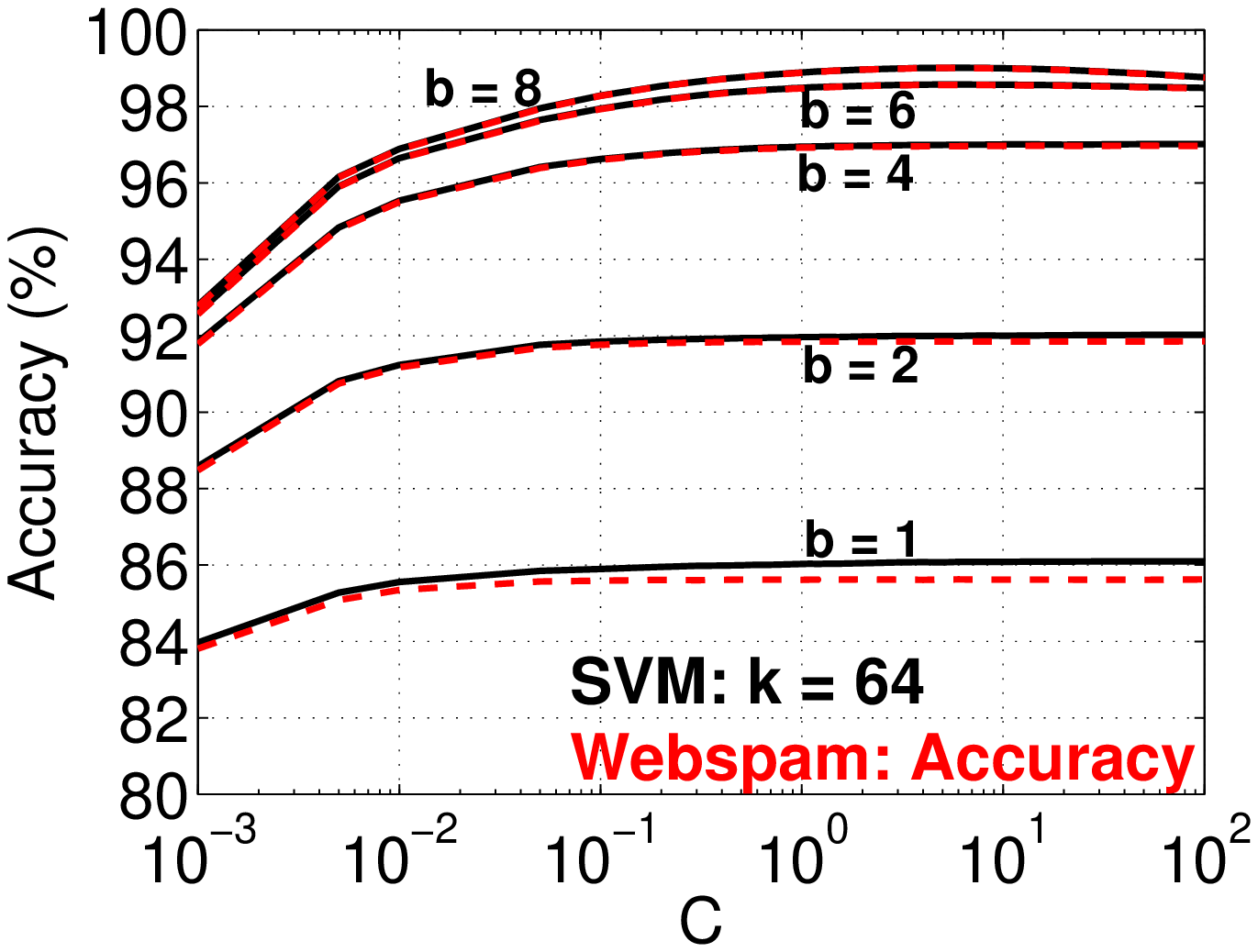}\hspace{-0.1in}
\includegraphics[width=1.5in]{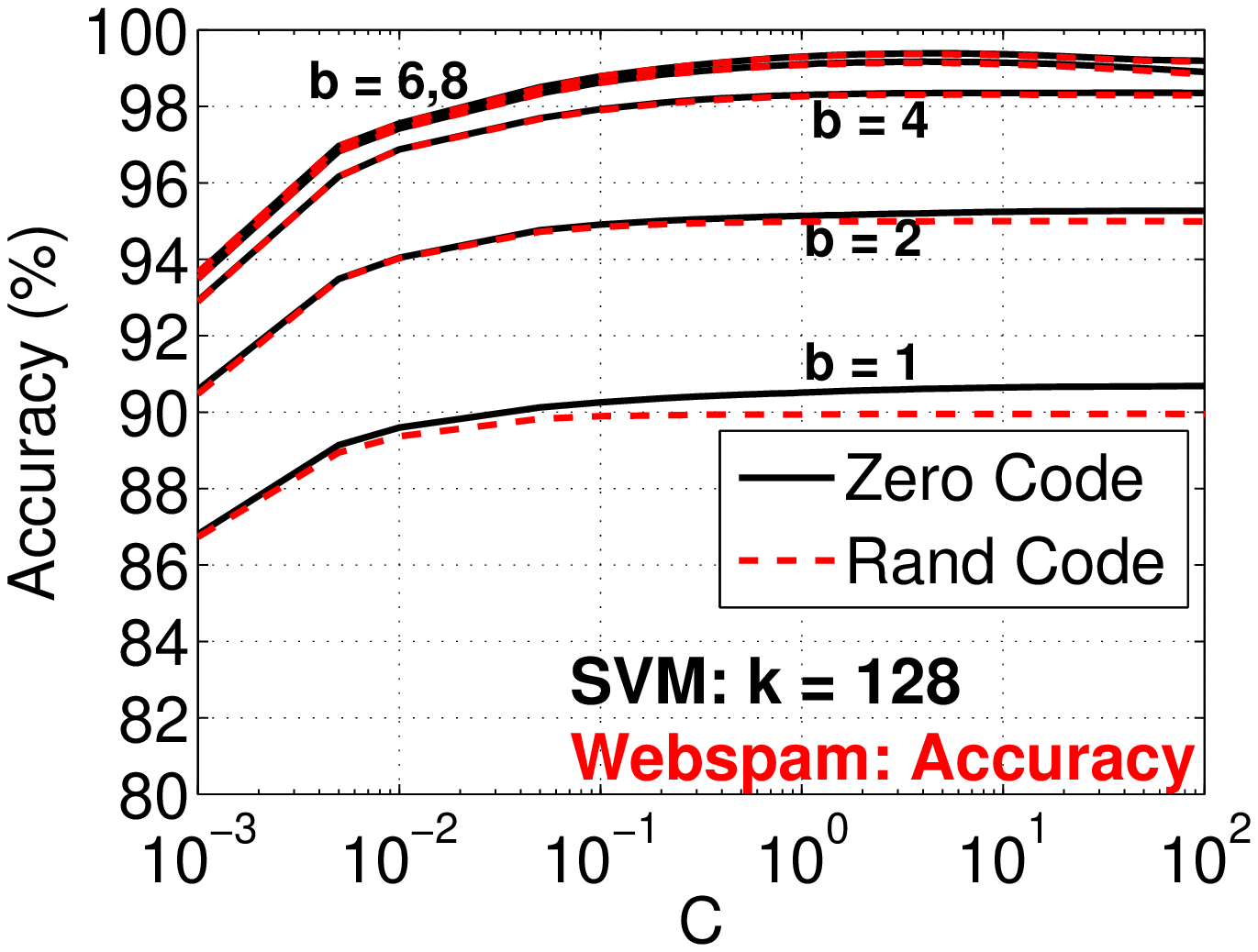}\hspace{-0.1in}
\includegraphics[width=1.5in]{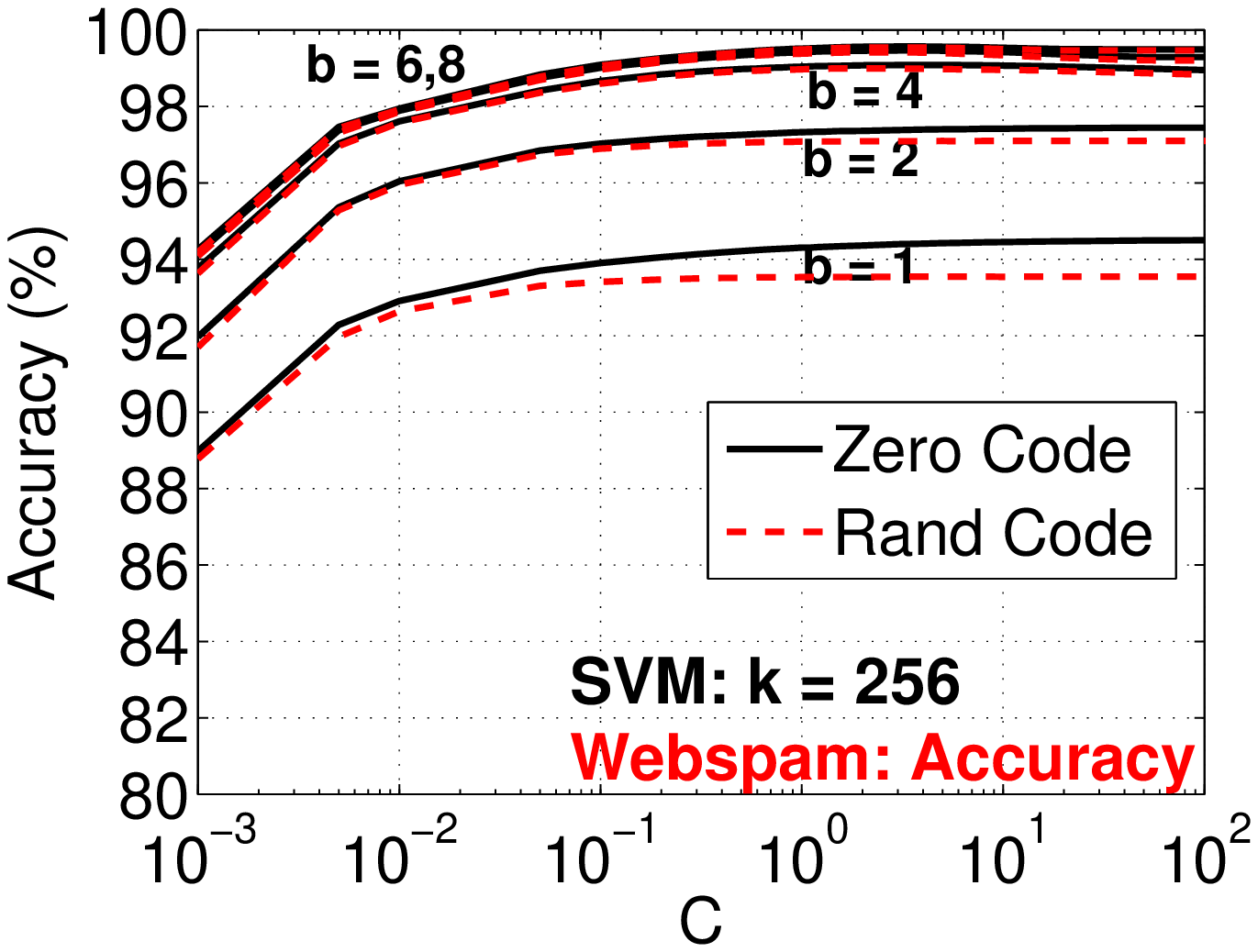}\hspace{-0.1in}
\includegraphics[width=1.5in]{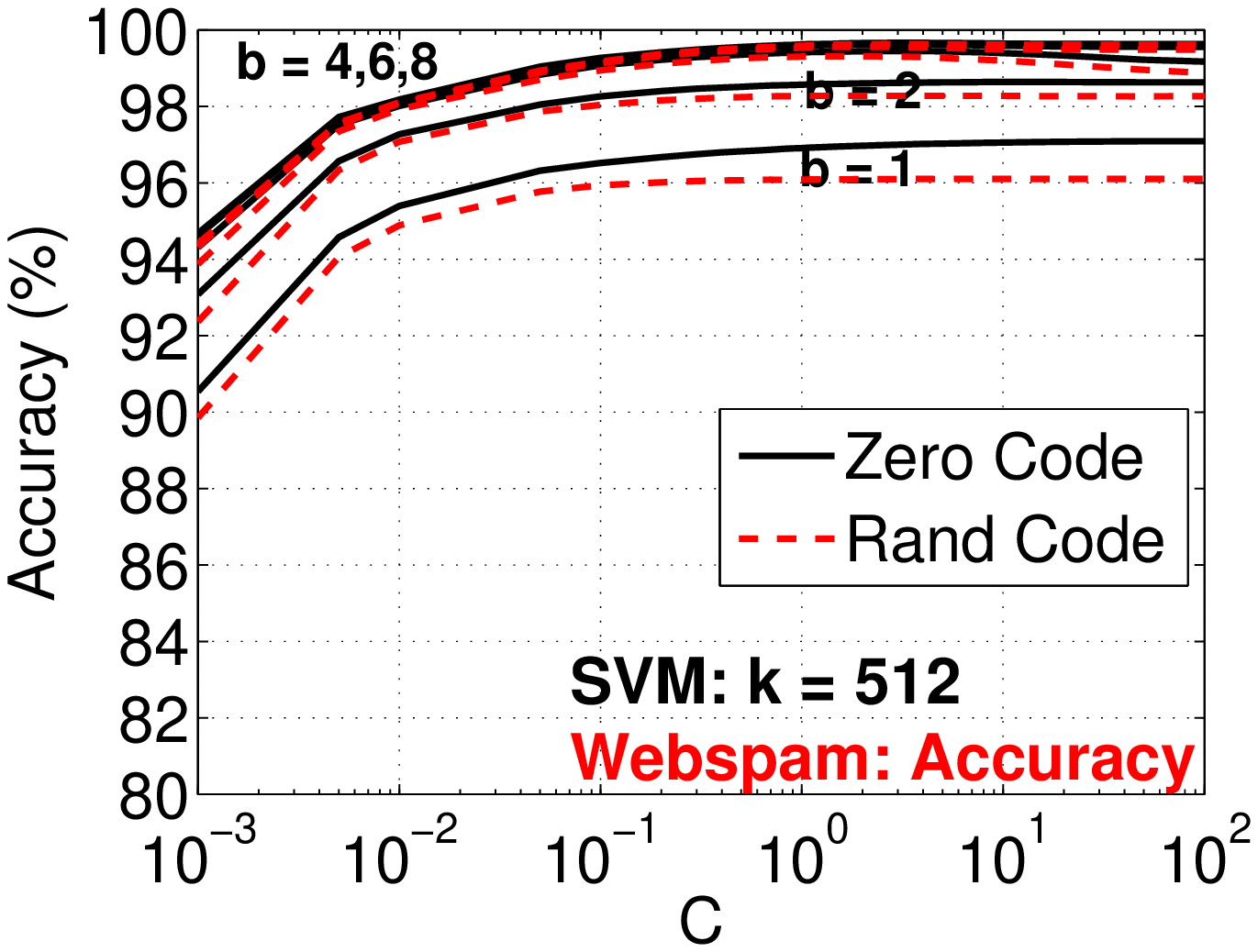}
}

\mbox{
\includegraphics[width=1.5in]{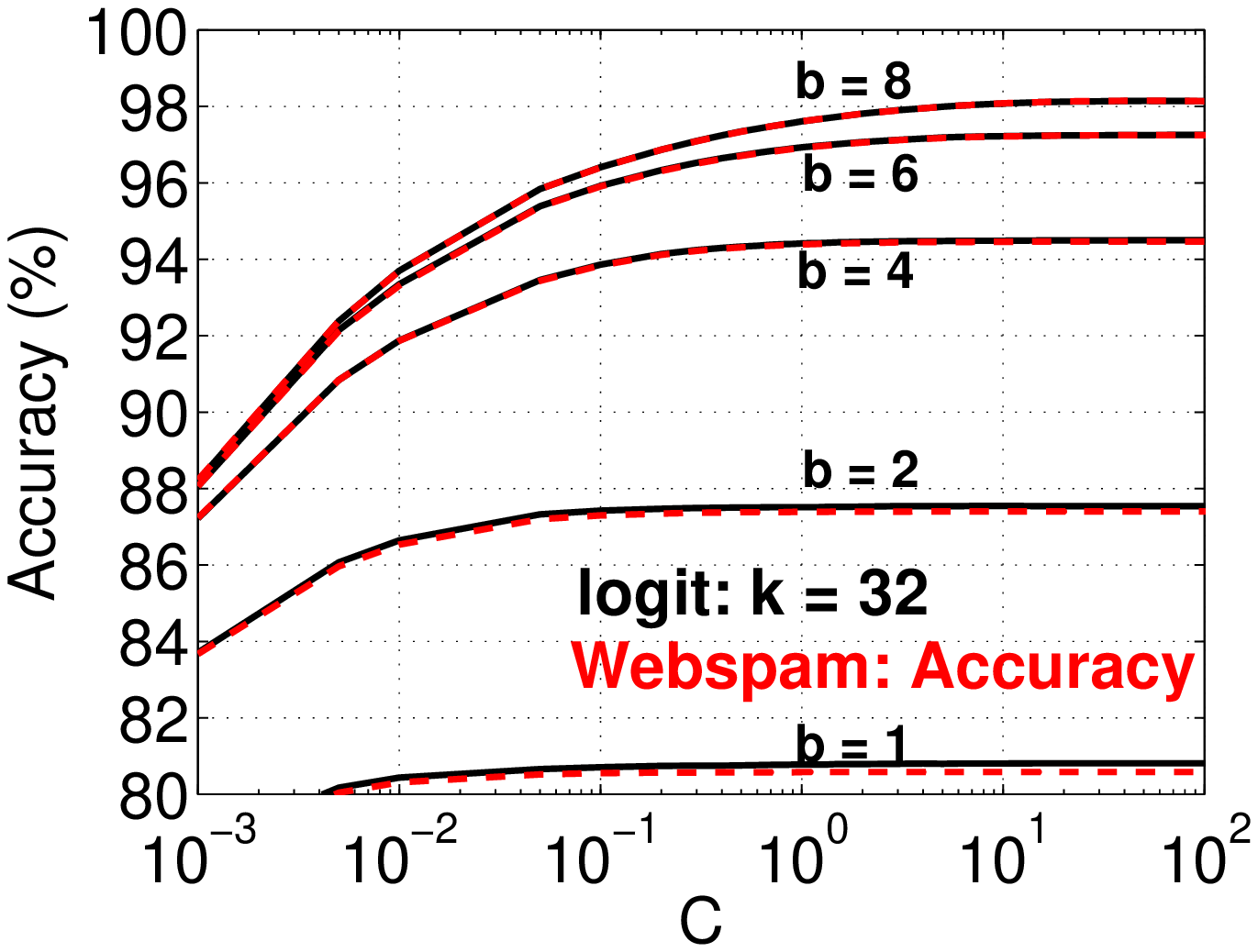}\hspace{-0.1in}
\includegraphics[width=1.5in]{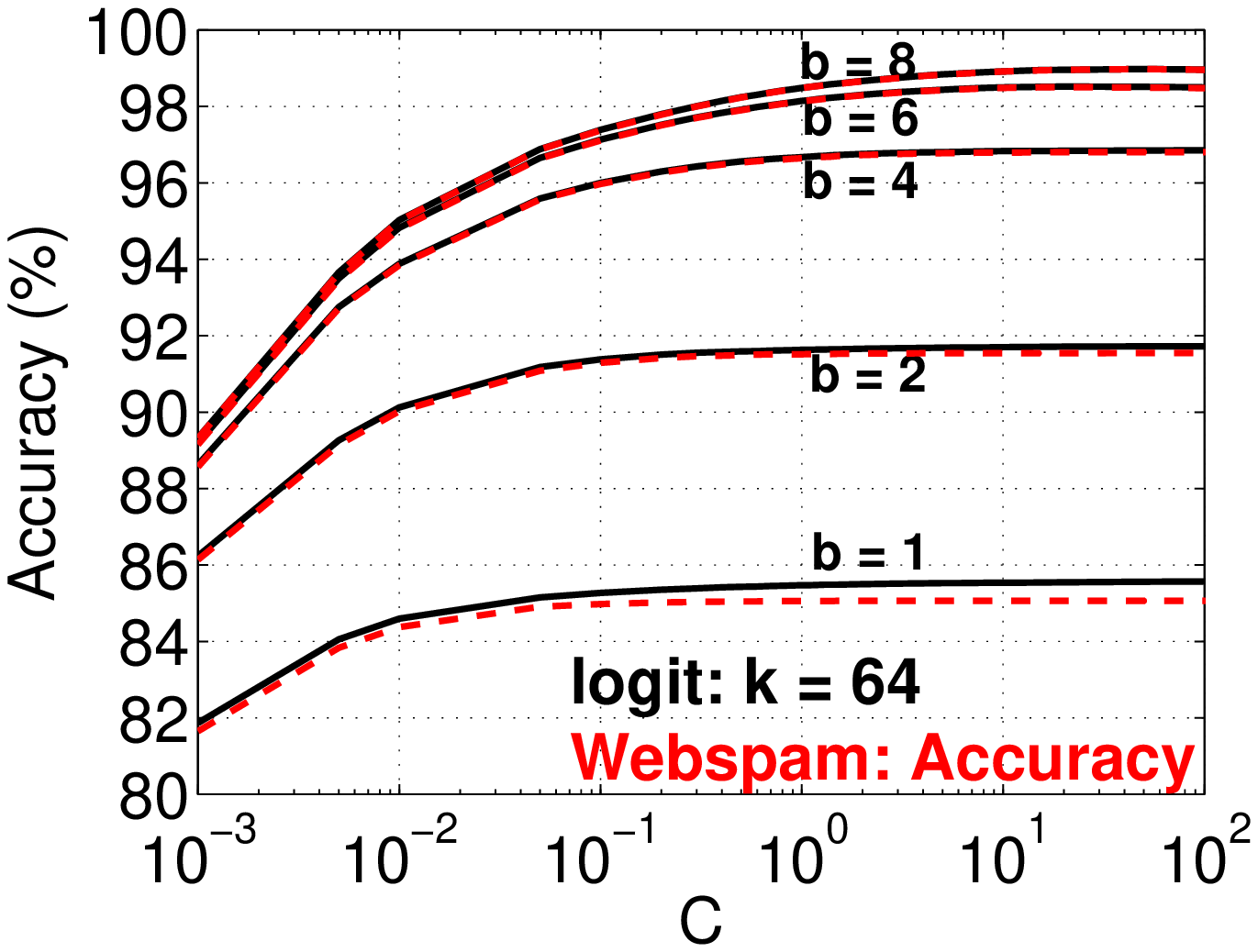}\hspace{-0.1in}
\includegraphics[width=1.5in]{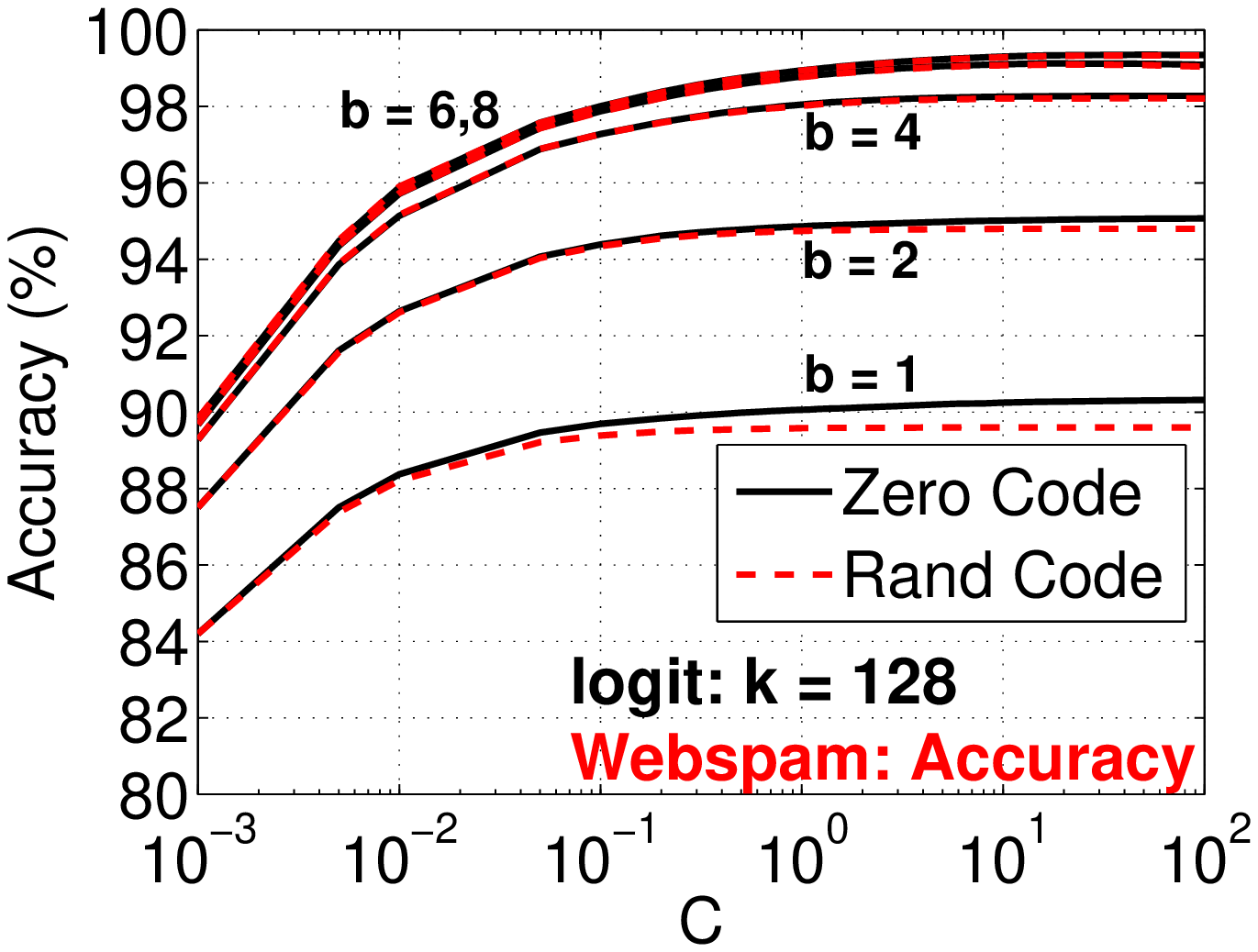}\hspace{-0.1in}
\includegraphics[width=1.5in]{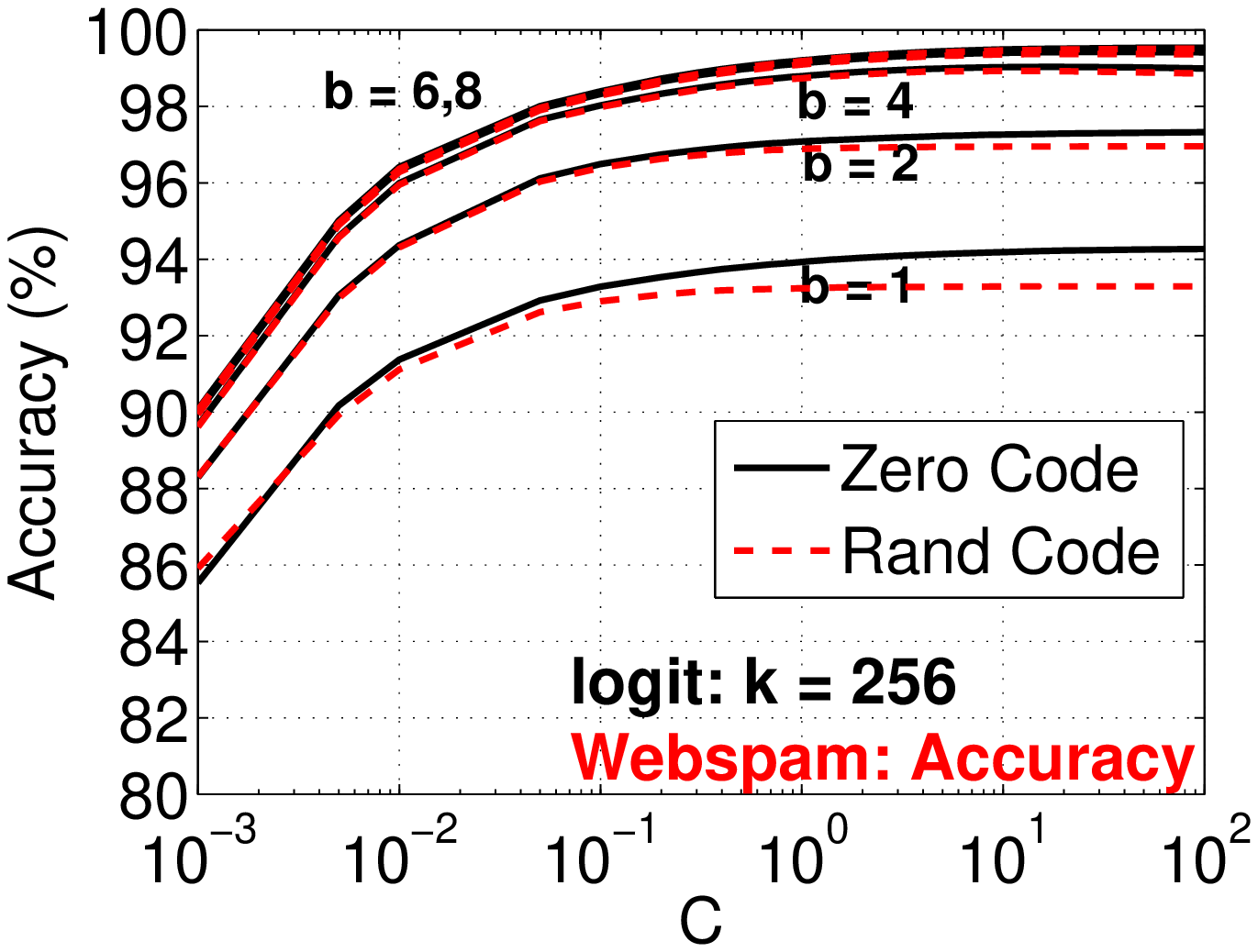}\hspace{-0.1in}
\includegraphics[width=1.5in]{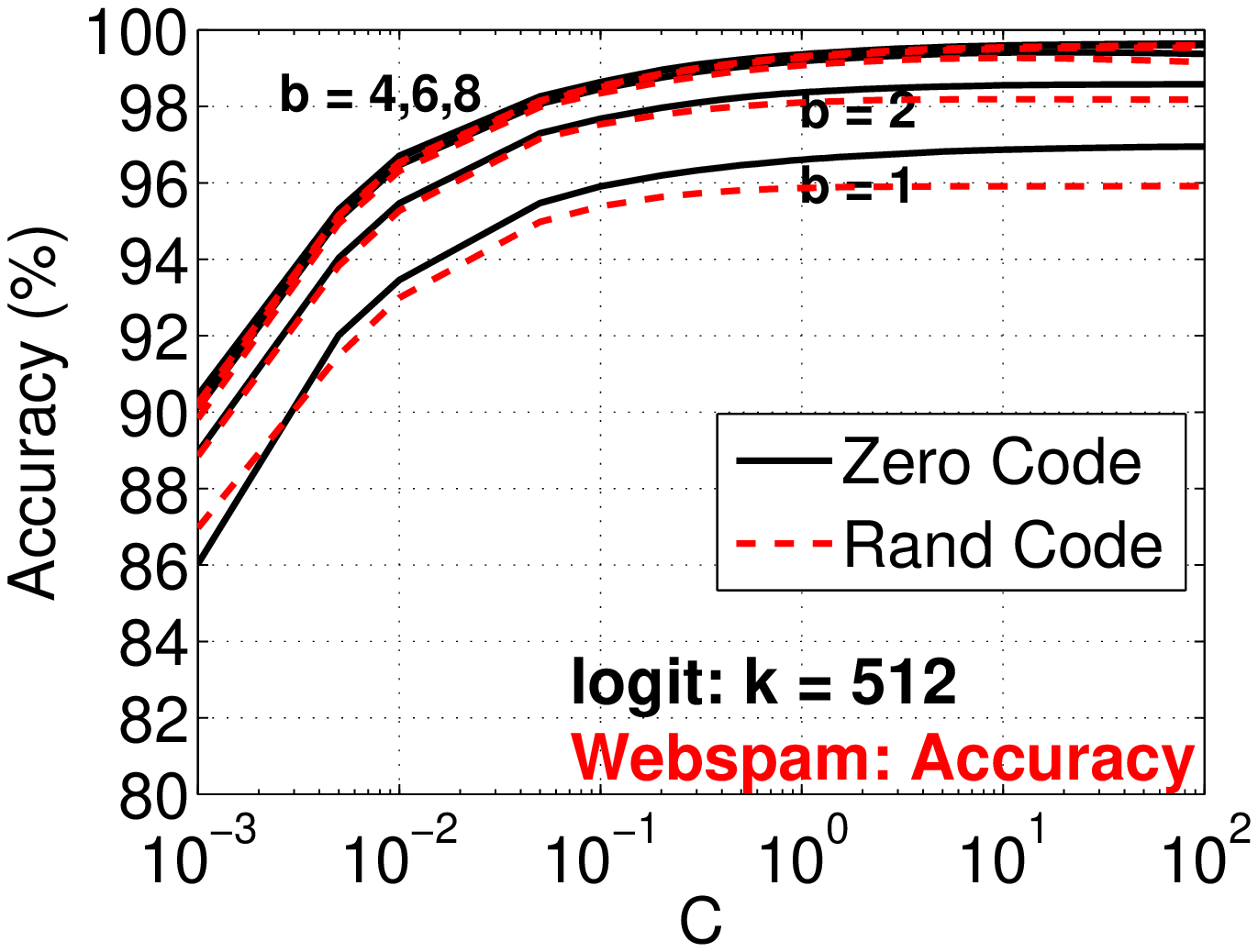}
}
\vspace{-0.3in}

\caption{Test accuracies of SVM (upper panels) and logistic regression (bottom panels),  averaged over 50 repetitions, for comparing the (recommended) zero coding strategy with the random coding strategy to deal with empty bins. We can see that the differences only become noticeable at $k=512$. }\label{fig_spam_accuracy_rand}
\end{figure}

The experimental results as shown in Figure~\ref{fig_spam_accuracy} are based on the ``zero coding'' strategy for dealing with empty bins.  Figure~\ref{fig_spam_accuracy_rand} plots the results for comparing  zero coding with the random coding.  When $k$ is large, zero coding is  superior to random coding, although  the differences remain small in this dataset. This is not surprising, of course. Random coding adds artificial nonzeros to the new (expanded) data matrix, which would not be desirable for learning algorithms.

\vspace{0.2in}

\noindent\textbf{Remark}:\ The empirical results on the {\em webspam} datasets are highly encouraging because they verify that our proposed one permutation hashing scheme works as well as (or even slightly better than) the original $k$-permutation scheme, at merely $1/k$ of the original preprocessing cost.  On the other hand, it would be more interesting, from the perspective of testing the robustness of our algorithm, to conduct experiments on a dataset where the empty bins will occur much more frequently.

\section{Experimental Results on the News20 Dataset}

The {\em news20} dataset (with 20,000 samples and 1,355,191 features) is a very small dataset in not-too-high dimensions. The average number of nonzeros per feature vector is about 500, which is also small. Therefore, this is more like a contrived example and we use it just to verify that our one permutation scheme (with the zero coding strategy) still works very well even when we let $k$ be as large as 4096 (i.e., most of the bins are empty). In fact, the one permutation schemes  achieves noticeably better accuracies than the original $k$-permutation scheme. We believe this is because the one permutation scheme is ``sample-without-replacement'' and provides a much better matrix sparsification strategy without ``contaminating'' the original data matrix too much.

\subsection{One Permutation  v.s. k-Permutation }

We experiment with $k \in \{2^3, 2^4, 2^5, 2^6, 2^7, 2^8, 2^9, 2^{10}, 2^{11}, 2^{12}\}$ and $b\in\{1, 2, 4, 6, 8\}$, for both  one permutation scheme and  $k$-permutation scheme. We use 10,000 samples for training and the other 10,000 samples for testing. For convenience, we let $D=2^{21}$ (which is larger than 1,355,191).

\begin{figure}[h!]
\mbox{
\includegraphics[width=1.5in]{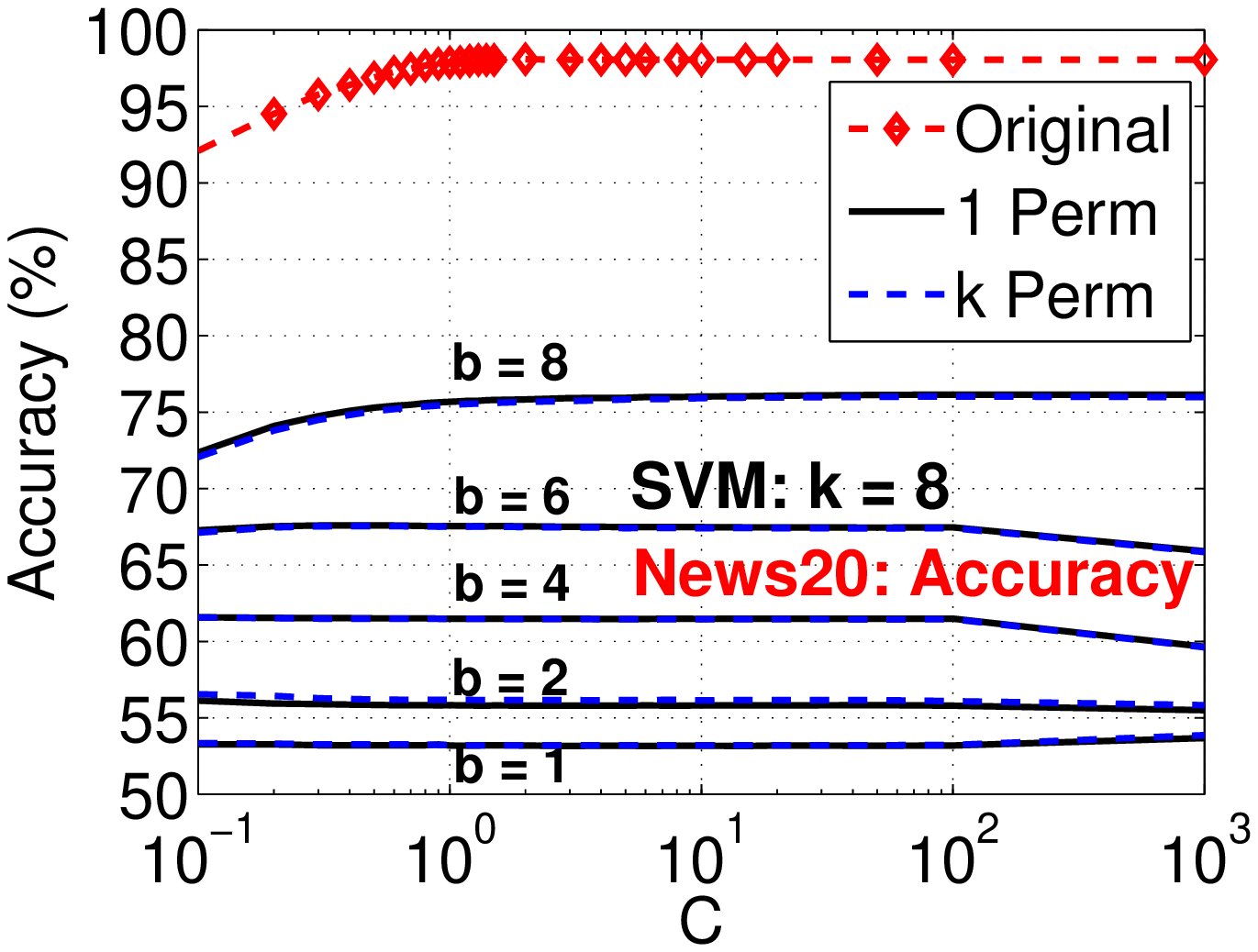}\hspace{-0.1in}
\includegraphics[width=1.5in]{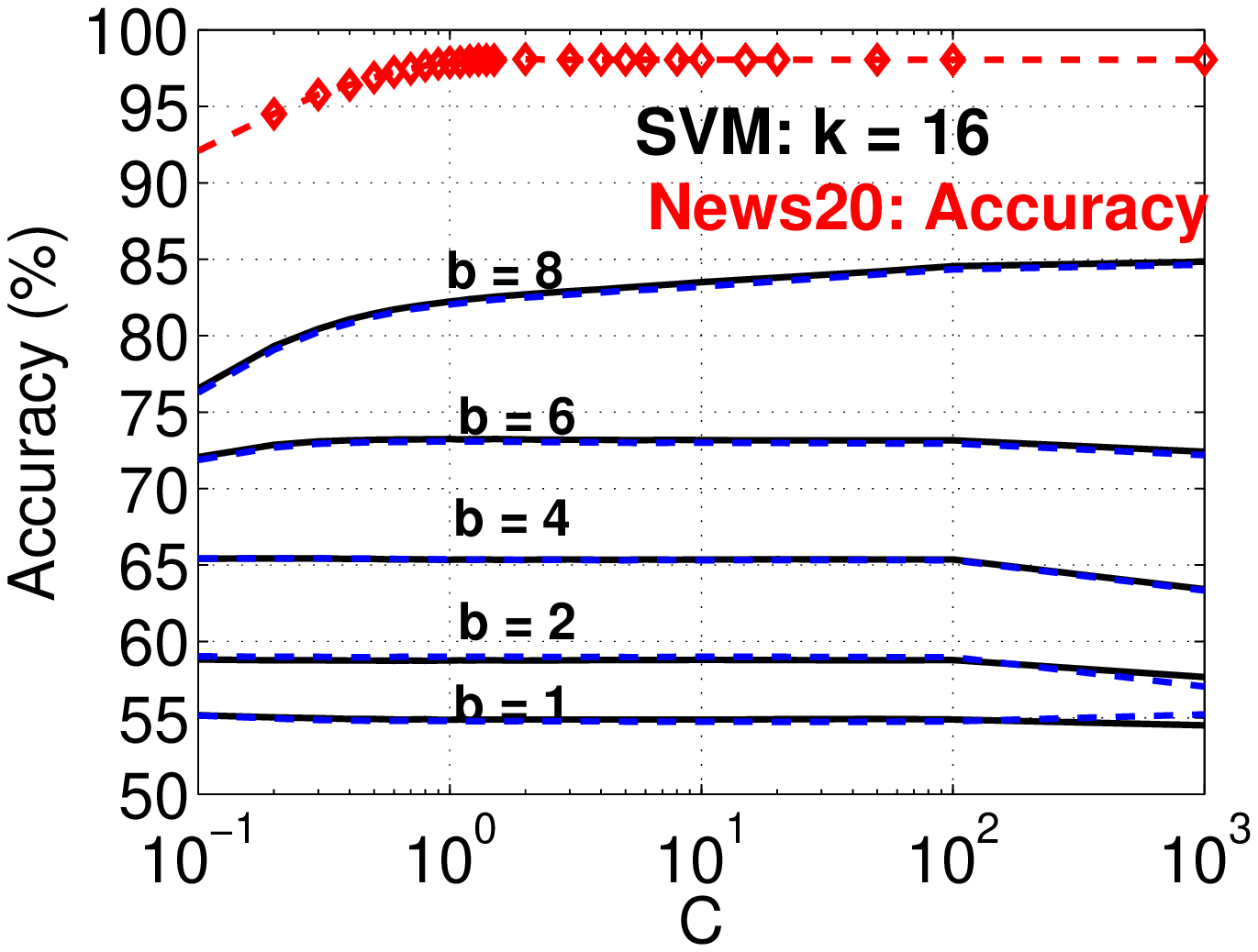}\hspace{-0.1in}
\includegraphics[width=1.5in]{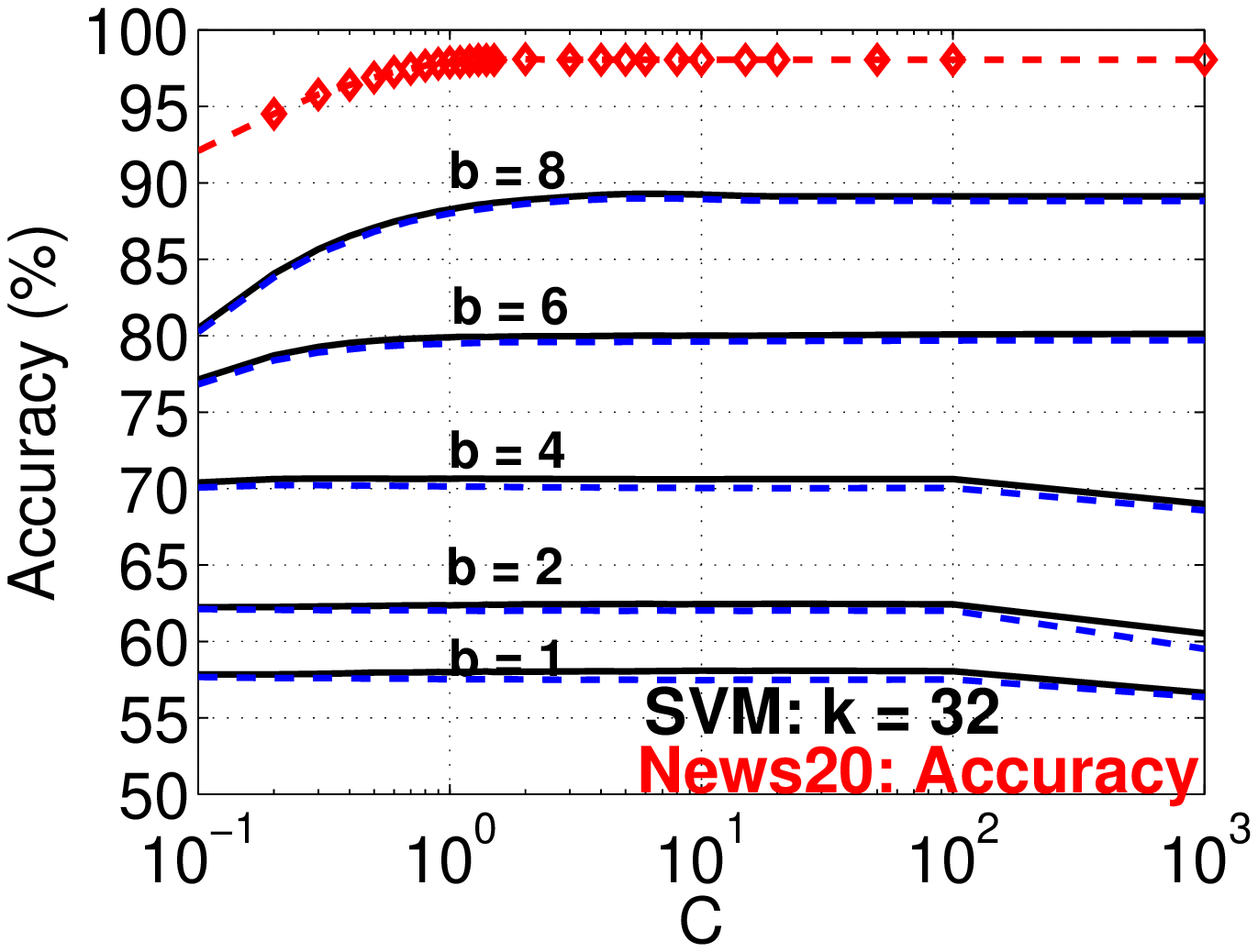}\hspace{-0.1in}
\includegraphics[width=1.5in]{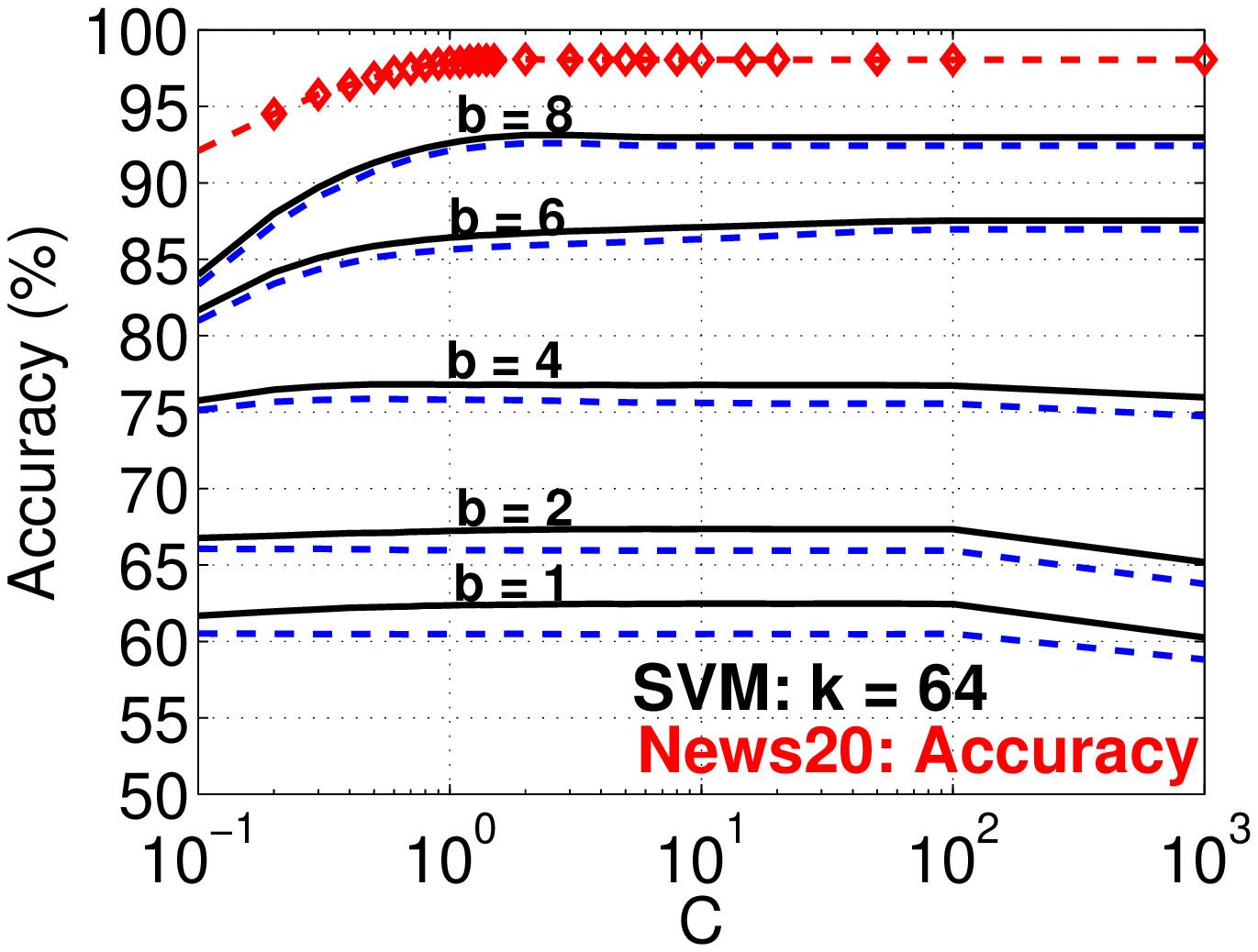}\hspace{-0.1in}
\includegraphics[width=1.5in]{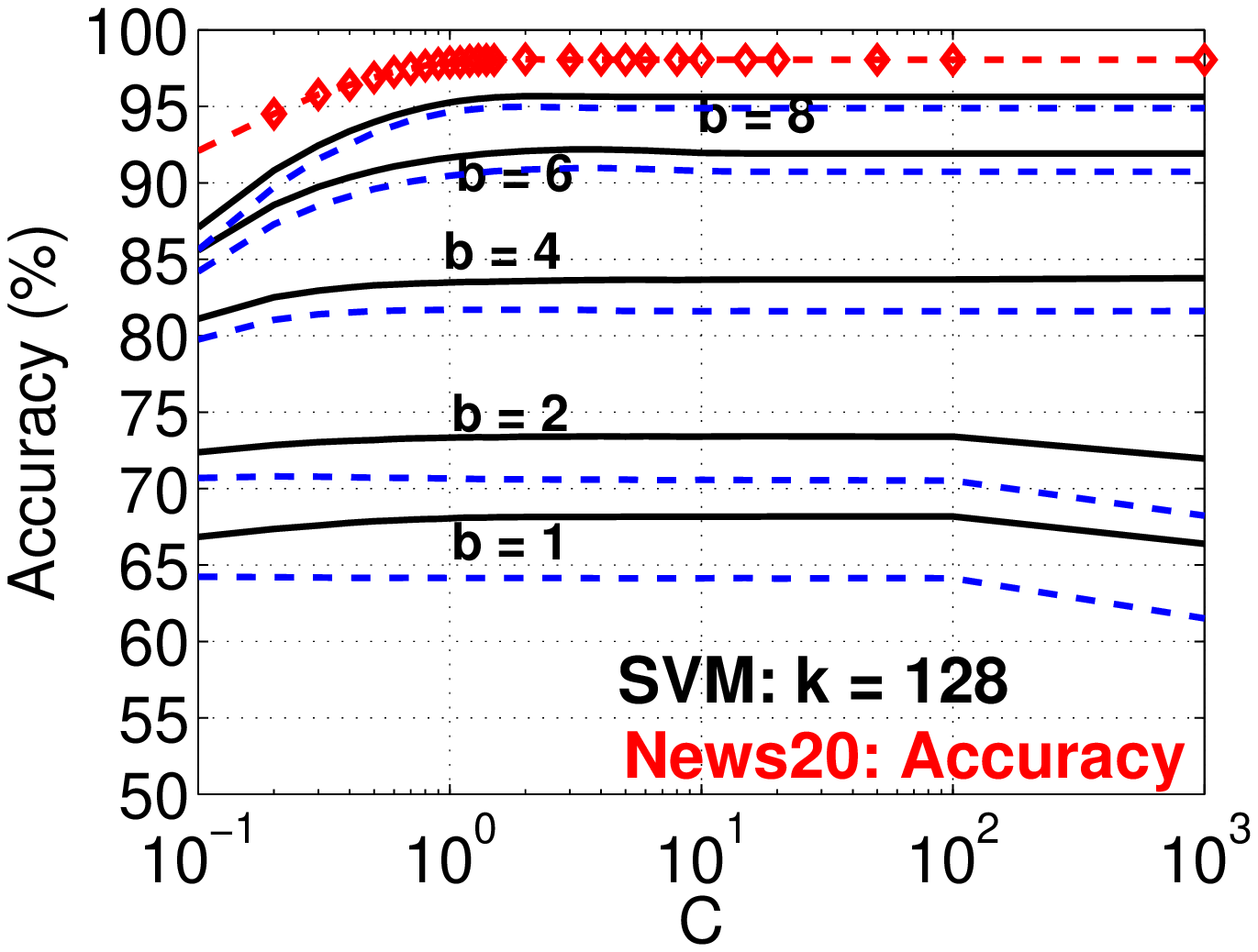}
}

\mbox{
\includegraphics[width=1.5in]{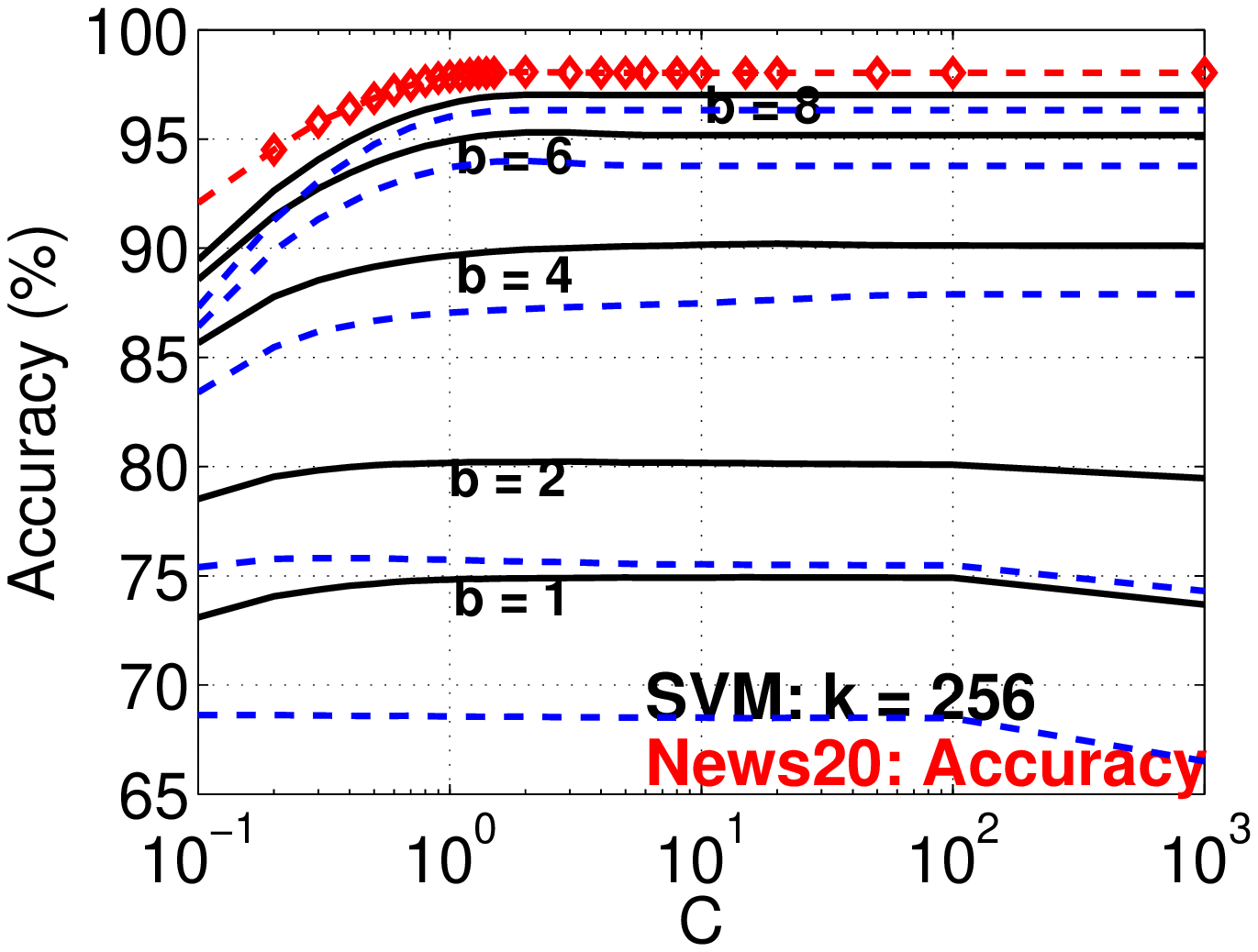}\hspace{-0.1in}
\includegraphics[width=1.5in]{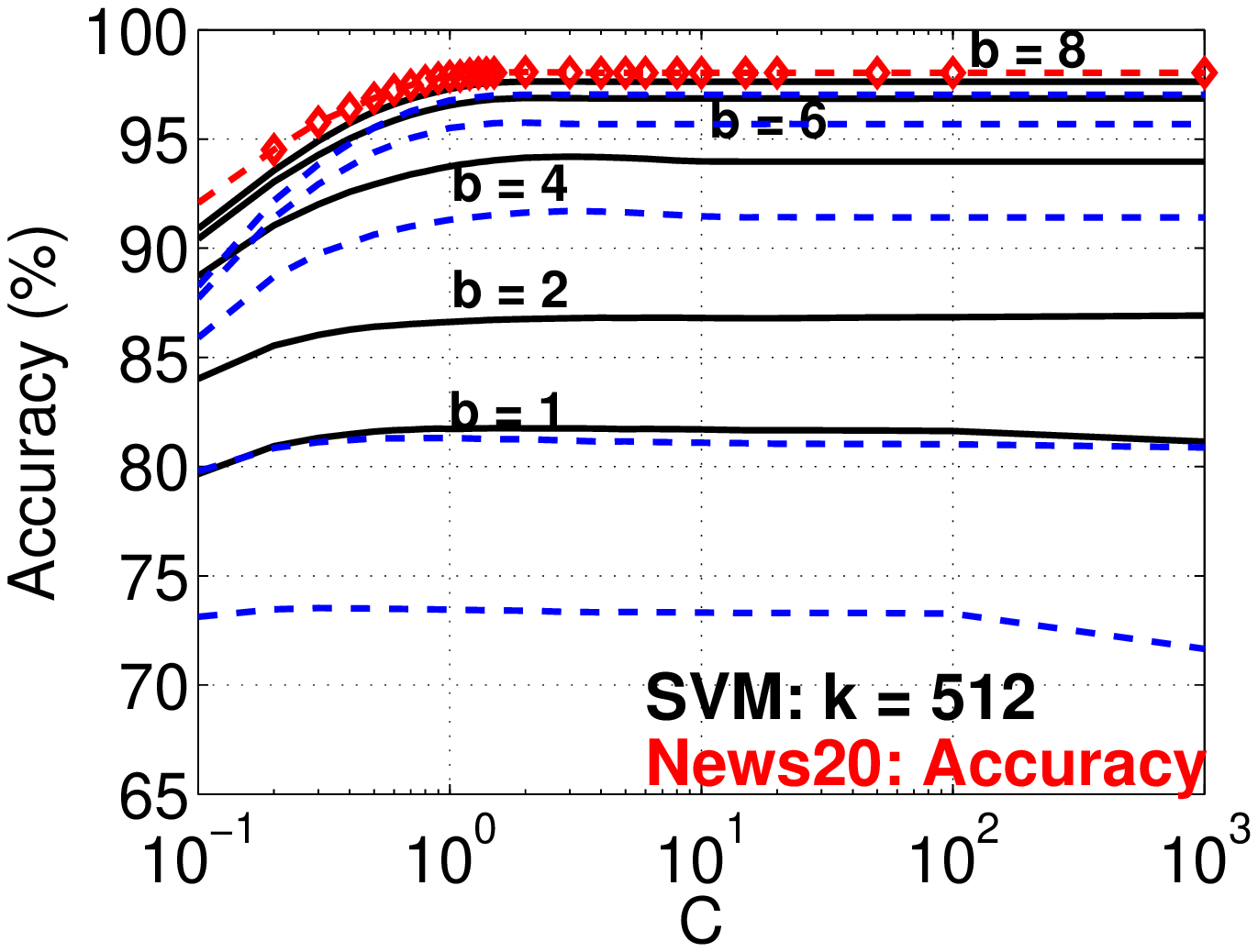}\hspace{-0.1in}
\includegraphics[width=1.5in]{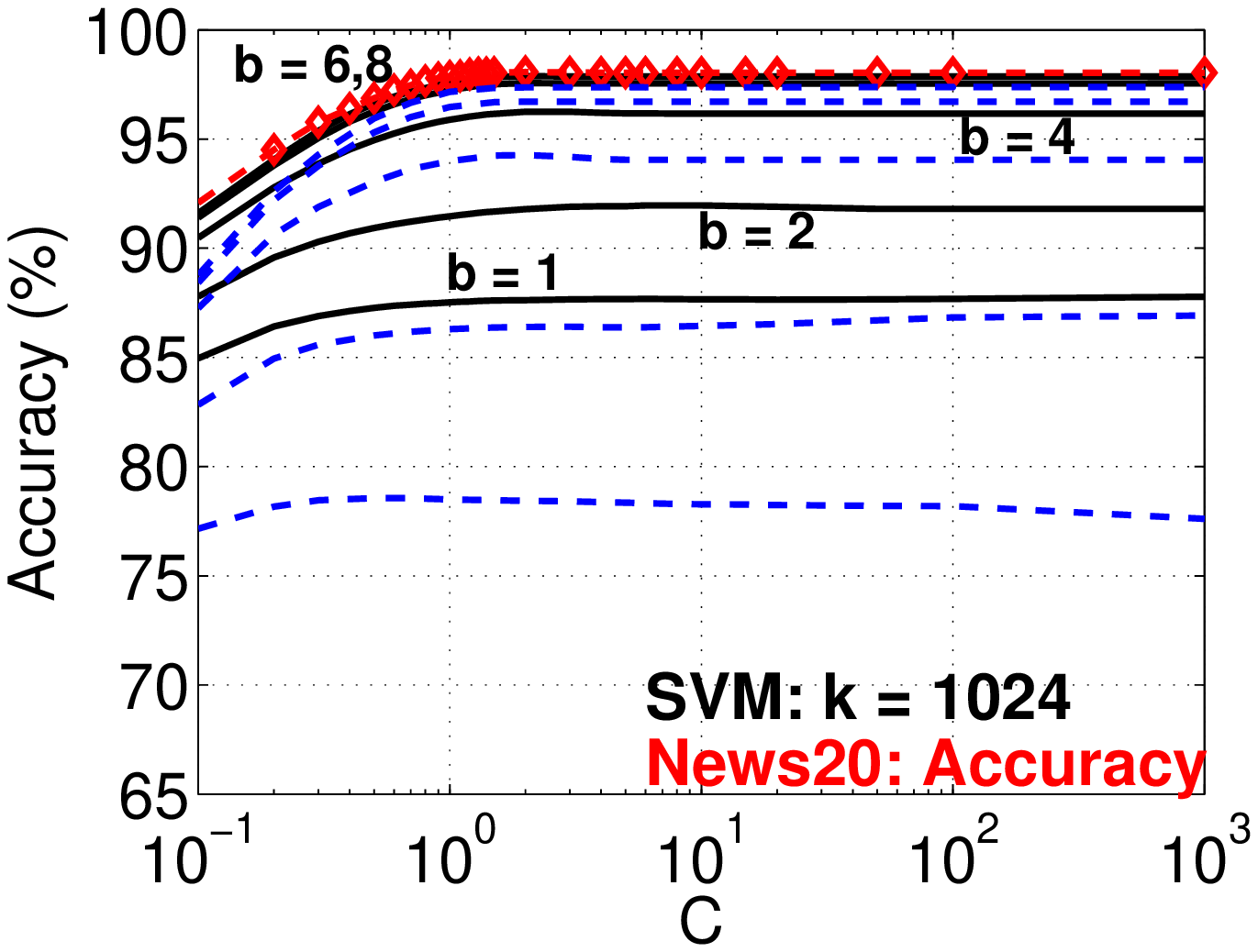}\hspace{-0.1in}
\includegraphics[width=1.5in]{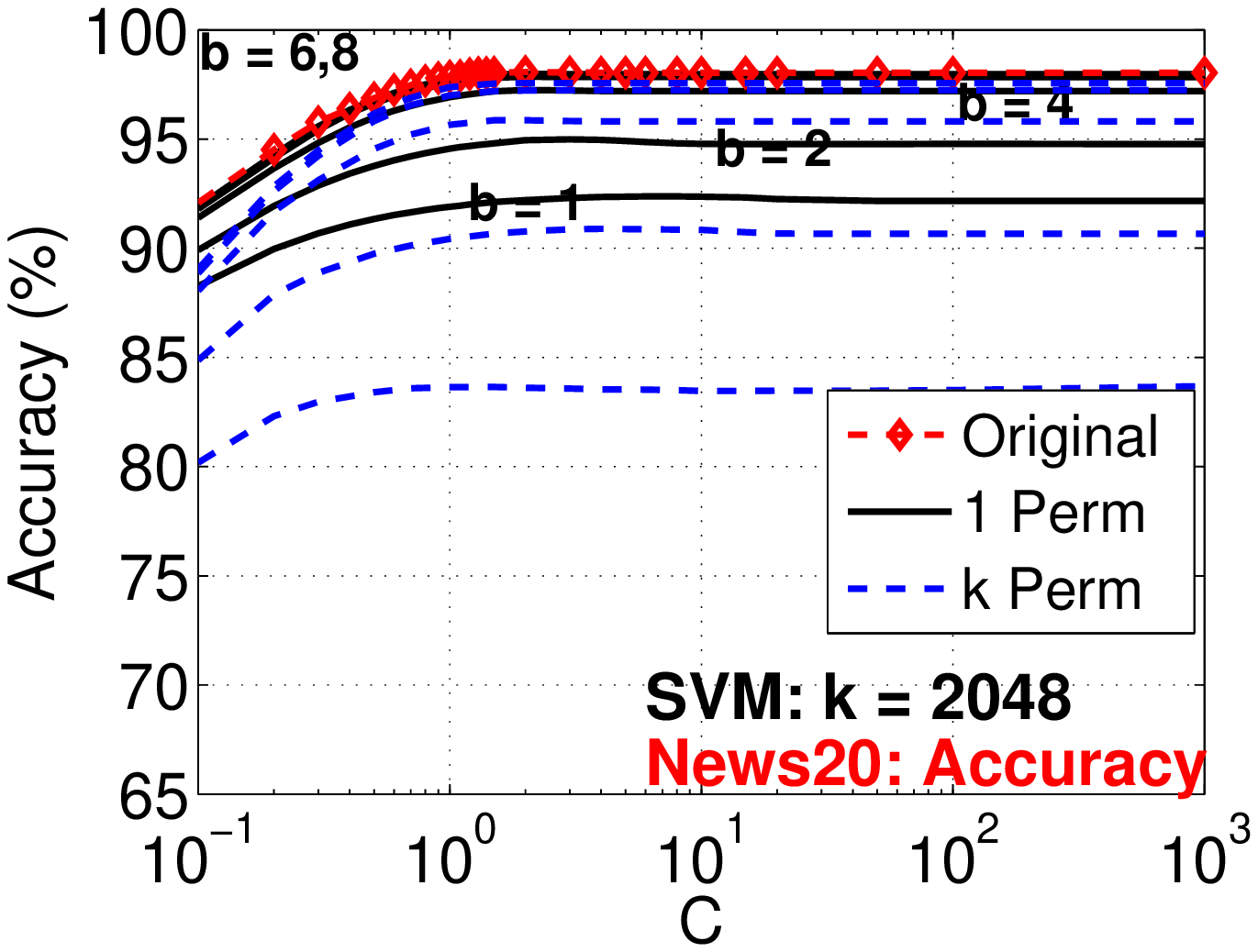}\hspace{-0.1in}
\includegraphics[width=1.5in]{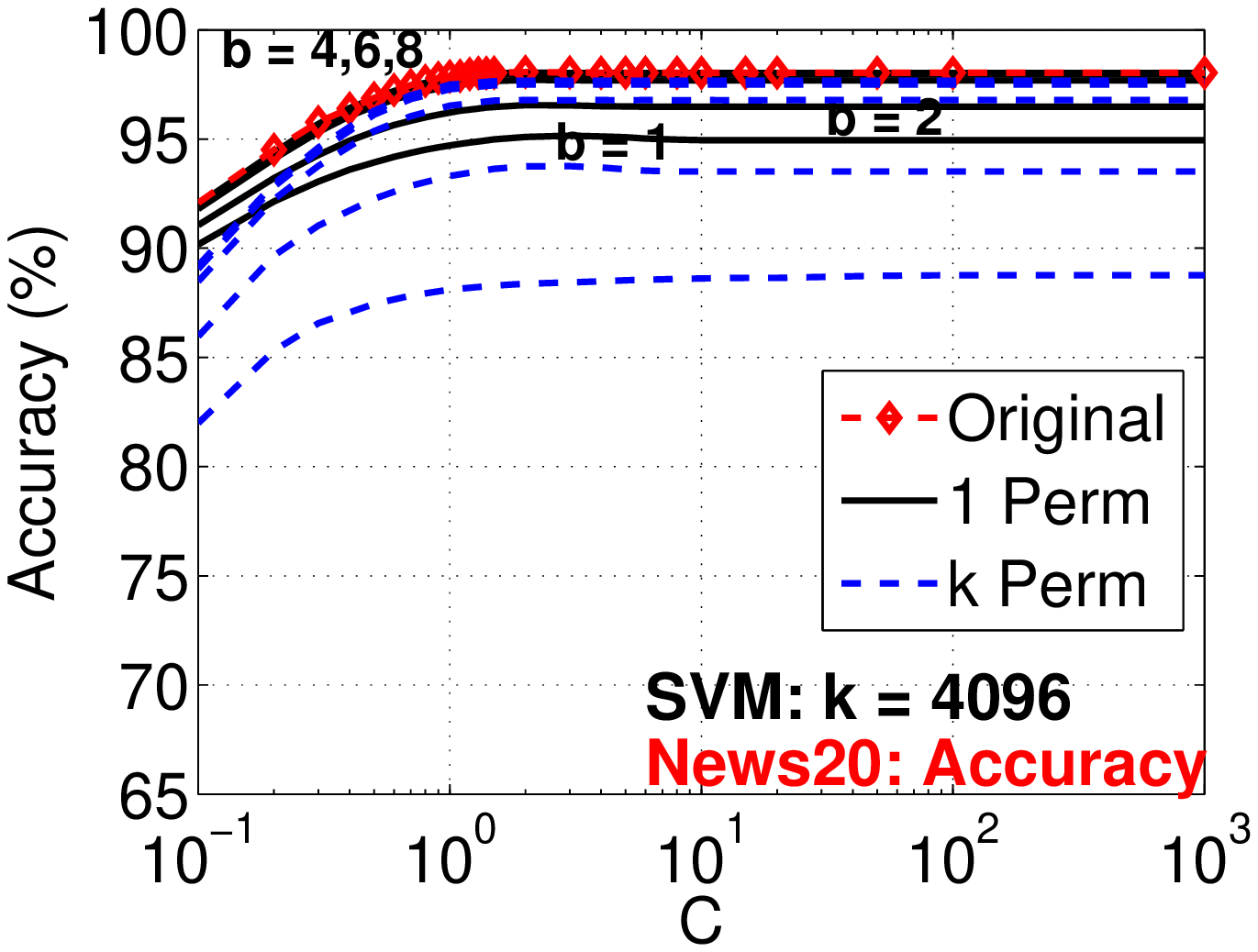}
}

\vspace{-0.2in}

\caption{Test accuracies of linear SVM  averaged over 100 repetitions. The proposed one permutation scheme noticeably outperforms the original $k$-permutation scheme especially when $k$ is not small.}\label{fig_news20_accuracy_svm}
\end{figure}

\begin{figure}[h!]
\mbox{
\includegraphics[width=1.5in]{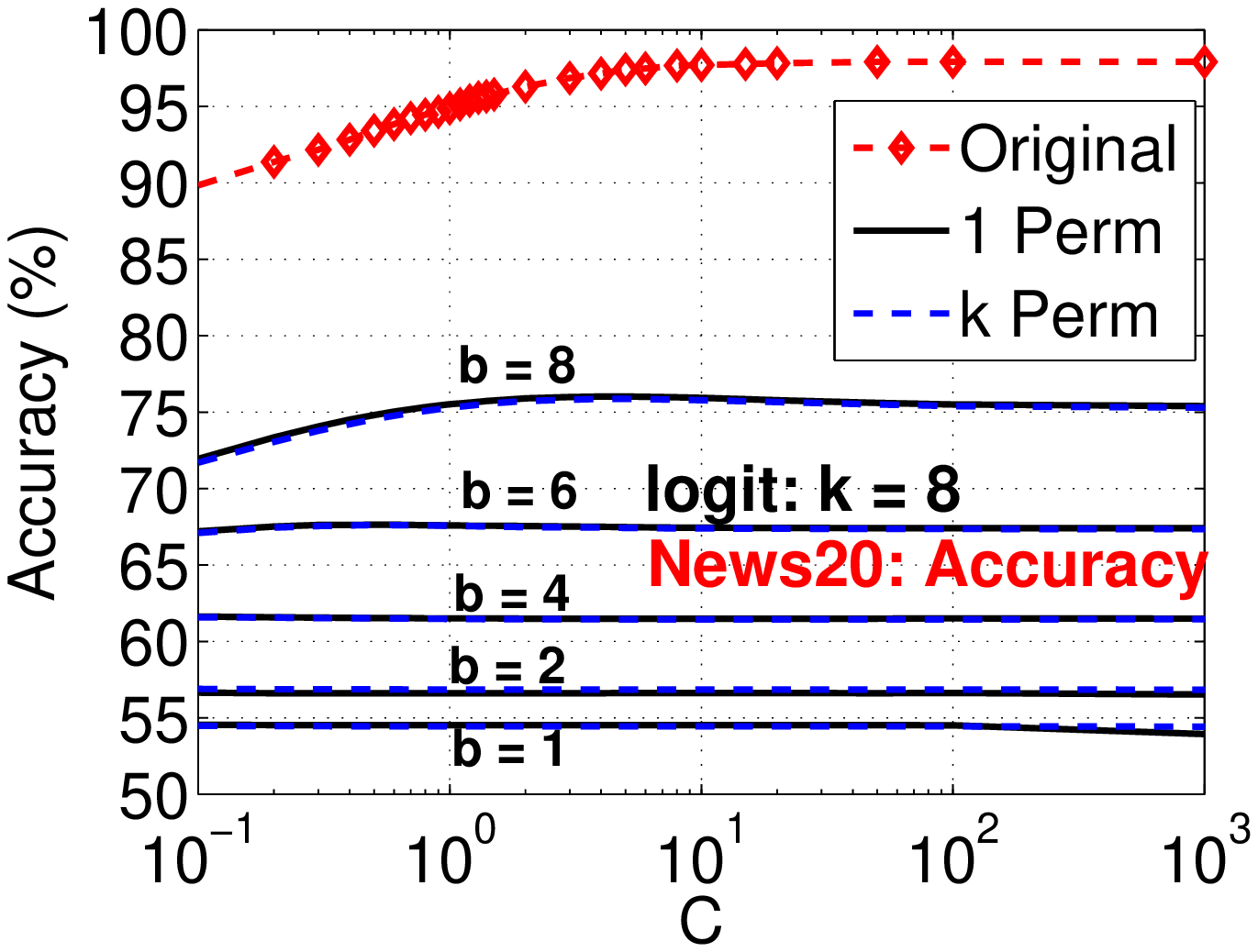}\hspace{-0.1in}
\includegraphics[width=1.5in]{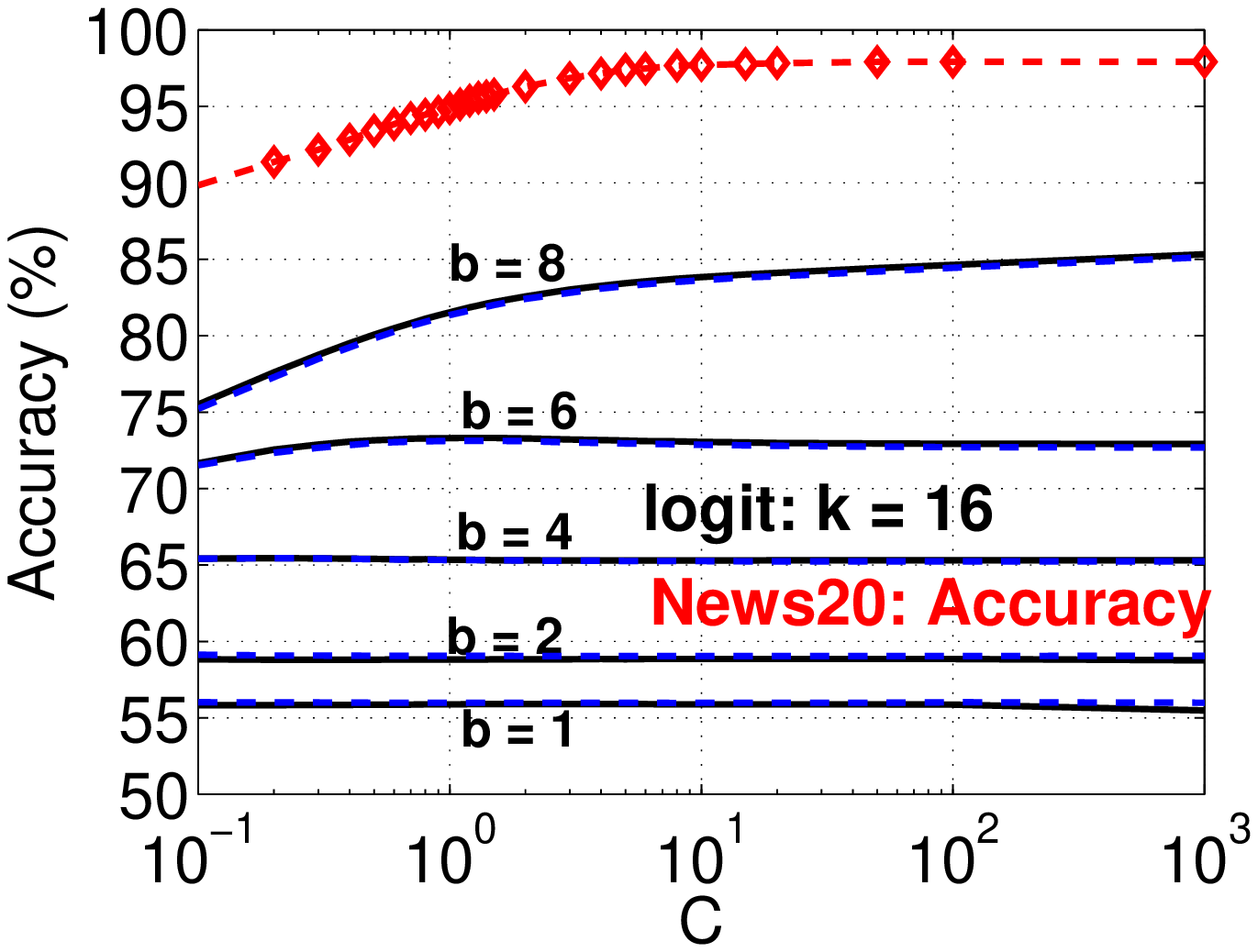}\hspace{-0.1in}
\includegraphics[width=1.5in]{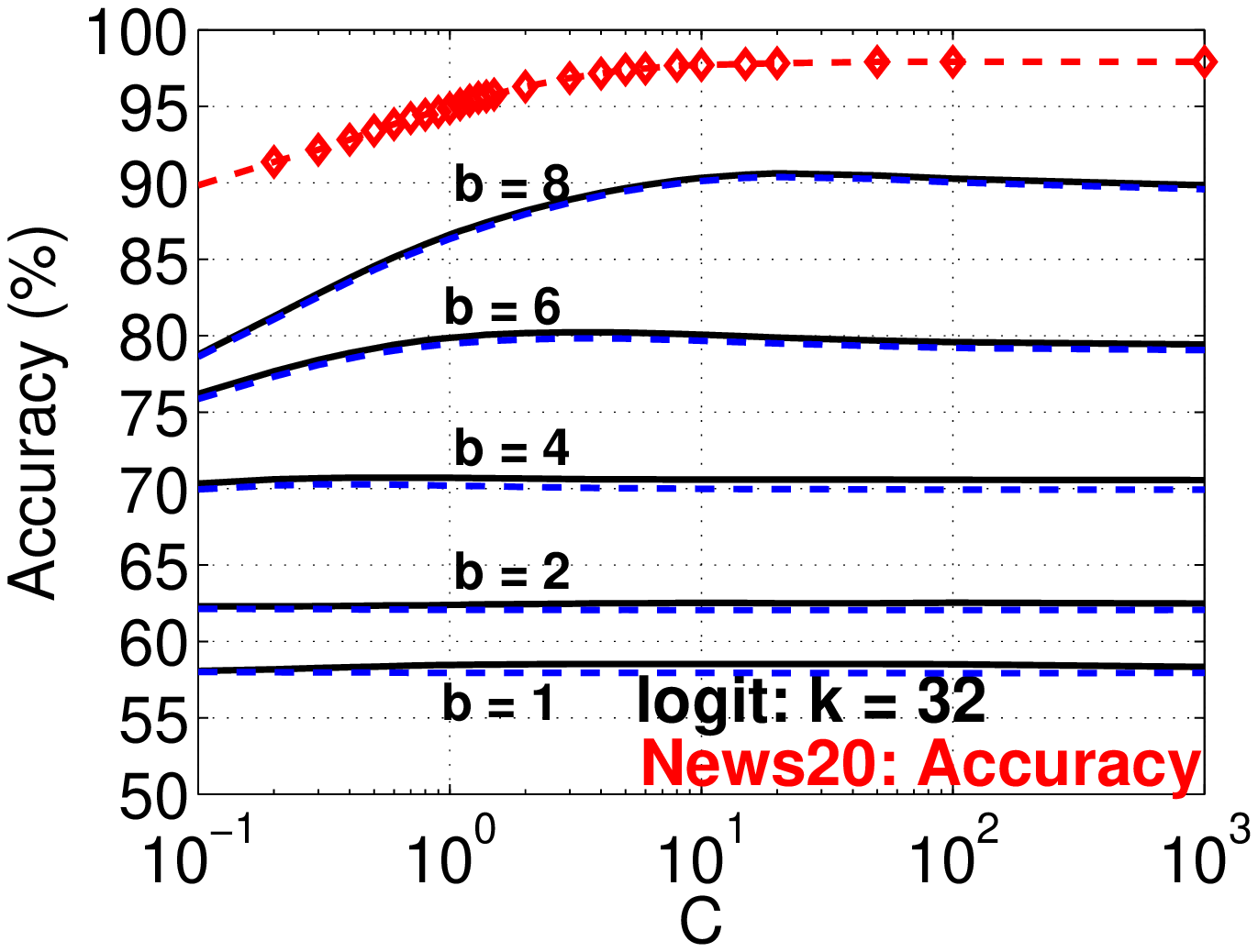}\hspace{-0.1in}
\includegraphics[width=1.5in]{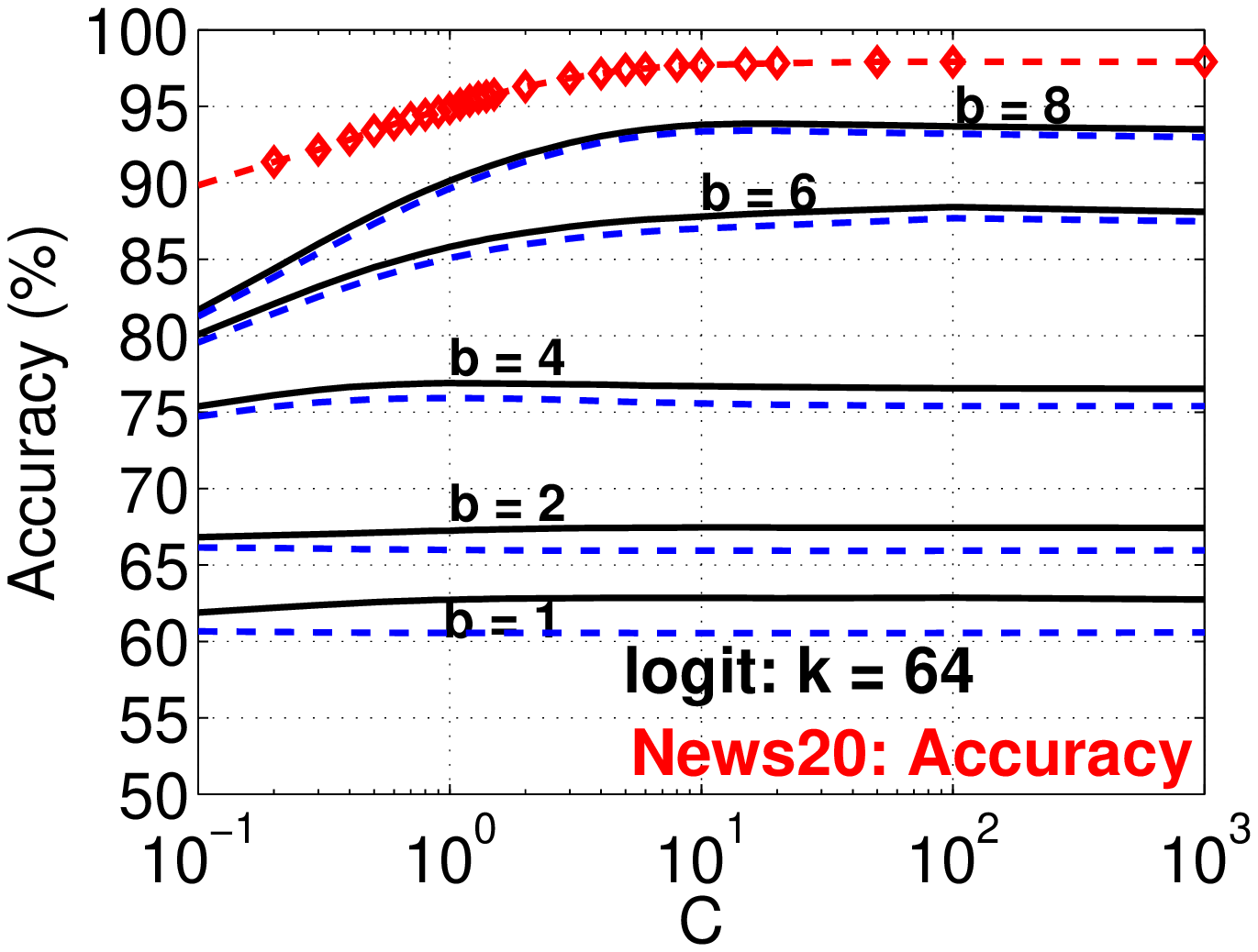}\hspace{-0.1in}
\includegraphics[width=1.5in]{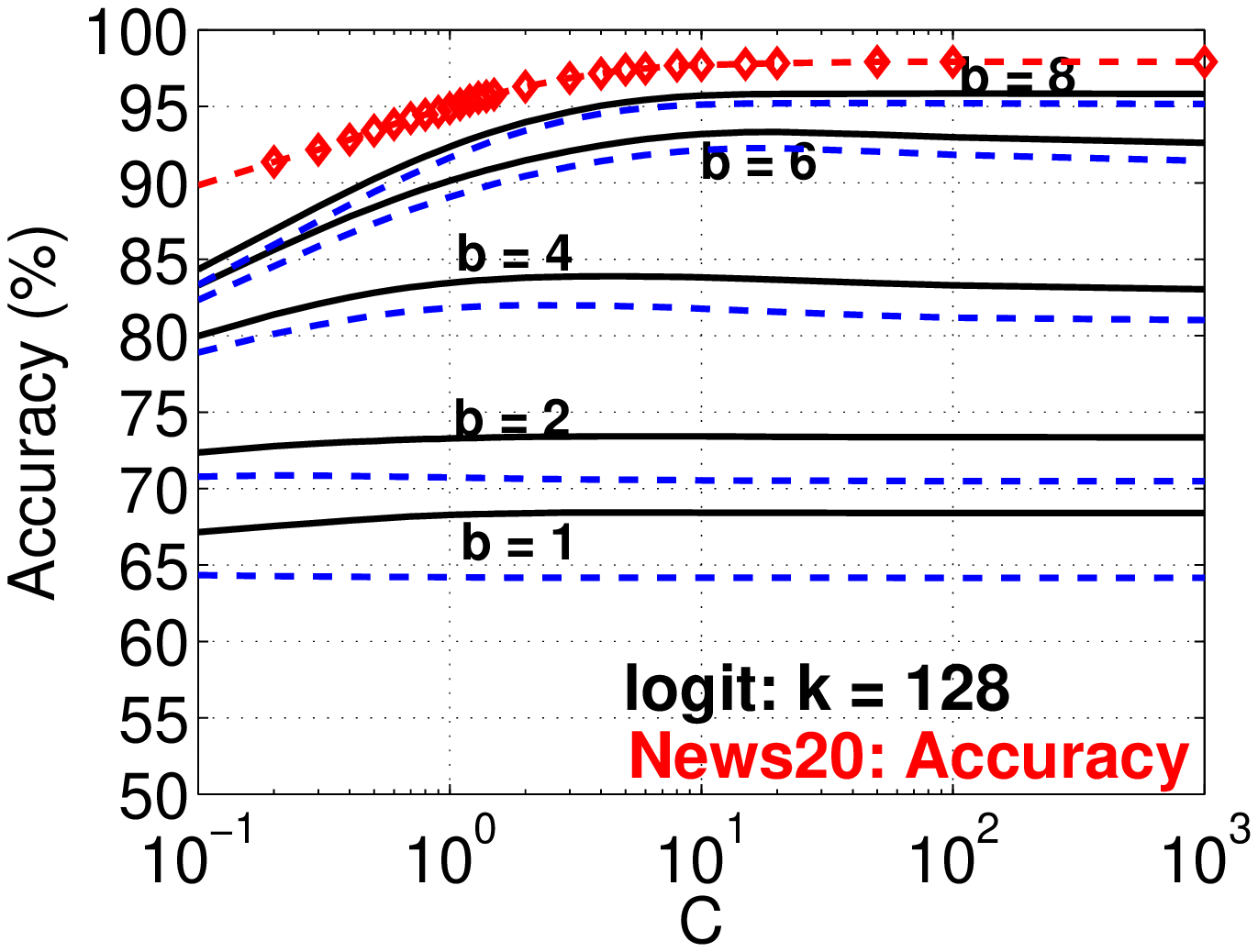}
}

\mbox{
\includegraphics[width=1.5in]{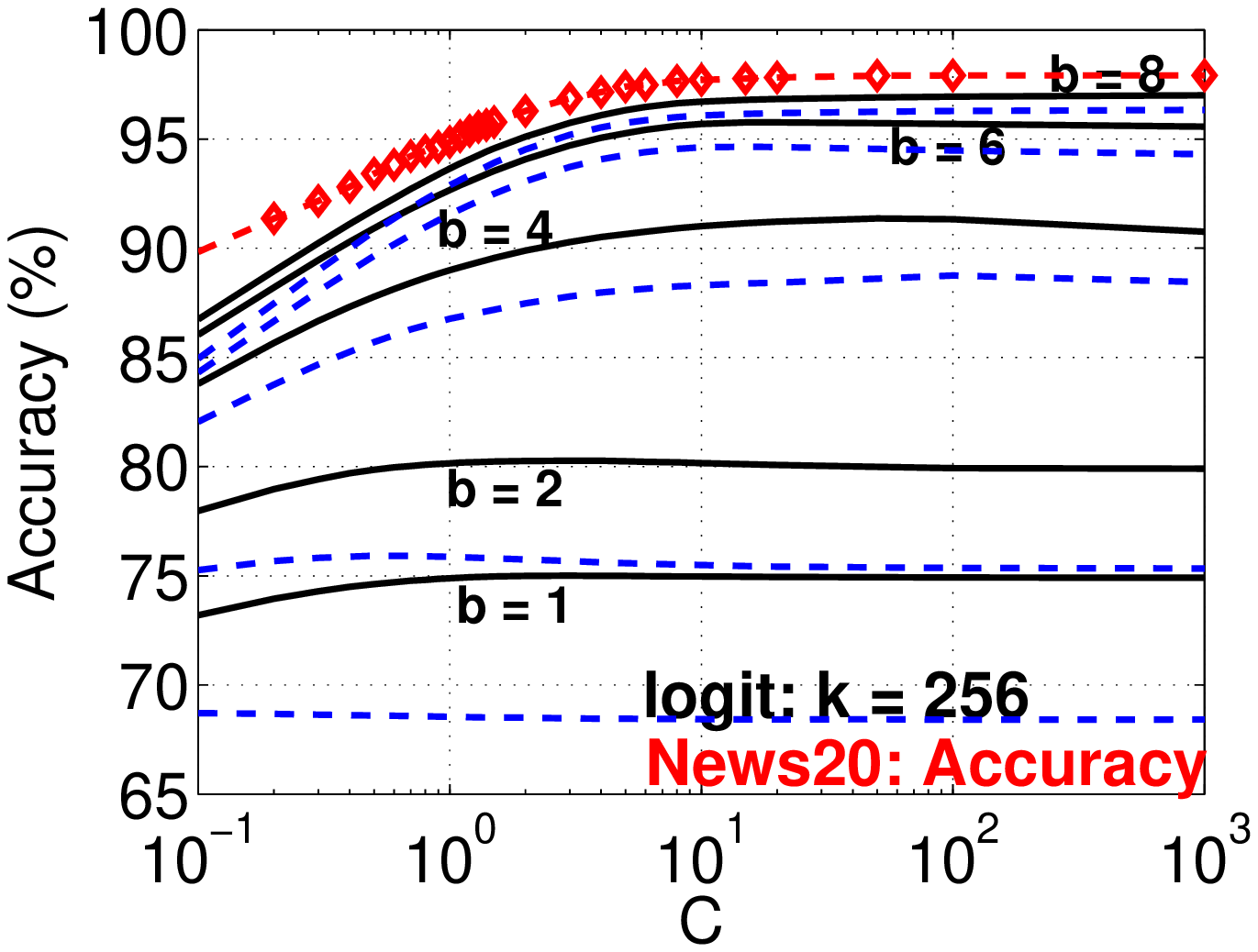}\hspace{-0.1in}
\includegraphics[width=1.5in]{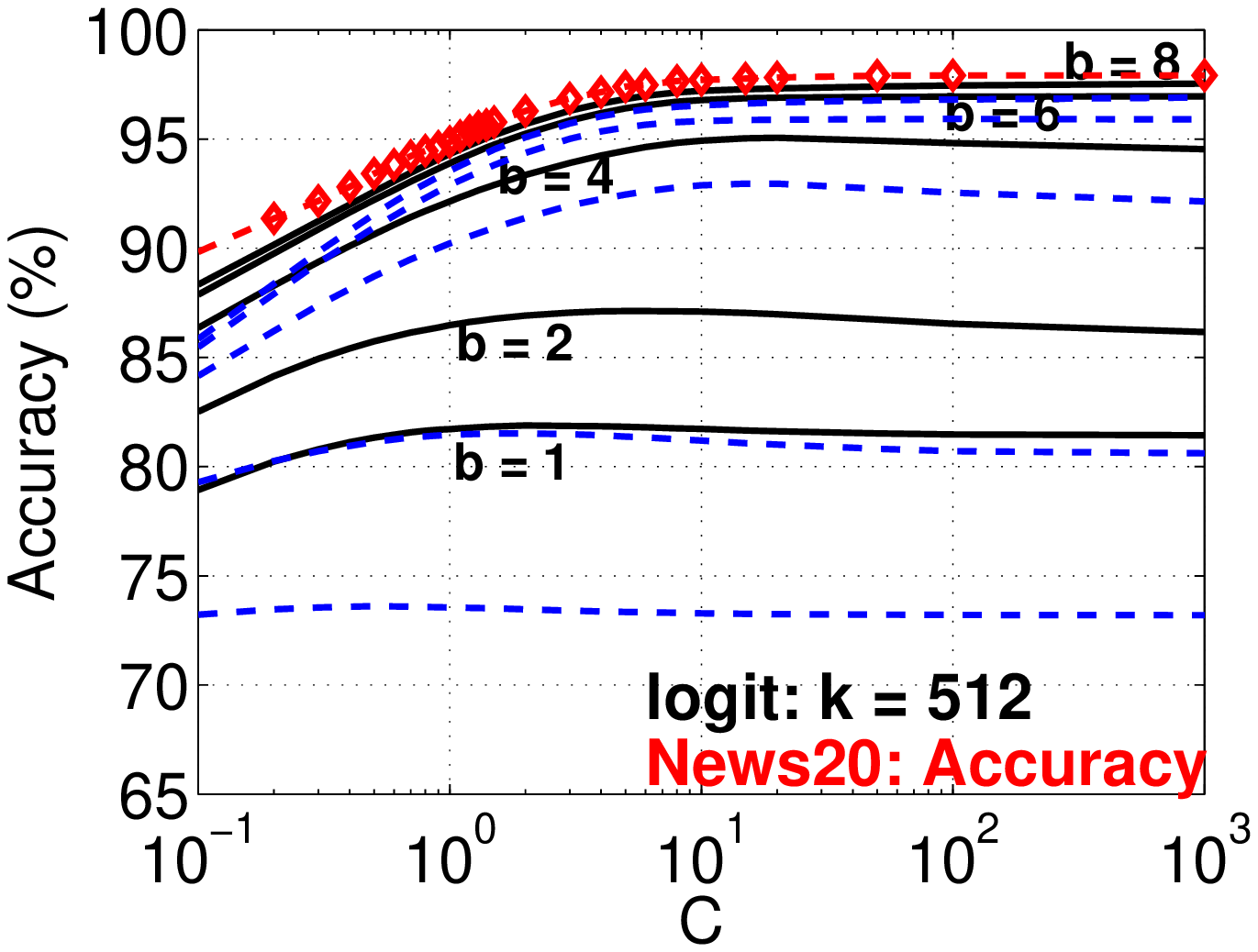}\hspace{-0.1in}
\includegraphics[width=1.5in]{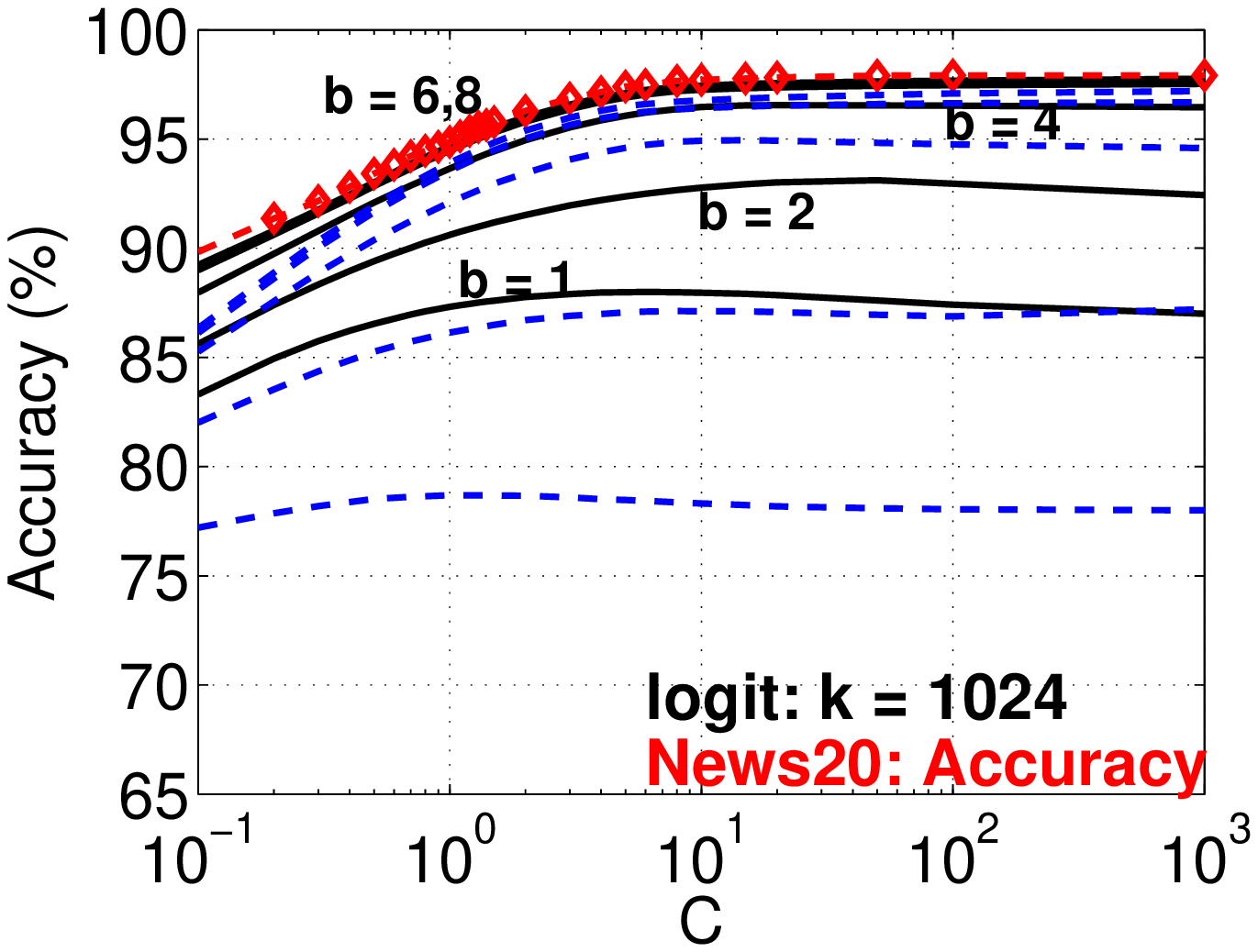}\hspace{-0.1in}
\includegraphics[width=1.5in]{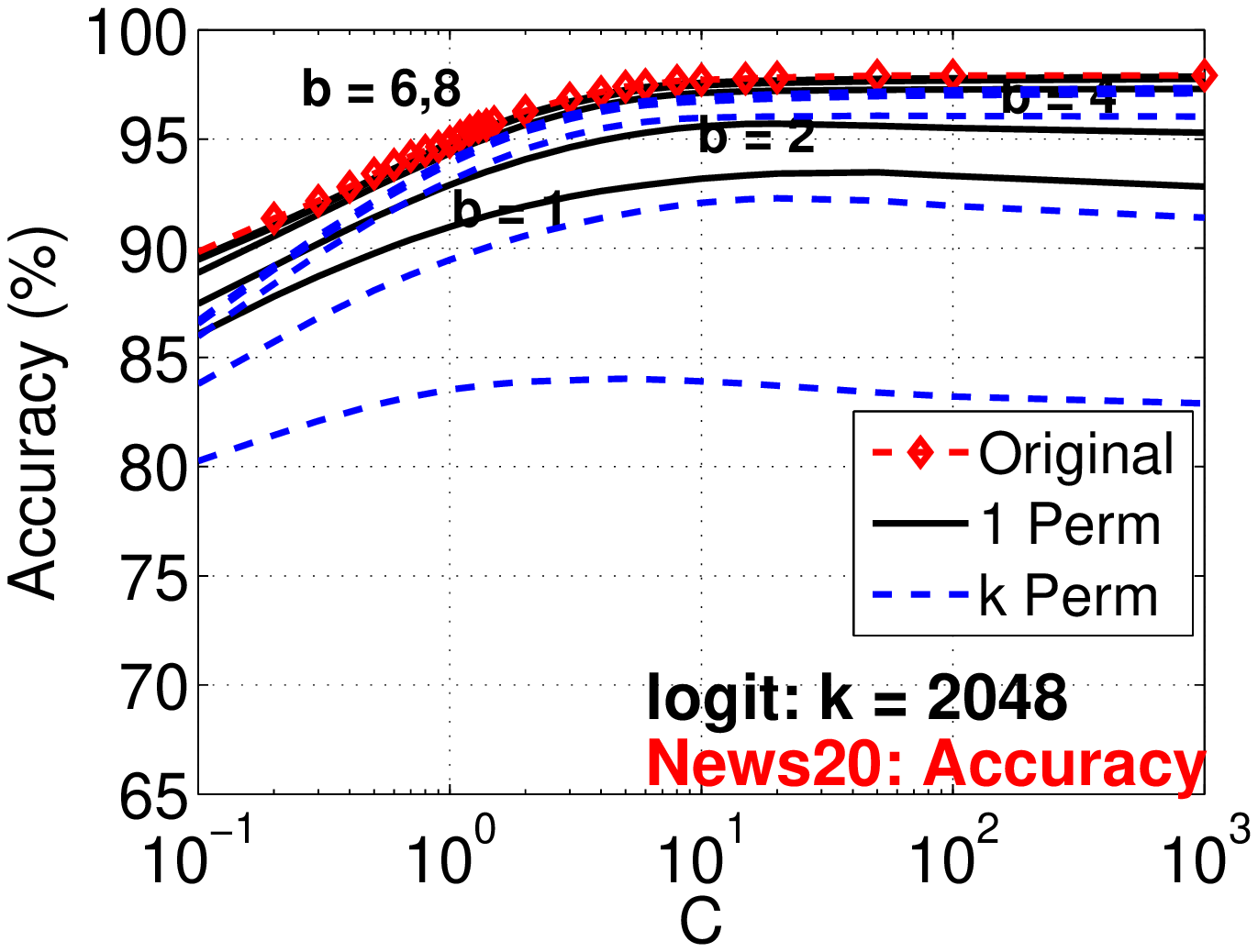}\hspace{-0.1in}
\includegraphics[width=1.5in]{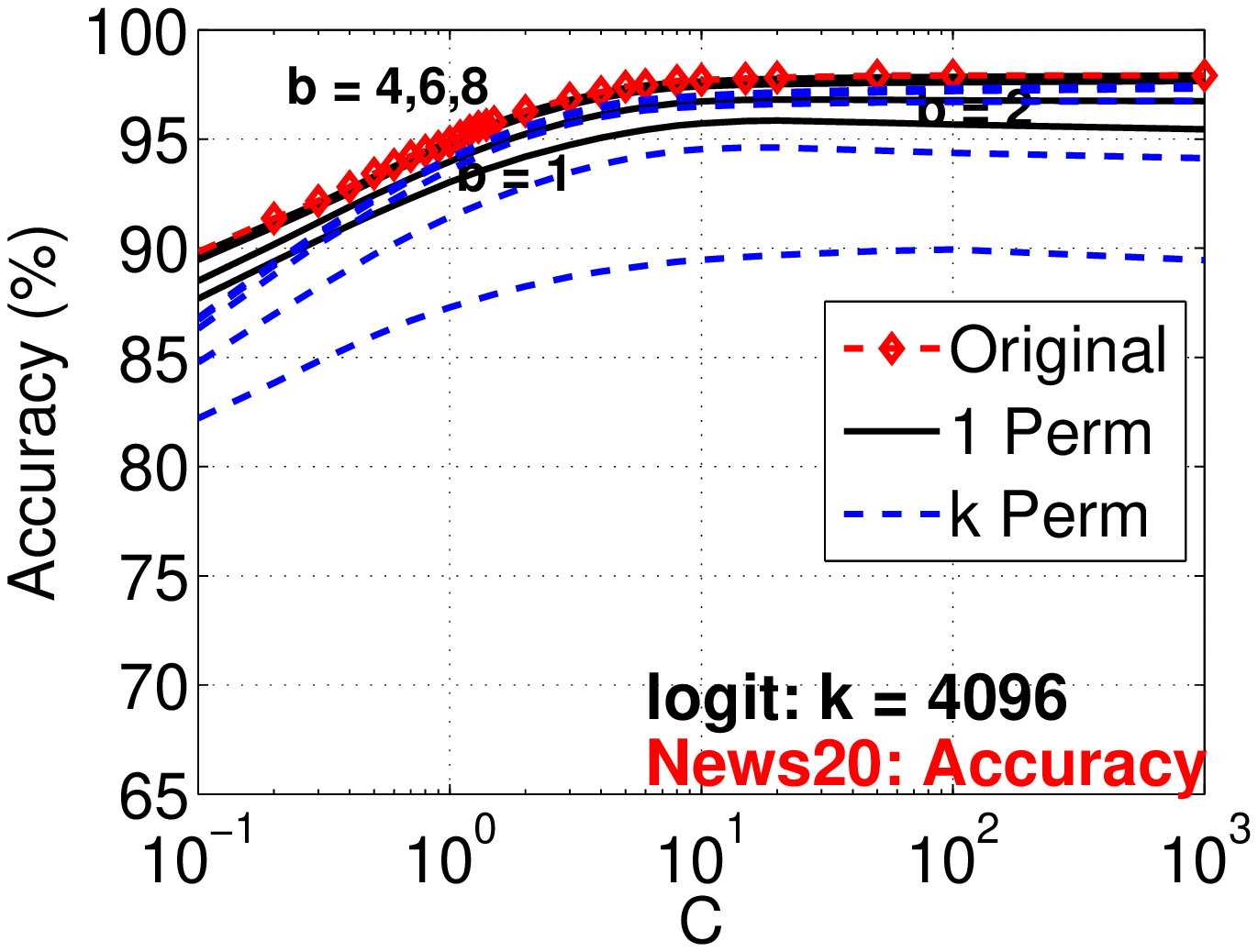}
}

\vspace{-0.2in}

\caption{Test accuracies of logistic regression  averaged over 100 repetitions.  The proposed one permutation scheme noticeably outperforms the original $k$-permutation scheme especially when $k$ is not small.}\label{fig_news20_accuracy_logit}
\end{figure}

Figure~\ref{fig_news20_accuracy_svm} and Figure~\ref{fig_news20_accuracy_logit} present the test accuracies for linear SVM and logistic regression, respectively. When $k$ is small (e.g., $k\leq 64$) both the one permutation scheme and  the original $k$-permutation scheme perform similarly. For larger $k$ values (especially as $k\geq 256$), however, our one permutation scheme noticeably outperforms the  $k$-permutation scheme. Using the original data, the test accuracies are about $98\%$. Our one permutation scheme with $k\geq 512$ and $b=8$ essentially achieves the original test accuracies, while the $k$-permutation scheme could only reach about $97.5\%$ even with $k=4096$.

\subsection{Zero Coding v.s. Random Coding for Empty Bins}

\begin{figure}[h!]
\mbox{
\includegraphics[width=1.5in]{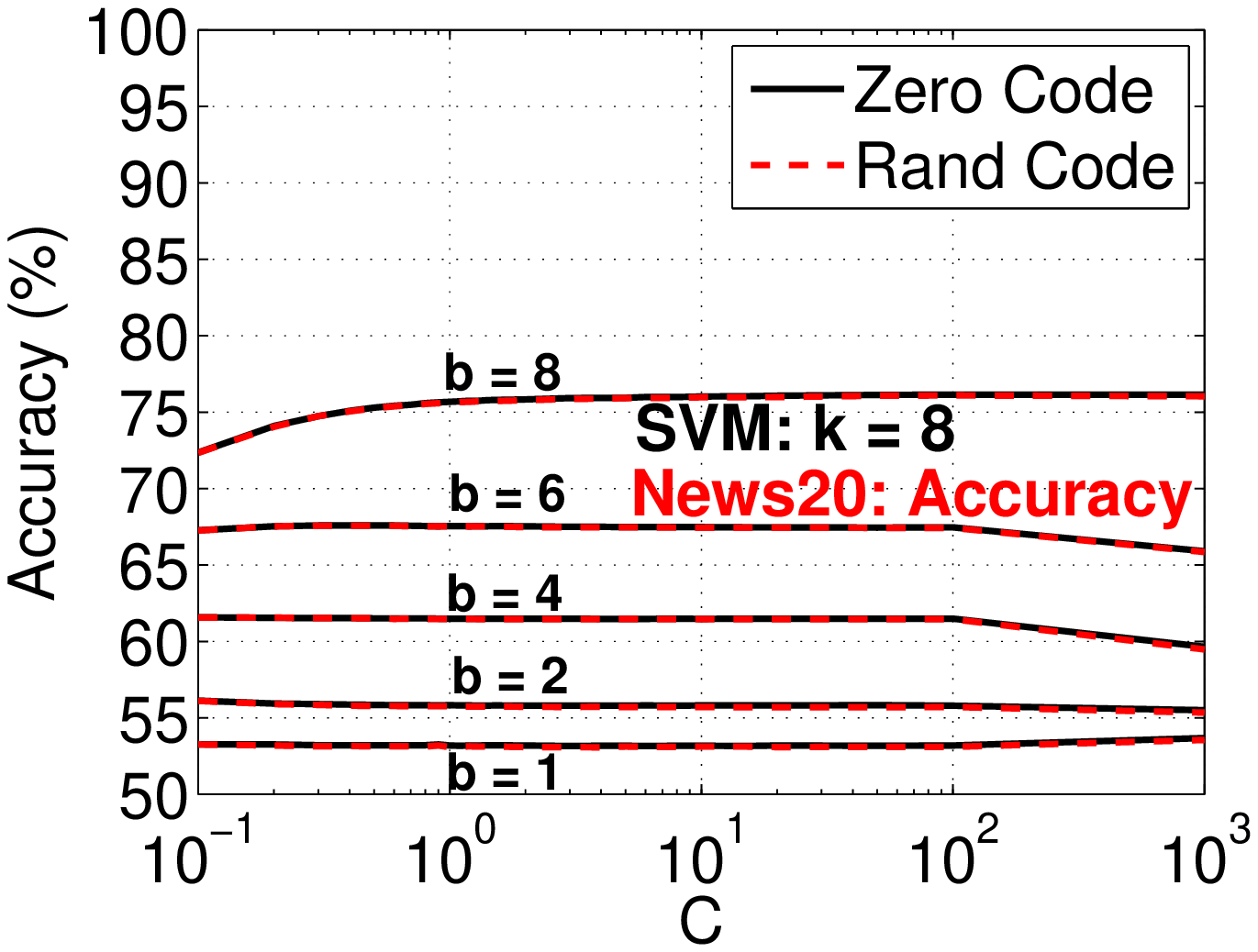}\hspace{-0.1in}
\includegraphics[width=1.5in]{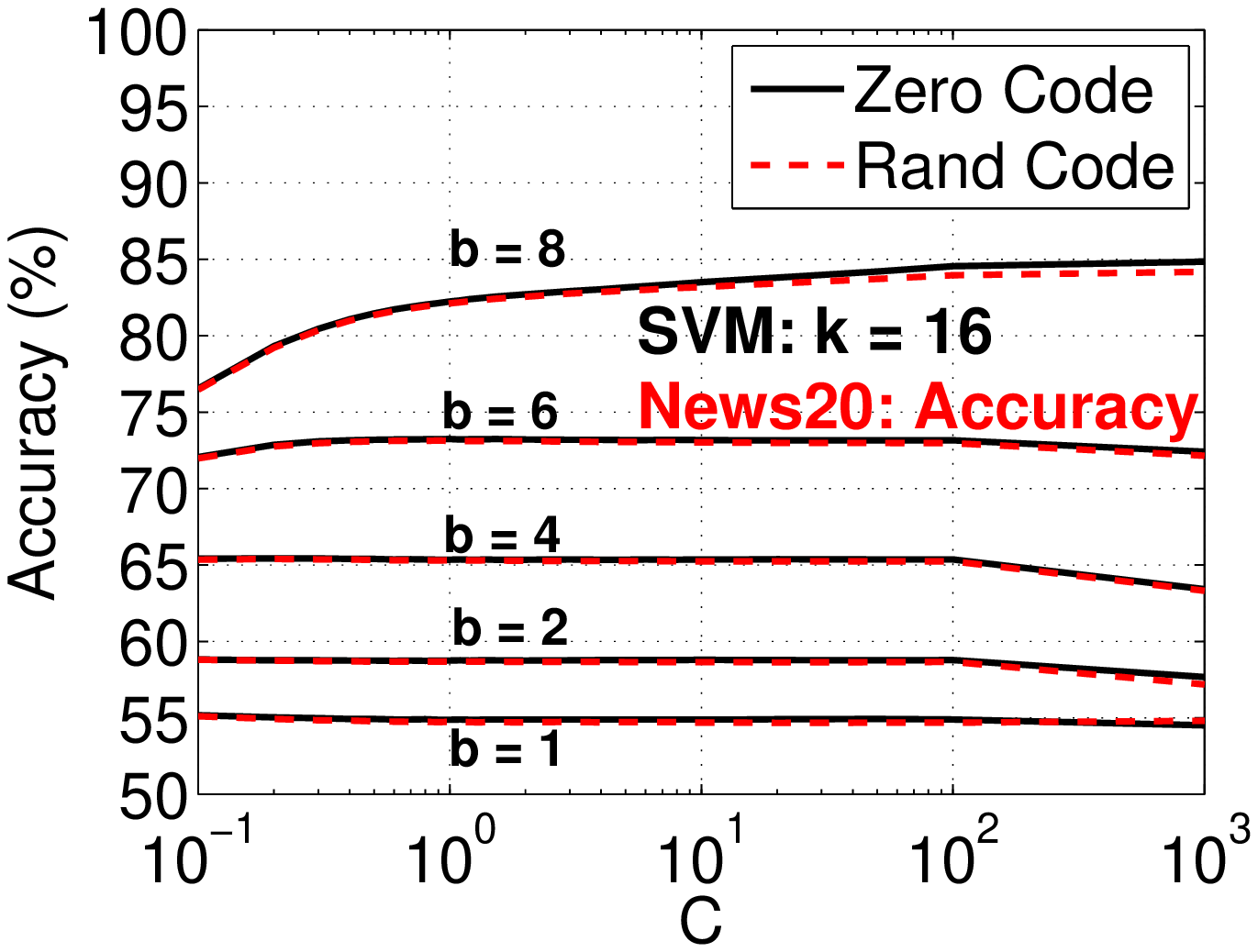}\hspace{-0.1in}
\includegraphics[width=1.5in]{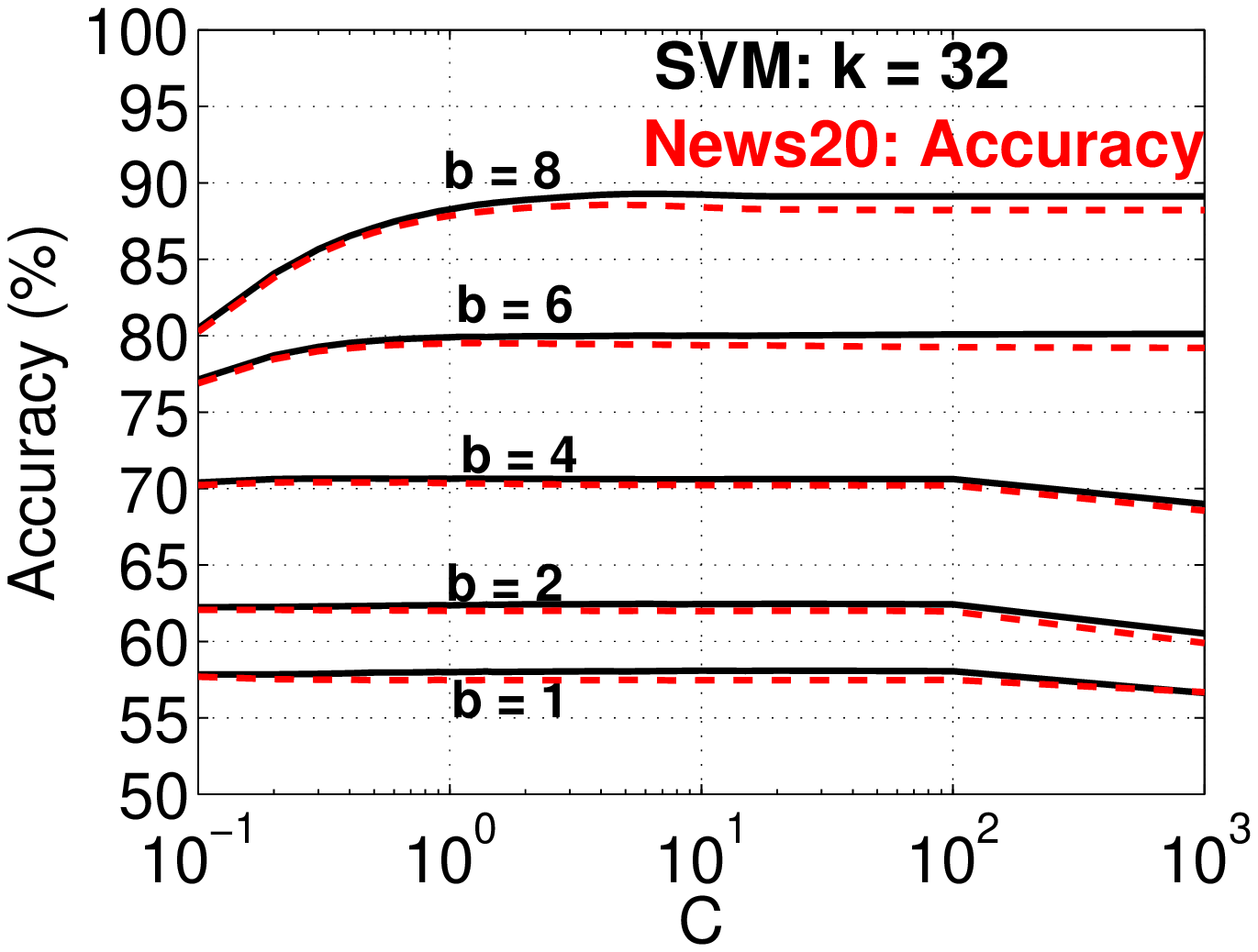}\hspace{-0.1in}
\includegraphics[width=1.5in]{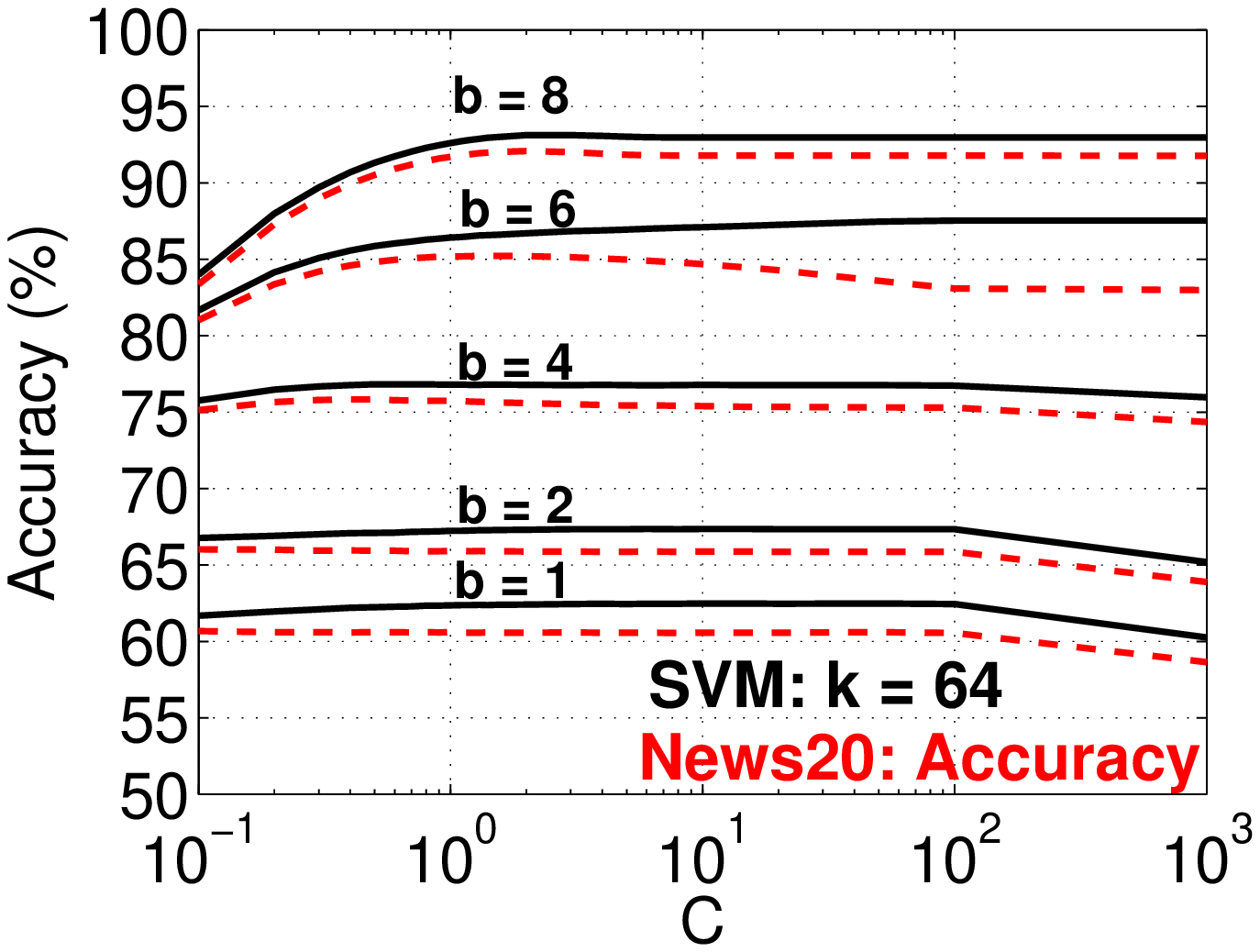}\hspace{-0.1in}
\includegraphics[width=1.5in]{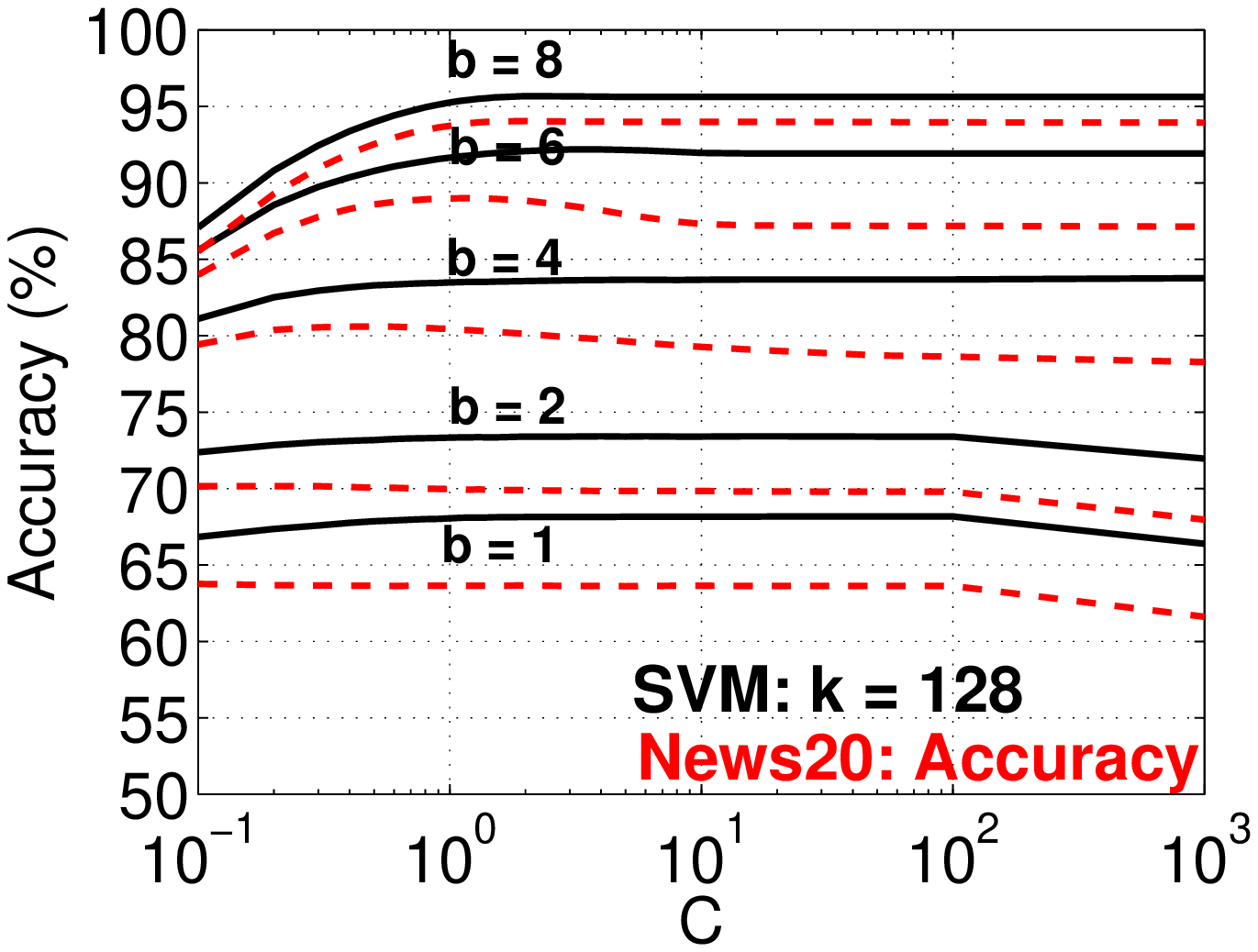}\hspace{-0.1in}
}

\mbox{
\includegraphics[width=1.5in]{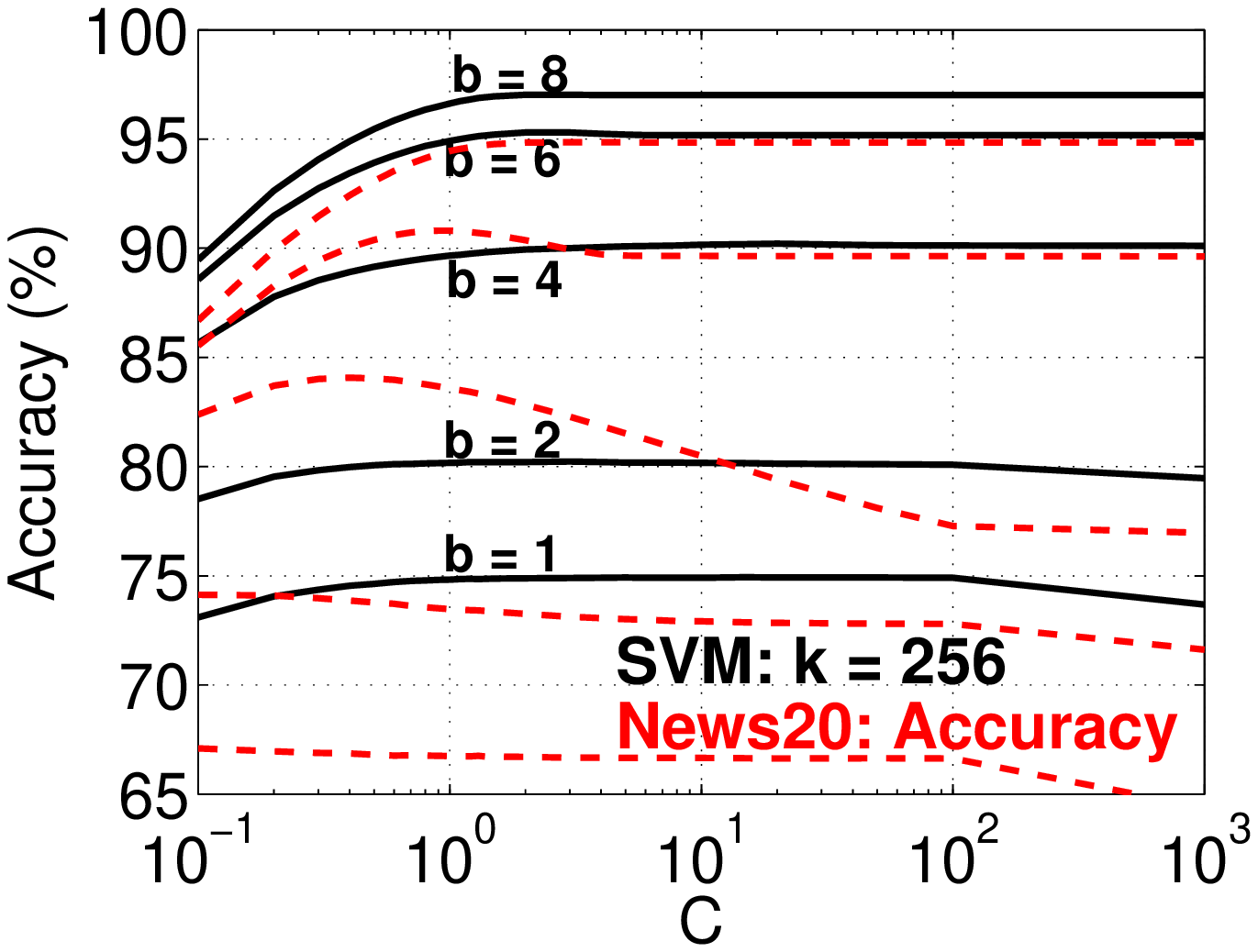}\hspace{-0.1in}
\includegraphics[width=1.5in]{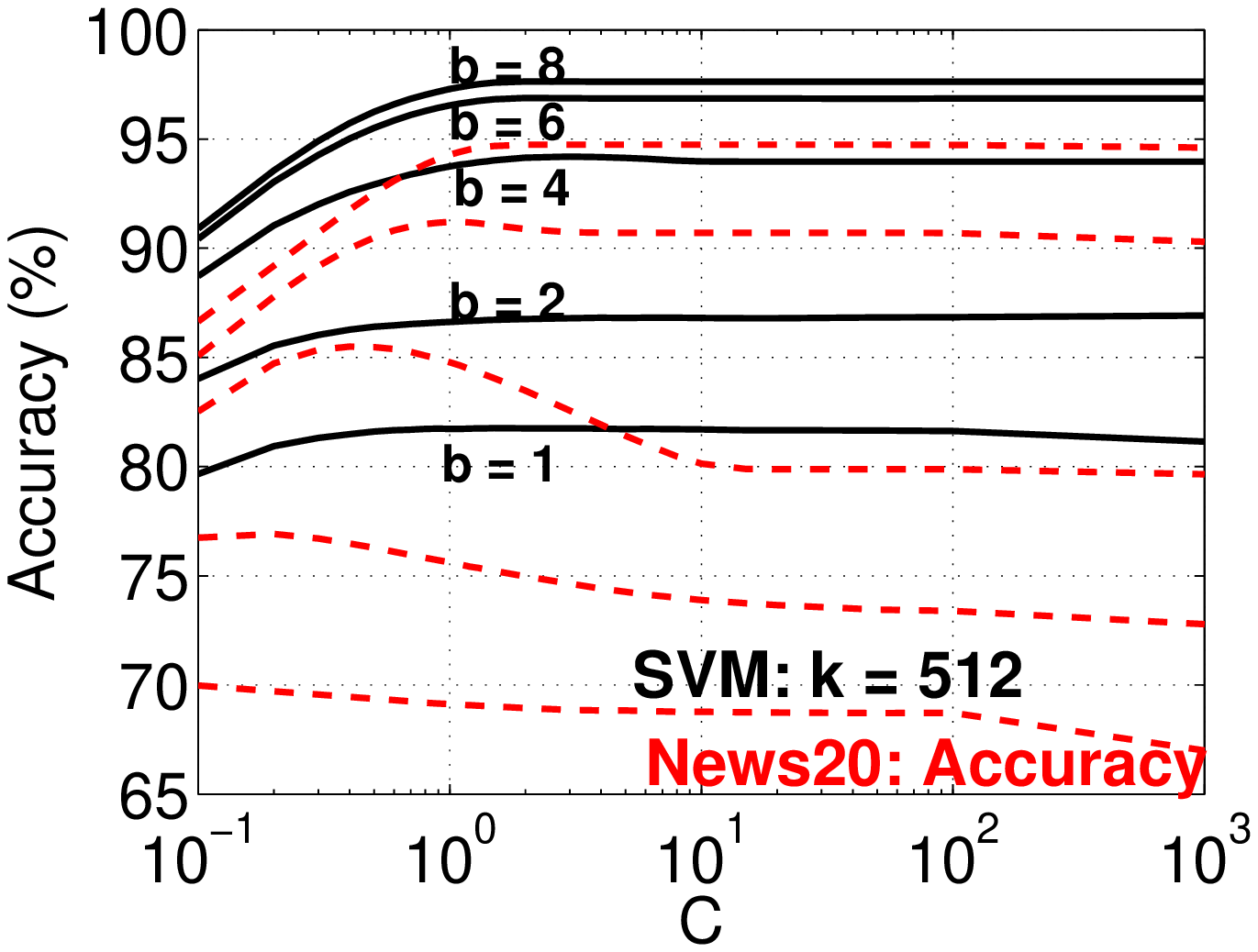}\hspace{-0.1in}
\includegraphics[width=1.5in]{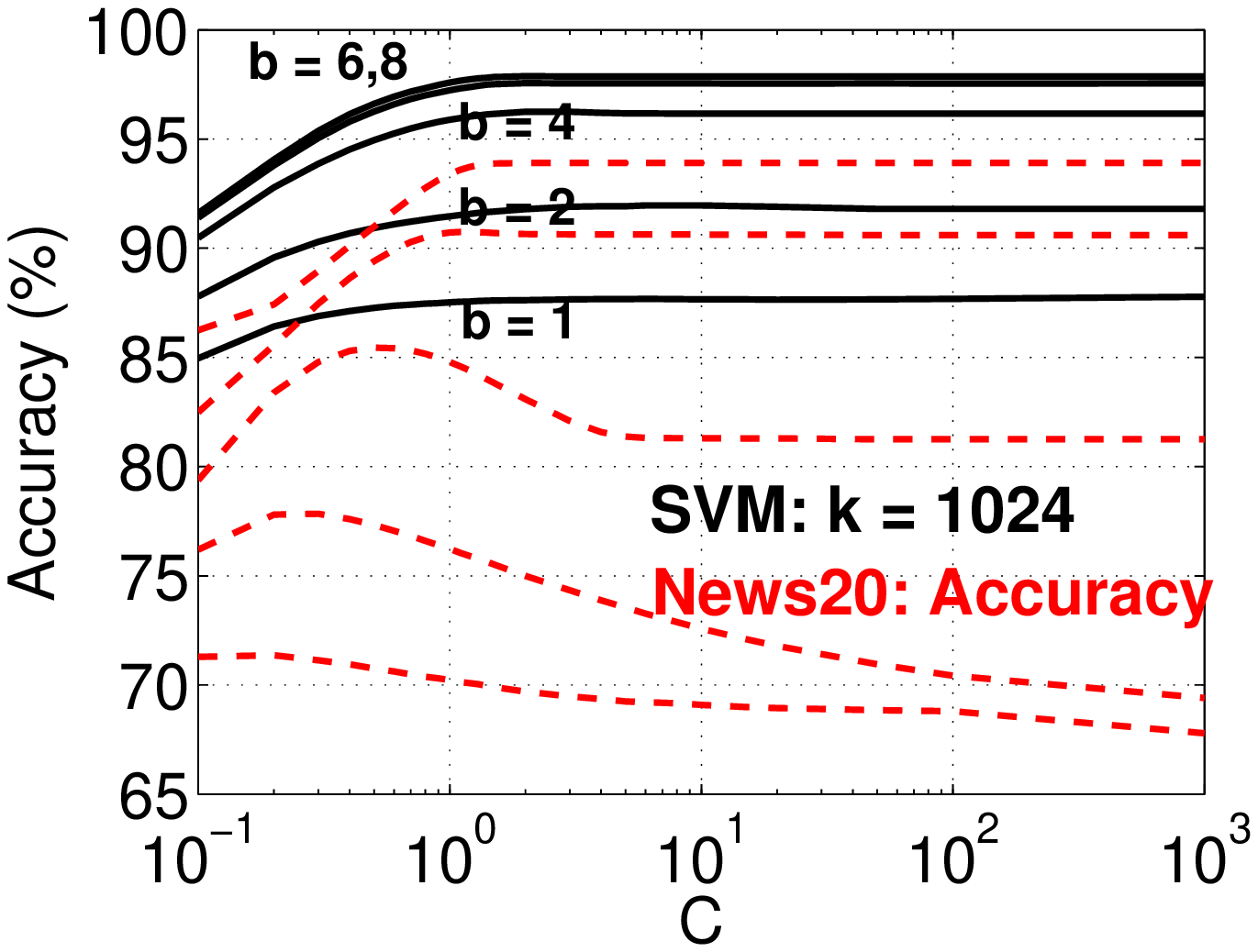}\hspace{-0.1in}
\includegraphics[width=1.5in]{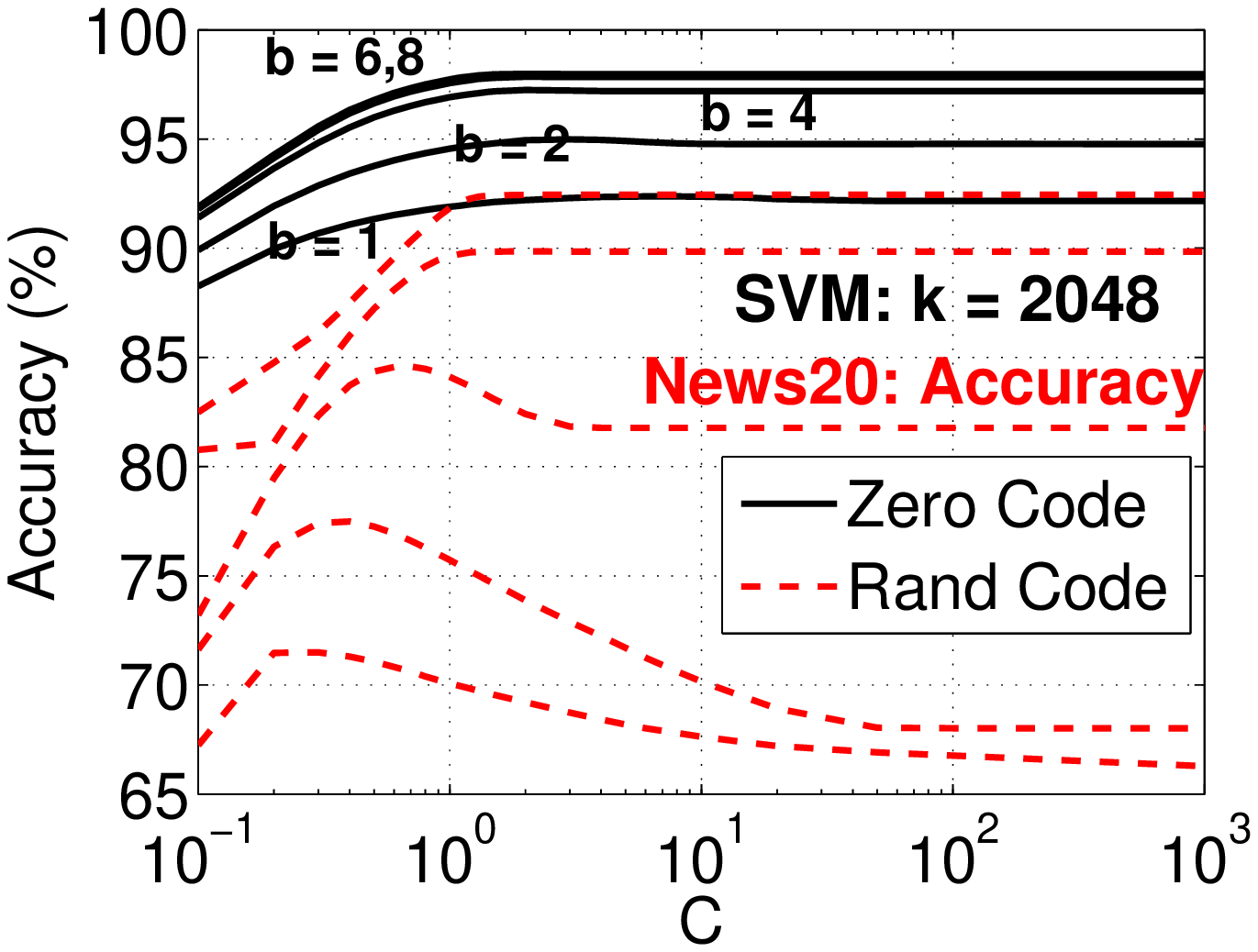}\hspace{-0.1in}
\includegraphics[width=1.5in]{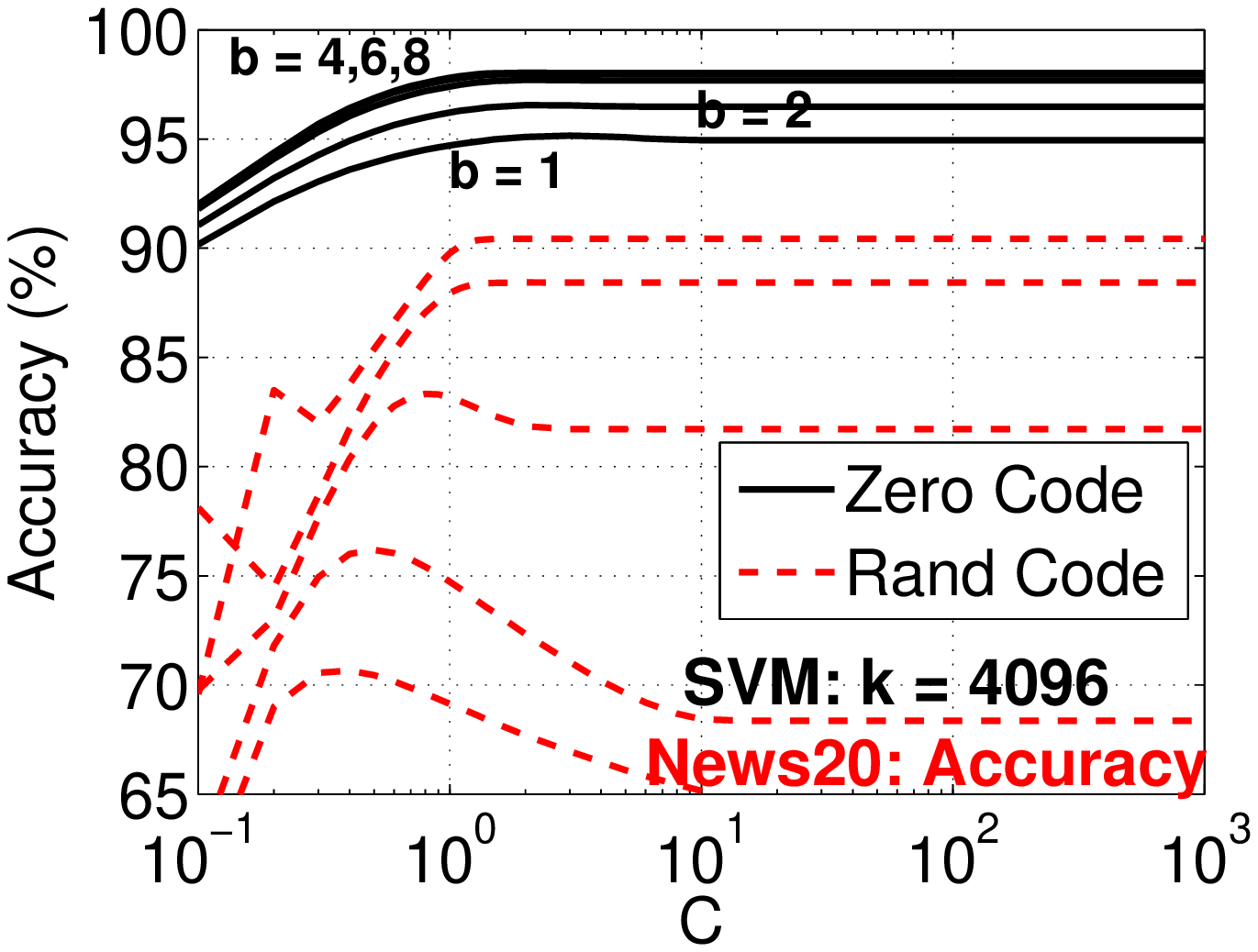}\hspace{-0.1in}
}

\vspace{-0.2in}

\caption{Test accuracies of linear SVM  averaged over 100 repetitions, for comparing the (recommended) zero coding strategy  with the random coding strategy to deal with empty bins. On this dataset, the performance of the random coding strategy can be bad.}\label{fig_news20_accuracy_svm_rand}\vspace{-0.in}
\end{figure}

\begin{figure}[h!]
\mbox{
\includegraphics[width=1.5in]{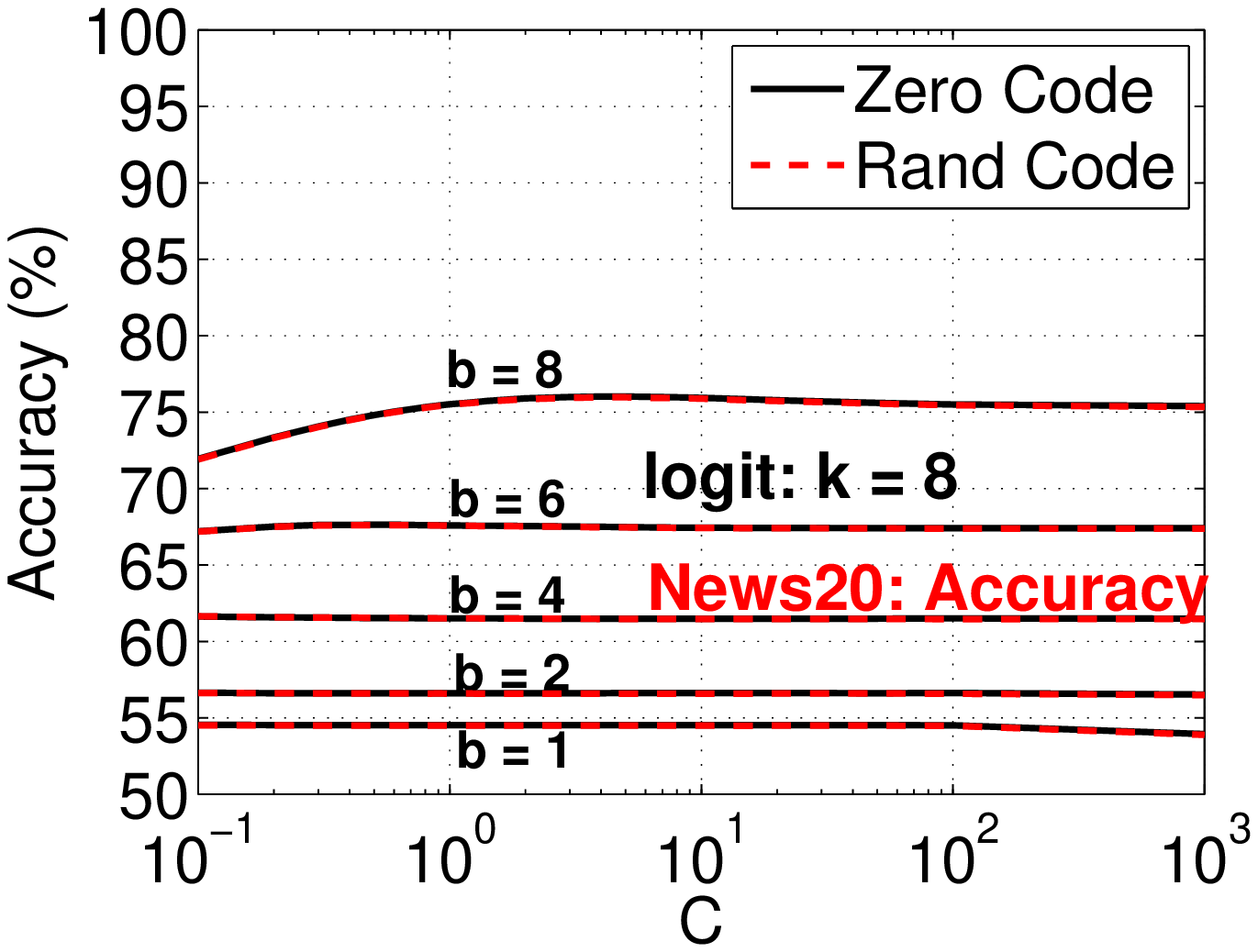}\hspace{-0.1in}
\includegraphics[width=1.5in]{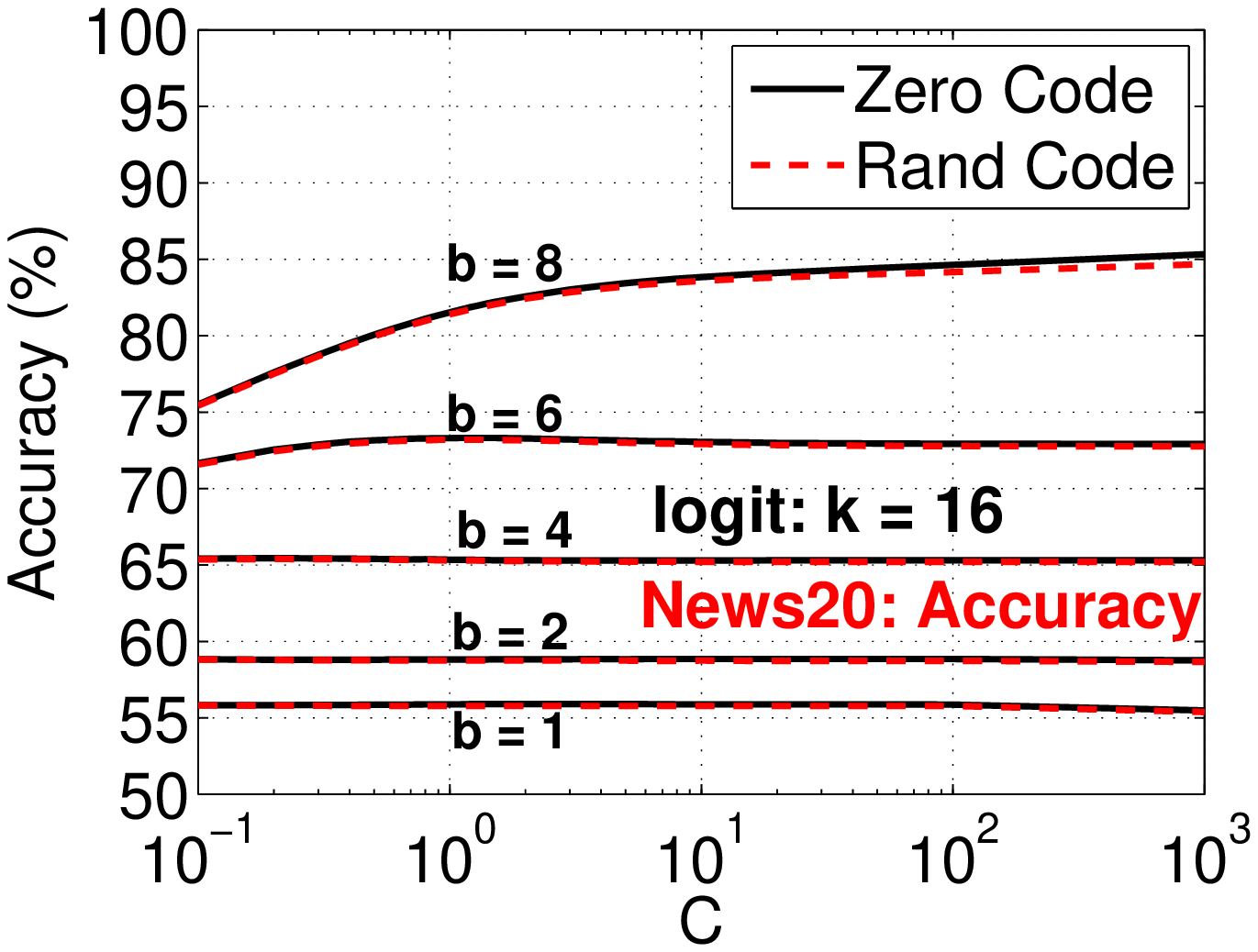}\hspace{-0.1in}
\includegraphics[width=1.5in]{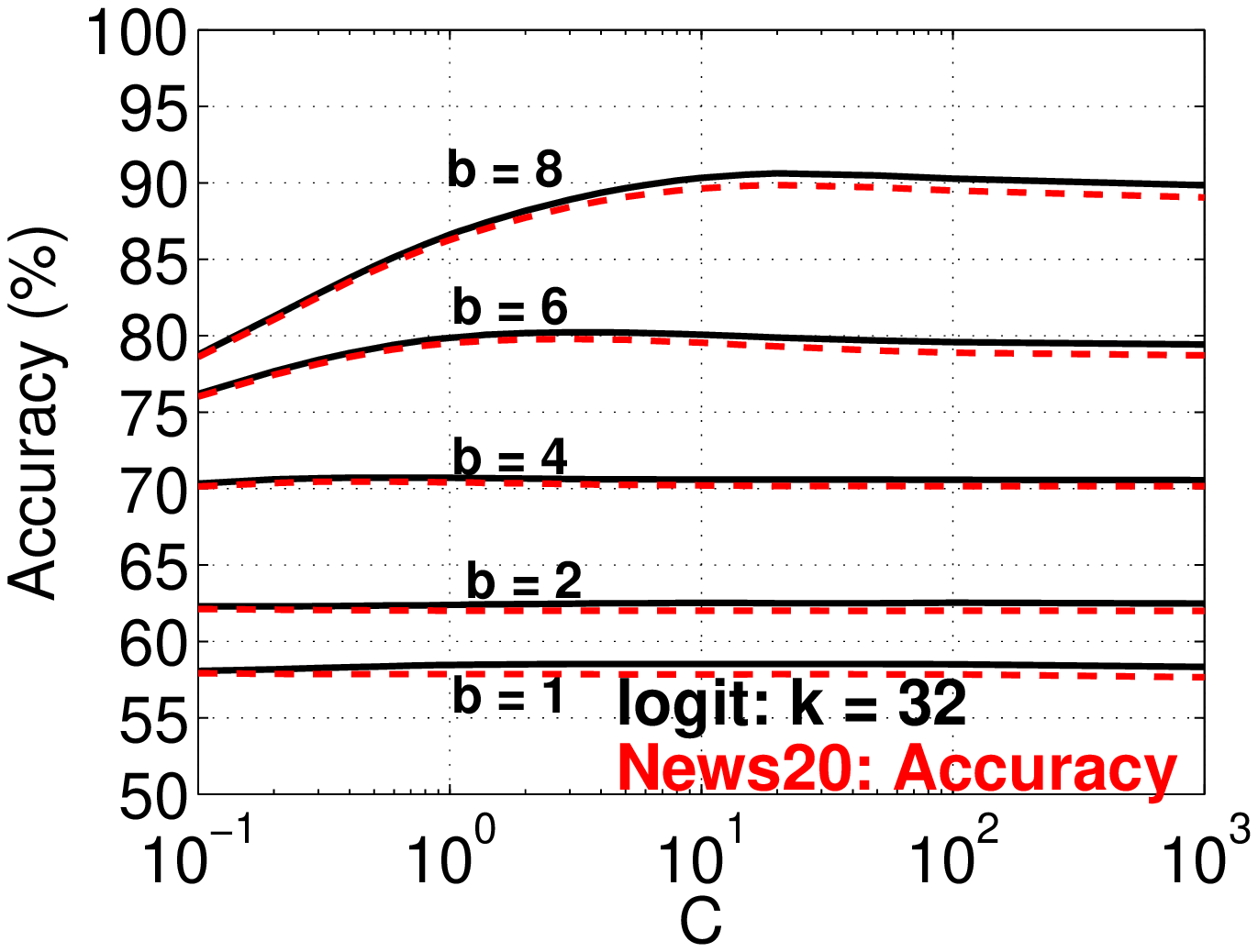}\hspace{-0.1in}
\includegraphics[width=1.5in]{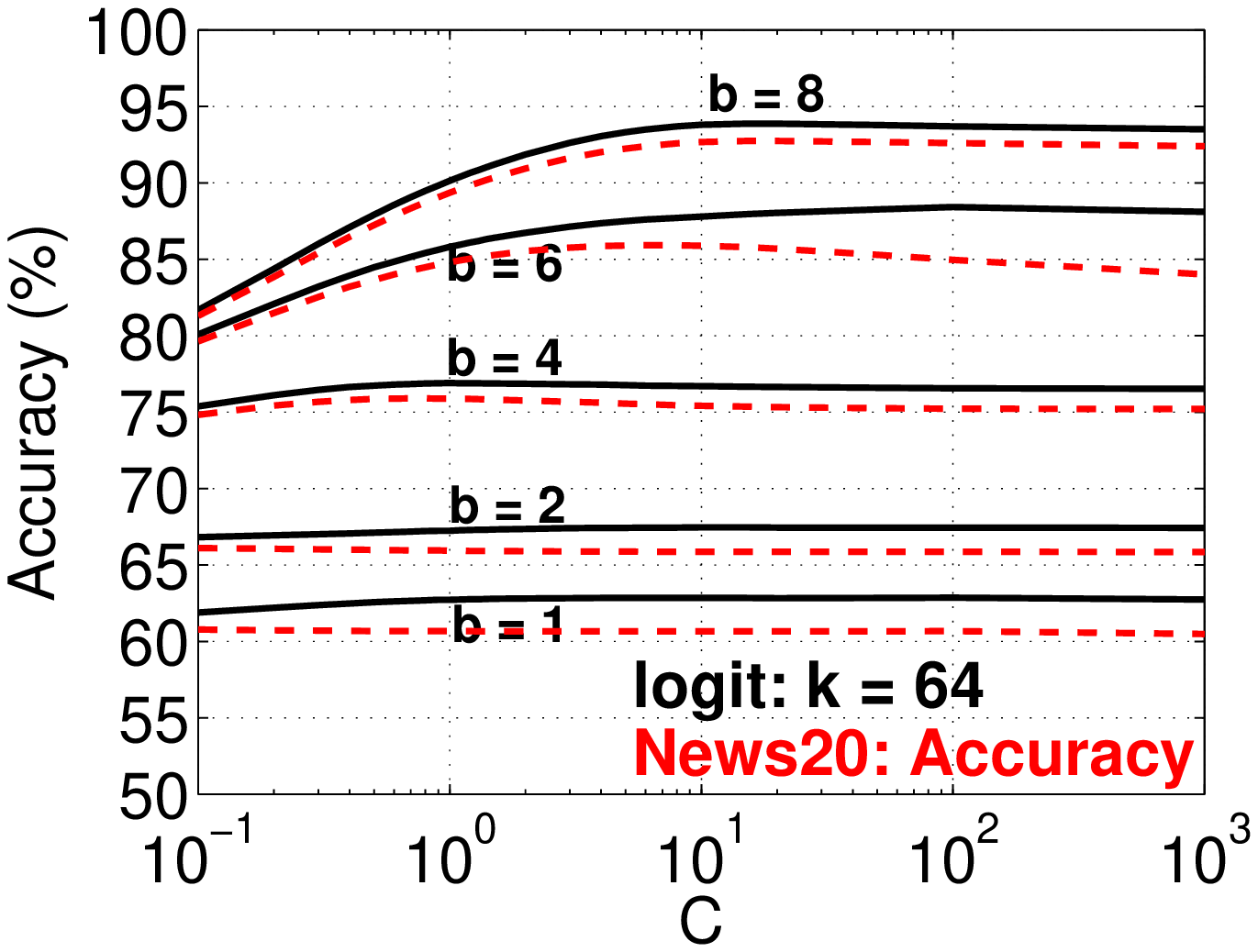}\hspace{-0.1in}
\includegraphics[width=1.5in]{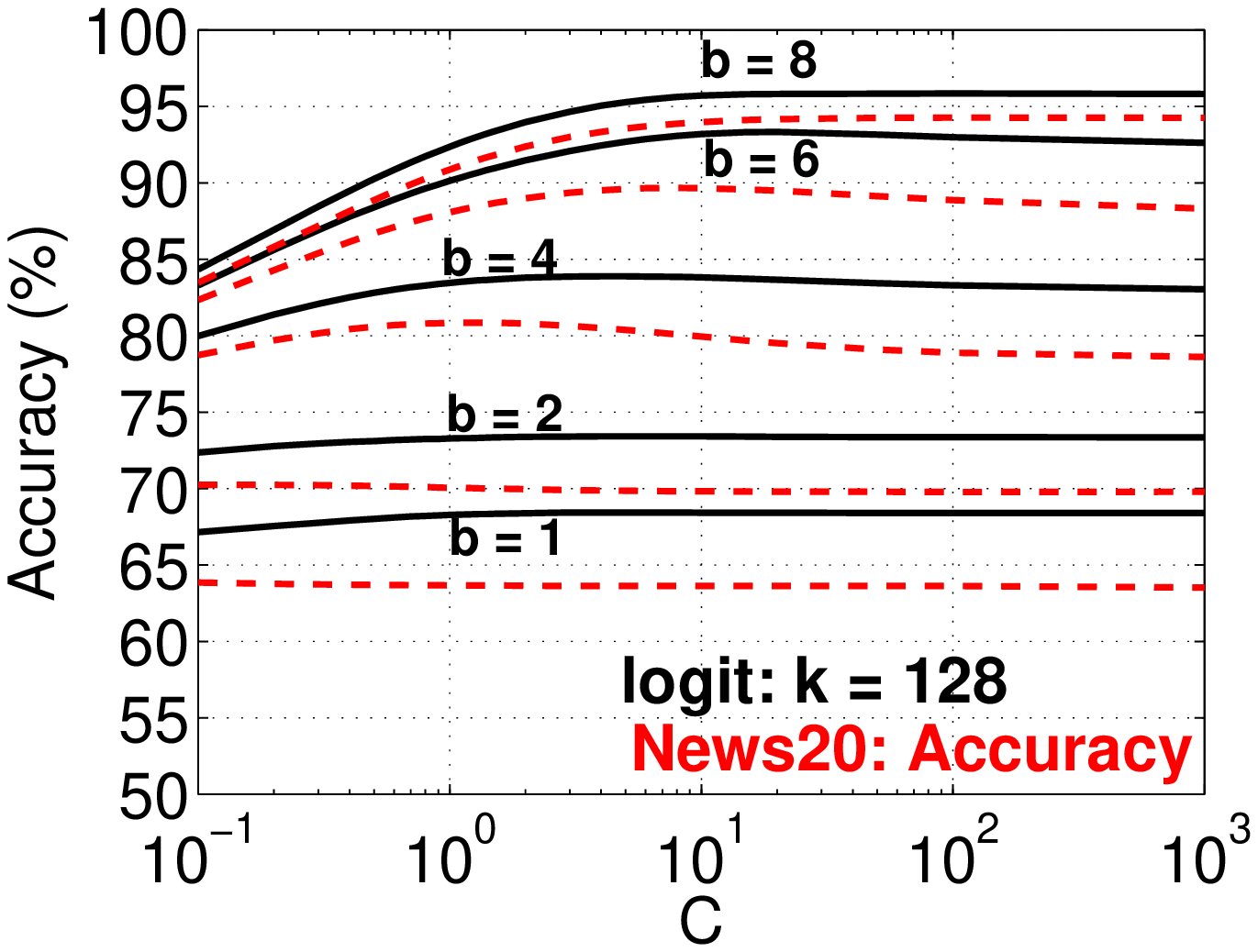}\hspace{-0.1in}
}

\mbox{
\includegraphics[width=1.5in]{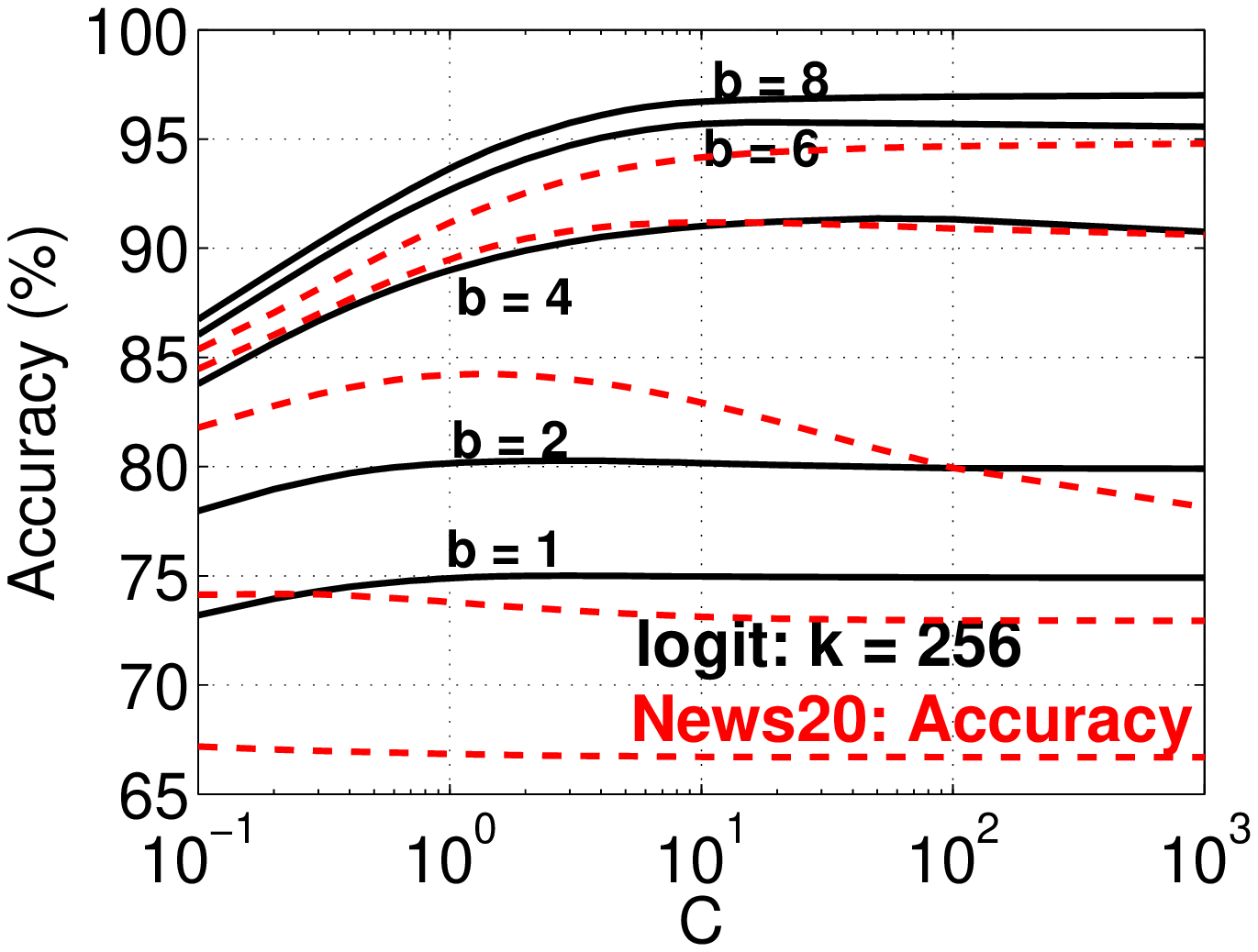}\hspace{-0.1in}
\includegraphics[width=1.5in]{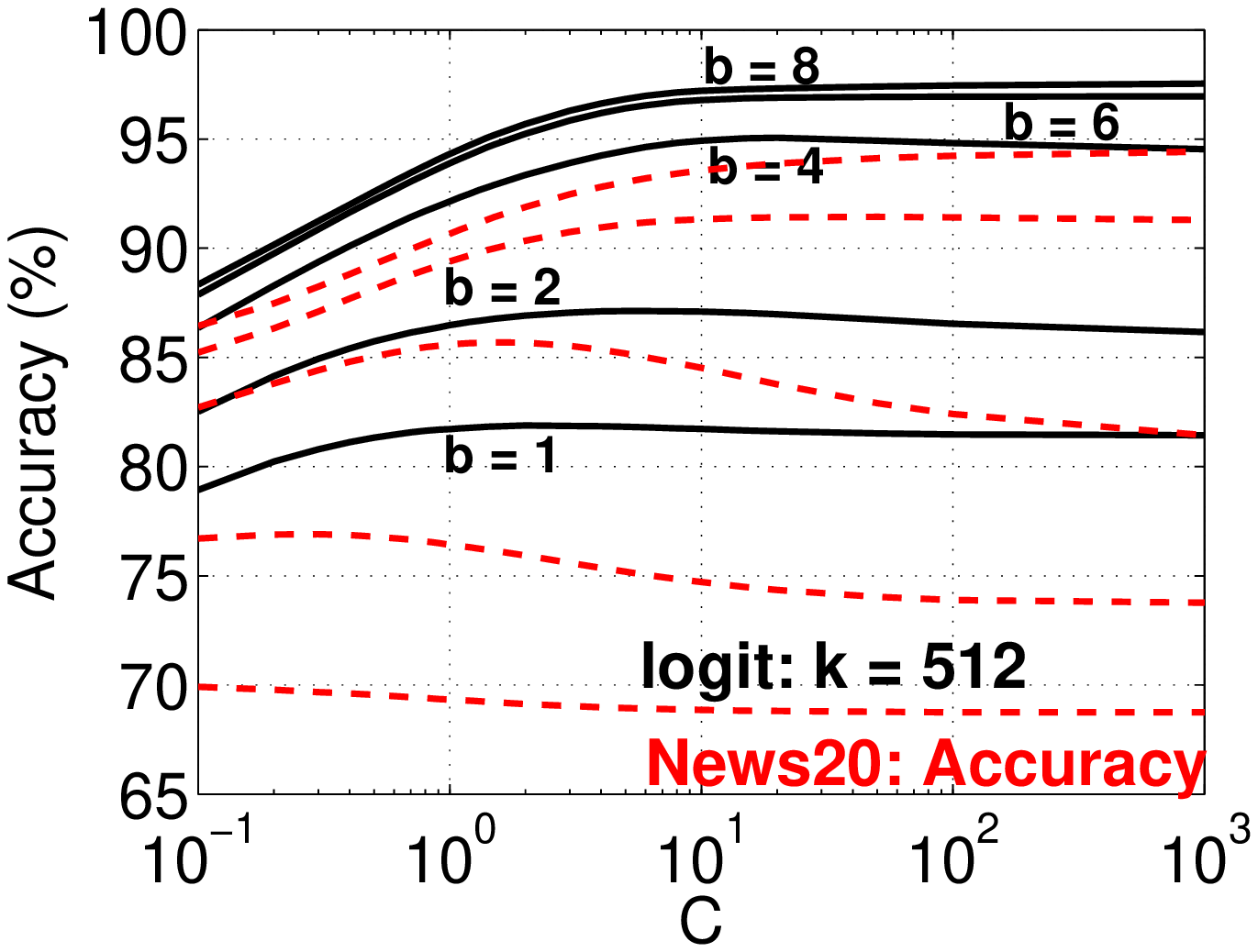}\hspace{-0.1in}
\includegraphics[width=1.5in]{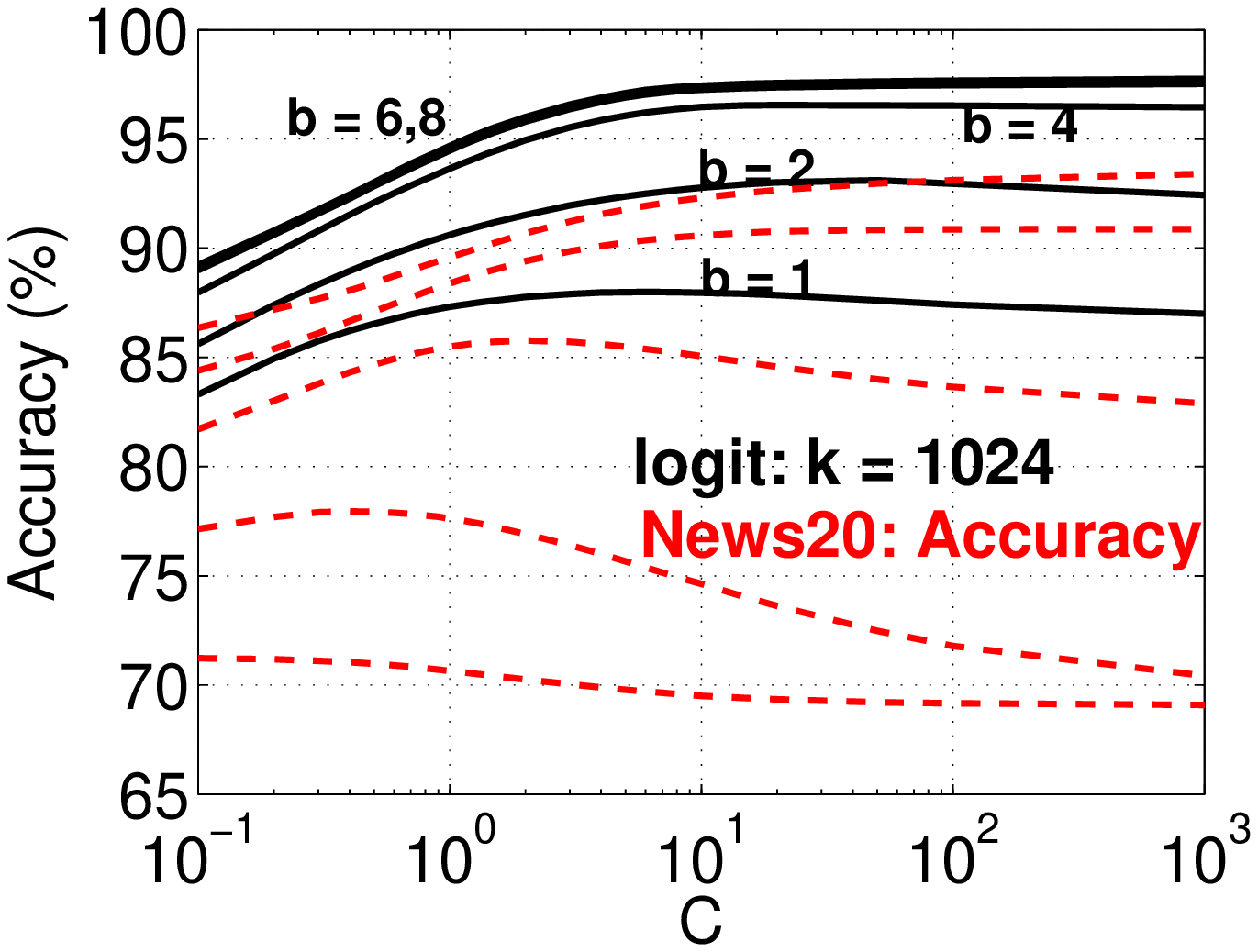}\hspace{-0.1in}
\includegraphics[width=1.5in]{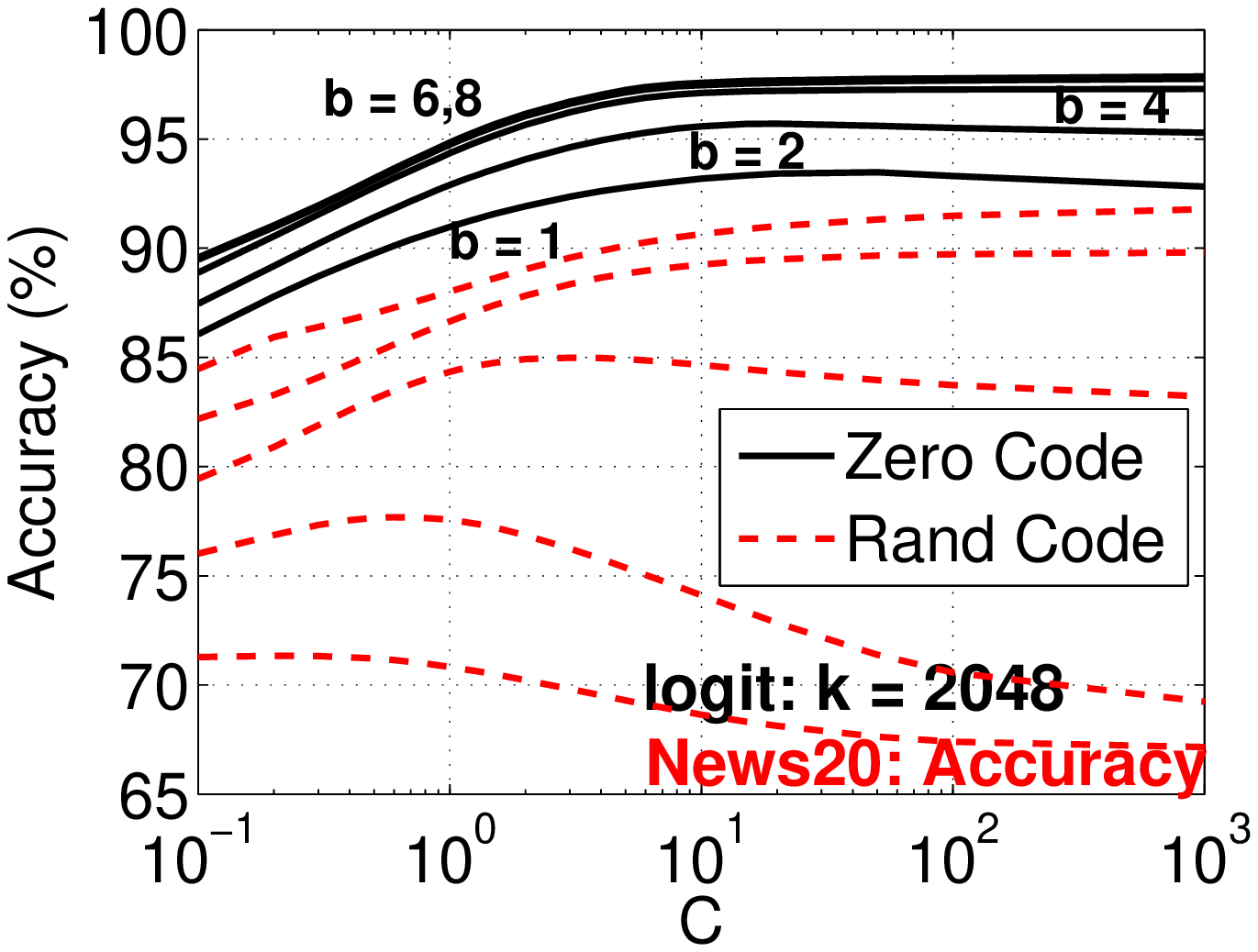}\hspace{-0.1in}
\includegraphics[width=1.5in]{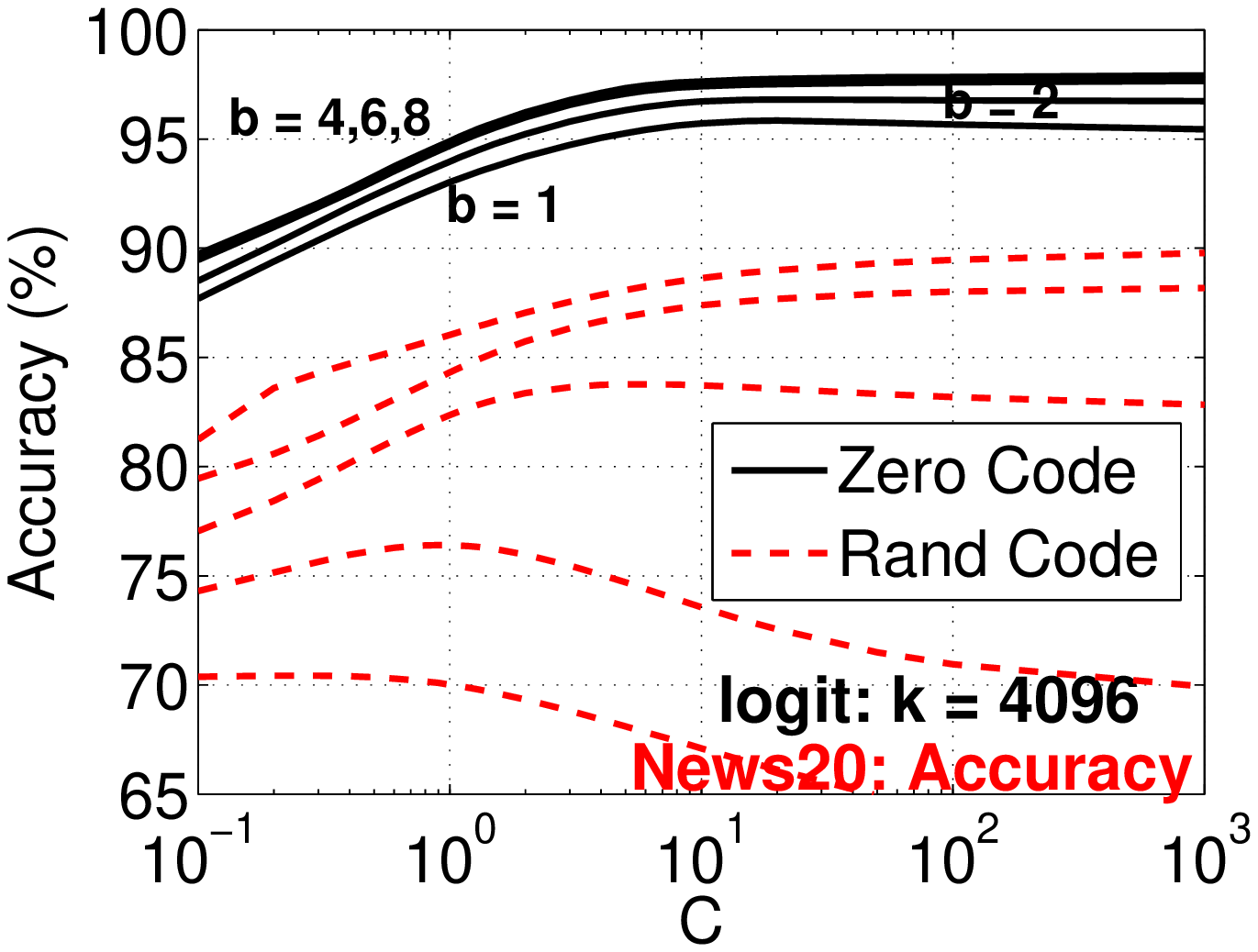}\hspace{-0.1in}
}

\vspace{-0.2in}

\caption{Test accuracies of logistic regression  averaged over 100 repetitions, for comparing the zero coding strategy (recommended) with the random coding strategy to deal with empty bins. On this dataset, the performance of the random coding strategy can be  bad.}\label{fig_news20_accuracy_logit_rand}\vspace{-0.in}
\end{figure}

Figure~\ref{fig_news20_accuracy_svm_rand} and Figure~\ref{fig_news20_accuracy_logit_rand} plot the results for comparing two coding strategies to deal with empty bins, respectively for linear SVM and logistic regression. Again, when $k$ is small (e.g., $k\leq 64$), both  strategies perform similarly. However, when $k$ is large, using the random coding scheme may be disastrous, which is of course also expected.  When $k=4096$, most of the nonzero entries in the new expanded data matrix fed to the solver are artificial, since the original {\em news20} dataset has merely about $500$ nonzero on average.

\vspace{0.1in}

\noindent\textbf{Remark}:\ We should re-iterate that the {\em news20} dataset is more like a contrived example, merely for testing the robustness of the one permutation scheme with the zero coding strategy. In more realistic industrial applications, we expect that numbers of nonzeros in many datasets should be significantly higher, and hence the performance differences between the one permutation scheme and the $k$-permutation scheme and the differences between the two strategies for empty bins should be  small.


\section{The Variable Length One Permutation Hashing Scheme}~\label{sec_variable}

While the fixed-length one permutation  scheme we have presented and analyzed should be simple to implement and easy to understand,  we would like to present a \textbf{variable-length} scheme which may more obviously connect with other known hashing methods such as the Count-Min (CM) sketch~\cite{Article:Cormode_05}.

As in the fixed-length scheme, we first conduct a permutation $\pi: \Omega \rightarrow \Omega$. Instead of dividing the space evenly, we vary the bin lengths according to a multinomial distribution $mult\left(D, \frac{1}{k}, \frac{1}{k}, ..., \frac{1}{k}\right)$.\\

This variable-length scheme is  equivalent to first uniformly grouping the original data entries into  $k$ bins and then applying permutations independently within each bin. The latter explanation connects our method with the Count-Min (CM) sketch~\cite{Article:Cormode_05} (but without the ``count-min'' step), which also hashes the elements uniformly to $k$ bins and the final (stored) hashed value in each bin is  the sum of all the elements in the bin. The bias of the CM estimate can be removed by subtracting a term. \cite{Article:Shi_JMLR09} adopted the CM sketch for linear learning.  Later,~\cite{Proc:Weinberger_ICML2009} proposed a novel idea (named ``VW'') to remove the bias, by pre-multiplying (element-wise) the original data vectors with a random vector whose entries are sampled i.i.d. from the  two-point distribution in $\{-1,1\}$ with equal probabilities. In a recent paper, \cite{Proc:HashLearning_NIPS11} showed that the variance of  the CM sketch and variants are equivalent to the variance of random projections~\cite{Proc:Li_Hastie_Church_KDD06}, which is substantially larger than the variance of the minwise hashing when the data are binary.

Since \cite{Proc:HashLearning_NIPS11} has already conducted (theoretical and empirical) comparisons with CM and VW methods, we do not include more comparisons in this paper. Instead, we have simply showed that with one permutation only, we are able to achieve essentially the same accuracy as using $k$ permutations.

We believe the fixed-length scheme is more convenient to implement. Nevertheless, we would like to present some theoretical results for the variable-length scheme, for better understanding the differences. The major difference is  the distribution of $N_{emp}$, the number of jointly empty bins.

\begin{lemma}\label{lem_Nemp_v}
{\small Under the variable-length scheme,
\begin{align}\label{eqn_Nemp_mean_v}
\frac{E\left(N_{emp}\right)}{k} =& \left(1-\frac{1}{k}\right)^{f_1+f_2-a}
\end{align}
\begin{align}\label{eqn_Nemp_var_v}
\frac{Var\left(N_{emp}\right)}{k^2}
=&\frac{1}{k}\left(\frac{E(N_{emp})}{k}\right)\left(1-\frac{E(N_{emp})}{k}\right)\\\notag
&-\left(1-\frac{1}{k}\right)\left( \left(1-\frac{1}{k}\right)^{2(f_1+f_2-a)} - \left(1-\frac{2}{k}\right)^{f_1+f_2-a}\right)\\\label{eqn_Nemp_var_ineq_v}
<&\frac{1}{k}\left(\frac{E(N_{emp})}{k}\right)\left(1-\frac{E(N_{emp})}{k}\right)
\end{align}}
\textbf{Proof}:\ See Appendix~\ref{app_lem_Nemp_v}. $\Box$
\end{lemma}

The other theoretical results for the fixed-length scheme which are expressed in terms $N_{emp}$ essentially hold for the variable-length scheme. For example, $\frac{N_{mat}}{k-N_{emp}}$ is still an unbiased estimator of $R$ and its variance is in the same form as (\ref{eqn_Rmat_Var}) in terms of $N_{emp}$.\\

\noindent\textbf{Remark:}\  \  \ The number of empty bins for the variable-length scheme as presented in (\ref{eqn_Nemp_mean_v}) is actually an upper bound of the number of empty bins for the fixed length scheme as shown in (\ref{eqn_Nemp_mean}). The difference between $\prod_{j=0}^{f-1} \frac{D\left(1-\frac{1}{k}\right)-j}{D-j}$ and $\left(1-\frac{1}{k}\right)^f$ (recall $f=f_1+f_2-a$) is small when the data are sparse, as shown in Lemma~\ref{lem_Nemp_approx}, although it is possible that $\prod_{j=0}^{f-1} \frac{D\left(1-\frac{1}{k}\right)-j}{D-j}\ll \left(1-\frac{1}{k}\right)^f$ in corner cases. Because smaller $N_{emp}$ implies potentially   better performance, we conclude that the fixed-length scheme should be sufficient and there are perhaps no practical needs to use the variable-length scheme.

\section{Conclusion}

A new hashing algorithm is developed for large-scale search and learning in massive binary data. Compared with the original $k$-permutation (e.g., $k=500$)  minwise hashing algorithm (which is the standard procedure in the context of search), our method requires only one permutation and can achieve similar or even better accuracies at merely $1/k$ of the original preprocessing cost. We expect that our proposed algorithm (or its variant) will be adopted in practice.

{


\begin{thebibliography}{10}

\bibitem{Article:Andoni_CACM08}
Alexandr Andoni and Piotr Indyk.
\newblock Near-optimal hashing algorithms for approximate nearest neighbor in
  high dimensions.
\newblock In {\em Commun. ACM}, volume~51, pages 117--122, 2008.

\bibitem{URL:Bottou_SGD}
Leon Bottou.
\newblock http://leon.bottou.org/projects/sgd.

\bibitem{Proc:Broder_STOC98}
Andrei~Z. Broder, Moses Charikar, Alan~M. Frieze, and Michael Mitzenmacher.
\newblock Min-wise independent permutations (extended abstract).
\newblock In {\em STOC}, pages 327--336, Dallas, TX, 1998.

\bibitem{Proc:Broder_WWW97}
Andrei~Z. Broder, Steven~C. Glassman, Mark~S. Manasse, and Geoffrey Zweig.
\newblock Syntactic clustering of the web.
\newblock In {\em WWW}, pages 1157 -- 1166, Santa Clara, CA, 1997.

\bibitem{Proc:Carter_STOC77}
J.~Lawrence Carter and Mark~N. Wegman.
\newblock Universal classes of hash functions (extended abstract).
\newblock In {\em STOC}, pages 106--112, 1977.

\bibitem{Article:Cormode_05}
Graham Cormode and S.~Muthukrishnan.
\newblock An improved data stream summary: the count-min sketch and its
  applications.
\newblock {\em Journal of Algorithm}, 55(1):58--75, 2005.

\bibitem{Article:Fan_JMLR08}
Rong-En Fan, Kai-Wei Chang, Cho-Jui Hsieh, Xiang-Rui Wang, and Chih-Jen Lin.
\newblock Liblinear: A library for large linear classification.
\newblock {\em Journal of Machine Learning Research}, 9:1871--1874, 2008.

\bibitem{Proc:Fetterly_WWW03}
Dennis Fetterly, Mark Manasse, Marc Najork, and Janet~L. Wiener.
\newblock A large-scale study of the evolution of web pages.
\newblock In {\em WWW}, pages 669--678, Budapest, Hungary, 2003.

\bibitem{Article:Friedman_75}
Jerome~H. Friedman, F.~Baskett, and L.~Shustek.
\newblock An algorithm for finding nearest neighbors.
\newblock {\em IEEE Transactions on Computers}, 24:1000--1006, 1975.

\bibitem{Book:Gradshteyn_00}
Izrail~S. Gradshteyn and Iosif~M. Ryzhik.
\newblock {\em Table of Integrals, Series, and Products}.
\newblock Academic Press, New York, sixth edition, 2000.

\bibitem{Proc:Hsieh_ICML08}
Cho-Jui Hsieh, Kai-Wei Chang, Chih-Jen Lin, S.~Sathiya Keerthi, and
  S.~Sundararajan.
\newblock A dual coordinate descent method for large-scale linear svm.
\newblock In {\em Proceedings of the 25th international conference on Machine
  learning}, ICML, pages 408--415, 2008.

\bibitem{Proc:Indyk_STOC98}
Piotr Indyk and Rajeev Motwani.
\newblock Approximate nearest neighbors: Towards removing the curse of
  dimensionality.
\newblock In {\em STOC}, pages 604--613, Dallas, TX, 1998.

\bibitem{Proc:Joachims_KDD06}
Thorsten Joachims.
\newblock Training linear svms in linear time.
\newblock In {\em KDD}, pages 217--226, Pittsburgh, PA, 2006.

\bibitem{Proc:Li_Church_EMNLP}
Ping Li and Kenneth~W. Church.
\newblock Using sketches to estimate associations.
\newblock In {\em HLT/EMNLP}, pages 708--715, Vancouver, BC, Canada, 2005 (The
  full paper appeared in Commputational Linguistics in 2007).

\bibitem{Proc:Li_Church_Hastie_NIPS08}
Ping Li, Kenneth~W. Church, and Trevor~J. Hastie.
\newblock One sketch for all: Theory and applications of conditional random
  sampling.
\newblock In {\em NIPS (Preliminary results appeared in NIPS 2006)}, Vancouver,
  BC, Canada, 2008.

\bibitem{Proc:Li_Hastie_Church_KDD06}
Ping Li, Trevor~J. Hastie, and Kenneth~W. Church.
\newblock Very sparse random projections.
\newblock In {\em KDD}, pages 287--296, Philadelphia, PA, 2006.

\bibitem{Article:Li_Konig_CACM11}
Ping Li and Arnd~Christian K\"onig.
\newblock Theory and applications b-bit minwise hashing.
\newblock {\em Commun. ACM}, 2011.

\bibitem{Proc:HashLearning_NIPS11}
Ping Li, Anshumali Shrivastava, Joshua Moore, and Arnd~Christian K\"onig.
\newblock Hashing algorithms for large-scale learning.
\newblock In {\em NIPS}, Granada, Spain, 2011.

\bibitem{Proc:Shalev-Shwartz_ICML07}
Shai Shalev-Shwartz, Yoram Singer, and Nathan Srebro.
\newblock Pegasos: Primal estimated sub-gradient solver for svm.
\newblock In {\em ICML}, pages 807--814, Corvalis, Oregon, 2007.

\bibitem{Article:Shi_JMLR09}
Qinfeng Shi, James Petterson, Gideon Dror, John Langford, Alex Smola, and
  S.V.N. Vishwanathan.
\newblock Hash kernels for structured data.
\newblock {\em Journal of Machine Learning Research}, 10:2615--2637, 2009.

\bibitem{Proc:Shrivastava_ECML12}
Anshumali Shrivastava and Ping Li.
\newblock Fast near neighbor search in high-dimensional binary data.
\newblock In {\em ECML}, 2012.

\bibitem{Proc:Sivic_ICCV03}
Josef Sivic and Andrew Zisserman.
\newblock Video google: a text retrieval approach to object matching in videos.
\newblock In {\em ICCV}, 2003.

\bibitem{GoogleBlog}
Simon Tong.
\newblock Lessons learned developing a practical large scale machine learning
  system.
\newblock
  http://googleresearch.blogspot.com/2010/04/lessons-learned-developing-practi%
cal.html, 2008.

\bibitem{Proc:Weinberger_ICML2009}
Kilian Weinberger, Anirban Dasgupta, John Langford, Alex Smola, and Josh
  Attenberg.
\newblock Feature hashing for large scale multitask learning.
\newblock In {\em ICML}, pages 1113--1120, 2009.

\end{thebibliography}
}

\appendix

\section{Proof of Lemma~\ref{lem_Nemp}}\label{app_lem_Nemp}

Recall $N_{emp} = \sum_{j=1}^k I_{emp,j}$,  where $I_{emp,j} = 1$ if, in the $j$-th bin, both $\pi(S_1)$ and $\pi(S_2)$ are empty, and $I_{emp,j}=0$ otherwise. Also recall $D=|\Omega|$, $f_1 = |S_1|$, $f_2 = |S_2|$, $a = |S_1\cap S_2|$. Obviously, if $D\left(1-\frac{1}{k}\right)< f_1+f_2-a$,  then none of the bins will be jointly empty, i.e., $E(N_{emp}) = Var(N_{emp})=0$. Next, assume $D\left(1-\frac{1}{k}\right)\geq f_1+f_2-a$, then by the linearity of expectation,
\begin{align}\notag
E\left(N_{emp}\right) = \sum_{j=1}^{k} \mathbf{Pr}\left(I_{emp,j}=1\right) = k\mathbf{Pr}\left(I_{emp,1}=1\right)
=k\frac{\binom{D\left(1-\frac{1}{k}\right)}{f_1+f_2-a}}{\binom{D}{f_1+f_2-a}} =  k \prod_{j=0}^{f_1+f_2-a-1} \frac{D\left(1-\frac{1}{k}\right)-j}{D-j}
\end{align}
To derive the variance, we first assume $D\left(1-\frac{2}{k}\right)\geq f_1+f_2-a$. Then
\begin{align}\notag
Var\left(N_{emp}\right) =& E\left(N_{emp}^2\right) - E^2\left(N_{emp}\right)\\\notag
 =& E\left(\sum_{j=1}^k I_{emp,j}^2 + \sum_{i\neq j} I_{emp,i}I_{emp,j}\right) -  E^2\left(N_{emp}\right)\\\notag
  =& k(k-1)\mathbf{Pr}\left(I_{emp,1}=1,I_{emp,2}=1\right) + E\left(N_{emp}\right)- E^2\left(N_{emp}\right)\\\notag
=& k(k-1)\times \prod_{j=0}^{f_1+f_2-a-1} \frac{D\left(1-\frac{2}{k}\right)-j}{D-j}\\\notag
&+k\times \prod_{j=0}^{f_1+f_2-a-1} \frac{D\left(1-\frac{1}{k}\right)-j}{D-j} - \left(k\times \prod_{j=0}^{f_1+f_2-a-1} \frac{D\left(1-\frac{1}{k}\right)-j}{D-j}\right)^2
\end{align}
If $D\left(1-\frac{2}{k}\right) < f_1+f_2-a \leq D\left(1-\frac{1}{k}\right)$, then $\mathbf{Pr}\left(I_{emp,1}=1,I_{emp,2}=1\right)=0$ and hence
\begin{align}\notag
Var\left(N_{emp}\right) = E\left(N_{emp}\right)- E^2\left(N_{emp}\right)=
k\times \prod_{j=0}^{f_1+f_2-a-1} \frac{D\left(1-\frac{1}{k}\right)-j}{D-j} - \left(k\times \prod_{j=0}^{f_1+f_2-a-1} \frac{D\left(1-\frac{1}{k}\right)-j}{D-j}\right)^2
\end{align}
Assuming $D\left(1-\frac{2}{k}\right)\geq f_1+f_2-a$, we obtain
\begin{align}\notag
\frac{Var\left(N_{emp}\right)}{k^2}
=& \frac{1}{k}\left(\prod_{j=0}^{f_1+f_2-a-1} \frac{D\left(1-\frac{1}{k}\right)-j}{D-j}\right)
 \left(1-\prod_{j=0}^{f_1+f_2-a-1} \frac{D\left(1-\frac{1}{k}\right)-j}{D-j}\right)\\\notag
&-\left(1-\frac{1}{k}\right)\left( \left(\prod_{j=0}^{f_1+f_2-a-1} \frac{D\left(1-\frac{1}{k}\right)-j}{D-j}\right)^2 - \prod_{j=0}^{f_1+f_2-a-1} \frac{D\left(1-\frac{2}{k}\right)-j}{D-j}\right)\\\notag
=&\frac{1}{k}\left(\frac{E(N_{emp})}{k}\right)\left(1-\frac{E(N_{emp})}{k}\right)\\\notag
&-\left(1-\frac{1}{k}\right)\left( \left(\prod_{j=0}^{f_1+f_2-a-1} \frac{D\left(1-\frac{1}{k}\right)-j}{D-j}\right)^2 - \prod_{j=0}^{f_1+f_2-a-1} \frac{D\left(1-\frac{2}{k}\right)-j}{D-j}\right)\\\notag
<&\frac{1}{k}\left(\frac{E(N_{emp})}{k}\right)\left(1-\frac{E(N_{emp})}{k}\right)
\end{align}
because
\begin{align}\notag
&\left( \frac{D\left(1-\frac{1}{k}\right)-j}{D-j}\right)^2 - \frac{D\left(1-\frac{2}{k}\right)-j}{D-j}>0\\\notag
\Longleftrightarrow&\left(D\left(1-\frac{1}{k}\right)-j\right)^2> (D-j)\left(D\left(1-\frac{2}{k}\right)-j\right)\\\notag
\Longleftrightarrow&\left(1-\frac{1}{k}\right)^2 = 1-\frac{2}{k}+\frac{1}{k^2} > 1-\frac{2}{k}
\end{align}

This completes the proof.

\section{Proof of Lemma~\ref{lem_Nemp_approx}}\label{app_lem_Nemp_approx}

The following  expansions will be  useful
\begin{align}
&\sum_{j=1}^{n-1} \frac{1}{j} = \log n + 0.577216-\frac{1}{2n}-\frac{1}{12n^2}+...\hspace{0.5in}(\text{\cite[8.367.13]{Book:Gradshteyn_00}})\\
&\log(1-x) = -x -\frac{x^2}{2} -\frac{x^3}{3} -... \hspace{0.5in} ( |x|<1)
\end{align}
Assume $D\left(1-\frac{1}{k}\right)\geq f_1+f_2-a$. We can write
\begin{align}\notag
\frac{E\left(N_{emp}\right)}{k} =&  \prod_{j=0}^{f_1+f_2-a-1} \frac{D\left(1-\frac{1}{k}\right)-j}{D-j}
 =\left(1-\frac{1}{k}\right)^{f_1+f_2-a}\times\prod_{j=0}^{f_1+f_2-a-1}\left(1-\frac{j}{(k-1)(D-j)}\right)
  \end{align}
Hence it suffices to study the error term
\begin{align}\notag
\prod_{j=0}^{f_1+f_2-a-1}\left(1-\frac{j}{(k-1)(D-j)}\right).
\end{align}

\begin{align}\notag
&\log \prod_{j=0}^{f_1+f_2-a-1}\left(1-\frac{j}{(k-1)(D-j)}\right) = \sum_{j=0}^{f_1+f_2-a-1}\log \left(1-\frac{j}{(k-1)(D-j)}\right)\\\notag
=&\sum_{j=0}^{f_1+f_2-a-1}\left\{-\frac{j}{(k-1)(D-j)} -\frac{1}{2}\left(\frac{j}{(k-1)(D-j)} \right)^2-\frac{1}{3}\left(\frac{j}{(k-1)(D-j)} \right)^3+...\right\}
\end{align}
Take the first term,
\begin{align}\notag
&\sum_{j=0}^{f_1+f_2-a-1}-\frac{j}{(k-1)(D-j)} =\frac{1}{k-1}\sum_{j=0}^{f_1+f_2-a-1}\frac{D-j}{D-j} - \frac{D}{D-j}  \\\notag
 =& \frac{1}{k-1}\left(f_1+f_2-a-D\sum_{j=0}^{f_1+f_2-a-1}\frac{1}{D-j}\right)\\\notag
 =&\frac{1}{k-1}\left(f_1+f_2-a-D\left(\sum_{j=1}^{D}\frac{1}{j} - \sum_{j=1}^{D-f_1-f_2+a}\frac{1}{j}\right)\right)\\\notag
 =& \frac{1}{k-1}\left(f_1+f_2-a-D\left(\log (D+1) - \frac{1}{2(D+1)} - \log (D - f_1-f_2+a+1) + \frac{1}{2(D-f_1-f_2+a+1)}\right)+...\right)
\end{align}
Thus, we obtain  (by ignoring a term $\frac{D}{D+1}$)
\begin{align}\notag
\prod_{j=0}^{f_1+f_2-a-1}\left(1-\frac{j}{(k-1)(D-j)}\right) =  \exp\left(\frac{-D\log \frac{D+1}{D - f_1-f_2+a+1} + (f_1+f_2-a)\left(1-\frac{1}{2(D-f_1-f_2+a+1)}\right)}{k-1} + ... \right)
\end{align}

Assuming $f_1+f_2-a \ll D$, we can further expand $\log \frac{D+1}{D - f_1-f_2+a+1}$ and obtain a more simplified approximation:
\begin{align}\notag
\frac{E\left(N_{emp}\right)}{k}  =&\left(1-\frac{1}{k}\right)^{f_1+f_2-a}\left(1-O\left(\frac{(f_1+f_2-a)^2}{kD}\right)\right)
  \end{align}

Next, we analyze the approximation of the variance by assuming $f_1+f_2-a \ll D$.  A similar analysis can show that
\begin{align}\notag
\prod_{j=0}^{f_1+f_2-a-1} \frac{D\left(1-\frac{2}{k}\right)-j}{D-j}  = \left(1-\frac{2}{k}\right)^{f_1+f_2-a}\left(1-O\left(\frac{(f_1+f_2-a)^2}{kD}\right)\right)
\end{align}
and hence we obtain, by using $1-\frac{2}{k} = \left(1-\frac{1}{k}\right)\left(1-\frac{1}{k-1}\right)$,
\begin{align}\notag
&\frac{Var\left(N_{emp}\right)}{k^2}\\\notag
=& \left(1-\frac{2}{k}\right)^{f_1+f_2-a} - \left(1-\frac{1}{k}\right)^{2(f_1+f_2-a)} + \frac{1}{k}\left( \left(1-\frac{1}{k}\right)^{f_1+f_2-a} - \left(1-\frac{2}{k}\right)^{f_1+f_2-a}\right) + O\left(\frac{(f_1+f_2-a)^2}{kD}\right)\\\notag
=&\frac{1}{k}\left(1-\frac{1}{k}\right)^{f_1+f_2-a}\left(1-\left(1-\frac{1}{k}\right)^{f_1+f_2-a}\right)\\\notag
&-\left(1-\frac{1}{k}\right)^{f_1+f_2-a+1} \left(\left(1-\frac{1}{k}\right)^{f_1+f_2-a}  - \left(1-\frac{1}{k-1}\right)^{f_1+f_2-a}\right)+ O\left(\frac{(f_1+f_2-a)^2}{kD}\right)
\end{align}

\section{Proof of Lemma~\ref{lem_Nemp_Pr}}~\label{app_lem_Nemp_Pr}

Let $q(D,k,f) = \mathbf{Pr}\left( N_{emp} = 0\right)$ and $D_{jk}=D(1-j/k)$. Then,
\begin{align}\notag
\mathbf{Pr}\left( N_{emp} = j\right) =& {k\choose j}P\{I_{emp,1}=\cdots=I_{emp,j}=1,I_{emp,j+1}=\cdots=I_{emp,k}=0\}\\\notag
 =& {k\choose j}\frac{P^{D_{jk}}_f}{P^D_f}q(D_{jk},k-j,f).
\end{align}
where $P_{f}^D$ is the ``permutation''  operator: $P_f^D = D(D-1)(D-2)...(D-f+1)$.

Thus, to derive $\mathbf{Pr}\left(N_{emp}=j\right)$, we just need to find  $q(D,k,f)$. By the union-intersection formula,
\begin{align}\notag
1 -  q(D,k,f)
= \sum_{j=1}^k (-1)^{j-1} {k\choose j} E \prod_{i=1}^j I_{emp,i}.
\end{align}
From Lemma~\ref{lem_Nemp}, we  can infer $E \prod_{i=1}^j I_{emp,i} = P^{D_{jk}}_f/P^D_f =\prod_{t=0}^{f-1}\frac{D\left(1-\frac{j}{k}\right)-t}{D-t}$. Thus we find
\begin{align}\notag
q(D,k,f) = 1 + \sum_{j=1}^k (-1)^j {k\choose j}\frac{P^{D_{jk}}_f}{P^D_f}
= \sum_{j=0}^k (-1)^j {k\choose j}\frac{P^{D_{jk}}_f}{P^D_f}.
\end{align}
It follows that
\begin{align}\notag
\mathbf{Pr}\left(N_{emp} = j\right) =& {k\choose j}\sum_{s =0}^{k-j} (-1)^s {k - j \choose s}
\frac{P^{D(1-j/k - s/k)}_f}{P^D_f}\\\notag
=&\sum_{s=0}^{k-j}(-1)^s\frac{k!}{j!s!(k-j-s)!}\prod_{t=0}^{f-1}\frac{D\left(1-\frac{j+s}{k}\right)-t}{D-t}
\end{align}

\section{Proof of Lemma~\ref{lem_Nmat}}\label{app_lem_Nmat}

 Define
\begin{align}\notag
&S_1 \cup S_2 = \{j_1, j_2, ..., j_{f_1+f_2-a}\}\\\notag
&J = \min\pi(S_1\cup S_2) = \underset{1\leq i\leq f_1+f_2-a}{\min} \pi(j_i)\\\notag
&T = \underset{i}{argmin}\ \ \pi(j_i), \text{  i.e., }  \pi(j_T) = J
\end{align}

Because $\pi$ is a random permutation, we know
\begin{align}\notag
\mathbf{Pr}\left(T = i\right)= \mathbf{Pr}\left(j_T = j_i\right) = \mathbf{Pr}\left(\pi(j_T) = \pi(j_i)\right)  = \frac{1}{f_1+f_2-a}, \ \ 1\leq i\leq f_1+f_2-a
\end{align}

Due to symmetry,
\begin{align}\notag
\mathbf{Pr}(T=i|J = t) = \mathbf{Pr}(\pi(j_i) = t|\underset{1\leq l\leq f_1+f_2-a}{\min} \pi(j_l)=t  ) =\frac{1}{f_1+f_2-a}
\end{align}
and hence we know that $J$ and $T$ are independent. Therefore,
\begin{align}\notag
E(N_{mat}) =& \sum_{j=1}^k \mathbf{Pr}(I_{mat,j}=1) = k\mathbf{Pr}(I_{mat,1}=1)\\\notag
=& k\mathbf{Pr}\left(j_T \in S_1\cap S_2, 0\leq J\leq D/k-1 \right)\\\notag
=&k\mathbf{Pr}\left(j_T \in S_1\cap S_2\right)\mathbf{Pr}\left(0\leq J\leq D/k-1 \right)\\\notag
=&kR\mathbf{Pr}\left(I_{emp,1}=0\right)\\\notag
=&kR\left(1-\frac{E\left(N_{emp}\right)}{k}\right)
\end{align}

\begin{align}\notag
E(N_{mat}^2) =& E\left(\left(\sum_{j=1}^k I_{mat,j}\right)^2\right) =  E\left(\sum_{j=1}^k I_{mat,j} + \sum_{i\neq j}^kI_{mat,i} I_{mat,j}\right)\\\notag
=&E(N_{mat}) + k(k-1)E(I_{mat,1}I_{mat,2})
\end{align}

\begin{align}\notag
&E(I_{mat,1}I_{mat,2}) = \mathbf{Pr}\left(I_{mat,1} = 1, I_{mat,2} = 1\right)\\\notag
=&\sum_{t=0}^{D/k-1} \mathbf{Pr}\left(I_{mat,1} = 1, I_{mat,2} = 1|J=t\right)\mathbf{Pr}\left(J=t\right)\\\notag
=& \sum_{t=0}^{D/k-1}\mathbf{Pr}\left(j_T \in S_1\cap S_2, I_{mat,2} = 1|J=t\right)\mathbf{Pr}\left(J=t\right)\\\notag
=& \sum_{t=0}^{D/k-1}\mathbf{Pr}\left(I_{mat,2} = 1|J=t, j_T \in S_1\cap S_2\right)\mathbf{Pr}\left(j_T \in S_1\cap S_2\right)\mathbf{Pr}\left(J=t\right)\\\notag
=& R\sum_{t=0}^{D/k-1}\mathbf{Pr}\left(I_{mat,2} = 1|J=t, j_T \in S_1\cap S_2\right)\mathbf{Pr}\left(J=t\right)
\end{align}
Note that, conditioning on $\{J=t, j_T \in S_1\cap S_2\}$, the problem (i.e., the event $\{I_{mat,2} = 1\}$) is actually the same as our original problem with $f_1+f_2-a-1$ elements whose locations are uniformly random on $\{t+1,t+2,...,D-1\}$.   Therefore,
\begin{align}\notag
&E(I_{mat,1}I_{mat,2})\\\notag
=&R\sum_{t=0}^{D/k-1}\frac{a-1}{f_1+f_2-a-1}\left(1-\prod_{j=0}^{f_1+f_2-a-2}\frac{D\left(1-\frac{1}{k}\right)-t-1-j}{D-t-1-j}\right)\mathbf{Pr}\left(J=t\right)\\\notag
=&R\frac{a-1}{f_1+f_2-a-1}\sum_{t=0}^{D/k-1}\mathbf{Pr}\left(J=t\right)\left(1-\prod_{j=0}^{f_1+f_2-a-2}\frac{D\left(1-\frac{1}{k}\right)-t-1-j}{D-t-1-j}\right)\\\notag
=&R\frac{a-1}{f_1+f_2-a-1}\left(\sum_{t=0}^{D/k-1}\mathbf{Pr}\left(J=t\right)-\sum_{t=0}^{D/k-1}\mathbf{Pr}\left(J=t\right)\prod_{j=0}^{f_1+f_2-a-2}\frac{D\left(1-\frac{1}{k}\right)-t-1-j}{D-t-1-j}\right)\\\notag
\end{align}
By observing that
\begin{align}\notag
&\mathbf{Pr}(J=t) = \frac{\binom{D-t-1}{f_1+f_2-a-1}}{\binom{D}{f_1+f_2-a}} = \frac{f_1+f_2-a}{D}\prod_{j=0}^{t-1}\frac{D-f_1-f_2+a-j}{D-1-j}  = \frac{f_1+f_2-a}{D}\prod_{j=1}^{f_1+f_2-a-1}\frac{D-t-j}{D-j}\\\notag
&\sum_{t=0}^{D/k-1}\mathbf{Pr}(J=t) = 1 - \mathbf{Pr}\left(I_{emp,1}=1\right) = 1-\frac{E\left(N_{emp}\right)}{k} = 1 - \prod_{j=0}^{f_1+f_2-a-1}\frac{D\left(1-\frac{1}{k}\right)-j}{D-j}
\end{align}
we obtain two interesting (combinatorial) identities
\begin{align}\notag
&\frac{f_1+f_2-a}{D}\sum_{t=0}^{D/k-1} \prod_{j=0}^{t-1}\frac{D-f_1-f_2+a-j}{D-1-j} = 1- \prod_{j=0}^{f_1+f_2-a-1}\frac{D\left(1-\frac{1}{k}\right)-j}{D-j}\\\notag
&\frac{f_1+f_2-a}{D}\sum_{t=0}^{D/k-1}\prod_{j=1}^{f_1+f_2-a-1}\frac{D-t-j}{D-j}= 1- \prod_{j=0}^{f_1+f_2-a-1}\frac{D\left(1-\frac{1}{k}\right)-j}{D-j}
\end{align}
which helps us simplify the expression:
\begin{align}\notag
&\sum_{t=0}^{D/k-1}\mathbf{Pr}\left(J=t\right)\prod_{j=0}^{f_1+f_2-a-2}\frac{D\left(1-\frac{1}{k}\right)-t-1-j}{D-t-1-j}\\\notag
=& \sum_{t=0}^{D/k-1} \frac{f_1+f_2-a}{D}\prod_{j=1}^{f_1+f_2-a-1}\frac{D-t-j}{D-j}\prod_{j=0}^{f_1+f_2-a-2}\frac{D\left(1-\frac{1}{k}\right)-t-1-j}{D-t-1-j}\\\notag
=& \sum_{t=0}^{D/k-1} \frac{f_1+f_2-a}{D}\prod_{j=1}^{f_1+f_2-a-1}\frac{D\left(1-\frac{1}{k}\right)-t-j}{D-j}\\\notag
=&\sum_{t=0}^{2D/k-1} \frac{f_1+f_2-a}{D}\prod_{j=1}^{f_1+f_2-a-1}\frac{D-t-j}{D-j}-\sum_{t=0}^{D/k-1} \frac{f_1+f_2-a}{D}\prod_{j=1}^{f_1+f_2-a-1}\frac{D-t-j}{D-j}   \\\notag
=&\left(1- \prod_{j=0}^{f_1+f_2-a-1}\frac{D\left(1-\frac{2}{k}\right)-j}{D-j}  \right) - \left(1- \prod_{j=0}^{f_1+f_2-a-1}\frac{D\left(1-\frac{1}{k}\right)-j}{D-j}  \right)\\\notag
=&- \prod_{j=0}^{f_1+f_2-a-1}\frac{D\left(1-\frac{2}{k}\right)-j}{D-j}  + \prod_{j=0}^{f_1+f_2-a-1}\frac{D\left(1-\frac{1}{k}\right)-j}{D-j}
\end{align}
Combining the results, we obtain
\begin{align}\notag
&E(I_{mat,1}I_{mat,2})\\\notag
=&R\frac{a-1}{f_1+f_2-a-1}\left(1-\prod_{j=0}^{f_1+f_2-a-1}\frac{D\left(1-\frac{1}{k}\right)-j}{D-j}
+ \prod_{j=0}^{f_1+f_2-a-1}\frac{D\left(1-\frac{2}{k}\right)-j}{D-j}  - \prod_{j=0}^{f_1+f_2-a-1}\frac{D\left(1-\frac{1}{k}\right)-j}{D-j} \right)\\\notag
=&R\frac{a-1}{f_1+f_2-a-1}\left(1-2\prod_{j=0}^{f_1+f_2-a-1}\frac{D\left(1-\frac{1}{k}\right)-j}{D-j}
+ \prod_{j=0}^{f_1+f_2-a-1}\frac{D\left(1-\frac{2}{k}\right)-j}{D-j}\right)
\end{align}
And hence
\begin{align}\notag
&Var(N_{mat}) =k(k-1)E(I_{mat,1}I_{mat,2}) + E(N_{mat}) - E^2(N_{mat})\\\notag
=&k(k-1)R\frac{a-1}{f_1+f_2-a-1}\left(1-2\prod_{j=0}^{f_1+f_2-a-1}\frac{D\left(1-\frac{1}{k}\right)-j}{D-j}
+ \prod_{j=0}^{f_1+f_2-a-1}\frac{D\left(1-\frac{2}{k}\right)-j}{D-j}\right)\\\notag
&+ kR\left(1-\prod_{j=0}^{f_1+f_2-a-1}\frac{D\left(1-\frac{1}{k}\right)-j}{D-j}\right) - k^2R^2
\left(1-\prod_{j=0}^{f_1+f_2-a-1}\frac{D\left(1-\frac{1}{k}\right)-j}{D-j}\right)^2
\end{align}
\begin{align}\notag
&\frac{Var(N_{mat})}{k^2}\\\notag
 =& \frac{1}{k}\left(\frac{E(N_{mat})}{k}\right)\left(1-\frac{E(N_{mat})}{k}\right)\\\notag
&+\left(1-\frac{1}{k}\right)R\frac{a-1}{f_1+f_2-a-1}\left(1-2\prod_{j=0}^{f_1+f_2-a-1}\frac{D\left(1-\frac{1}{k}\right)-j}{D-j}
+ \prod_{j=0}^{f_1+f_2-a-1}\frac{D\left(1-\frac{2}{k}\right)-j}{D-j}\right)\\\notag
&-\left(1- \frac{1}{k}\right)R^2
\left(1-\prod_{j=0}^{f_1+f_2-a-1}\frac{D\left(1-\frac{1}{k}\right)-j}{D-j}\right)^2\\\notag
<&\frac{1}{k}\left(\frac{E(N_{mat})}{k}\right)\left(1-\frac{E(N_{mat})}{k}\right)\\\notag
&+\left(1-\frac{1}{k}\right)R^2\left(1-2\prod_{j=0}^{f_1+f_2-a-1}\frac{D\left(1-\frac{1}{k}\right)-j}{D-j}
+ \left(\prod_{j=0}^{f_1+f_2-a-1}\frac{D\left(1-\frac{1}{k}\right)-j}{D-j}\right)^2\right)\\\notag
&-\left(1- \frac{1}{k}\right)R^2\left(1-\prod_{j=0}^{f_1+f_2-a-1}\frac{D\left(1-\frac{1}{k}\right)-j}{D-j}\right)^2\\\notag
=&\frac{1}{k}\left(\frac{E(N_{mat})}{k}\right)\left(1-\frac{E(N_{mat})}{k}\right)
\end{align}

To see the inequality, note that $\frac{a-1}{f_1+f_2-a-1} < R = \frac{a}{f_1+f_2-a}$, and $ \frac{D\left(1-\frac{2}{k}\right)-j}{D-j}< \left( \frac{D\left(1-\frac{1}{k}\right)-j}{D-j}\right)^2$ as proved towards the end of Appendix~\ref{app_lem_Nemp}. This completes the proof.

\section{Proof of Lemma~\ref{lem_CovN}}\label{app_lem_CovN}

\begin{align}\notag
E\left(N_{mat} N_{emp}\right) =& E\left(\sum_{j=1}^k I_{mat,j} \sum_{j=1}^k I_{emp,j} \right)
=\sum_{j=1}^k E\left(I_{mat,j}I_{emp,j}\right) + \sum_{i\neq j} E\left(I_{mat,i}I_{emp,j}\right)\\\notag
=&0+\sum_{i\neq j} E\left(I_{mat,i}I_{emp,j}\right)= k(k-1)E\left(I_{emp,1}I_{mat,2}\right)
\end{align}

\begin{align}\notag
E\left(I_{emp,1}I_{mat,2}\right) =& \mathbf{Pr}\left(I_{emp,1} = 1, I_{mat,2}=1\right)
= \mathbf{Pr}\left(I_{mat,2}=1|I_{emp,1} = 1\right)\mathbf{Pr}\left(I_{emp,1} = 1\right)\\\notag
=& R\left(1-\prod_{j=0}^{f_1+f_2-a-1}\frac{D\left(1-\frac{2}{k}\right)-j}{D\left(1-\frac{1}{k}\right)-j}\right)\prod_{j=0}^{f_1+f_2-a-1}
\frac{D\left(1-\frac{1}{k}\right)-j}{D-j}
\end{align}
\begin{align}\notag
&Cov\left(N_{mat},\ N_{emp}\right)
 = E\left(N_{mat} N_{emp}\right)-E\left(N_{mat}\right)E\left(N_{emp}\right)\\\notag
=&k(k-1)R\left(1-\prod_{j=0}^{f_1+f_2-a-1}\frac{D\left(1-\frac{2}{k}\right)-j}{D\left(1-\frac{1}{k}\right)-j}\right)\left(\prod_{j=0}^{f_1+f_2-a-1}
\frac{D\left(1-\frac{1}{k}\right)-j}{D-j}\right)\\\notag
&-kR\left(1-\prod_{j=0}^{f_1+f_2-a-1}\frac{D\left(1-\frac{1}{k}\right)-j}{D-j}\right)k\left(\prod_{j=0}^{f_1+f_2-a-1}\frac{D\left(1-\frac{1}{k}\right)-j}{D-j}\right)\\\notag
=&k^2R\left(\prod_{j=0}^{f_1+f_2-a-1}\frac{D\left(1-\frac{1}{k}\right)-j}{D-j}\right)\left(\prod_{j=0}^{f_1+f_2-a-1}\frac{D\left(1-\frac{1}{k}\right)-j}{D-j}
-\prod_{j=0}^{f_1+f_2-a-1}\frac{D\left(1-\frac{2}{k}\right)-j}{D\left(1-\frac{1}{k}\right)-j}\right)\\\notag
&-kR\left(1-\prod_{j=0}^{f_1+f_2-a-1}\frac{D\left(1-\frac{2}{k}\right)-j}{D\left(1-\frac{1}{k}\right)-j}\right)\left(\prod_{j=0}^{f_1+f_2-a-1}
\frac{D\left(1-\frac{1}{k}\right)-j}{D-j}\right)
\leq0
\end{align}
To see the inequality, it suffices to show that $g(k)<0$, where
\begin{align}\notag
g(k) =& k\left(\prod_{j=0}^{f_1+f_2-a-1}\frac{D\left(1-\frac{1}{k}\right)-j}{D-j}
-\prod_{j=0}^{f_1+f_2-a-1}\frac{D\left(1-\frac{2}{k}\right)-j}{D\left(1-\frac{1}{k}\right)-j}\right)
-\left(1-\prod_{j=0}^{f_1+f_2-a-1}\frac{D\left(1-\frac{2}{k}\right)-j}{D\left(1-\frac{1}{k}\right)-j}\right)\\\notag
=&k\left(\prod_{j=0}^{f_1+f_2-a-1}\frac{D\left(1-\frac{1}{k}\right)-j}{D-j}\right) -1
-(k-1)\left(\prod_{j=0}^{f_1+f_2-a-1}\frac{D\left(1-\frac{2}{k}\right)-j}{D\left(1-\frac{1}{k}\right)-j}\right)\\\notag
\end{align}

Because $g(k=\infty)=0$, it suffices to show that $g(k)$ is increasing in $k$.

\begin{align}\notag
g(f;k) =&k\left(\prod_{j=0}^{f-1}\frac{D\left(1-\frac{1}{k}\right)-j}{D-j}\right) -1
-(k-1)\left(\prod_{j=0}^{f-1}\frac{D\left(1-\frac{2}{k}\right)-j}{D\left(1-\frac{1}{k}\right)-j}\right)\\\notag
\end{align}
\begin{align}\notag
g(f+1;k) =&k\left(\prod_{j=0}^{f-1}\frac{D\left(1-\frac{1}{k}\right)-j}{D-j}\right) \left(\frac{D\left(1-\frac{1}{k}\right)-f}{D-f}\right)-1
-(k-1)\left(\prod_{j=0}^{f-1}\frac{D\left(1-\frac{2}{k}\right)-j}{D\left(1-\frac{1}{k}\right)-j}\right)
\left(\frac{D\left(1-\frac{2}{k}\right)-f}{D\left(1-\frac{1}{k}\right)-f}\right)\\\notag
=&g(f;k) - \left(\prod_{j=0}^{f-1}\frac{D\left(1-\frac{1}{k}\right)-j}{D-j}\right) \left(\frac{D}{D-f}\right)
+\left(\prod_{j=0}^{f-1}\frac{D\left(1-\frac{2}{k}\right)-j}{D\left(1-\frac{1}{k}\right)-j}\right)
\left(\frac{D\left(1-\frac{1}{k}\right)}{D\left(1-\frac{1}{k}\right)-f}\right)
\end{align}
Thus, it suffices to show
\begin{align}\notag
&- \left(\prod_{j=0}^{f-1}\frac{D\left(1-\frac{1}{k}\right)-j}{D-j}\right) \left(\frac{D}{D-f}\right)
+\left(\prod_{j=0}^{f-1}\frac{D\left(1-\frac{2}{k}\right)-j}{D\left(1-\frac{1}{k}\right)-j}\right)
\left(\frac{D\left(1-\frac{1}{k}\right)}{D\left(1-\frac{1}{k}\right)-f}\right)\leq0\\\notag
\Longleftrightarrow&h(f;k) = \left(\prod_{j=0}^{f-1}\frac{\left(D\left(1-\frac{2}{k}\right)-j\right)\left(D-j\right)}
{\left(D\left(1-\frac{1}{k}\right)-j\right)^2}\right)
\left(\frac{\left(1-\frac{1}{k}\right)(D-f)}{D\left(1-\frac{1}{k}\right)-f}\right)\leq 1
\end{align}
$h(f;k)\leq 1$ holds because one can check that $h(1;k)\leq 1$ and $\frac{\left(D\left(1-\frac{2}{k}\right)-j\right)\left(D-j\right)}
{\left(D\left(1-\frac{1}{k}\right)-j\right)^2}<1$.

This completes the proof.

\section{Proof of Lemma~\ref{lem_Rmat}}~\label{app_lem_Rmat}
We first prove that $\hat{R}_{mat}=\frac{N_{mat}}{k-N_{emp}}$ is unbiased,
\begin{align}\notag
&I_{emp,j}=1 \Rightarrow I_{mat,j} =0\\\notag
& E\Big(I_{mat,j} \Big| I_{emp,j}=0\Big) = R\\\notag
& E\Big(I_{mat,j} \Big| k-N_{emp} = m\Big) = (m/k)R,\ m>0\\\notag
& P\{k-N_{emp}>0\}=1\\\notag
& E\Big(N_{mat} \Big| k-N_{emp} \Big) = R(k - N_{emp})\\\notag
& E\Big(N_{mat}/(k-N_{emp}) \Big| k-N_{emp} \Big) = R \hspace{0.5in} \text{independent of } N_{emp}\\\notag
& E\left(\hat{R}_{mat}\right) = R
\end{align}

Next, we compute the variance.  To simplify the notation, denote  $f = f_1+f_2-a$ and $\tilde{R} = \frac{a-1}{f-1}$. Note that
\begin{align}\notag
& E\Big(I_{mat,1} I_{mat,2} \Big| I_{emp,1}=I_{emp,2}=0\Big) = R(a-1)/(f-1) = R\tilde{R}\\\notag
& R^2-R\tilde{R} = R\{a(f-1)-f(a-1)\}/\{f(f-1)\} = R(1-R)/(f-1)\\\notag
& E\Big(I_{mat,1} I_{mat,2} \Big| I_{emp,1}+I_{emp,2}>0\Big) = 0
\end{align}
By conditioning on $k - N_{emp}$, we obtain
\begin{align}\notag
& E\Big(N_{mat}^2 \Big| k-N_{emp} =m \Big)\\\notag
& = k E\Big(I_{mat,1} \Big| k-N_{emp} =m \Big) + k(k-1)
E\Big(I_{mat,1} I_{mat,2} \Big| k-N_{emp} =m\Big)\\\notag
& = Rm + k(k-1)R\tilde{R}\mathbf{Pr}\Big(I_{emp,1}=I_{emp,2}=0\Big| k-N_{emp} =m\Big)\\\notag
& = Rm + k(k-1)R\tilde{R}{m\choose 2}\Big/{k\choose 2}\\\notag
& = Rm + m(m-1)R\tilde{R}
\end{align}
and
\begin{align}\notag
& E\Big({\hat{R}_{mat}}^2 \Big| k-N_{emp} =m \Big) = R\tilde{R} + (R - R\tilde{R})/m\\\notag
& E{\hat{R}_{mat}}^2 = R\tilde{R} + (R - R\tilde{R})E(k-N_{emp})^{-1}
\end{align}
Combining the above results, we obtain
\begin{align}\notag
 Var\Big({\hat{R}_{mat}}\Big) =& R\tilde{R}-R^2 + (R - R\tilde{R})E(k-N_{emp})^{-1}\\\notag
 =& R(1-R)E(k-N_{emp})^{-1} - (R^2-R\tilde{R})(1-E(k-N_{emp})^{-1})\\\notag
=& R(1-R)E(k-N_{emp})^{-1} - R(1-R)(f-1)^{-1}(1-E(k-N_{emp})^{-1})\\\notag
=&R(1-R)\left\{E(k-N_{emp})^{-1} - (f-1)^{-1} + (f-1)^{-1}E(k-N_{emp})^{-1})\right\}
\end{align}
%
%
%

\section{Proof of Lemma~\ref{lem_var_ratio}}~\label{app_lem_var_ratio}

\begin{align}\notag
g(f;k) = \frac{1}{1-\left(1-\frac{1}{k}\right)^{f}}\left(1+\frac{1}{f-1}\right) -\frac{k}{f-1}
\end{align}

To show $g(f;,k) \leq 1$, it suffices to show
\begin{align}\notag
h(f;k) = (f+k-1)\left(1-\left(1-\frac{1}{k}\right)^f\right) -f \geq 0\hspace{0.5in}
\text{(note that } h(1;k)=0,\ h(2;k)>0)
\end{align}
for which it suffices to show
\begin{align}\notag
&\frac{\partial h(f;k)}{\partial f} = \left(1-\left(1-\frac{1}{k}\right)^f\right) + (f+k-1)\left(-\left(1-\frac{1}{k}\right)^f\log\left(1-\frac{1}{k}\right)\right) -1 \geq 0
\end{align}
and hence it suffices to show  $-1-(f+k-1)\log\left(1-\frac{1}{k}\right) \geq 0$, which is true because
$\log\left(1-\frac{1}{k}\right)<-\frac{1}{k}$.  This completes the proof.

\section{Proof of Lemma~\ref{lem_Nemp_v}}\label{app_lem_Nemp_v}

Recall we first divide the $D$ elements into $k$ bins whose lengths are multinomial distributed with equal probability $\frac{1}{k}$. We denote their lengths by $L_j$, $j = 1$ to $k$. In other words,
\begin{align}\notag
(L_1, L_2, ..., L_k) \sim multinomial\left(D, \frac{1}{k}, \frac{1}{k}, ..., \frac{1}{k}\right)
\end{align}
and we know
\begin{align}\notag
E(L_j) = \frac{D}{k}, \hspace{0.2in} Var(L_j) = D\frac{1}{k}\left(1-\frac{1}{k}\right), \hspace{0.2in} Cov(L_i,\ L_j) = -\frac{D}{k^2}
\end{align}

Define
\begin{align}
I_{i,j} = \left\{\begin{array}{cc}
1 &\text{ if the } i \text{-th element is hashed to the } j \text{-th bin}\\
0 &\text{otherwise}
\end{array}
\right.
\end{align}
We know
\begin{align}\notag
&E(I_{i,j}) = \frac{1}{k}, \hspace{0.2in}E(I_{i,j}^2) = \frac{1}{k},\hspace{0.2in}E(I_{i,j}I_{i,j^\prime}) = 0, \hspace{0.2in}E(I_{i,j}I_{i^\prime,j}) = \frac{1}{k^2},\\\notag
&E(1-I_{i,j}) = 1-\frac{1}{k},\hspace{0.2in}E(1-I_{i,j})^2 = 1-\frac{1}{k},\hspace{0.2in} E(1-I_{i,j})(1-I_{i,j^\prime}) = 1-\frac{2}{k}
\end{align}
Thus
\begin{align}\notag
N_{emp} = \sum_{j=1}^k \prod_{i\in S_1 \cup S_2} \left(1-I_{i,j}\right)
\end{align}

\begin{align}\notag
E\left(N_{emp}\right) = \sum_{j=1}^k \prod_{i\in S_1 \cup S_2} E\left(\left(1-I_{i,j}\right)\right) = k\left(1-\frac{1}{k}\right)^{f_1+f_2-a}
\end{align}

\begin{align}\notag
E\left(N_{emp}^2\right) =& \sum_{j=1}^k \prod_{i\in S_1 \cup S_2} \left(1-I_{i,j}\right)^2
+\sum_{j\neq j^\prime} \prod_{i\in S_1 \cup S_2} \left(1-I_{i,j}\right)\left(1-I_{i,j^\prime}\right)\\\notag
=&k\left(1-\frac{1}{k}\right)^{f_1+f_2-a} + k(k-1)\left(1-\frac{2}{k}\right)^{f_1+f_2-a}
\end{align}
\begin{align}\notag
Var\left(N_{emp}\right)=&k\left(1-\frac{1}{k}\right)^{f_1+f_2-a} + k(k-1)\left(1-\frac{2}{k}\right)^{f_1+f_2-a} - k^2\left(1-\frac{1}{k}\right)^{2(f_1+f_2-a)}
\end{align}
Therefore,
\begin{align}\notag
\frac{Var\left(N_{emp}\right)}{k^2}=&\frac{1}{k}\left(1-\frac{1}{k}\right)^{f_1+f_2-a} \left(1-\left(1-\frac{1}{k}\right)^{f_1+f_2-a}\right)\\\notag
  &-\left(1-\frac{1}{k}\right)\left(\left(1-\frac{1}{k}\right)^{2(f_1+f_2-a)} -\left(1-\frac{2}{k}\right)^{f_1+f_2-a}\right)\\\notag
<& \frac{1}{k}\left(1-\frac{1}{k}\right)^{f_1+f_2-a} \left(1-\left(1-\frac{1}{k}\right)^{f_1+f_2-a}\right)
\end{align}

This completes the proof of  Lemma~\ref{lem_Nemp_v}.

%

\end{document}